\documentclass[conference]{IEEEtran}
\IEEEoverridecommandlockouts
\usepackage{cite}
\usepackage{amsmath,amssymb,amsfonts}
\usepackage{algorithmic}
\usepackage{graphicx}
\usepackage{textcomp}
\usepackage{xcolor}
\def\BibTeX{{\rm B\kern-.05em{\sc i\kern-.025em b}\kern-.08em
    T\kern-.1667em\lower.7ex\hbox{E}\kern-.125emX}}

\usepackage{graphicx}
\usepackage{amsmath}
\usepackage{amssymb}
\usepackage{booktabs}
\usepackage{multirow}
\usepackage{float} 
\usepackage{subcaption} 
\usepackage[mathscr]{eucal}
\usepackage{verbatim}
\usepackage{makecell}
\usepackage{url}
\usepackage{cite}
\usepackage{enumitem}
\usepackage{caption}
\usepackage{amsmath, bm}
\usepackage[pagebackref,breaklinks,colorlinks]{hyperref}

\makeatletter
\newcommand{\linebreakand}{%
  \end{@IEEEauthorhalign}
  \hfill\mbox{}\par
  \mbox{}\hfill\begin{@IEEEauthorhalign}
}
\makeatother

\begin{document}\sloppy
\title{Depth and DOF Cues Make A Better Defocus Blur Detector}
\author{Yuxin Jin$^{*}$\thanks{*~Equal contribution}\quad Ming Qian$^{*}$\quad Jincheng Xiong\quad Nan Xue\quad Gui-Song Xia$^{\dag}$\thanks{\dag~Corresponding author}\\ 
Wuhan University\\
{\tt\small \{jinyuxin,mingqian,JinchengXiong\}@whu.edu.cn,xuenan@ieee.org,guisong.xia@whu.edu.cn}
}

\maketitle
\vspace{-1em}
\noindent
\begin{abstract}
Defocus blur detection (DBD) separates in-focus and out-of-focus regions in an image. Previous approaches mistakenly mistook homogeneous areas in focus for defocus blur regions, likely due to not considering the internal factors that cause defocus blur. Inspired by the law of depth, depth of field (DOF), and defocus, we propose an approach called D-DFFNet, which incorporates depth and DOF cues in an implicit manner. This allows the model to understand the defocus phenomenon in a more natural way. Our method proposes a {\em depth feature distillation} strategy to obtain depth knowledge from a pre-trained monocular depth estimation model and uses a {\em DOF-edge loss} to understand the relationship between DOF and depth. Our approach outperforms state-of-the-art methods on public benchmarks and a newly collected large benchmark dataset, EBD.
Source codes and EBD dataset are available at: \href{https:github.com/yuxinjin-whu/D-DFFNet}{github.com/yuxinjin-whu/D-DFFNet}.
\end{abstract}
\begin{IEEEkeywords}
Defocus blur detection, Depth feature distillation, Depth of field
\end{IEEEkeywords}

\vspace{-0.5em}
\section{Introduction} 
\noindent
Given an image, DBD aims to distinguish the out-of-focus regions from it and facilitates a wide range of applications,
such as salient object detection, monocular depth estimation, bokeh rendering, image deblurring, and quality assessment, {\em etc}.

\begin{figure}[t!]
	\begin{subfigure}{0.28\linewidth}
	    \captionsetup{font={footnotesize}}
		\centering
		\includegraphics[width=\linewidth]{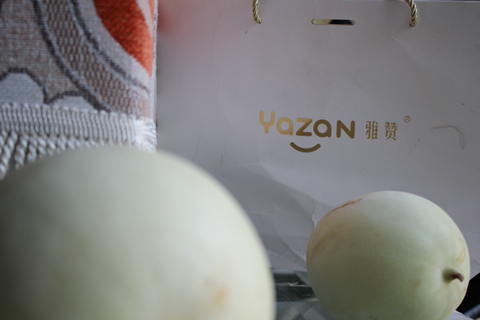}
		\caption{Image}
	\end{subfigure}
	\centering
	\begin{subfigure}{0.28\linewidth}
	    \captionsetup{font={footnotesize}}
		\centering
		\includegraphics[width=\linewidth]{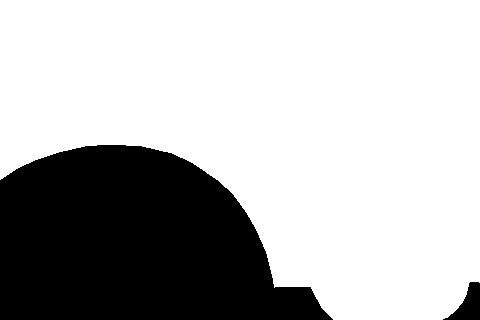}
		\caption{GT}
	\end{subfigure}
	\centering
	\begin{subfigure}{0.28\linewidth}
	    \captionsetup{font={footnotesize}}
		\centering
		\includegraphics[width=\linewidth]{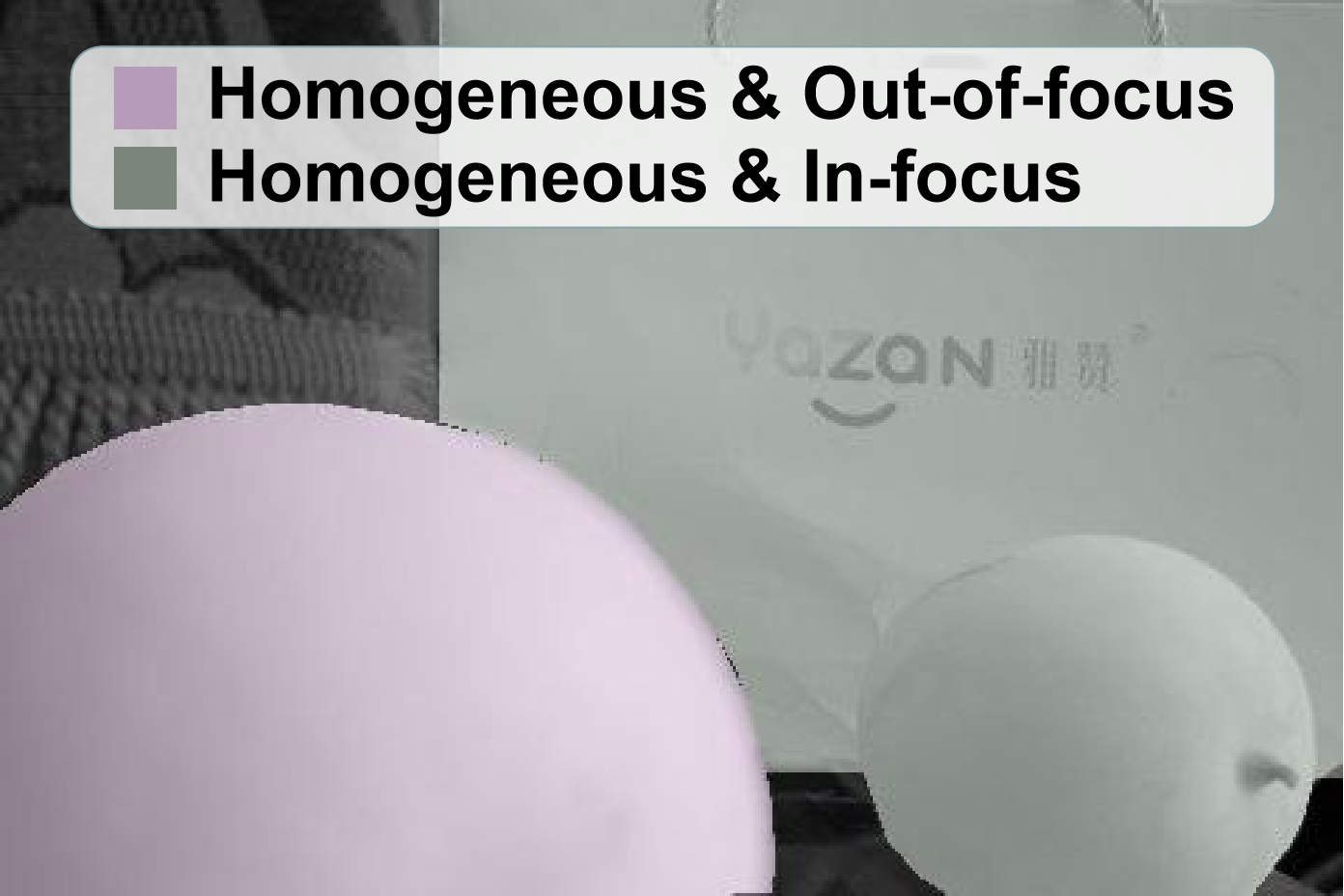}
		\caption{Homogeneous}
	\end{subfigure}

	\centering
	\begin{subfigure}{0.28\linewidth}
	    \captionsetup{font={footnotesize}}
		\centering
		\includegraphics[width=\linewidth]{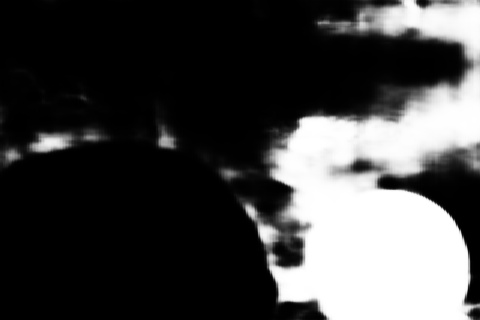}
		\caption{DD~\cite{cun2020defocus}}
	\end{subfigure}   
	\centering
	\begin{subfigure}{0.28\linewidth}
	    \captionsetup{font={footnotesize}}
		\centering
		\includegraphics[width=\linewidth]{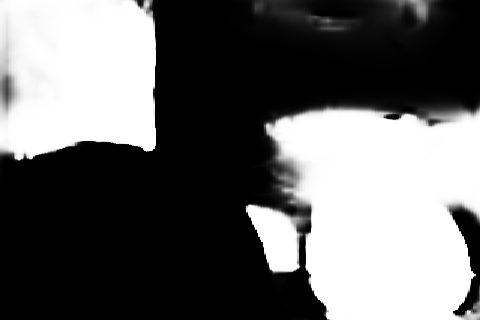}
    	\caption{DFFNet}
	\end{subfigure}
	\centering
	\begin{subfigure}{0.28\linewidth}
	    \captionsetup{font={footnotesize}}
		\centering
		\includegraphics[width=\linewidth]{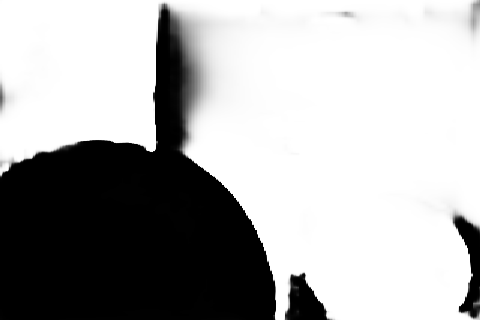}
		\caption{D-DFFNet}
	\end{subfigure}
	\centering
	\captionsetup{font={small}}
	\caption{In (a) and (b), we see the image and ground truth. In (b), the areas that are in focus are visualized as white. The paper bag and fruits are homogeneous regions, but only the foreground fruit is out of focus, as shown in (c). Our proposed DFFNet with depth feature distillation (D-DFFNet) can accurately detect the homogeneous but in-focus regions, while DD~\cite{cun2020defocus} (d) and DFFNet (e) tend to identify them as defocus blur regions.}
	\label{Homogenous regions}
\vspace{-1em}
\end{figure}

For an image of the bokeh effect, an intuitive observation, {\em i.e., homogeneousness}, tells us that the image structures in defocus regions are often smoothed due to blurring and being homogeneous. 
With this assumption, traditional methods~\cite{shi2014discriminative,yi2016lbp,alireza2017spatially} usually rely on low-level cues of image ({\em e.g.}, image gradient) to model defocus regions and well perform on images with rich scene structures. 
However, it is worth noticing that there is no one-to-one correspondence between homogeneous regions and defocus regions, saying that defocus regions are often smoothed but smooth regions might be in-focus. One can check Fig.~\ref{Homogenous regions} for an instance, where both the regions of the paper bag and fruits are homogeneous, while only the region containing the front fruit is out-of-focus.
Thus, it is demanded to pursue better models for defocus regions. 

Recent studies alleviate this issue by relying on deep convolutional neural networks (CNN) and have reported promising performance thanks to the strong representation capability of deep models~\cite{zhao2018defocus,zhao2019defocus,zhao2019enhancing,tang2020br,zhao2021defocus,zhao2021image,cun2020defocus,tang2020defusionnet,li2021layer,jiang2022ma}.
Although the defocus regions are better described with implicitly learned deep features, it is still difficult to distinguish smooth and in-focus regions from defocus ones.

Notice that, in an ideal case, if the camera model and the 3D geometry of the scene are known, the DBD problem will turn out to be well-posed and easy to approach. 
Although it is always infeasible to achieve this with only a single image, we argue that hinting at the detector image depth and partial camera model (DOF) would be an alternative.
To implement the above hints, we present two approaches separately:
\par- A {\em depth feature distillation} method to get depth knowledge.
\par- A {\em DOF-edge loss} helps to recognize the Depth of Field (DOF)~\cite{timmurphy.org2} in each image in an implicit way.

To obtain depth representation from the input images, we introduce {\em depth feature distillation} using a knowledge distillation strategy~\cite{gou2021knowledge} to transfer multi-level feature knowledge from a pre-trained depth estimation model to a defocus model. Our approach uses feature-alignment projectors and pair-wise loss to jointly distill knowledge from a pre-trained depth teacher model and a defocus teacher model to a defocus student model, ensuring a balanced representation of depth and defocus information. This is illustrated in Fig.~\ref{method}.

In addition to obtaining implicit 3D geometry information in the defocus blur detector, we also introduce the DOF parameter in a simple but effective way. 
DOF is the range of distances in a scene that appears in focus, and it is determined by the boundary between the in-focus and defocus regions. To incorporate DOF in our model, we use a `DOF-edge loss' to provide additional supervision on the edge of DOF regions during the learning of depth. This helps the model to understand the relationship between DOF and depth, and to be more accurate in detecting defocus blur.
The implicit depth and DOF cues make the detector more robust and help it resolve the `homogeneous region' problem.

A recent work called DD~\cite{cun2020defocus} also introduced depth information to the DBD task. However, compared to our method, they only used a response-based knowledge distillation strategy~\cite{gou2021knowledge} to obtain depth representation, which is easy to implement but not as effective. Additionally, they ignored the importance of the camera model in detecting defocus regions. In contrast, our method considers the depth of the feature aspect and takes into account the DOF at the same time, providing more accurate and robust results.

Last but not least, we recognize that there are some limitations in current DBD benchmarks~\cite{shi2014discriminative,zhao2018defocus,tang2020defusionnet}, such as too few pictures, the absence of high-resolution images, and the lack of images captured with wide or shallow DOF. To avail evaluation, we collect a new test dataset called EBD, consisting of 1605 high-resolution images selected from the EBB!~\cite{ignatov2020rendering} dataset and manually labeled with pixel-wise DBD maps. 1305 images of EBD were captured with a shallow DOF, resulting in a strong bokeh effect, and the other 300 images were shot with a wide DOF, resulting in sharp photos, as shown in Fig.~\ref{EBD}.

Our main contributions are summarized below.
\begin{itemize}[leftmargin=*]
\item We propose a {\em depth feature distillation} strategy and a {\em DOF-edge loss} to hint at the defocus blur detector depth knowledge and camera model information separately, which significantly and intuitively improves the detector.
\item We build a large-scale benchmark named {\em EBD} with 1605 high-resolution, well-annotated images, featuring more complex scenes and a wider range of DOF settings.
\item We conduct experiments with 14 state-of-the-art methods, and results show that our method produces much better results.
\end{itemize}
\label{sec:intro}

\section{Related Works}
\subsection{CNN-based Defocus Blur Detection Methods}
\noindent
Many CNN-based methods for DBD focused on network design~\cite{zhao2018defocus,tang2020defusionnet,zhao2019enhancing,zhao2021defocus,tang2020br,zhao2021image,li2021layer,jiang2022ma}. However, some other methods explored using auxiliary information from related tasks to guide DBD. For example, some works~\cite{qian2020defocus,zhang2020rethinking} used salient information to guide their approach. While salient regions are often the regions that people pay attention to, they do not necessarily coincide with the region of the camera's depth of field, which can introduce bias when there is no overlap.
In contrast, the 3D geometry of the scene (image depth) is more relevant to the defocus blur detection task. Cun~\emph{et al.}\cite{cun2020defocus} proposed depth distillation (DD), which uses a distillation strategy to introduce depth knowledge from response-based distillation\cite{gou2021knowledge}. While this approach is easy to implement, it is less effective.
In this work, we propose a depth feature distillation strategy that distills multi-level depth feature knowledge. In addition, we introduce the camera model (DOF) to defocus blur detection for the first time, using a DOF-edge loss. This allows our model to better understand the underlying factors that cause defocus blur and improve its performance.

\begin{figure}[t!]
\centering
{
\includegraphics[width=\linewidth]{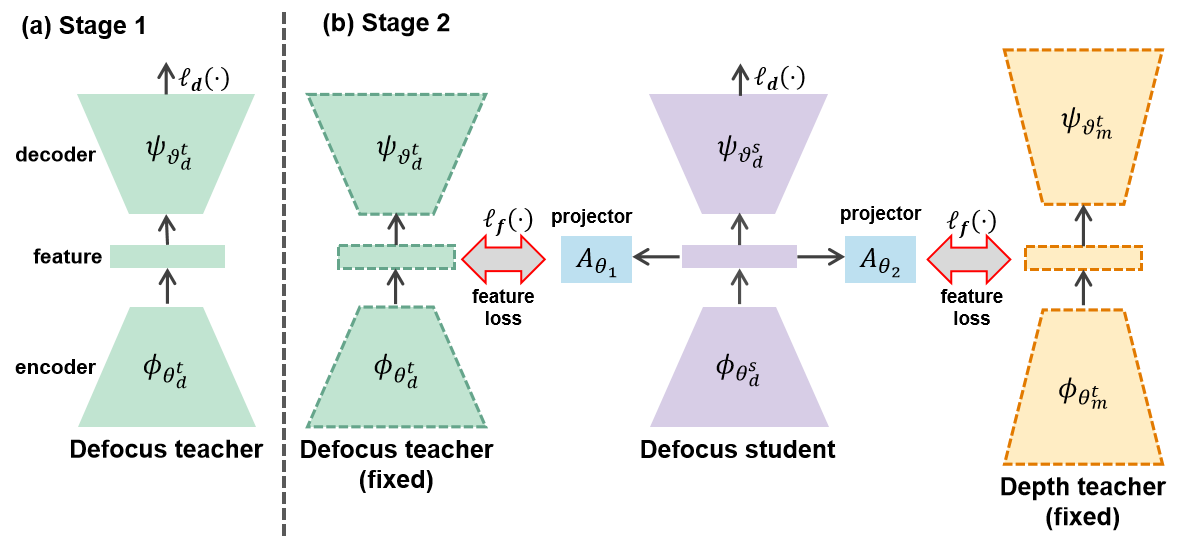}}
\hspace{0in}
\vspace{-1.5em}
\caption{Diagram of depth feature distillation strategy.}
\label{method}
\vspace{-1.5em}
\end{figure}

\subsection{Knowledge Distillation}
\noindent
The goal of vanilla knowledge distillation is to train a small student model using the knowledge from a larger teacher model. Response-based knowledge distillation~\cite{gou2021knowledge} directly mimics the final predictions of the teacher model. Romero~\cite{romero2014fitnets} proposed using the intermediate representations learned by the teacher as hints to improve the training of the student model. This has led to the development of feature-based knowledge distillation.
Our proposed method, called depth feature distillation, is related to feature-based knowledge distillation. However, instead of distilling a compact student model from a large teacher model, we use the feature knowledge from a depth estimation model to guide the training of the DBD model, helping it to learn depth knowledge. This allows our DBD model to better understand the 3D geometry of the scene, which is important for accurately detecting defocus blur.

\section{Methods}
\subsection{Physical Relation of DOF, Depth and Defocus}
\noindent
DOF in photography refers to the range of distances in a photo that appear sharp. The amount of DOF is determined by factors such as the aperture size and the distance to the subject (depth). Defocus, or bokeh, is the aesthetic quality of the out-of-focus areas of a photograph. A shallow DOF means that only a small range of distances from the camera is in focus, resulting in a strong bokeh effect. A wide DOF means that a large range of distances is in focus, resulting in a sharp photograph with little or no bokeh. Our DBD model leverages DOF information and depth representation to detect defocus regions in a natural way.

\subsection{Depth Feature Distillation Strategy}
\noindent
To incorporate this relationship into our DBD algorithm, we propose the {\em depth feature distillation}  strategy that introduces 3D information (image depth) to the DBD model. This is done using a two-stage training process, as illustrated in Fig.~\ref{method}. 

In the first stage (Fig.~\ref{method}(a)), we train a defocus blur detector from scratch without using any depth information.
The defocus model can predict defocus blur independently. We train the model to map the input image $\boldsymbol{x}$ to the target label $\boldsymbol{y}$.

We describe the encoding process as $\phi(\boldsymbol{x};\theta_d^t)$, which takes an image as input and outputs a high-dimensional feature. The decoding process is described as $\psi(\phi(\boldsymbol{x};\theta_d^t);\vartheta_d^t)$, which takes the encoded feature as input and predicts the final defocus blur output. Here, $\theta_d^t$ and $\vartheta_d^t$ are the parameters of the encoder and decoder, respectively. The subscript $d$ denotes the defocus blur detector, and the superscript $t$ denotes the teacher model.

The network parameters $\theta_d^t$ and $\vartheta_d^t$ are optimized using a DBD loss function $\ell_d$ over the training samples, which measures the distance between the ground truth and the predictions. This is formulated as follows:
\vspace{-0.3em}
\begin{equation}
\underset{\theta_d^t\vartheta_d^t}{min}\sum_{\boldsymbol{x},\boldsymbol{y}\in \emph{$\mathcal{D}$}}{\ell_d(\psi(\phi(\boldsymbol{x};\theta_d^t);\vartheta_d^t),\boldsymbol{y})}
\vspace{-0.5em}
\end{equation}

\par
In the second stage (shown in Fig.~\ref{method}(b)), we implement our depth feature distillation strategy. In this stage, we use two teacher models: the pre-trained defocus model from the first stage as the defocus teacher model, and a pre-trained depth estimation model called Midas~\cite{ranftl2020towards} as the depth teacher model. In this stage of training, we simultaneously distill feature knowledge from both the defocus teacher model and the depth teacher model into the defocus student model, while training the student model.
This is achieved by minimizing the distance between the encoder features  of the teacher and student models. In specific, we adopt the squared difference to formulate the pair-wise similarity distillation loss, as follows:

\vspace{-1em}
\begin{equation}
\ell_2(\boldsymbol{u},\boldsymbol{v})=\left\|\frac{\boldsymbol{u}}{\left\|\boldsymbol{u}\right\|_2}-\frac{\boldsymbol{v}}{\left\|\boldsymbol{v}\right\|_2}\right\|_2^2
\vspace{-0.3em}
\end{equation}

Due to differences in feature dimension between the defocus and depth features, we use two projectors (1×1 convolutional layer), denoted as $A_{\theta_1}$ and $A_{\theta_2}$, for feature alignment, as in the work of Li~\cite{li2020knowledge}. $\theta_1$ and $\theta_2$ are parameters of the two projectors, respectively.  
We optimize the parameters of the defocus student network (encoder: $\theta_d^s$, decoder: $\vartheta_d^s$) and the parameters of the two projectors ($\theta_1$ and $\theta_2$) using the feature loss $\ell_f$ and the DBD loss $\ell_d$, the feature loss $\ell_f$ is defined as:
\vspace{-0.5em}
\begin{equation}
\begin{split}
\ell_f = 
\ell_2(A_{\theta_1}(\phi(\boldsymbol{x};\theta_d^s)),\phi(\boldsymbol{x};\theta_d^t))+\\
\ell_2(A_{\theta_2}(\phi(\boldsymbol{x};\theta_d^s)),\phi(\boldsymbol{x};\theta_m^t)) 
\end{split}
\vspace{-0.3em}
\end{equation}
where the subscript $m$ refers to monocular depth estimation, the subscript $d$ refers to DBD as above, the superscript $s$ indicates the student model, and the superscript $t$ indicates the teacher model as above. $\theta_d^t$ and $\theta_m^t$ are frozen parameters from the defocus teacher encoder and the depth teacher encoder, respectively.

Our carefully-designed distillation strategy eliminates the need to alter the architecture of the DBD model, enabling our method to be applied to any similar architecture with no impact on inference speed. The only additional cost is training a defocus student model for distillation, compared to using an existing DBD model.
\begin{figure}[t!]
\centering
{
\includegraphics[width=\linewidth]{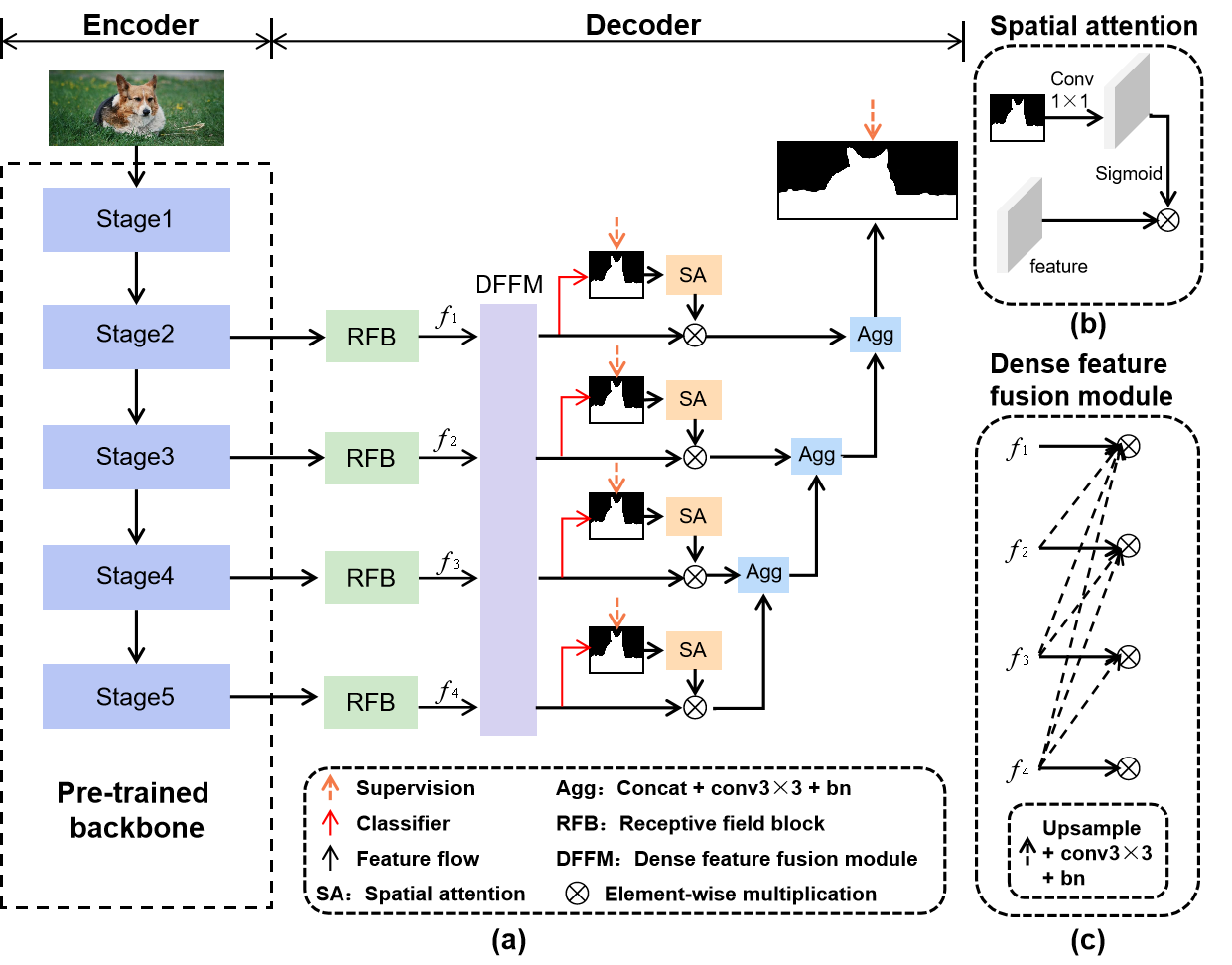}}
\hspace{0in}
\vspace{-1.5em}
\caption{Overview of proposed DFFNet architecture.}
\label{network}
\vspace{-1.5em}
\end{figure}

\begin{table*}[t!]\scriptsize
\centering
\caption{
\centering
Quantitative comparison with methods training on CUHK-TR-1. EBD(1305) indicates all 1305 shallow DOF images from EBD dataset. DFFNet refers to our proposed model without depth feature distillation, D-DFFNet represents DFFNet with depth feature distillation, and R-DFFNet represents DFFNet with response-based depth knowledge distillation proposed in DD.  The best results are marked in bold. $\uparrow$ : larger is better, $\downarrow$ : smaller is better.}
\vspace{-0.3em}
\begin{tabular}{|c|ccc|ccc|ccc|ccc|ccc|}\hline
\multirow{2}{*}{Method} & \multicolumn{3}{c}{CUHK-TE-1} \vline & \multicolumn{3}{c}{DUT-TE} \vline & \multicolumn{3}{c}{CTCUG} \vline & \multicolumn{3}{c}{EBD} \vline & \multicolumn{3}{c}{EBD(1305)} \vline\\
                         & MAE$\downarrow$     & $F_\beta\uparrow$      & IoU$\uparrow$    & MAE$\downarrow$    & $F_\beta\uparrow$      & IoU$\uparrow$   & MAE$\downarrow$     & $F_\beta\uparrow$      & IoU$\uparrow$    & MAE$\downarrow$    & $F_\beta\uparrow$      & IoU$\uparrow$   & MAE$\downarrow$      & $F_\beta\uparrow$        & IoU$\uparrow$     \\\hline
DBDF~\cite{shi2014discriminative}                    & 0.292   & 0.806  & 0.549  & 0.384  & 0.754  & 0.519 & 0.360   & 0.668  & 0.438  & 0.389  & 0.603  & 0.435 & 0.507    & 0.590    & 0.357   \\
LBP~\cite{yi2016lbp}                      & 0.154   & 0.917  & 0.755  & 0.199  & 0.873  & 0.783 & 0.291   & 0.732  & 0.638  & 0.334  & 0.705  & 0.611 & 0.353    & 0.757    & 0.620   \\
HiFST~\cite{alireza2017spatially}                    & 0.223   & 0.803  & 0.743  & 0.313  & 0.839  & 0.640 & 0.274   & 0.775  & 0.618  & 0.376  & 0.603  & 0.498 & 0.501    & 0.594    & 0.407   \\
BTBNet~\cite{zhao2018defocus}                   & 0.110    & 0.949  & 0.914  & 0.196  & 0.861  & 0.803 & 0.177   & 0.809  & 0.762  & --     & --     & --    & --       & --       & --      \\
BTBNet2~\cite{zhao2019defocus}                  & 0.085   & 0.935  & 0.908  & 0.145  & 0.873  & 0.837 & --      & --     & --     & --     & --     & --    & --       & --       & --      \\
CENet~\cite{zhao2019enhancing}                    & 0.061   & 0.945  & 0.932  & 0.141  & 0.869  & 0.833 & 0.117   & 0.845  & 0.820  & 0.072  & 0.810  & 0.899 & 0.062    & 0.945    & 0.924   \\
BR2Net~\cite{tang2020br}                   & 0.059   & 0.966  & 0.927  & 0.083  & 0.943 & 0.893 & 0.140   & 0.834  & 0.788  & 0.087  & 0.813  & 0.880 & 0.063    & 0.948    & 0.924   \\
DD~\cite{cun2020defocus}                       & 0.045  & 0.966  & 0.941  & 0.074  & 0.935  &0.903& 0.155   & 0.797  & 0.775  & 0.118  & 0.812  & 0.842 & 0.048    & 0.950    & 0.940   \\
DefuNet~\cite{tang2020defusionnet}                  & --      & --     & --     & 0.085  & \textbf{0.952}  & 0.889 & 0.132   & 0.828  & 0.798  & --     & --     & --    & --       & --       & --      \\
IS2CNet~\cite{zhao2021image}                  & 0.049   & 0.964  & 0.937  & 0.142  & 0.868  & 0.831 & 0.112   & 0.858  & 0.826  & \textbf{0.070}  & 0.809  & \textbf{0.901} & 0.062    & 0.944    & 0.923   \\\hline
DFFNet                   & 0.039   & 0.971  & 0.947  & 0.072  & 0.938  & 0.903 & 0.082   & 0.879  & 0.868  & 0.091  & 0.823  & 0.872 &\textbf{0.038}&0.961&0.951\\
R-DFFNet                      & 0.041   & 0.968  & 0.945  & 0.091  & 0.920  & 0.881 & 0.082   & 0.895  & 0.871  & 0.088  & 0.819  & 0.879 & 0.045 & 0.957 & 0.944\\
D-DFFNet            & \textbf{0.036}  & \textbf{0.973}  & \textbf{0.951}  & \textbf{0.070}  & 0.939  & \textbf{0.906} & \textbf{0.074}  & \textbf{0.892}  & \textbf{0.878}  & 0.084  & \textbf{0.826}  & 0.882 & \textbf{0.038}    & \textbf{0.963}   & \textbf{0.952} \\
\hline
\end{tabular}
\label{comparison with methods training on CUHK-TR-1}
\end{table*}

\begin{table*}[t!]\scriptsize
\centering
\caption{
\centering
Quantitative comparison with methods training on CUHK-TR-1 \& DUT-TR.}
\vspace{-0.3em}
\begin{tabular}{|c|ccc|ccc|ccc|ccc|ccc|}\hline
\centering
\multirow{2}{*}{Method} & \multicolumn{3}{c}{CUHK-TE-1} \vline & \multicolumn{3}{c}{DUT-TE}\vline & \multicolumn{3}{c}{CTCUG} \vline& \multicolumn{3}{c}{EBD}\vline & \multicolumn{3}{c}{EBD(1305)}\vline \\
                         & MAE$\downarrow$   &$F_\beta\uparrow$   & IoU$\uparrow$   & MAE$\downarrow$    & $F_\beta\uparrow$    & IoU$\uparrow$   & MAE$\downarrow$     & $F_\beta\uparrow$     & IoU$\uparrow$    & MAE$\downarrow$    & $F_\beta\uparrow$      & IoU$\uparrow$   & MAE$\downarrow$      &$F_\beta\uparrow$    & IoU$\uparrow$     \\\hline
AENet~\cite{zhao2021defocus}                    & 0.049 & 0.964 & 0.935 & 0.120  & 0.886  & 0.855 & 0.114   & 0.845  & 0.823  & \textbf{0.069}  & 0.811  & \textbf{0.901} & 0.053    & 0.946    & 0.933   \\
EFENet~\cite{zhao2021defocus}                   & 0.052 & 0.958 & 0.935 & 0.091  & 0.927  & 0.883 & 0.116   & 0.837  & 0.827  & 0.113  & 0.794  & 0.852 & 0.069    & 0.928    & 0.919   \\
LOCAL~\cite{li2021layer}                    & \textbf{0.039} & --    & --    & 0.069  & --     & --    & 0.093   & --     & --     & --     & --     & --    & --       & --       & --      \\
D-DFFNet                     & \textbf{0.039}  & \textbf{0.970} & \textbf{0.947} & \textbf{0.060}  & \textbf{0.950}  & \textbf{0.919} & \textbf{0.075}   & \textbf{0.893}  & \textbf{0.878}  & 0.094  & \textbf{0.825}  & 0.868 & \textbf{0.036}    & \textbf{0.963}    & \textbf{0.954}  \\\hline
\end{tabular}
\label{comparison with methods training on CUHK-TR-1 & DUT-TR}
\vspace{-1em}
\end{table*}

\begin{table}[t!]\scriptsize
\centering
\caption{
\centering
Quantitative comparison with methods training on CUHK-TR-2 \& DUT-TR.}
\vspace{-0.3em}
\begin{tabular}{|c|cc|cc|cc|}\hline
\centering
\multirow{2}{*}{Method} & \multicolumn{2}{c}{CUHK-TE-2}\vline & \multicolumn{2}{c}{DUT-TE}\vline & \multicolumn{2}{c}{CTCUG} \vline\\
                         & MAE$\downarrow$        & IoU$\uparrow$         & MAE$\downarrow$        & IoU$\uparrow$        & MAE$\downarrow$         & IoU$\uparrow$         \\\hline
MA-GANet~\cite{jiang2022ma}                 & 0.084       & 0.886       & 0.070      & 0.907      & 0.105       & 0.802       \\
D-DFFNet            & \textbf{0.078}       & \textbf{0.889}       & \textbf{0.061}     & \textbf{0.917}    & \textbf{0.071}      & \textbf{0.883}      \\\hline
\end{tabular}
\label{comparison with methods training on CUHK-TR-2 & DUT-TR}
\vspace{-1.2em}
\end{table}

\subsection{DOF-edge Loss}
\noindent
We introduce DOF information using DOF-edge loss.
Specifically, we use a dice-based loss function $\ell_e$ to measure the distance between edge predictions and their corresponding labels, as described by Zheng~\cite{zheng2020parsing}:
\vspace{-0.3em}
\begin{equation}
\ell_e(\boldsymbol{\hat{E}},\boldsymbol{E})=1-Dice(\boldsymbol{\hat{E}},\boldsymbol{E})
\vspace{-0.3em}
\end{equation}
where $\boldsymbol{\hat{E}}$ represents the edge of DBD prediction, while $\boldsymbol{E}$ denotes the edge of DBD label.
In addition to the DOF-edge loss, we use binary cross entropy loss (BCE) to calculate the pixel-wise difference between the DBD maps $\boldsymbol{\hat{y}}$ and their corresponding labels $\boldsymbol{y}$, as shown below:
\vspace{-0.3em}
\begin{equation}
\ell_{bce}(\boldsymbol{\hat{y}},\boldsymbol{y})=\boldsymbol{y}\log \boldsymbol{\hat{y}}+(1-\boldsymbol{y})\log(1-\boldsymbol{\hat{y}})
\vspace{-0.3em}
\end{equation}

We use both DOF-edge loss and BCE loss for the final defocus predictions $\boldsymbol{\hat{y}}$ and the four side outputs $\boldsymbol{\hat{y_k^{\prime}}}$ ($k\in[1,..,4] $), using a hyper-parameter $\lambda$. In our experiments, we set $\lambda$ experimentally to 0.5. The final DBD loss $\ell_d$ is defined as follows:
\vspace{-0.3em}
\begin{equation}
\ell_d = 
\ell_{bce}
+\lambda \ell_e
\vspace{-0.3em}
\end{equation}

\subsection{DFFNet Model Architecture}
\noindent
The model DFFNet is used as both the defocus teacher model and the defocus student model.
The structure of the DFFNet is illustrated in Fig.~\ref{network}(a). For feature extraction, we use a pre-trained ResNeSt101~\cite{zhang2022resnest} as the backbone. In each decoder branch, receptive field blocks (RFBs)~\cite{liu2018receptive} are used to enhance and refine the features. The features from different branches are then densely fused using a dense feature fusion module (DFFM). Multi-scale predictions are generated from each branch using side classifiers (1x1 convolutional layers). After that, we use a spatial attention block (Fig.~\ref{network}(b)) to generate attention maps based on the predictions. These attention maps are then used to weight the features via element-wise multiplication. Finally, we aggregate the features of each branch in the bottom-top pathway to generate the final prediction. More details can be found in the supplemental material.\\
\vspace{-1.5em}
\subsection{Optimization}
\noindent
Stage 1: the loss function is defined as follows:
\vspace{-0.3em}
\begin{equation}
\ell=
\ell_d^f+
\sum_k \alpha_k \ell_d^k
\vspace{-1em}
\end{equation}
\vspace{0.3em}
\par
\noindent
Stage 2: the loss function is defined as follows:
\begin{equation}
\ell=
\ell_d^f+
\sum_k \alpha_k \ell_d^k+ \beta\ell_f
\vspace{-0.5em}
\end{equation}

where $\ell_d^f$ represents the $\ell_d$ loss for the final defocus predictions, and $\ell_d^k$ represents the $\ell_d$ loss for the k-level side outputs. $\alpha_k$($k\in[1,..,4] $) and $\beta$ are trade-off parameters. We set $\alpha_k$($k\in[1,..,4] $) to 1, and modify the value of $\beta$ with the training epoch:
\vspace{-0.3em}
\begin{equation}
\begin{aligned}
\beta =\left\{\begin{matrix} 
  $3$ \quad epoch \leq 15\\  
  3\ast (\frac{epoch-15}{lastepoch}) \quad 15 < epoch \leq lastepoch
\end{matrix}\right. 
\end{aligned}
\end{equation}

\begin{figure*}[t!]
\captionsetup{font={small}}
	\begin{subfigure}{0.066\linewidth}
		\centering
		\includegraphics[width=\linewidth]{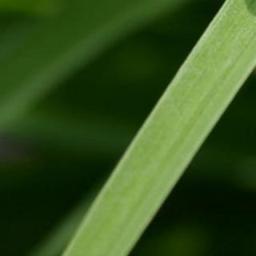}
	\end{subfigure}
	\begin{subfigure}{0.066\linewidth}
		\centering
		\includegraphics[width=\linewidth]{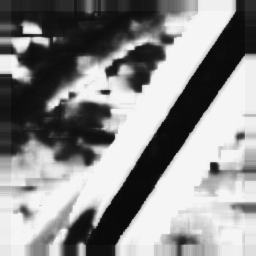}
	\end{subfigure}
	\begin{subfigure}{0.066\linewidth}
		\centering
		\includegraphics[width=\linewidth]{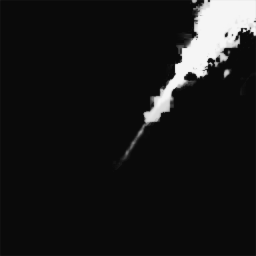}
	\end{subfigure}
	\begin{subfigure}{0.066\linewidth}
		\centering
		\includegraphics[width=\linewidth]{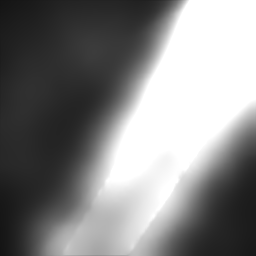}
	\end{subfigure}
	\begin{subfigure}{0.066\linewidth}
		\centering
		\includegraphics[width=\linewidth]{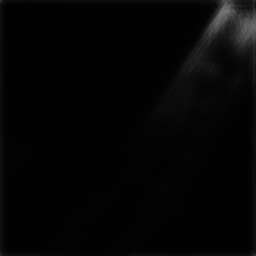}
	\end{subfigure}
	\begin{subfigure}{0.066\linewidth}
		\centering
		\includegraphics[width=\linewidth]{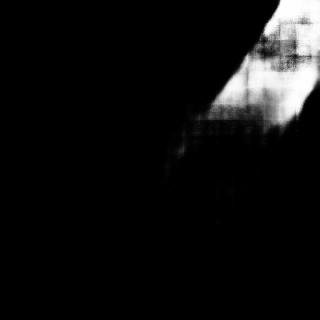}
	\end{subfigure}
	\begin{subfigure}{0.066\linewidth}
		\centering
		\includegraphics[width=\linewidth]{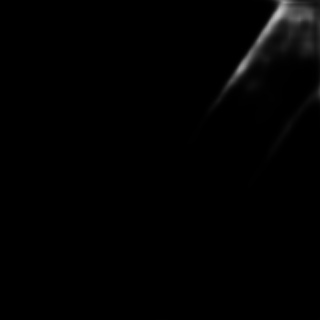}
	\end{subfigure}
	\begin{subfigure}{0.066\linewidth}
		\includegraphics[width=\linewidth]{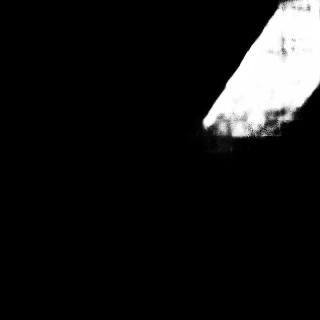}
	\end{subfigure}
	\begin{subfigure}{0.066\linewidth}
		\includegraphics[width=\linewidth]{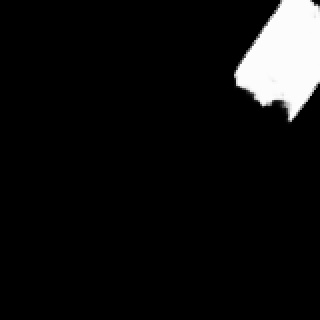}
	\end{subfigure}
	\begin{subfigure}{0.066\linewidth}
		\centering
		\includegraphics[width=\linewidth]{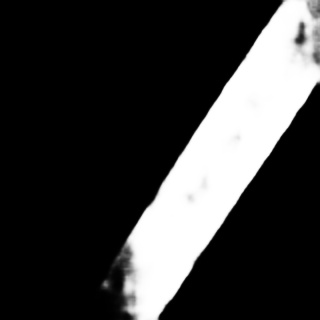}
	\end{subfigure}
	\begin{subfigure}{0.066\linewidth}
		\centering
		\includegraphics[width=\linewidth]{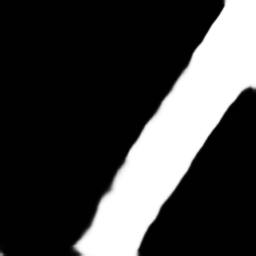}
	\end{subfigure}
	\begin{subfigure}{0.066\linewidth}
		\centering
		\includegraphics[width=\linewidth]{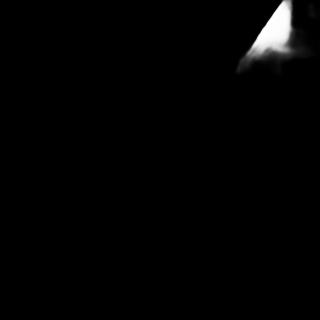}
	\end{subfigure}
	\begin{subfigure}{0.066\linewidth}
		\centering
		\includegraphics[width=\linewidth]{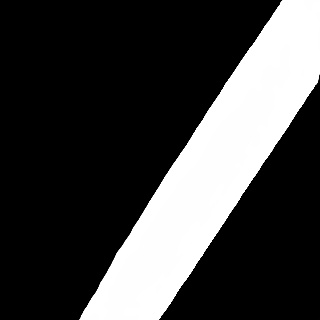}
	\end{subfigure}
	\begin{subfigure}{0.066\linewidth}
		\centering
		\includegraphics[width=\linewidth]{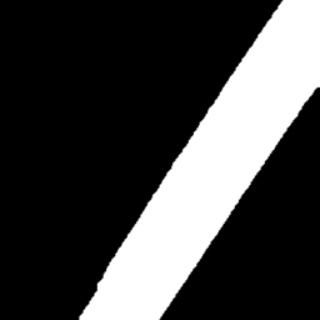}
	\end{subfigure}

	\begin{subfigure}{0.066\linewidth}
		\captionsetup{font={tiny}}
		\centering
		\includegraphics[width=\linewidth]{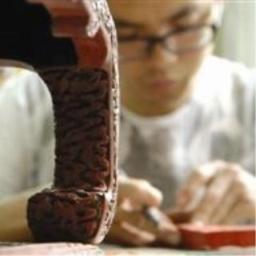}
		\caption{Images}
	\end{subfigure}
	\begin{subfigure}{0.066\linewidth}
	\captionsetup{font={tiny}}
		\centering
		\includegraphics[width=\linewidth]{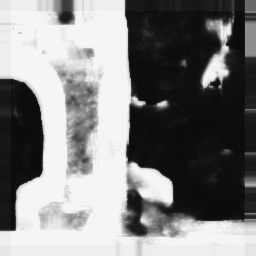}
		\caption{DBDF}
	\end{subfigure}
	\begin{subfigure}{0.066\linewidth}
	\captionsetup{font={tiny}}
		\centering
		\includegraphics[width=\linewidth]{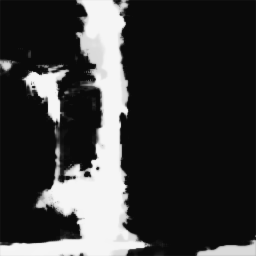}
		\caption{LBP}
	\end{subfigure}
	\begin{subfigure}{0.066\linewidth}
		\captionsetup{font={tiny}}
		\centering
		\includegraphics[width=\linewidth]{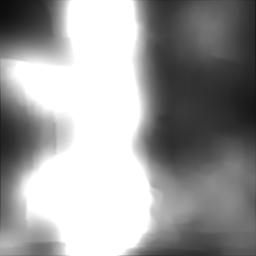}
		\caption{HiFST}
	\end{subfigure}
	\begin{subfigure}{0.066\linewidth}
		\captionsetup{font={tiny}}
		\centering
		\includegraphics[width=\linewidth]{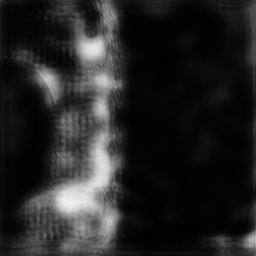}
		\caption{BTBNet}
	\end{subfigure}
	\begin{subfigure}{0.066\linewidth}
		\captionsetup{font={tiny}}
		\centering
		\includegraphics[width=\linewidth]{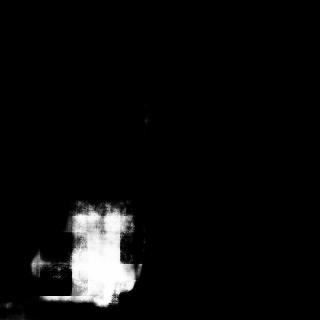}
		\caption{CENet}
	\end{subfigure}
	\begin{subfigure}{0.066\linewidth}
		\captionsetup{font={tiny}}
		\centering
		\includegraphics[width=\linewidth]{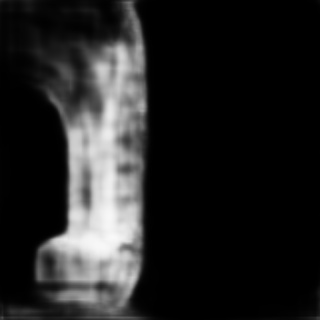}
		\caption{BTBNet2}
	\end{subfigure}
	\begin{subfigure}{0.066\linewidth}
		\captionsetup{font={tiny}}
		\includegraphics[width=\linewidth]{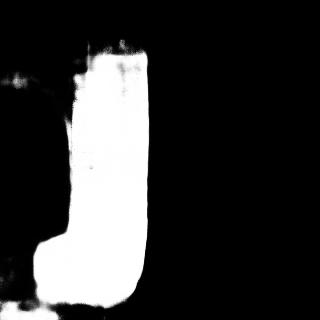}
		\caption{AENet}
	\end{subfigure}
	\begin{subfigure}{0.066\linewidth}
		\captionsetup{font={tiny}}
		\includegraphics[width=\linewidth]{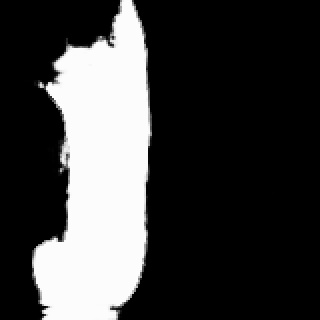}
		\caption{EFENet}
	\end{subfigure}
	\begin{subfigure}{0.066\linewidth}
		\captionsetup{font={tiny}}
		\centering
		\includegraphics[width=\linewidth]{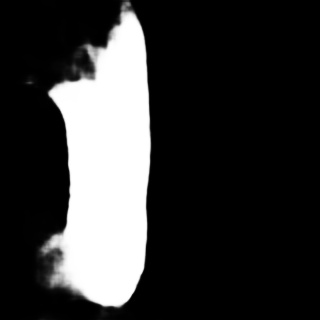}
		\caption{DD}
	\end{subfigure}
	\begin{subfigure}{0.066\linewidth}
		\captionsetup{font={tiny}}
		\centering
		\includegraphics[width=\linewidth]{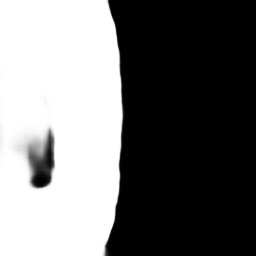}
		\caption{DeFuNet}
	\end{subfigure}
	\begin{subfigure}{0.066\linewidth}
		\captionsetup{font={tiny}}
		\centering
		\includegraphics[width=\linewidth]{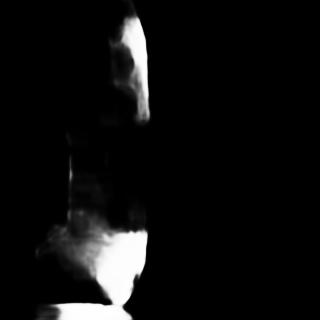}
		\caption{IS2CNet}
	\end{subfigure}
	\begin{subfigure}{0.066\linewidth}
		\captionsetup{font={tiny}}
		\centering
		\includegraphics[width=\linewidth]{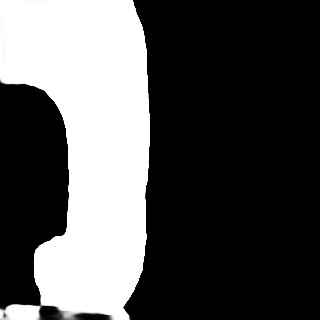}
		\caption{D-DFFNet}
	\end{subfigure}
	\begin{subfigure}{0.066\linewidth}
		\captionsetup{font={tiny}}
		\centering
		\includegraphics[width=\linewidth]{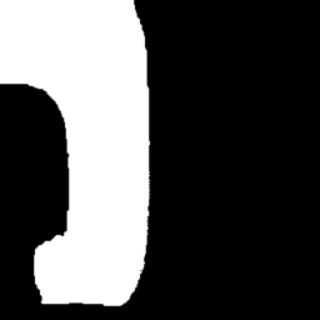}
		\caption{GTs}
	\end{subfigure}
\caption{Qualitative comparison of methods on DUT-TE dataset.}
\label{comparison on DUT}
\end{figure*}

\begin{table*}[t!]
\scriptsize
\centering
\caption{
\centering
Ablation study. PDNet is a simplified DFFNet. `-' in the loss stage 2 means one stage of training without distillation. bce represents using BCE loss, and bce\&el represents using BCE and DOF-edge loss.}
\vspace{-0.5em}
\begin{tabular}{|c|c|c|ccc|ccc|ccc|ccc|}\hline
\multirow{2}{*}{Model}  & \multirow{2}{*}{\makecell[c]{stage1\\loss}} & \multirow{2}{*}{\makecell[c]{stage2\\loss}} & \multicolumn{3}{c}{CUHK-TE-1}    \vline                    & \multicolumn{3}{c}{DUT-TE}      \vline                    & \multicolumn{3}{c}{CTCUG}  \vline                     & \multicolumn{3}{c}{EBD}  \vline                        \\
                                                     &                              &                              & MAE$\downarrow$            & $F_\beta\uparrow$              & IoU$\uparrow$            & MAE$\downarrow$            & $F_\beta\uparrow$              & IoU$\uparrow$            & MAE$\downarrow$            & $F_\beta\uparrow$             & IoU$\uparrow$            & MAE$\downarrow$            & $F_\beta\uparrow$             & IoU$\uparrow$            \\\hline
\multirow{2}{*}{\makecell[c]{PDNet}} & bce\&el& --&0.041   & 0.968  & 0.944  & 0.076  & 0.934  & 0.898 & 0.093   & 0.859  & 0.851  & 0.097  & 0.822  & 0.864 \\
& bce\&el                     & bce\&el   & 0.039   & 0.969  & 0.946 & 0.074  & 0.934  &0.900 & 0.089   & 0.870  & 0.857  & 0.102  & 0.822  & 0.859\\\hline
\multirow{5}{*}{DFFNet}                    & bce                          &   --                           & 0.045          & 0.964          & 0.941          & 0.073          & 0.937          & 0.904          & 0.087          & 0.872          & 0.864          & 0.099          & 0.821          & 0.863          \\
                                       & bce                          & bce                          & 0.042          & 0.968          & 0.945          & 0.073          & 0.938          & 0.904          & 0.077          & 0.886          & 0.877          & 0.094          & 0.823          & 0.869          \\
                                    & bce                          & bce\&el                     & 0.040          & 0.969          & 0.945          & 0.070 & \textbf{0.941} & 0.905          & \textbf{0.073} & \textbf{0.898} & \textbf{0.880} & 0.089          & 0.825          & 0.875          \\
                                       & bce\&el                     &     --                         & 0.039          & 0.971          & 0.947          & 0.072          & 0.938          & 0.903          & 0.082          & 0.879          & 0.868          & 0.091          & 0.823          & 0.872          \\
                            & bce\&el                     & bce\&el                     & \textbf{0.036} & \textbf{0.973} & \textbf{0.951} & \textbf{0.070} & 0.939          & \textbf{0.906} & 0.074          & 0.892          & 0.878          & \textbf{0.084} & \textbf{0.826} & \textbf{0.882}\\\hline
\end{tabular}
\label{main ablation study}
\end{table*}

\begin{table*}[t!]
\scriptsize
\centering
\caption{
Backbone analysis in D-DFFNet.}
\vspace{-0.5em}
\begin{tabular}{|c|ccc|ccc|ccc|ccc|ccc|}\hline
\multirow{2}{*}{Backbone} & \multicolumn{3}{c}{CUHK-TE-1}    \vline                    & \multicolumn{3}{c}{DUT-TE}     \vline                     & \multicolumn{3}{c}{CTCUG}   \vline                     & \multicolumn{3}{c}{EBD}  \vline                        & \multicolumn{3}{c}{EBD(1305)}      \vline              \\
                           & MAE$\downarrow$            & $F_\beta\uparrow$              & IoU$\uparrow$            & MAE$\downarrow$            & $F_\beta\uparrow$              & IoU$\uparrow$            & MAE$\downarrow$            & $F_\beta\uparrow$              & IoU$\uparrow$            & MAE$\downarrow$            & $F_\beta\uparrow$              & IoU$\uparrow$            & MAE$\downarrow$            & $F_\beta\uparrow$              & IoU$\uparrow$            \\ \hline
Vgg16                      & 0.045          & 0.963          & 0.939          & 0.130          & 0.882          & 0.838          & 0.099          & 0.858          & 0.844          & \textbf{0.076} & 0.813          & \textbf{0.892} & 0.049          & 0.947          & 0.938          \\
ResNeSt101                 & \textbf{0.036} & \textbf{0.973} & \textbf{0.951} & \textbf{0.070} & \textbf{0.939} & \textbf{0.906} & \textbf{0.074} & \textbf{0.892} & \textbf{0.878} & 0.084          & \textbf{0.826} & 0.882          & \textbf{0.038} & \textbf{0.963} & \textbf{0.952} \\\hline
\end{tabular}
\label{backbone}
\vspace{-1em}
\end{table*}

\begin{figure}[t!]
	\begin{subfigure}{0.23\linewidth}
		\centering
		\includegraphics[width=0.9\linewidth]{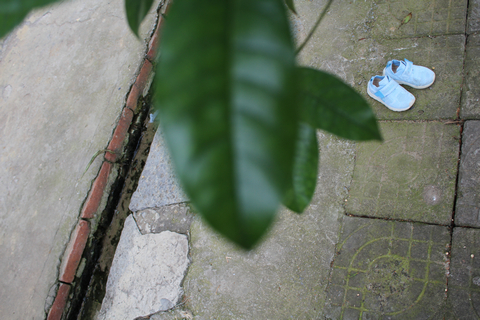}
	\end{subfigure}
	\centering
	\begin{subfigure}{0.23\linewidth}
		\centering
		\includegraphics[width=0.9\linewidth]{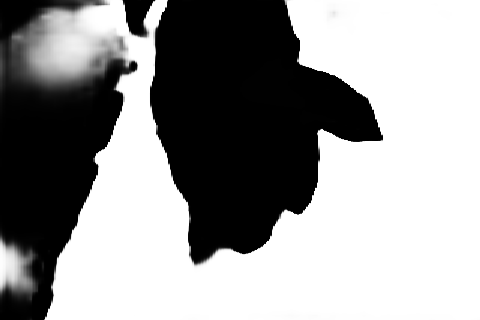}
	\end{subfigure}
	\begin{subfigure}{0.23\linewidth}
		\centering
		\includegraphics[width=0.9\linewidth]{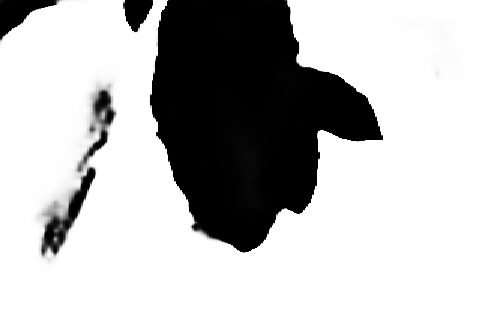}
	\end{subfigure}
	\begin{subfigure}{0.23\linewidth}
		\centering
		\includegraphics[width=0.9\linewidth]{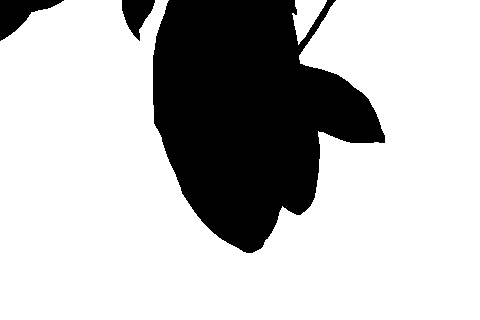}
	\end{subfigure}

	\begin{subfigure}{0.23\linewidth}
	\captionsetup{font={footnotesize}}
		\centering
		\includegraphics[width=0.9\linewidth]{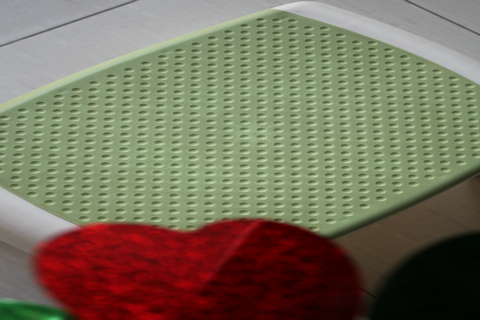}
		\caption{Images}
	\end{subfigure}
	\centering
	\begin{subfigure}{0.23\linewidth}
	\captionsetup{font={footnotesize}}
		\centering
		\includegraphics[width=0.9\linewidth]{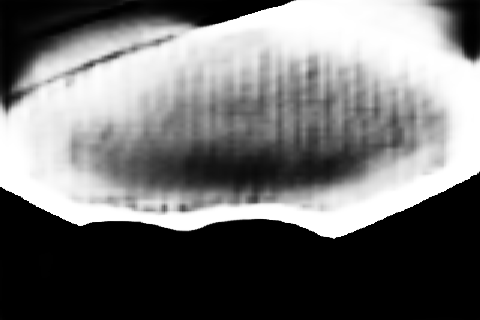}
		\caption{DFFNet}
	\end{subfigure}
	\begin{subfigure}{0.23\linewidth}
	\captionsetup{font={footnotesize}}
		\centering
		\includegraphics[width=0.9\linewidth]{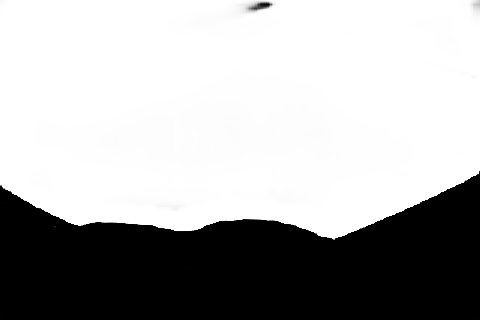}
		\caption{D-DFFNet}
	\end{subfigure}
	\begin{subfigure}{0.23\linewidth}
	\captionsetup{font={footnotesize}}
		\centering
		\includegraphics[width=0.9\linewidth]{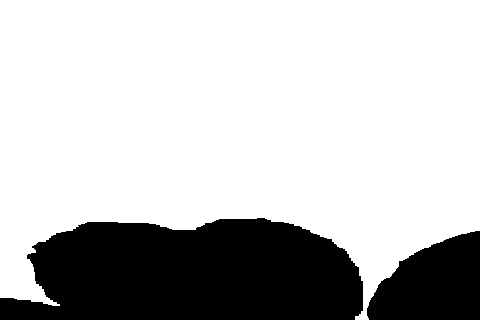}
		\caption{GTs}
	\end{subfigure}
\caption{Ablation study of depth feature distillation.}
\label{Ablation of depth feature distillation}
\end{figure}

\section{Experiments}
\subsection{Experimental setup}
\noindent
\textbf{Training dataset.} There are three different training splits in previous CNN-based methods: CUHK-TR-1~\cite{zhao2018defocus},  CUHK-TR-1 \& DUT-TR~\cite{zhao2021defocus}, and CUHK-TR-2 \& DUT-TR~\cite{jiang2022ma}.
\par
\begin{figure}[t]
	\begin{subfigure}{0.22\linewidth}
		\centering
		\includegraphics[width=\linewidth]{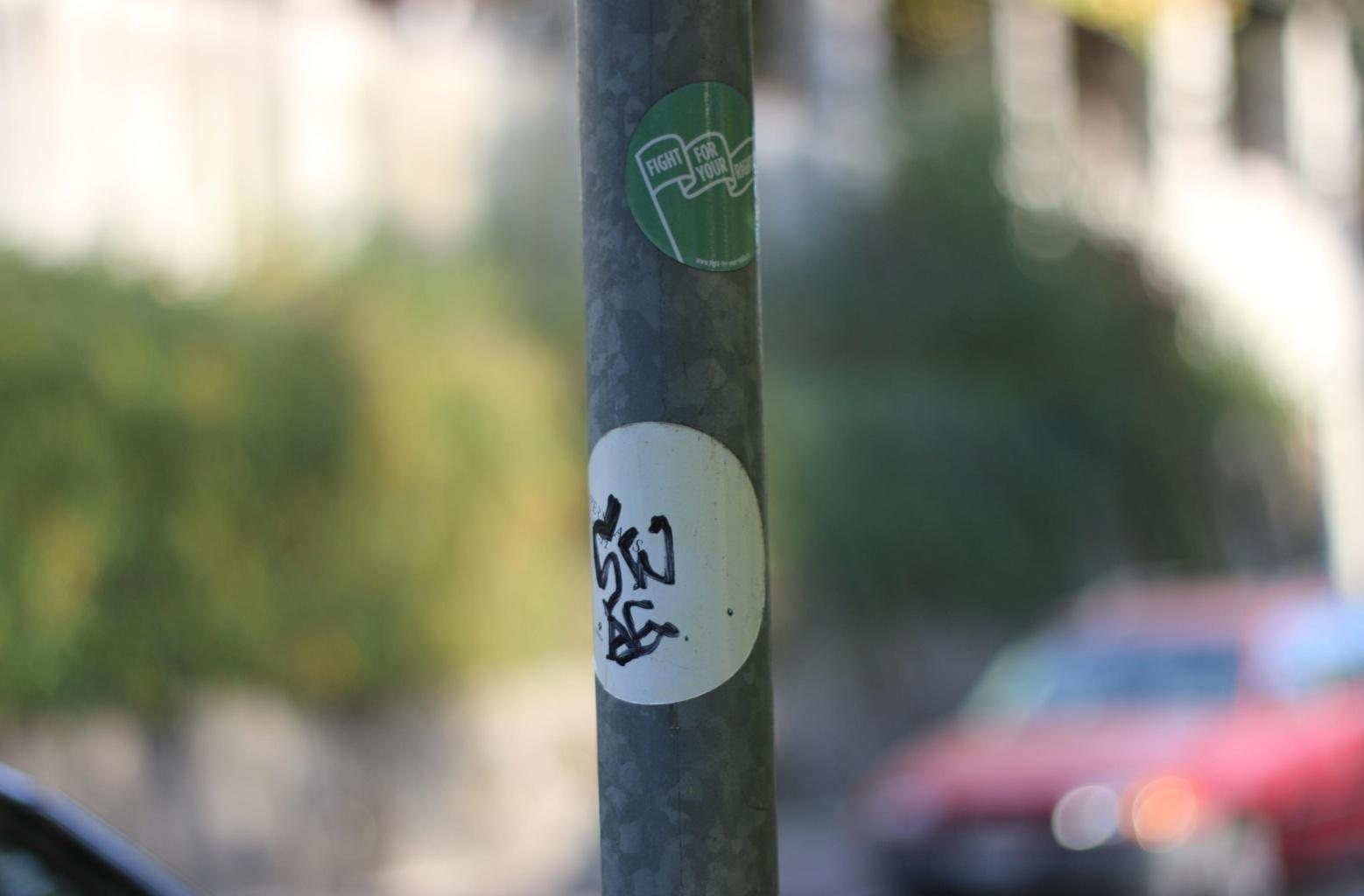}
	\end{subfigure}
	\centering
	\begin{subfigure}{0.22\linewidth}
		\centering
		\includegraphics[width=\linewidth]{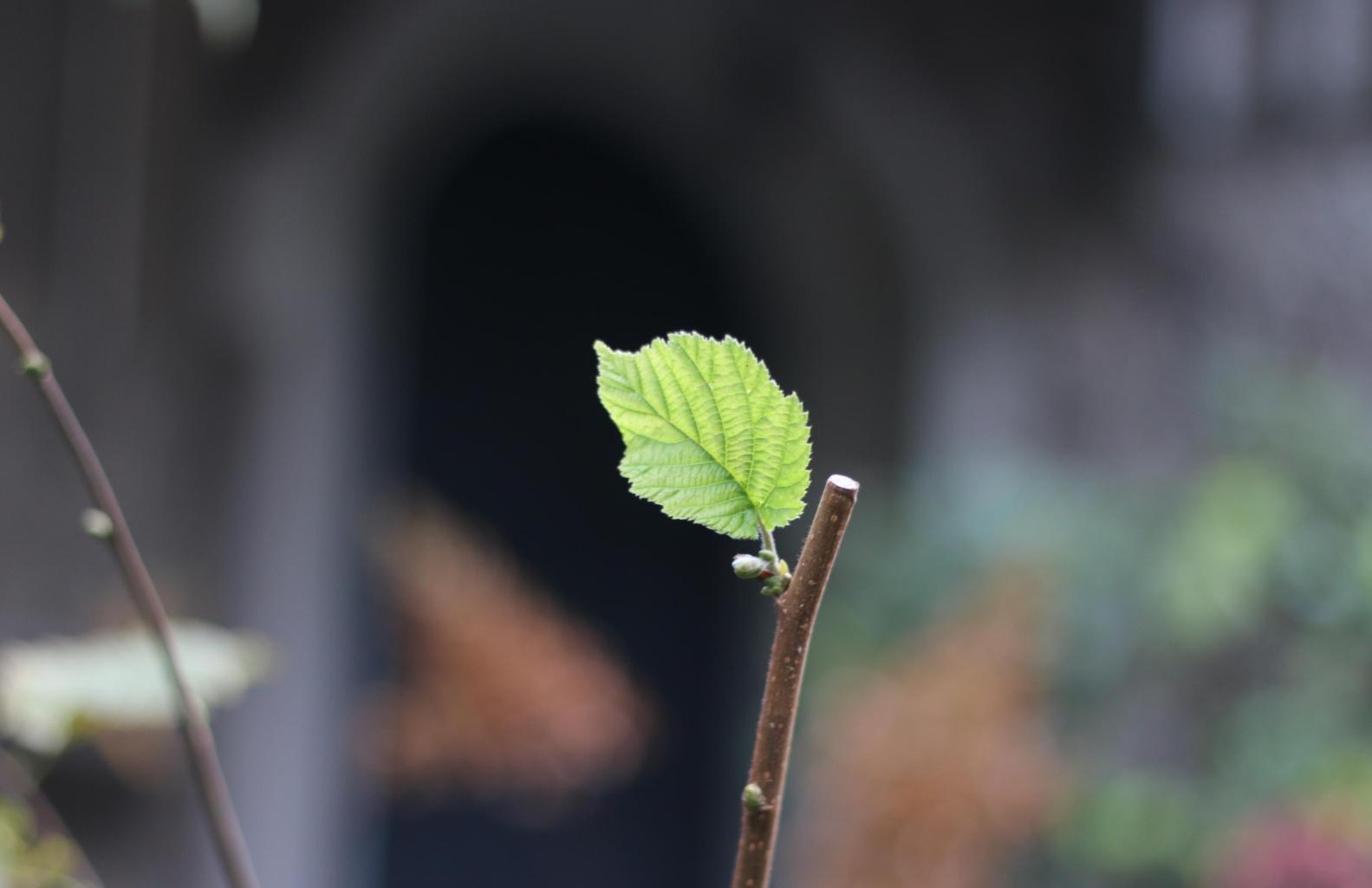}
	\end{subfigure}
	\centering
	\begin{subfigure}{0.22\linewidth}
		\centering
		\includegraphics[width=\linewidth]{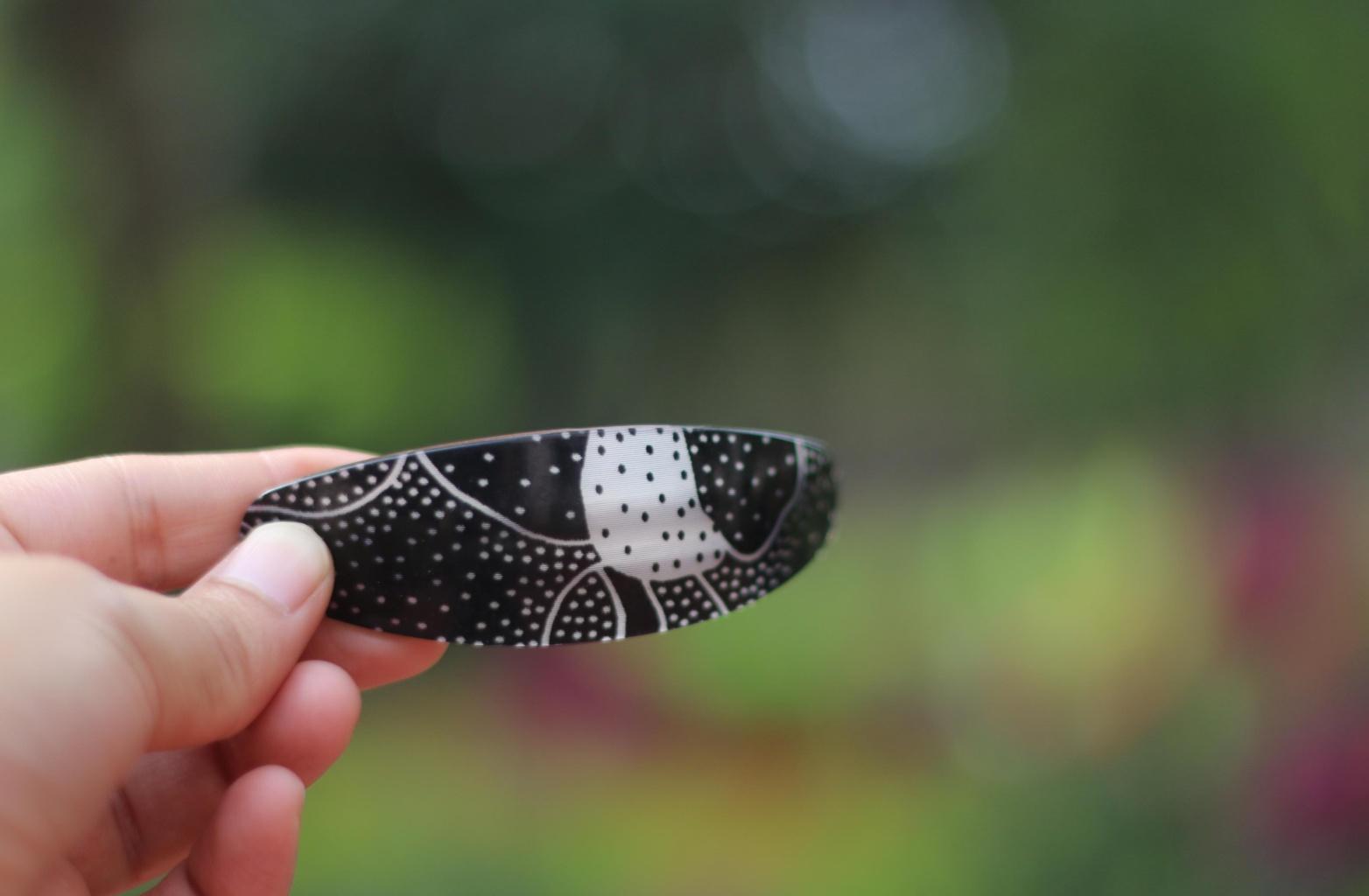}
	\end{subfigure}
	\begin{subfigure}{0.22\linewidth}
		\centering
		\includegraphics[width=\linewidth]{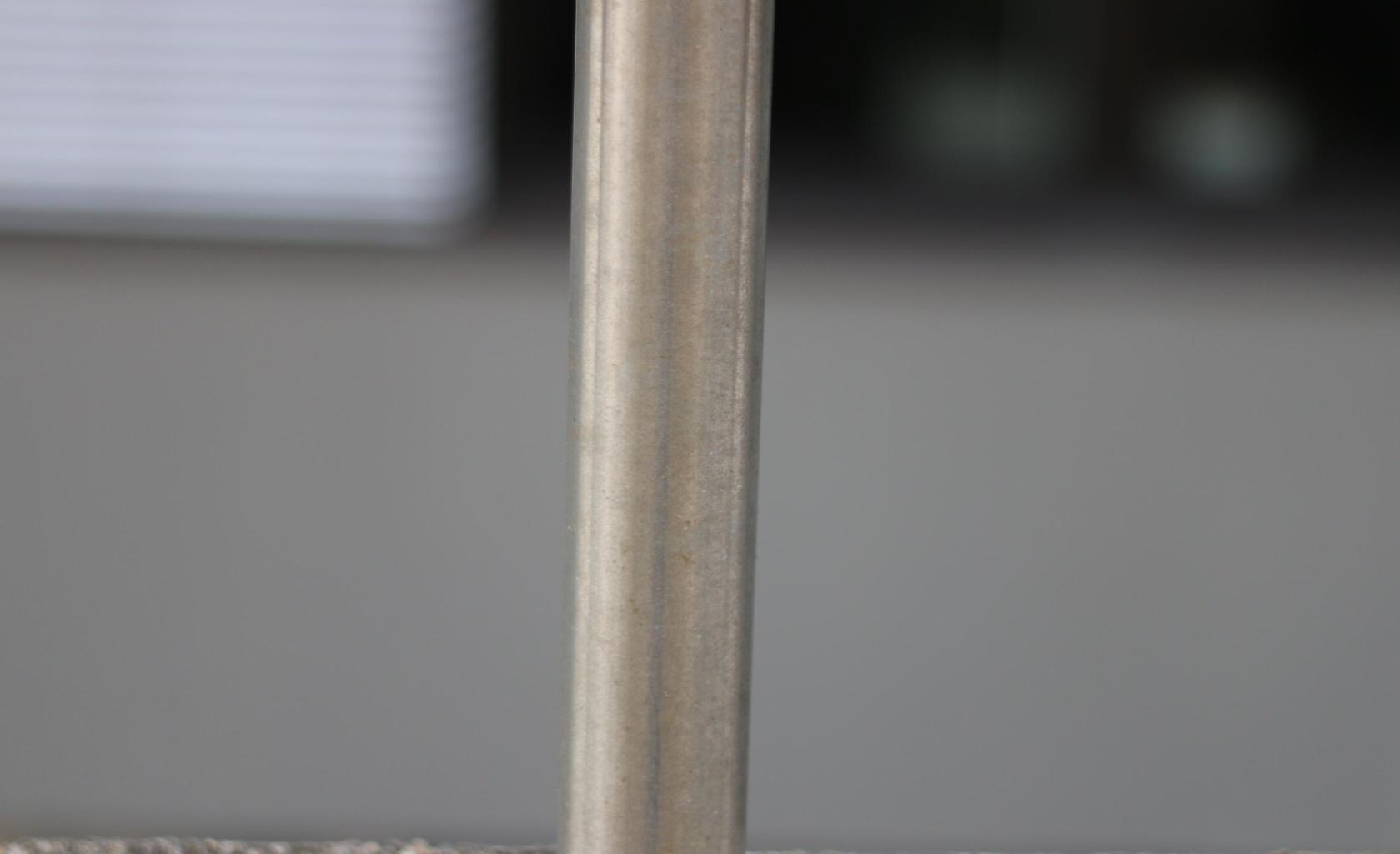}
	\end{subfigure}

	\begin{subfigure}{0.22\linewidth}
		\centering
		\includegraphics[width=\linewidth]{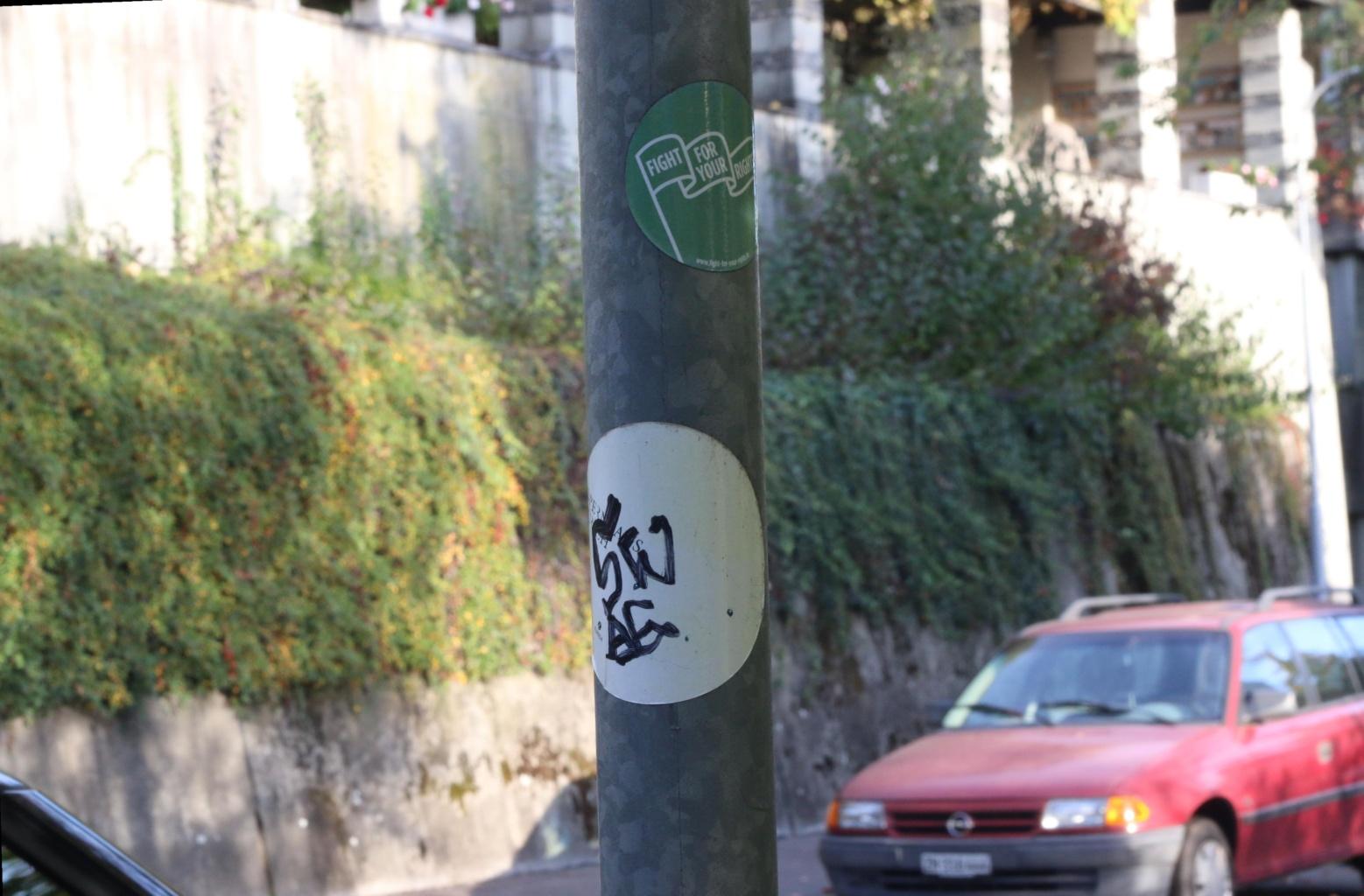}
	\end{subfigure}
	\begin{subfigure}{0.22\linewidth}
		\centering
		\includegraphics[width=\linewidth]{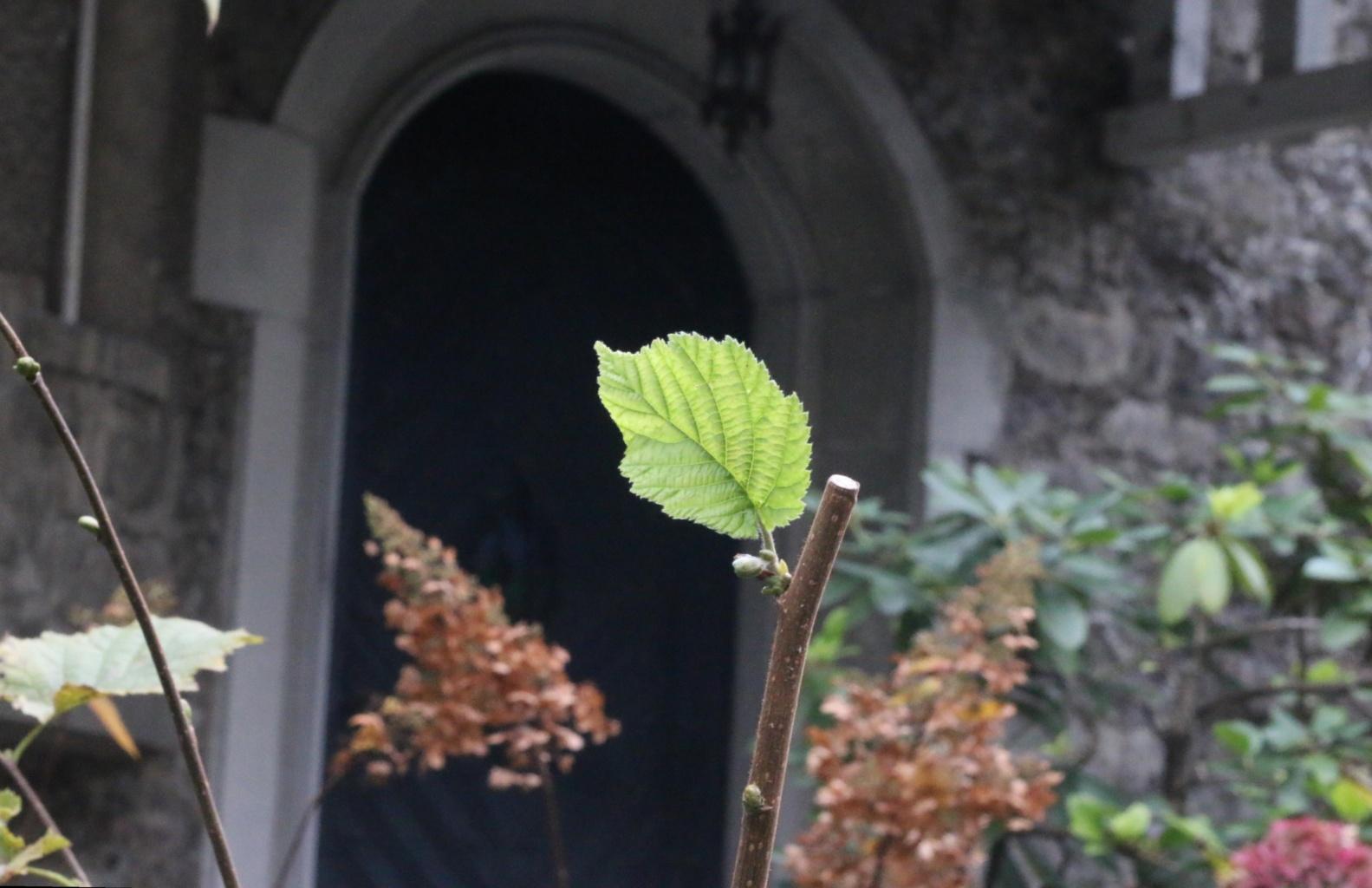}
	\end{subfigure}
	\begin{subfigure}{0.22\linewidth}
		\centering
		\includegraphics[width=\linewidth]{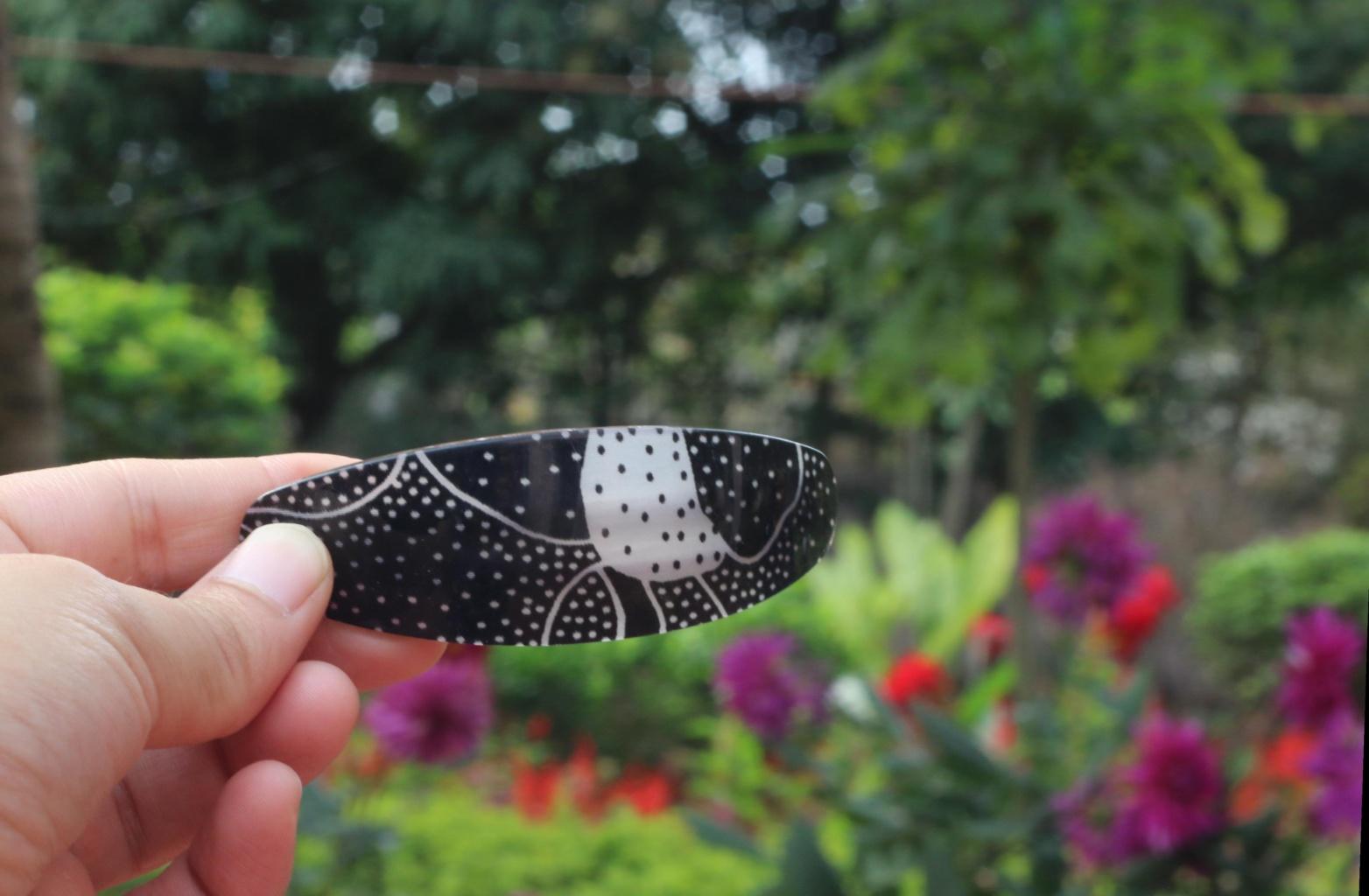}
	\end{subfigure}
	\begin{subfigure}{0.22\linewidth}
		\centering
		\includegraphics[width=\linewidth]{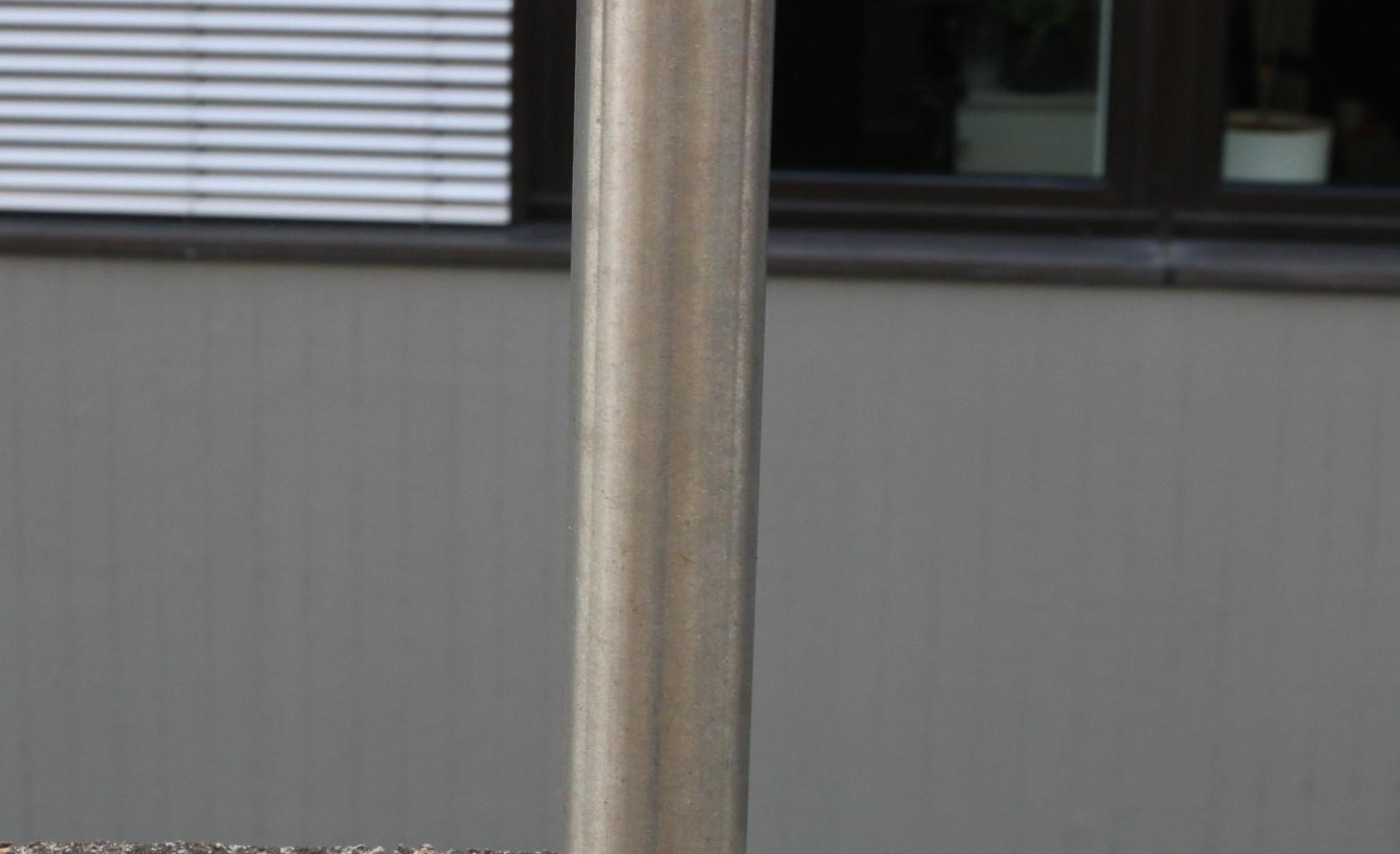}
	\end{subfigure}
	
\caption{Samples from EBD dataset. The first row shows images with shallow DOF. The second row shows images with wide DOF. }
\label{EBD}
\end{figure}

 The CUHK~\cite{shi2014discriminative} dataset consists of 704 images, with 604 used for training (`-TR') and the remaining 100 for testing (`-TE'). The `-1' and `-2' designations indicate two different splits in BTBNet~\cite{zhao2018defocus} and MA-GANet~\cite{jiang2022ma}, respectively. The DUT~\cite{zhao2018defocus} dataset, consists of 600 images for training (`-TR') and 500 images for testing (`-TE'). `\&' denotes training the datasets together.
To provide a fair comparison with previous methods, we trained our model separately on all three splits. \\
\textbf{Testing dataset.} 
Three frequently used benchmark datasets are utilized to evaluate our method: CUHK-TE (including CUHK-TE-1 and CUHK-TE-2), DUT-TE, and CTCUG~\cite{tang2020defusionnet}. While CTCUG is a small dataset with only 150 images, it is known to be particularly challenging. It should be noted that all three benchmarks have limited quantities of images, with fewer than 500 each.\\
\textbf{EBD dataset.}
The test datasets mentioned above have several limitations, including a lack of high-resolution images and a limited number of images with wide or shallow depth of field. In response, we collected a new DBD test dataset (EBD) composed of 1605 high-resolution images, selected from the EBB!~\cite{ignatov2020rendering} dataset and manually annotated with pixel-wise labels to produce defocus maps. As for the images in the EBD dataset, 1305 have a shallow depth of field (achieved using an aperture size of f/1.8), resulting in a strong bokeh effect. The remaining 300 images have a wide depth of field (achieved using an aperture size of f/16), resulting in sharp photos. All images have a resolution around $1600\times 1024$. Some samples from the EBD dataset are shown in Fig.~\ref{EBD}. This more challenging benchmark may assist future researchers in more robustly exploring the DBD task.\\
\textbf{Implementation Details.}  
In the training phase (both stage 1 and stage 2), the images are resized to 320$\times$320, with a vertical flip in a probability of 50$\%$, and random color jittering.
The batch size is 6, and the maximum epoch is 75. The defocus model is optimized using Adam optimizer, with an initial learning rate of 1e-4, and a `poly' policy with a power of 0.9. 
Supplementary material contains more detailed information.\\
\textbf{Evaluation metrics.} We use four commonly-used evaluation metrics: mean absolute error (MAE), $F_\beta$ value, intersect over union (IoU), and precision-recall (PR) curve. The $F_\beta$ value is computed using the same parameter settings as BTBNet~\cite{zhao2018defocus}, and we use a binarized DBD map (with blur regions as positive) and a threshold of 0.5 to compute the $F_\beta$ value and IoU. The precision-recall curves are included in the supplemental material.

\subsection{Comparison With the State-of-the-Arts}
\noindent
We compare our method with 14 recent methods,
including DBDF~\cite{shi2014discriminative}, LBP~\cite{yi2016lbp}, HiFST~\cite{alireza2017spatially}, BTBNet~\cite{zhao2018defocus} and its later version BTBNet2~\cite{zhao2019defocus}, CENet~\cite{zhao2019enhancing}, BR2Net~\cite{tang2020br}, AENet~\cite{zhao2021defocus}, EFENet~\cite{zhao2021defocus}, DD~\cite{cun2020defocus}, DefusionNet2~\cite{tang2020defusionnet}, IS2CNet~\cite{zhao2021image}, LOCAL~\cite{li2021layer}, and MA-GANet~\cite{jiang2022ma}. 
We retrained the BR2Net model and used data from the papers for LOCAL and MA-GANet, as they did not provide codes and results. For the other methods, we downloaded their results and tested them using our metrics.\\
\textbf{Quantitative comparison of methods.}  
We provide numerical comparisons against state-of-the-art methods on four datasets using three metrics in Table~\ref{comparison with methods training on CUHK-TR-1}, Table~\ref{comparison with methods training on CUHK-TR-1 & DUT-TR}, and Table~\ref{comparison with methods training on CUHK-TR-2 & DUT-TR}. Our D-DFFNet outperforms existing DBD methods. Notably, we achieve excellent performance on the CTCUG dataset, which is a highly challenging dataset and has proven difficult for previous works, as they have mentioned~\cite{jiang2022ma}. For more detailed quantitative results and precision-recall curves, we refer the readers to the supplemental material.\\
\textbf{Qualitative comparison of methods.}
In Fig.~\ref{comparison on DUT}, we provide a qualitative comparison of our proposed method with all competing methods. We would like to draw attention to the first image's foreground grass and the second image's top part of the wood substance: both are homogeneous and in-focus regions, but are considered defocus blur regions by most previous methods. Our method, which incorporates depth feature distillation, shows superior prediction to these regions. Additional benchmarks and their respective results can be found in the supplemental material.

\subsection{Ablation study}
\label{Ablation study}
\noindent
\textbf{Effectiveness of depth feature distillation.}  
Table~\ref{main ablation study} shows the effectiveness of the depth feature distillation strategy on four test datasets. We also present visual comparisons in Fig.~\ref{Ablation of depth feature distillation} and Fig.~\ref{Homogenous regions}. In Fig.~\ref{Ablation of depth feature distillation}, the cement ground on the left (the first image) and the white region (the second image) are homogeneous but in-focus regions. With depth feature distillation (Fig.~\ref{Ablation of depth feature distillation}(c)), DFFNet can accurately predict these regions to be in-focus, while without distillation (Fig.~\ref{Ablation of depth feature distillation}(b)), it mistakenly consider them to defocus blur regions. We also present a comparison of depth feature distillation (D-DFFNet) and response-based depth distillation (R-DFFNet) on Table~\ref{comparison with methods training on CUHK-TR-1} for a fair comparison with DD~\cite{cun2020defocus}, which shows that depth feature distillation is a better way to introduce depth representation. \\
\textbf{Effectiveness of DOF-edge loss.}  From Table~\ref{main ablation study}, adding DOF-edge loss in the second stage when applying depth feature distillation can achieve better results than using only BCE loss, this demonstrates that with both depth representation and DOF-edge loss, we can get a more robust defocus blur detector. Besides, adding DOF-edge loss when training the model in the first stage can still help the results, as it helps recognize DOF regions.\\
\textbf{Generalization of depth feature distillation.}
To test the generalizability of our depth feature distillation on similar architectures, we simplified the decoder of our model, keeping only the Agg module (from Fig.~\ref{network}) for basic feature fusion, and we named the resulting model PDNet. Table~\ref{main ablation study} shows that even the simple PDNet, with depth feature distillation, can achieve better scores on most benchmarks. This demonstrates that the depth feature distillation can be generalized to any method with a similar architecture.\\

\section{Future work}
\noindent
From Table~\ref{comparison with methods training on CUHK-TR-1} and Table~\ref{comparison with methods training on CUHK-TR-1 & DUT-TR}, we can see that our method (D-DFFNet) performs best on all shallow-DOF images of the EBD dataset (EBB(1305)), but does not perform as well on the complete EBD dataset with both wide- and shallow-DOF images. We have observed that the choice of the backbone can influence the generalization ability to wide-DOF images, as shown in Table~\ref{backbone}. The Vgg16~\cite{simonyan2014very} backbone does not perform well on most datasets, but it shows great generalizability to wide-DOF images, even though such images are rarely seen in the training data. This observation inspires us to find detectors that not only perform well on in-distribution data but also have great generalization ability.

\section{Conclusion}
\noindent
In this paper, we explore how to improve defocus blur detection with a single image. Specifically, we propose a depth feature distillation strategy to help the DBD model learn depth information at the feature level and a dice-based DOF-edge loss as an auxiliary loss function to incorporate DOF information. By combining depth and DOF cues, defocus blur detector could recognize defocus phenomenon in a natural way. Additionally, we collect a new test dataset (EBD) consisting of 1605 image-label pairs, including images with shallow and wide depth of field to avail evaluation. Our extensive experimental results show that our method outperforms state-of-the-art approaches on all benchmarks.

\section*{Acknowledgement}
\noindent
This work was supported in part by the Fundamental Research Funds for the Central Universities 2042022kf1193, and
National Nature Science Foundation of China U22B2011, 61871299. 

\bibliographystyle{IEEEtran}
\bibliography{icme2023template}

\clearpage
\begin{appendices}
\begin{table*}[!h] 
\scriptsize
\centering
\caption{
\centering
Ablation study. `--' in the loss stage 2 means one stage training without distillation. bce represents using BCE loss, and bce\&el represents using BCE and DOF-edge loss.}
\begin{tabular}{|c|c|c|ccc|ccc|ccc|ccc|}\hline
\multirow{2}{*}{Model}  & \multirow{2}{*}{\makecell[c]{stage1\\loss}} & \multirow{2}{*}{\makecell[c]{stage2\\loss}} & \multicolumn{3}{c}{CUHK-TE-1}    \vline                    & \multicolumn{3}{c}{DUT-TE}      \vline                    & \multicolumn{3}{c}{CTCUG}  \vline                     & \multicolumn{3}{c}{EBD}  \vline                        \\
&                              &                              & MAE$\downarrow$            & $F_\beta\uparrow$              & IoU$\uparrow$            & MAE$\downarrow$            & $F_\beta\uparrow$              & IoU$\uparrow$            & MAE$\downarrow$            & $F_\beta\uparrow$             & IoU$\uparrow$            & MAE$\downarrow$            & $F_\beta\uparrow$             & IoU$\uparrow$            \\\hline
\multirow{2}{*}{\makecell[c]{DFFNet \\w/o DFFM}}                & bce\&el                     &     --                         & 0.040          & 0.969          & 0.946          & 0.083          & 0.925          & 0.890          & 0.090          & 0.872          & 0.855          & 0.091          & 0.820          & 0.873          \\
                                      & bce\&el                     & bce\&el                     & 0.038          & 0.970          & 0.948          & 0.081          & 0.927          & 0.892          & 0.082          & 0.884          & 0.867          & 0.090          & 0.821          & 0.873          \\\hline
\multirow{2}{*}{DFFNet}                   
& bce\&el                     &     --                         & 0.039          & 0.971          & 0.947          & 0.072          & 0.938          & 0.903          & 0.082          & 0.879          & 0.868          & 0.091          & 0.823          & 0.872          \\
& bce\&el                     & bce\&el                     & \textbf{0.036} & \textbf{0.973} & \textbf{0.951} & \textbf{0.070} & \textbf{0.939}          & \textbf{0.906} & \textbf{0.074}          & \textbf{0.892}          & \textbf{0.878}          & \textbf{0.084} & \textbf{0.826} & \textbf{0.882}\\\hline
\end{tabular}
\label{main ablation study}
\end{table*}
\section{Dataset Comparison}
\noindent
We present additional comparisons of our EBD dataset and other DBD test datasets in Table~\ref{datasets}.
From Table~\ref{datasets}, we can see that previous test datasets have limitations in quantity, with a total of only 750 images, while our EBD dataset contains 1605 images.
Additionally, we notice that previous datasets lacked high-resolution images, whereas all of our collected images are high-resolution (around 1600×1024).
Previous datasets only contained partial-focus images, which have both focused and defocused regions. However, in reality, people adjust camera parameters to capture images with different DOF, images with wide DOF can be all-in-focus, while images with shallow DOF have a strong bokeh effect.
Therefore, we selected images with both wide and shallow DOF, with aperture sizes of f/16 and f/1.8 respectively.
The last 300 images in our EBD dataset are clear images with a wide DOF, where most regions of the images are in focus. In contrast, the other 1305 images from EBD are bokeh images with a shallow DOF. It is worth noting that recent studies have highlighted limitations in using synthetic images compared to real images. Like the CUHK and CTCUG datasets, all images in our EBD dataset are real. However, in stark contrast, images in the DUT-TE dataset are synthetic images generated from all-in-focus images.
\begin{table*}[t!]
\small
\centering
\caption{
\centering
Comparison with existing DBD test datasets. Clear images are those with a wide DOF where most regions are in focus. Partial-focus images have defocused regions. Image type real means images captured by a camera. Image type
synthetic means images generated from all-in-focus images. `--' in Aperture sizes means we do not know their aperture sizes.}
\begin{tabular}{|c|cccccc|}\hline
Name    & Number & Resolution & Aperture sizes & Clear images & Partial-focus images & Image type\\\hline
CUHK-TE & 100    & 640×427    & f/5.0          & 0                   & 100     &real             \\
DUT-TE  & 500    & 320×320    & --               & 0                   & 500     &synthetic              \\
CTCUG   & 150    & 480×360    & --               & 0                   & 150     &real              \\
EBD     & 1605   & 1600×1024  & f/1.8\&f/16    & 300                 & 1305    &real          \\\hline   
\end{tabular}
\label{datasets}
\end{table*}
\section{Implementation Details}
\subsection{Dense feature fusing module}
\noindent
Handling multi-level features is essential for DBD models.
We present a dense feature fusing module (DFFM) to propagate useful information about the features at all higher-level stages over features at each stage and aggregate them level by level.
Since low-level features with high resolution usually contain noise due to a small receptive field, and high-level features after down-sampling, are much cleaner. Our proposed DFFM can help filter out the noise and emphasize valuable information about low-level features with the help of all higher-level features by dense fusion.

\subsection{More Implementation Details of D-DFFNet}
\noindent
We implement our method by Pytorch, and our model runs on one GPU of NVIDIA RTX 3090 with CUDA version 11.7.
In both two stages, the defocus model is optimized using Adam optimizer, with an initial learning rate of 1e-4, and a `poly' policy with power by 0.9. In stage 2, we use an extra Adam optimizer with a learning rate fixed to 1e-1 and with 5e-4 weight decay to optimize two projectors. Note that, our training process (including two stages) takes less than two hours.

\subsection{More Implementation Details of R-DFFNet}
\noindent
When comparing our depth feature distillation with response-based depth distillation, we implement R-DFFNet in the same way as DD~\cite{cun2020defocus}. We add extra classifiers to generate depth predictions in the last layer of our network and in all decoder branches (the same position with DBD side classifiers), and we use a pre-trained depth estimation model Midas~\cite{ranftl2020towards} to provide fake labels for depth supervision. Unlike our D-DFFNet, implementing R-DFFNet only need one-stage training. We optimize this model using DBD loss (the same as our work) and depth loss (we use $\ell_2$ loss):
\begin{equation}
\ell=
\ell_d^f+\sum_k \alpha_k \ell_d^k + \lambda (\ell_2^f +\sum_k \beta_k \ell_2^k)
\vspace{-0.5em}
\end{equation}
where $\ell_d^f$ represents $\ell_d$ loss for final defocus prediction, $\ell_d^k$ represents $\ell_d$ loss for k-level side defocus outputs, $\ell_2^f$ represents $\ell_2$ loss for final depth prediction, and $\ell_2^k$ represents $\ell_2$ loss for k-level side depth outputs. $\alpha_k$($k\in[1,..,4] $), $\beta_k$($k\in[1,..,4] $) and $\lambda$ are trade-off parameters, we set $\alpha_k$($k\in[1,..,4] $), $\beta_k$($k\in[1,..,4] $), and $\lambda$ to 1.
The optimizer, batch size, and training epoch are the same with D-DFFNet.

\section{Experiments}
\subsection{More Quantitative Comparison}
\begin{figure*}[t!]
	\begin{subfigure}{0.20\linewidth}
		\centering
		\includegraphics[width=\linewidth]{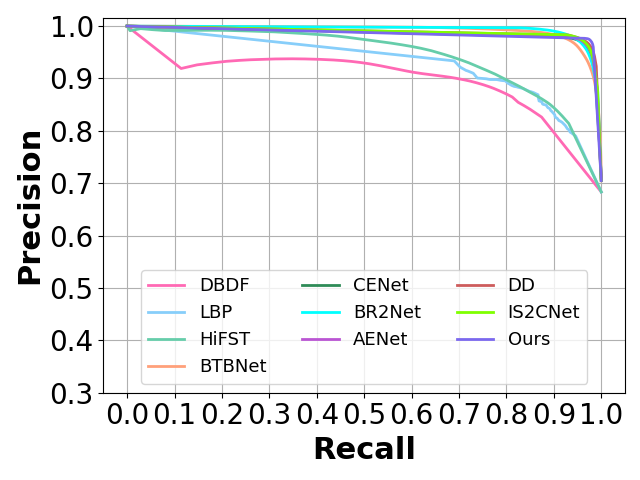}
	\end{subfigure}
	\centering
	\begin{subfigure}{0.20\linewidth}
		\centering
		\includegraphics[width=\linewidth]{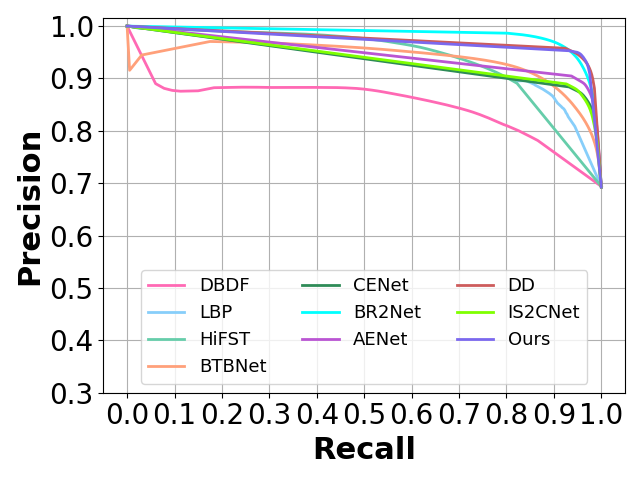}
	\end{subfigure}
	\centering
	\begin{subfigure}{0.20\linewidth}
		\centering
		\includegraphics[width=\linewidth]{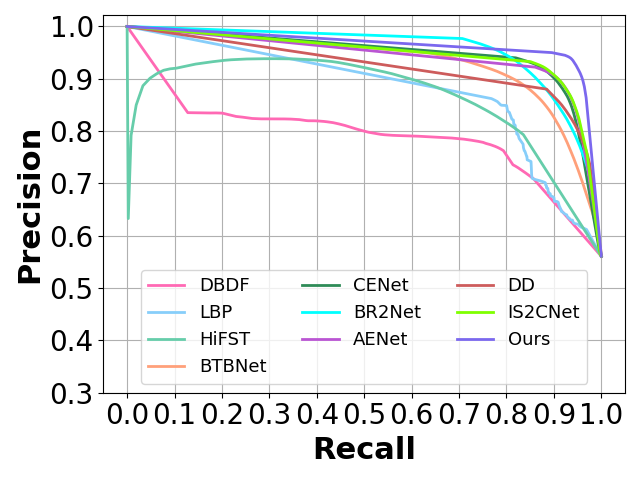}
	\end{subfigure}
	\centering
	\begin{subfigure}{0.20\linewidth}
		\centering
		\includegraphics[width=\linewidth]{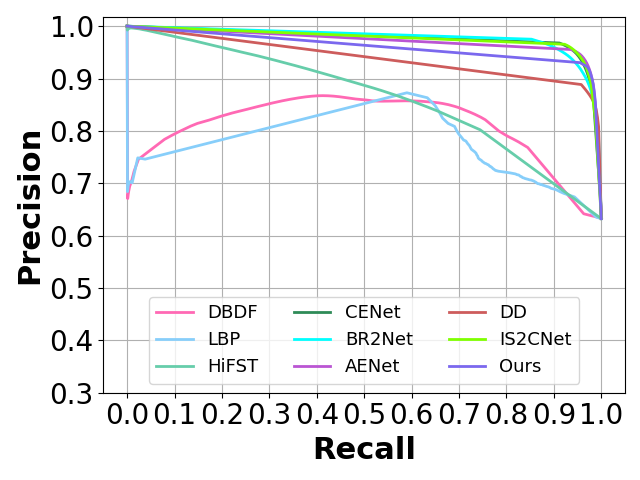}
	\end{subfigure}
	
	\centering
	\begin{subfigure}{0.20\linewidth}
		\centering
		\includegraphics[width=\linewidth]{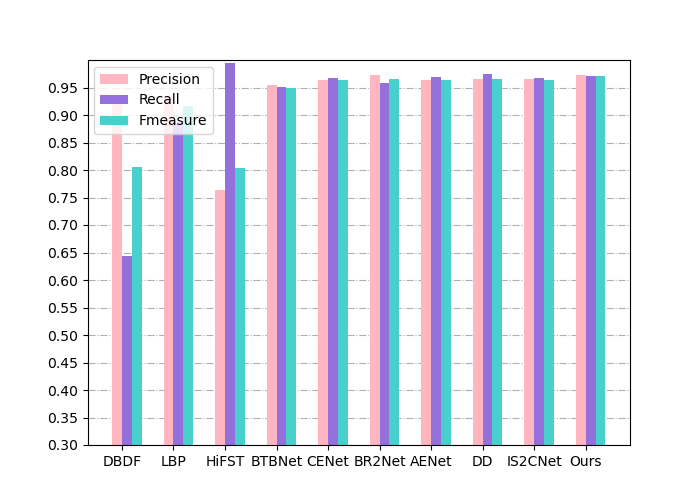}
    	\caption{CUHK-TE-1}
	\end{subfigure}
	\centering
	\begin{subfigure}{0.20\linewidth}
		\centering
		\includegraphics[width=\linewidth]{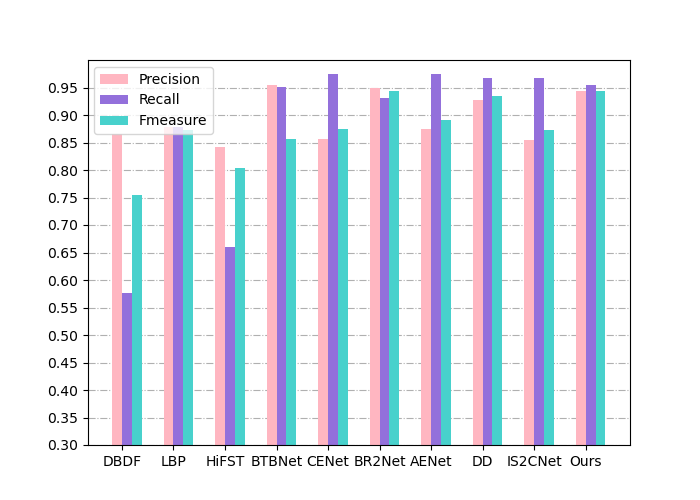}
		\caption{DUT-TE}
	\end{subfigure}
	\centering
	\begin{subfigure}{0.20\linewidth}
		\centering
		\includegraphics[width=\linewidth]{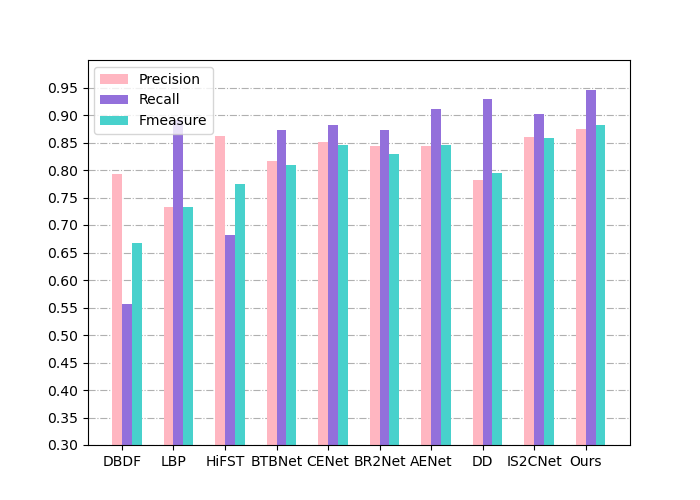}
		\caption{CTCUG}
	\end{subfigure}
	\begin{subfigure}{0.20\linewidth}
		\centering
		\includegraphics[width=\linewidth]{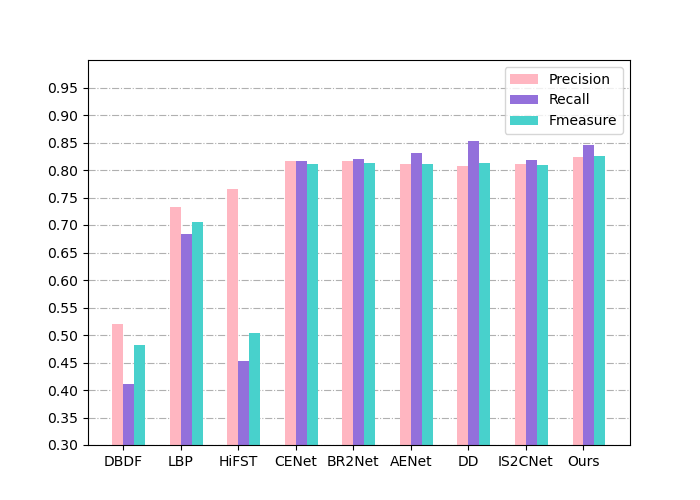}
		\caption{EBD}
	\end{subfigure}
	\centering
	\caption{Precision-Recall curves and comparison of Precision, Recall, and $F_\beta $ on CUHK-TE-1, DUT-TE, CTCUG, and EBD datasets. Ours represents our D-DFFNet using the training data CUHK-TR-1.}
	\label{PR curve}
\end{figure*}

\noindent
We present Precision-Recall curves on four test datasets in Fig.~\ref{PR curve}. The results demonstrate that our method outperforms the others on most of the test datasets.

\subsection{More Qualitative Comparison}
\begin{figure*}[t]
    \begin{subfigure}{0.12\linewidth}
		\centering
		\includegraphics[width=\linewidth]{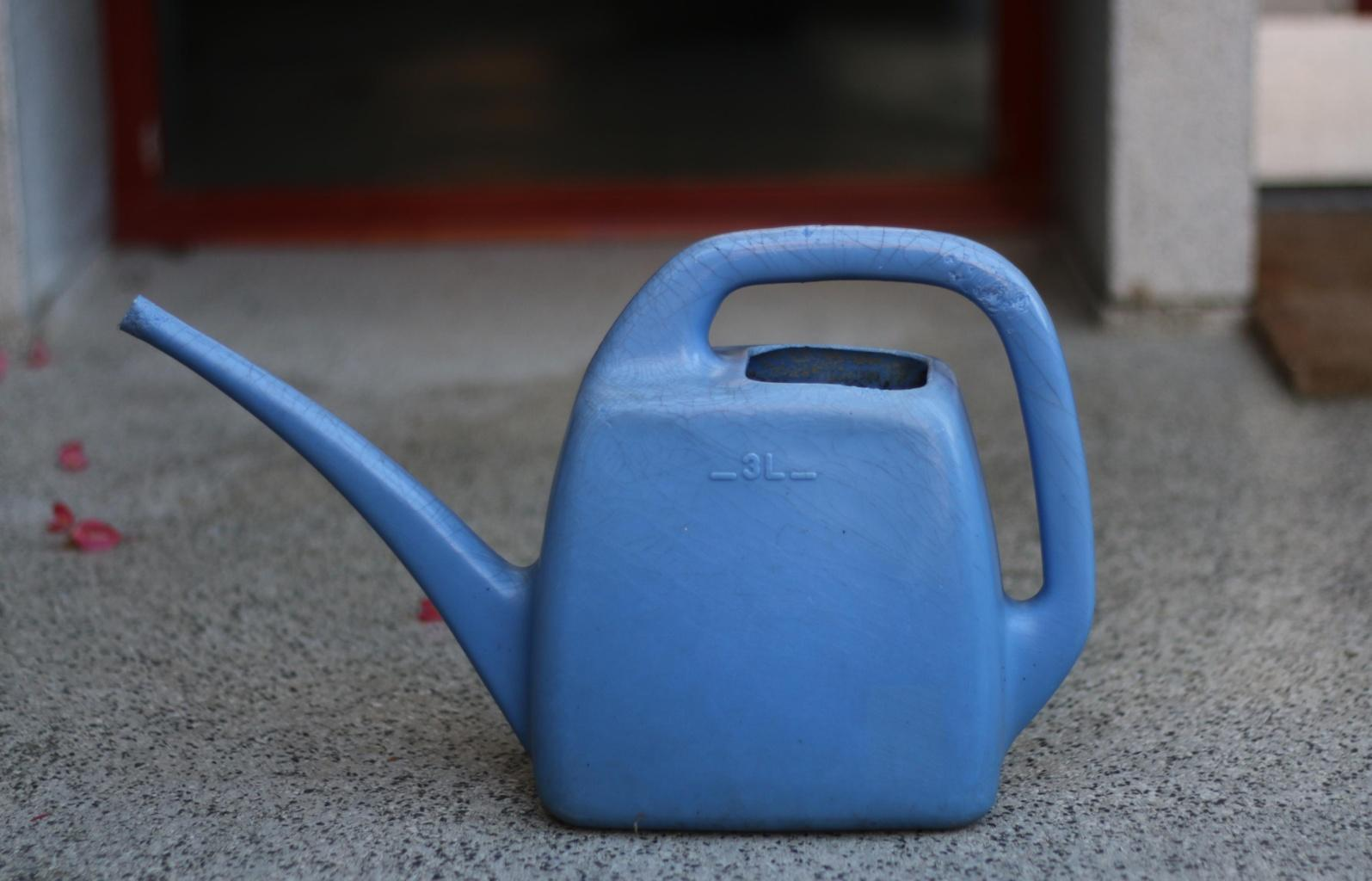}
	\end{subfigure}
    \begin{subfigure}{0.12\linewidth}
		\centering
		\includegraphics[width=\linewidth]{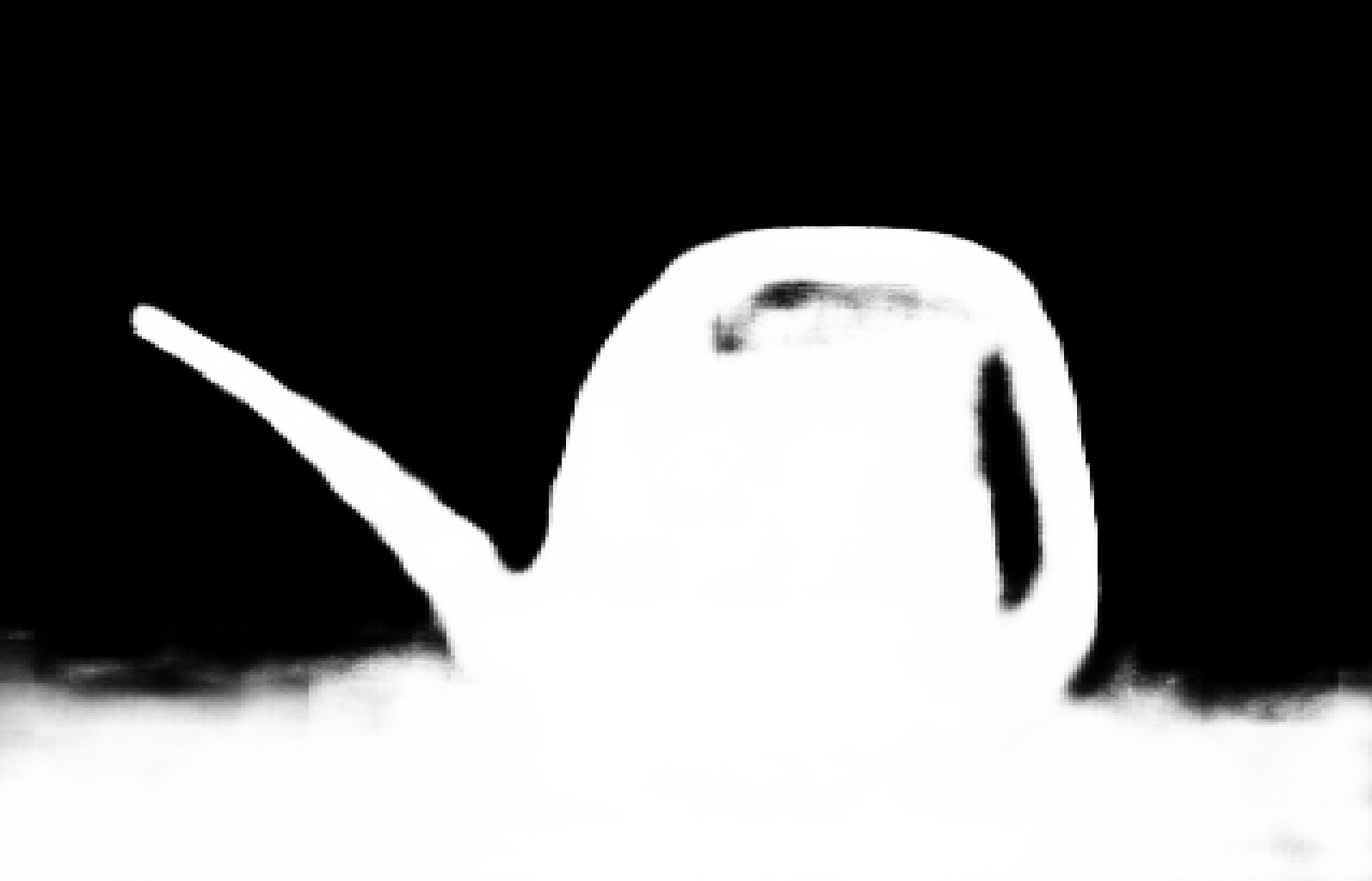}
	\end{subfigure}
    \begin{subfigure}{0.12\linewidth}
		\centering
		\includegraphics[width=\linewidth]{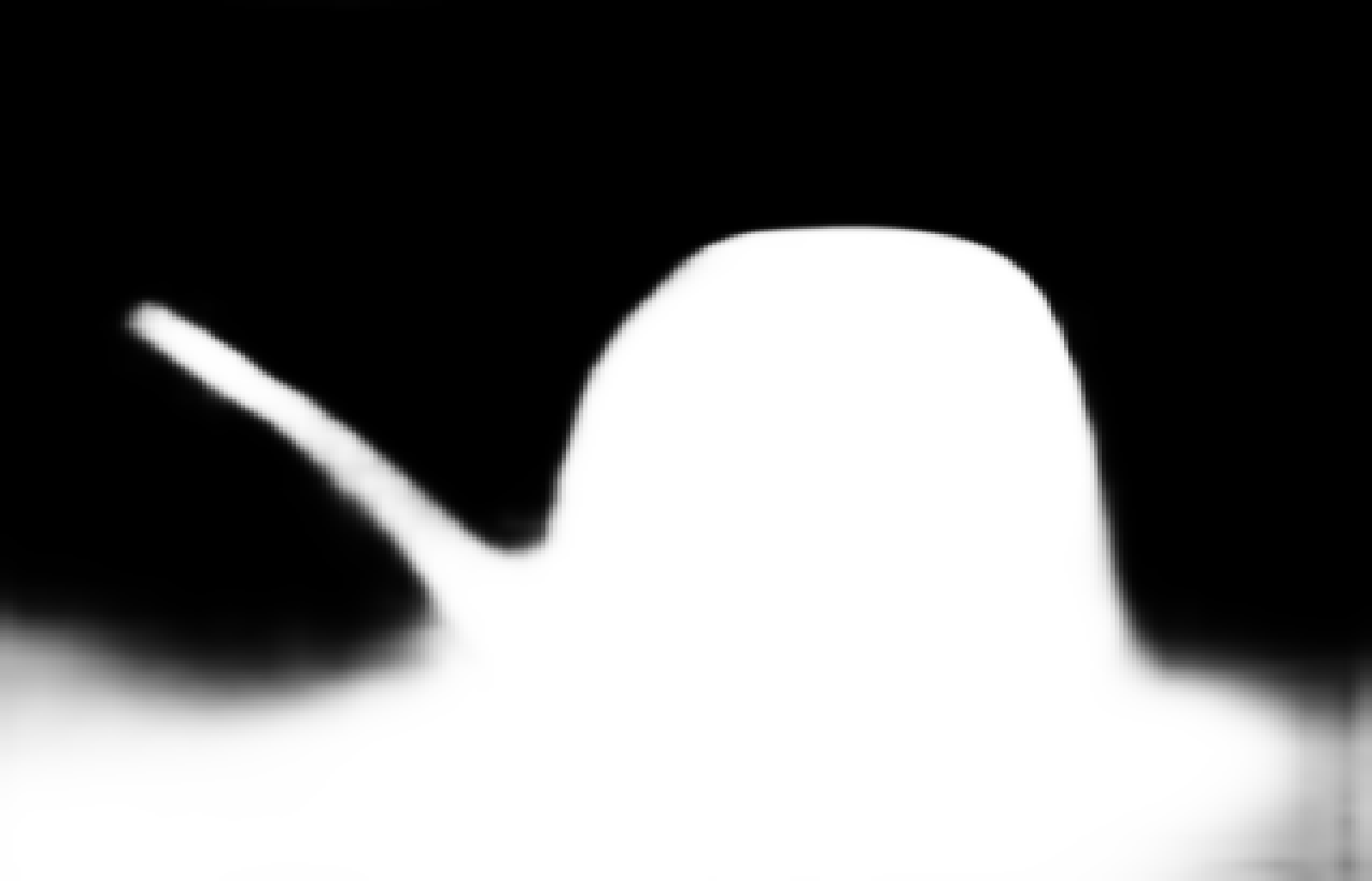}
	\end{subfigure}
    \begin{subfigure}{0.12\linewidth}
		\centering
		\includegraphics[width=\linewidth]{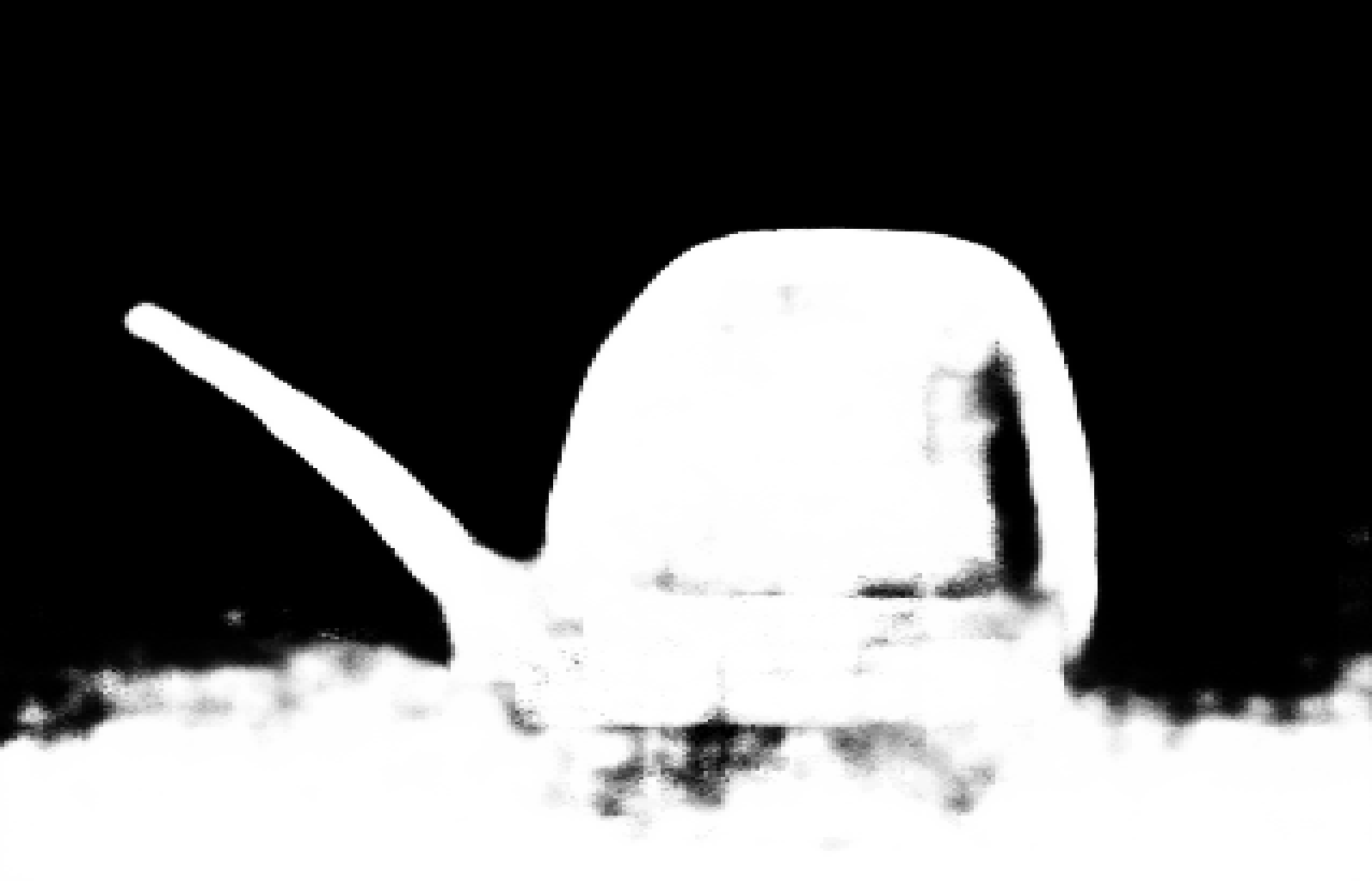}
	\end{subfigure}
    \begin{subfigure}{0.12\linewidth}
		\centering
		\includegraphics[width=\linewidth]{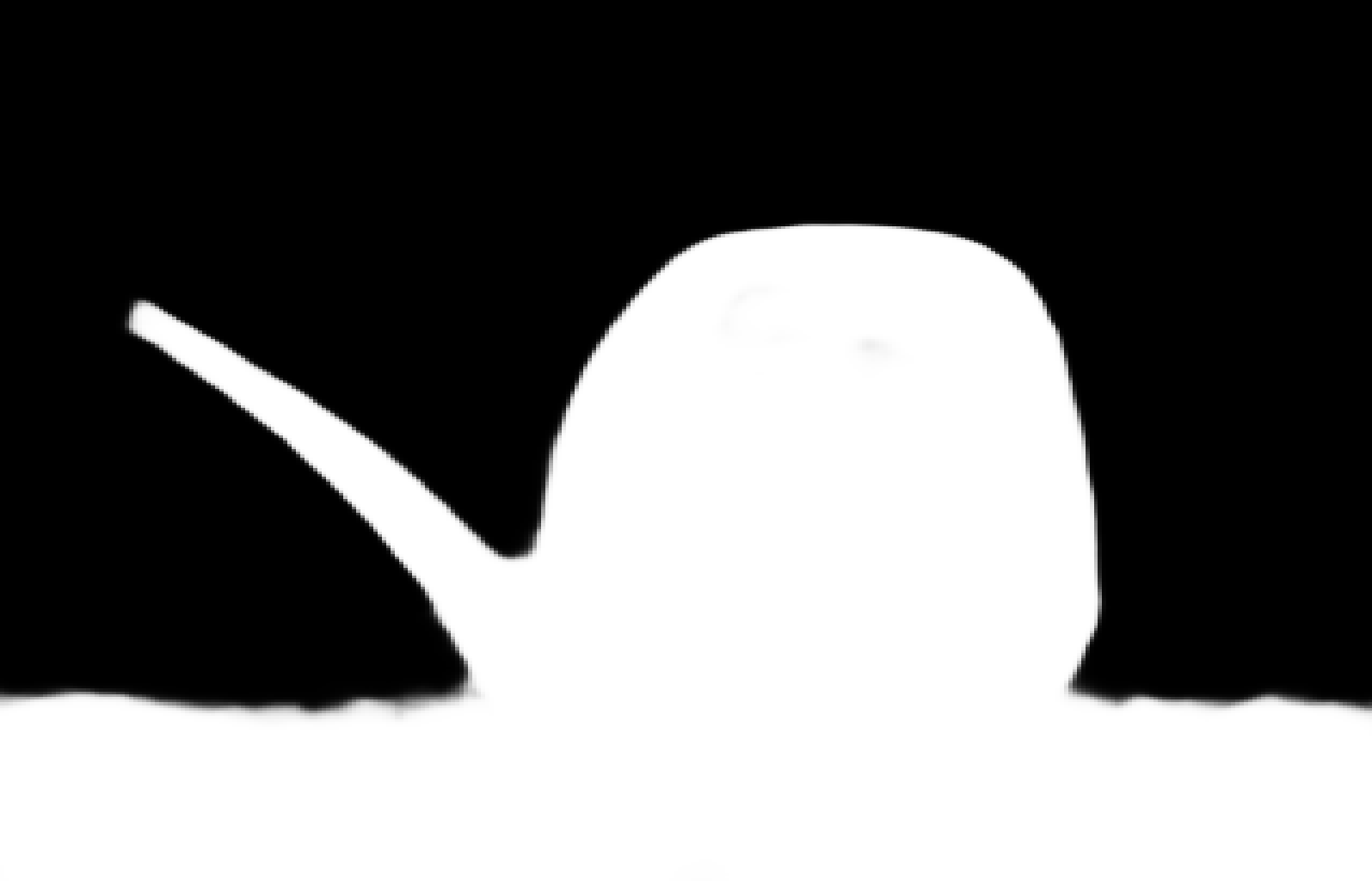}
	\end{subfigure}
    \begin{subfigure}{0.12\linewidth}
		\centering
		\includegraphics[width=\linewidth]{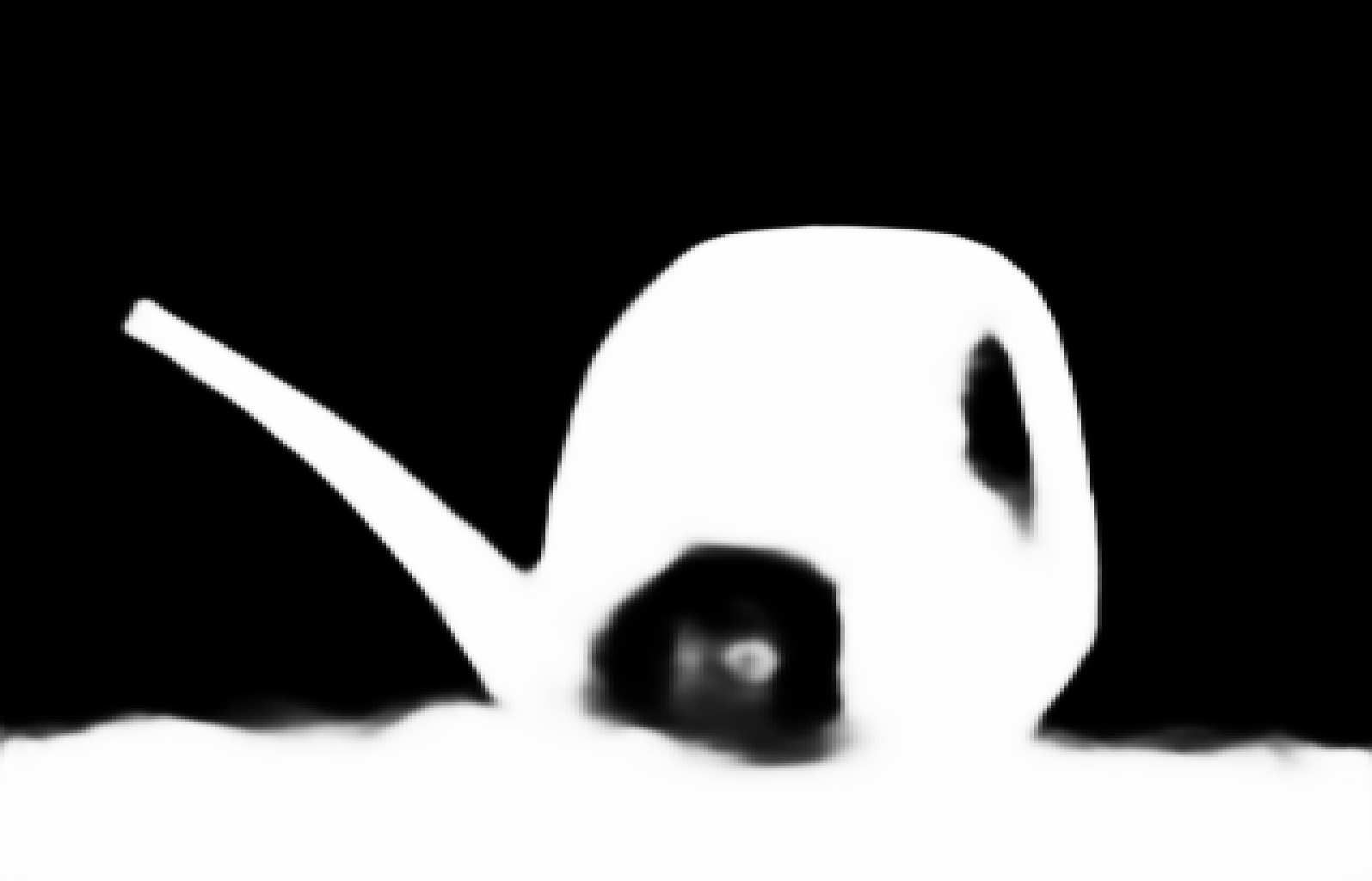}
	\end{subfigure}
    \begin{subfigure}{0.12\linewidth}
		\centering
		\includegraphics[width=\linewidth]{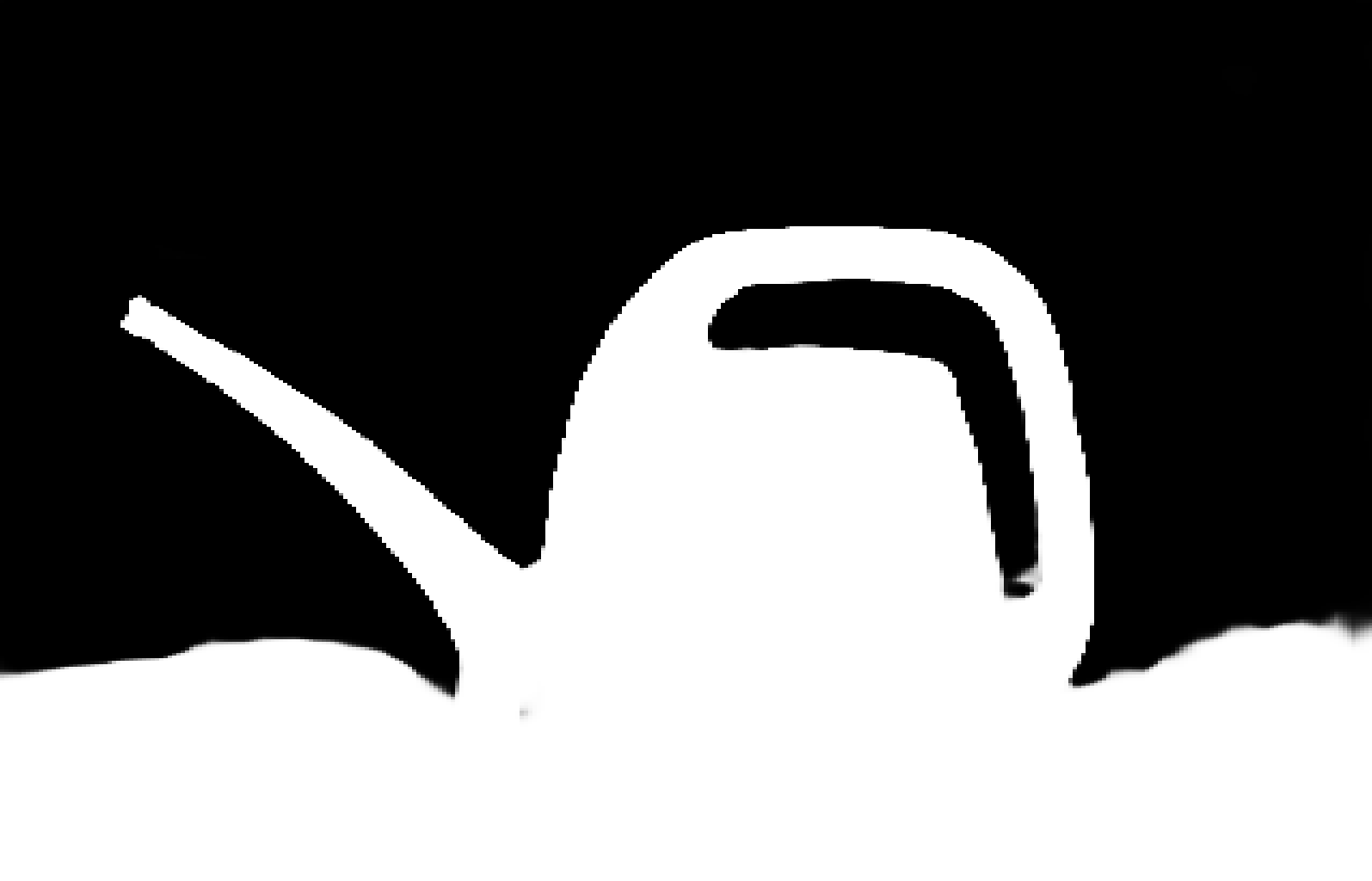}
	\end{subfigure}
    \begin{subfigure}{0.12\linewidth}
		\centering
		\includegraphics[width=\linewidth]{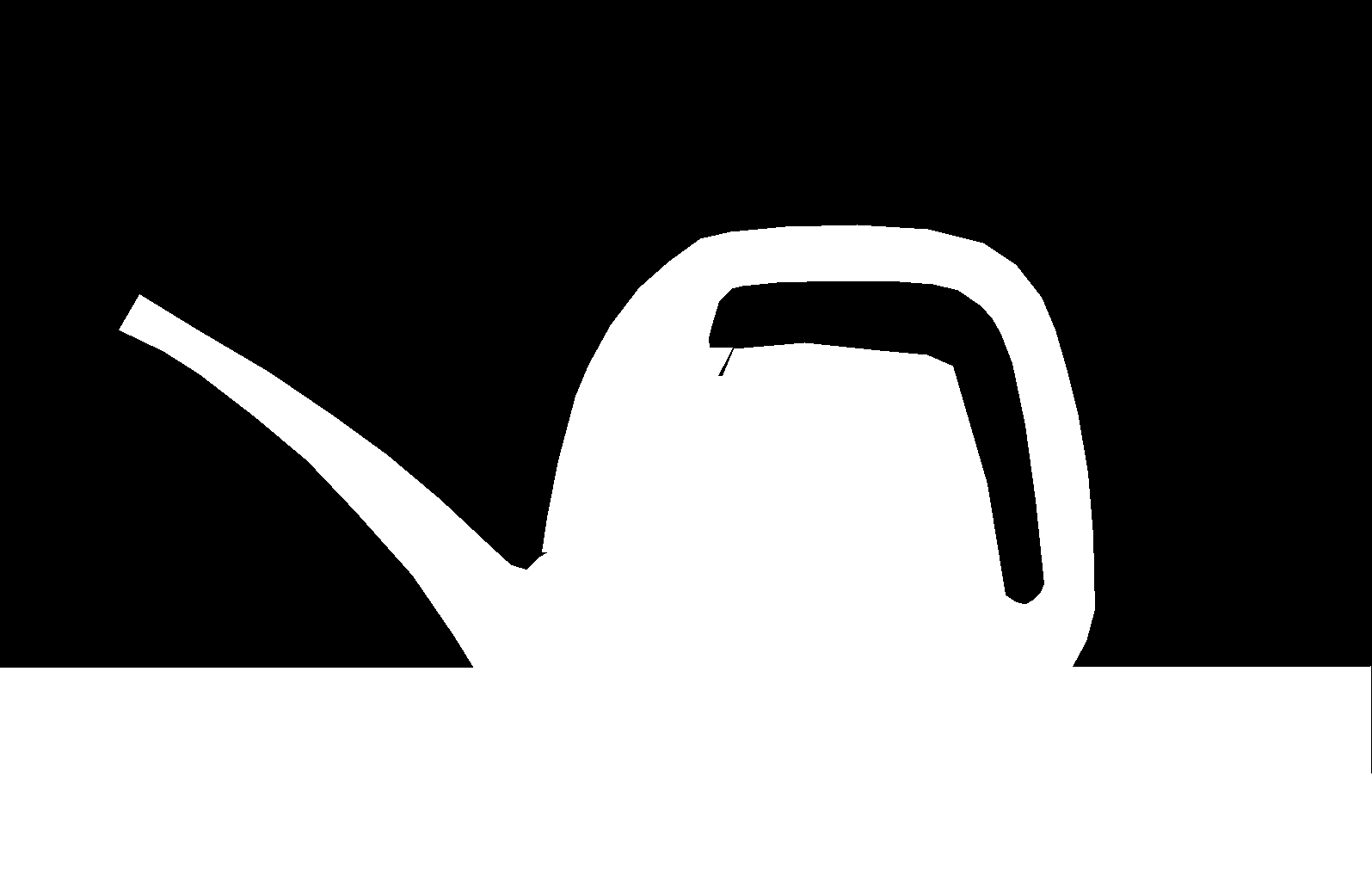}
	\end{subfigure}

    \begin{subfigure}{0.12\linewidth}
		\centering
		\includegraphics[width=\linewidth]{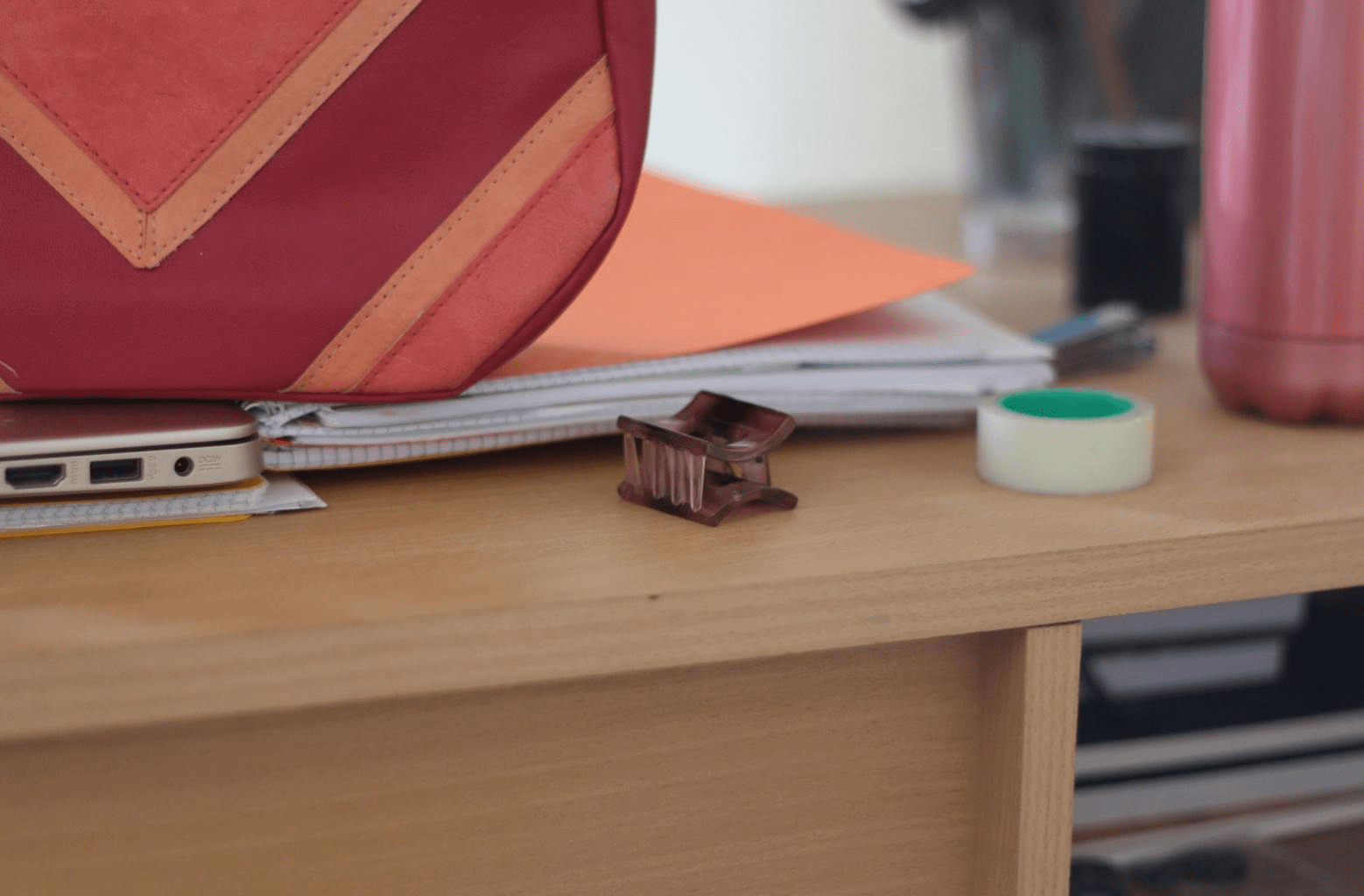}
	\end{subfigure}
    \begin{subfigure}{0.12\linewidth}
		\centering
		\includegraphics[width=\linewidth]{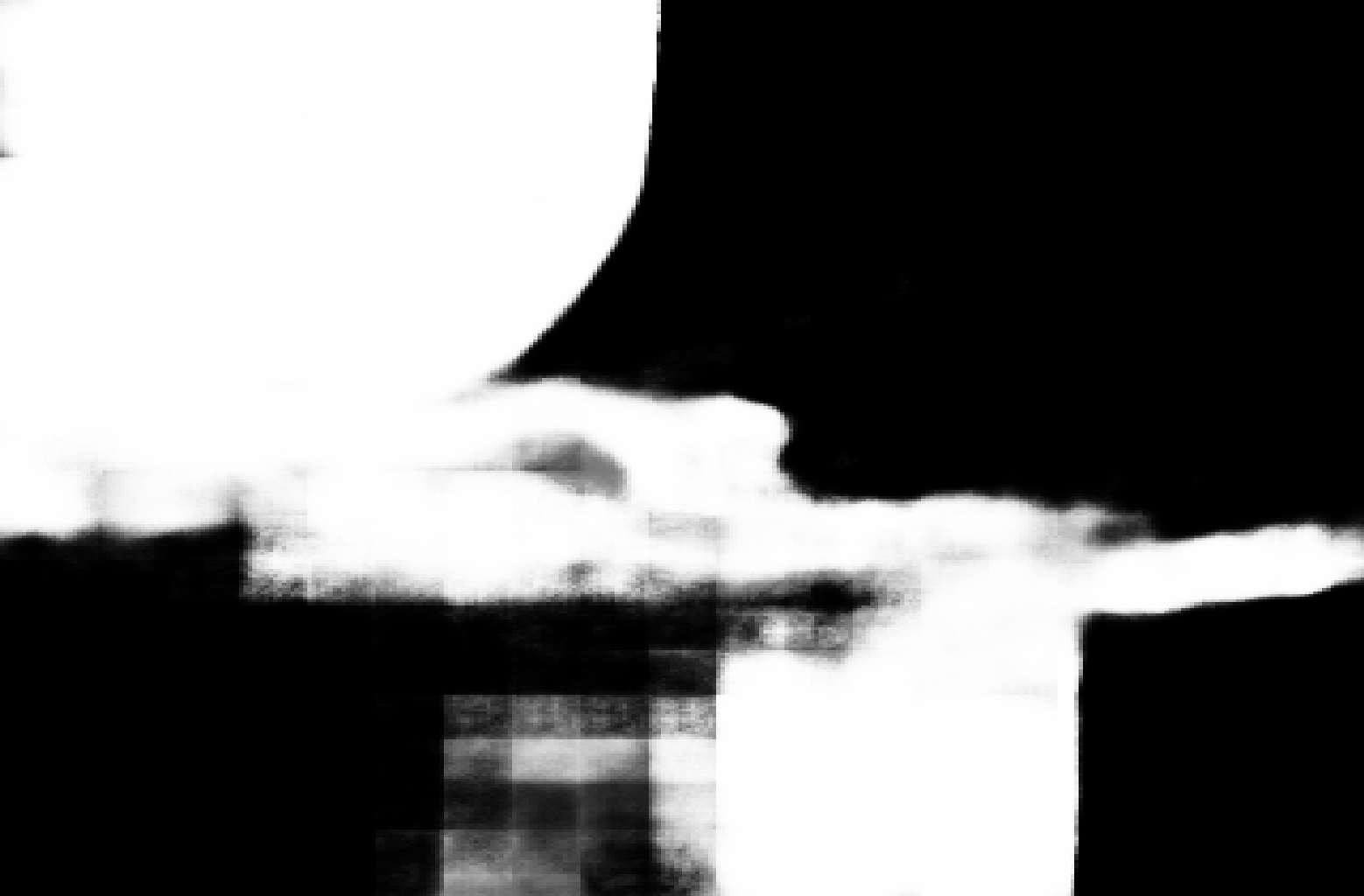}
	\end{subfigure}
    \begin{subfigure}{0.12\linewidth}
		\centering
		\includegraphics[width=\linewidth]{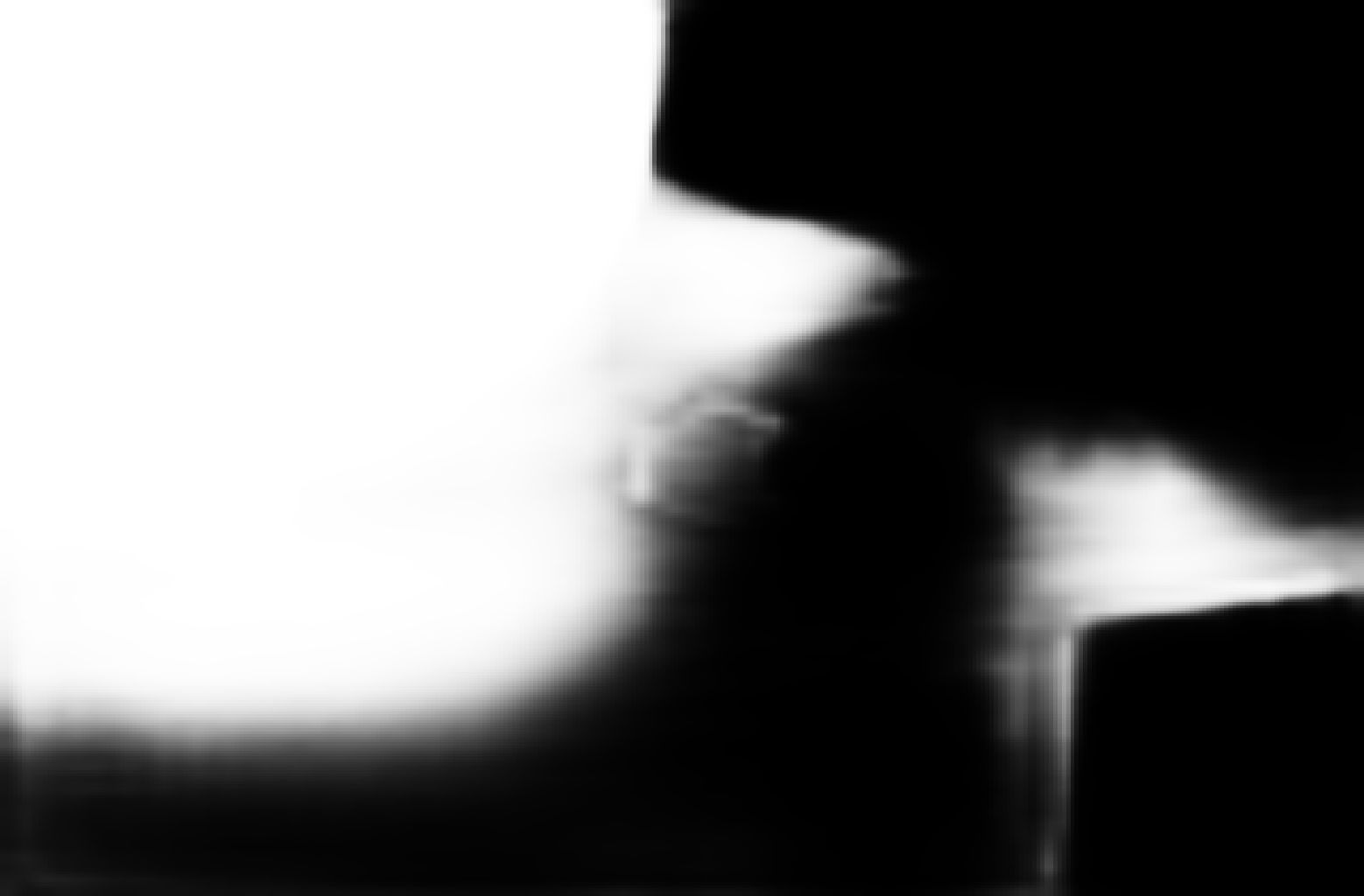}
	\end{subfigure}
    \begin{subfigure}{0.12\linewidth}
		\centering
		\includegraphics[width=\linewidth]{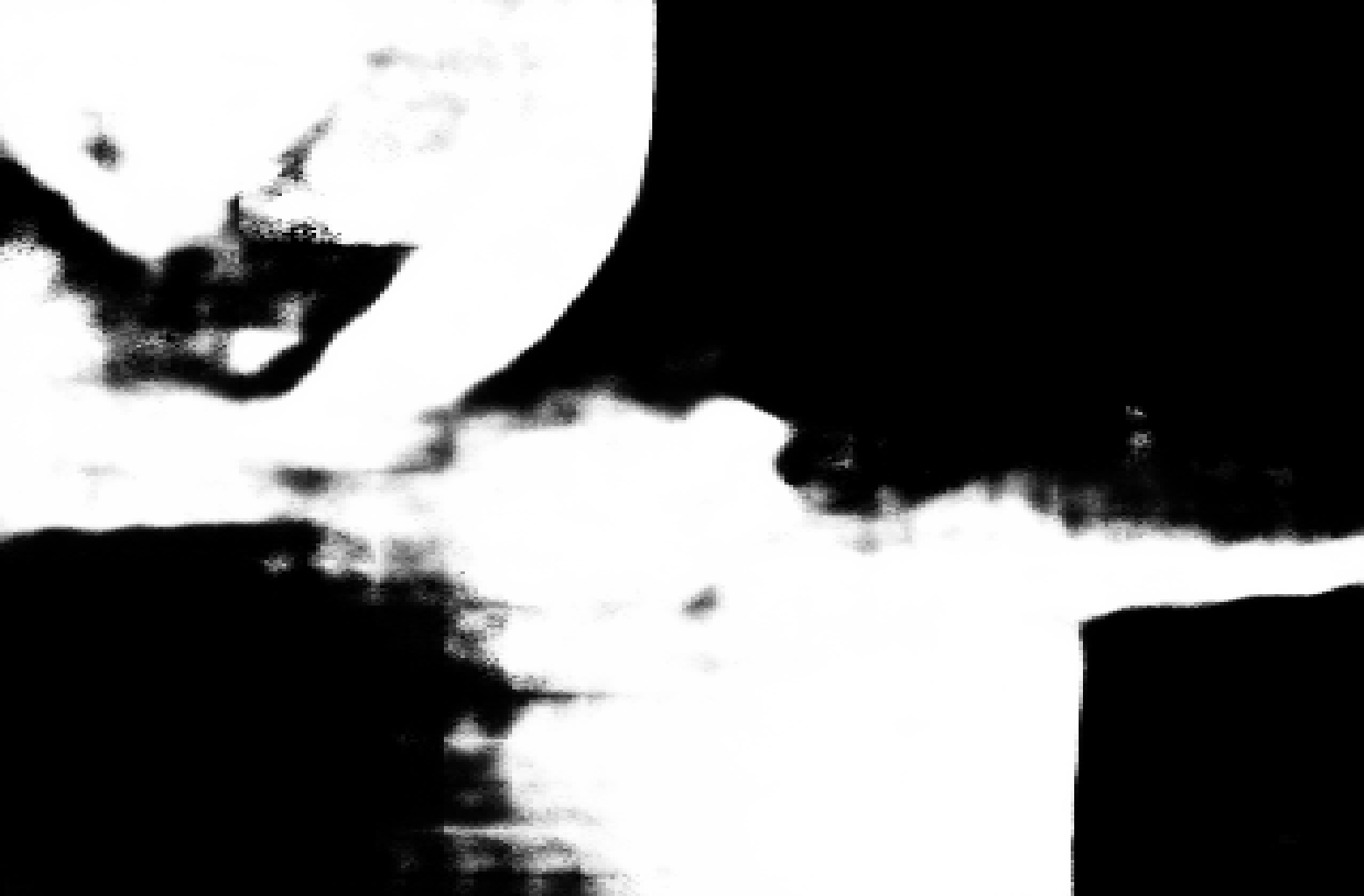}
	\end{subfigure}
    \begin{subfigure}{0.12\linewidth}
		\centering
		\includegraphics[width=\linewidth]{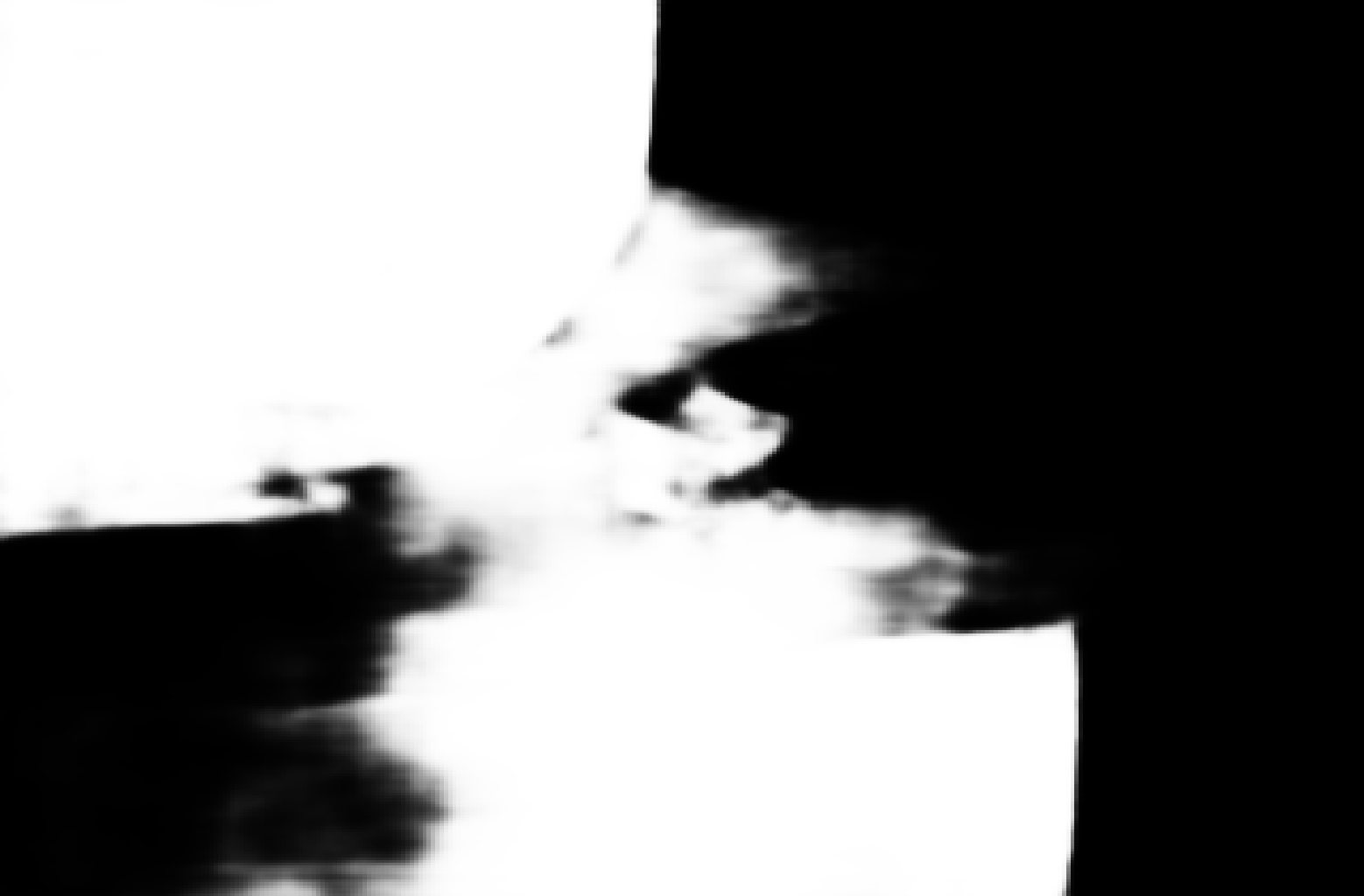}
	\end{subfigure}
    \begin{subfigure}{0.12\linewidth}
		\centering
		\includegraphics[width=\linewidth]{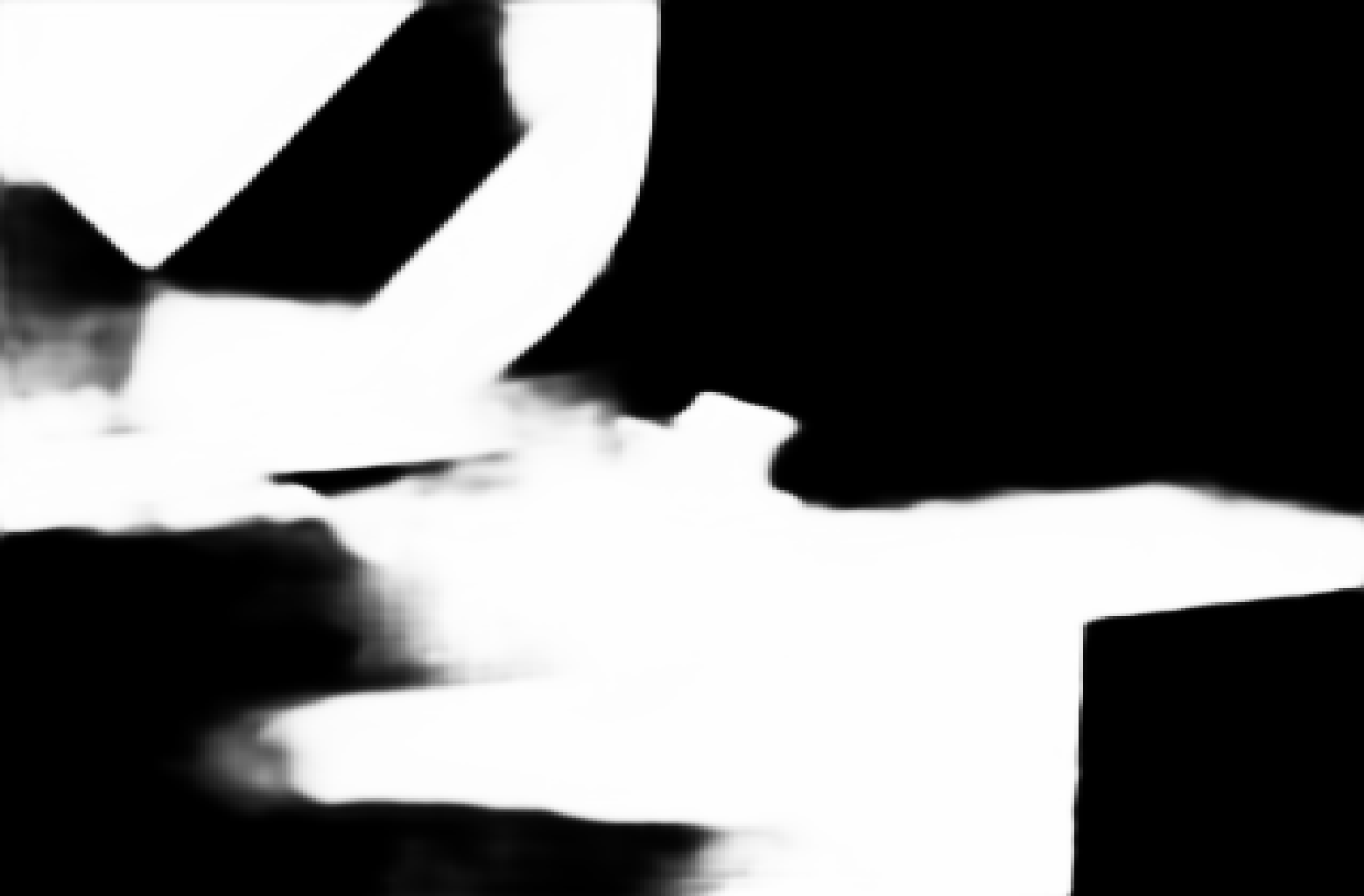}
	\end{subfigure}
    \begin{subfigure}{0.12\linewidth}
		\centering
		\includegraphics[width=\linewidth]{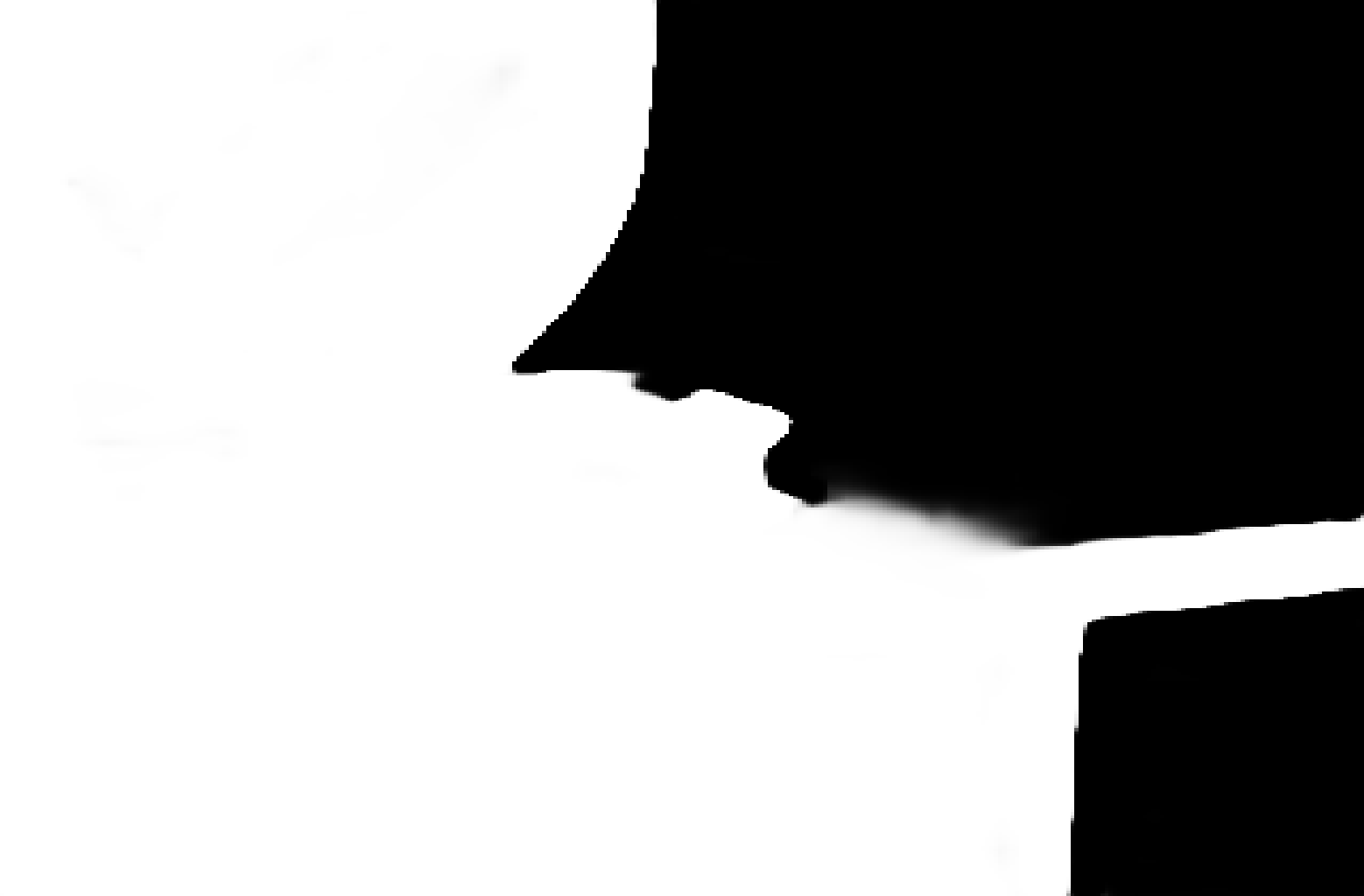}
	\end{subfigure}
    \begin{subfigure}{0.12\linewidth}
		\centering
		\includegraphics[width=\linewidth]{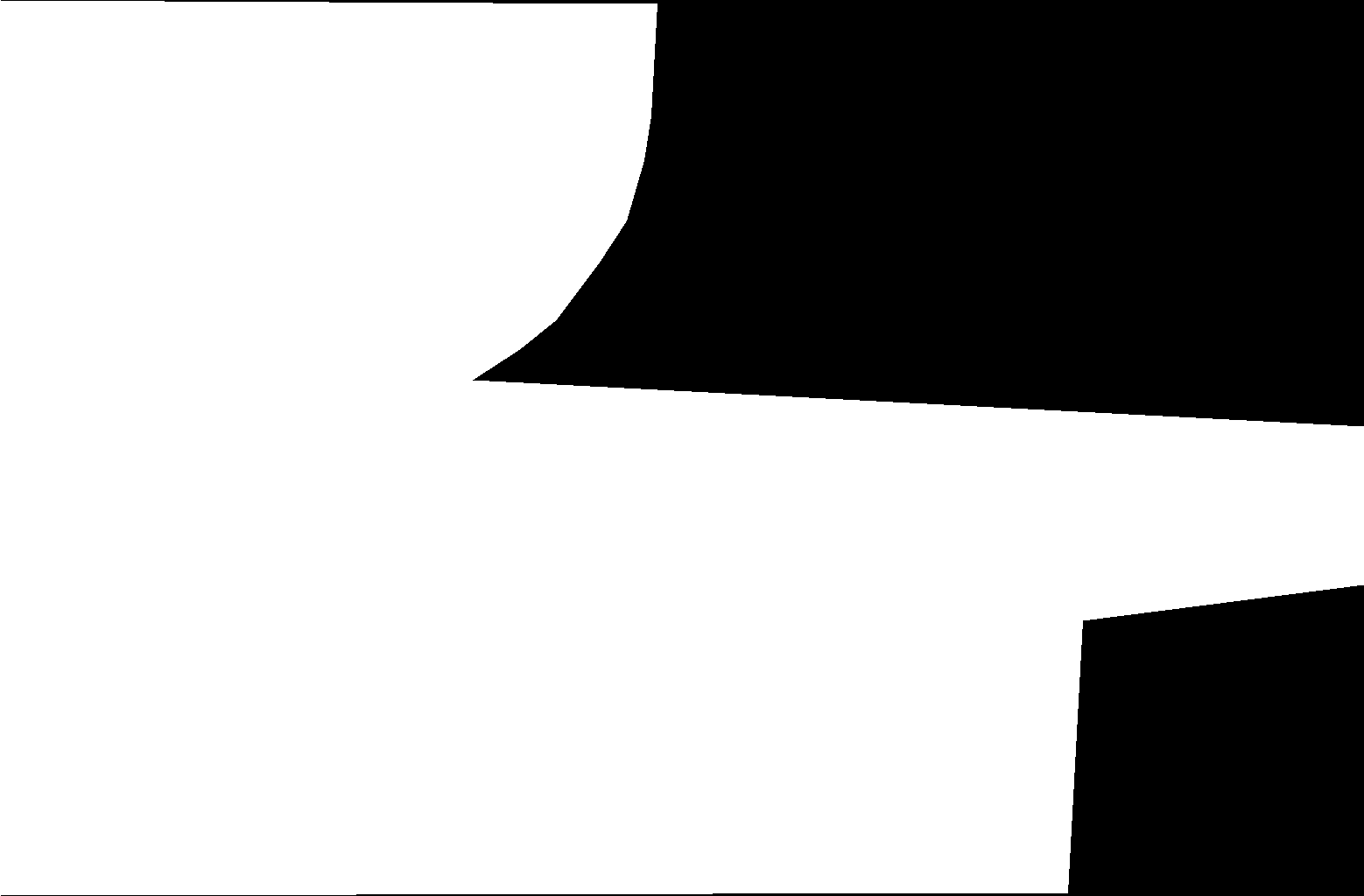}
	\end{subfigure}

    \begin{subfigure}{0.12\linewidth}
		\centering
		\includegraphics[width=\linewidth]{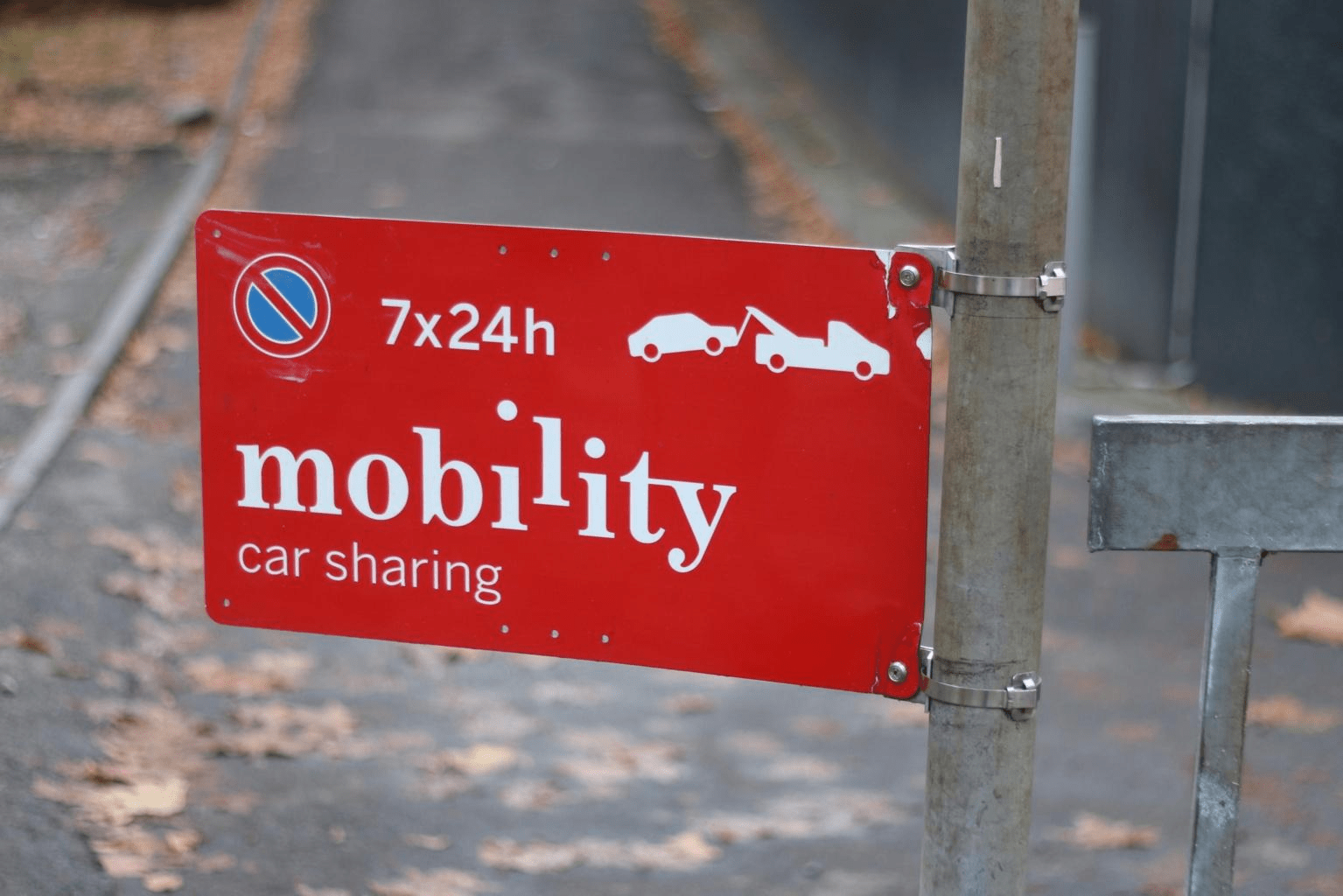}
	\end{subfigure}
    \begin{subfigure}{0.12\linewidth}
		\centering
		\includegraphics[width=\linewidth]{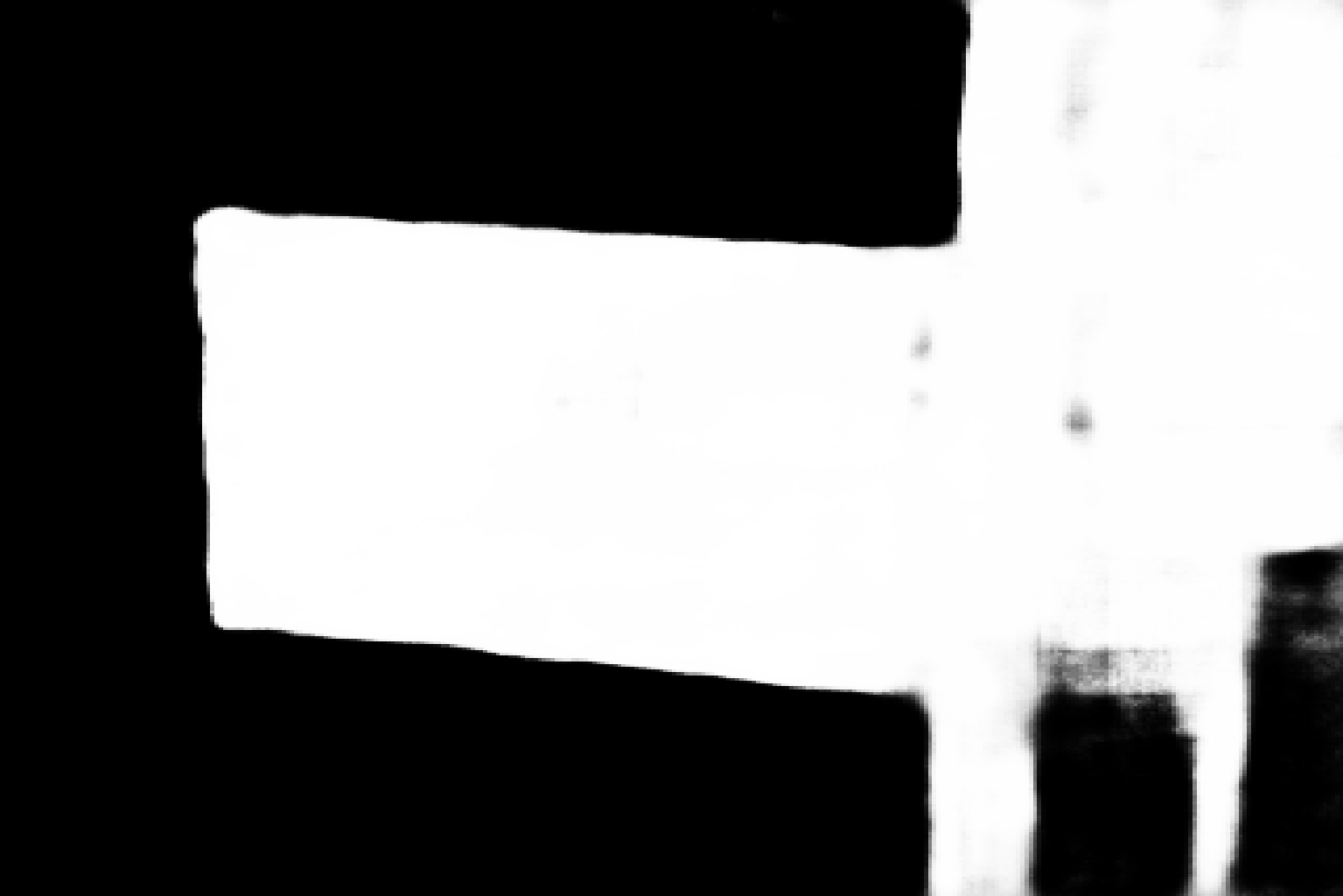}
	\end{subfigure}
    \begin{subfigure}{0.12\linewidth}
		\centering
		\includegraphics[width=\linewidth]{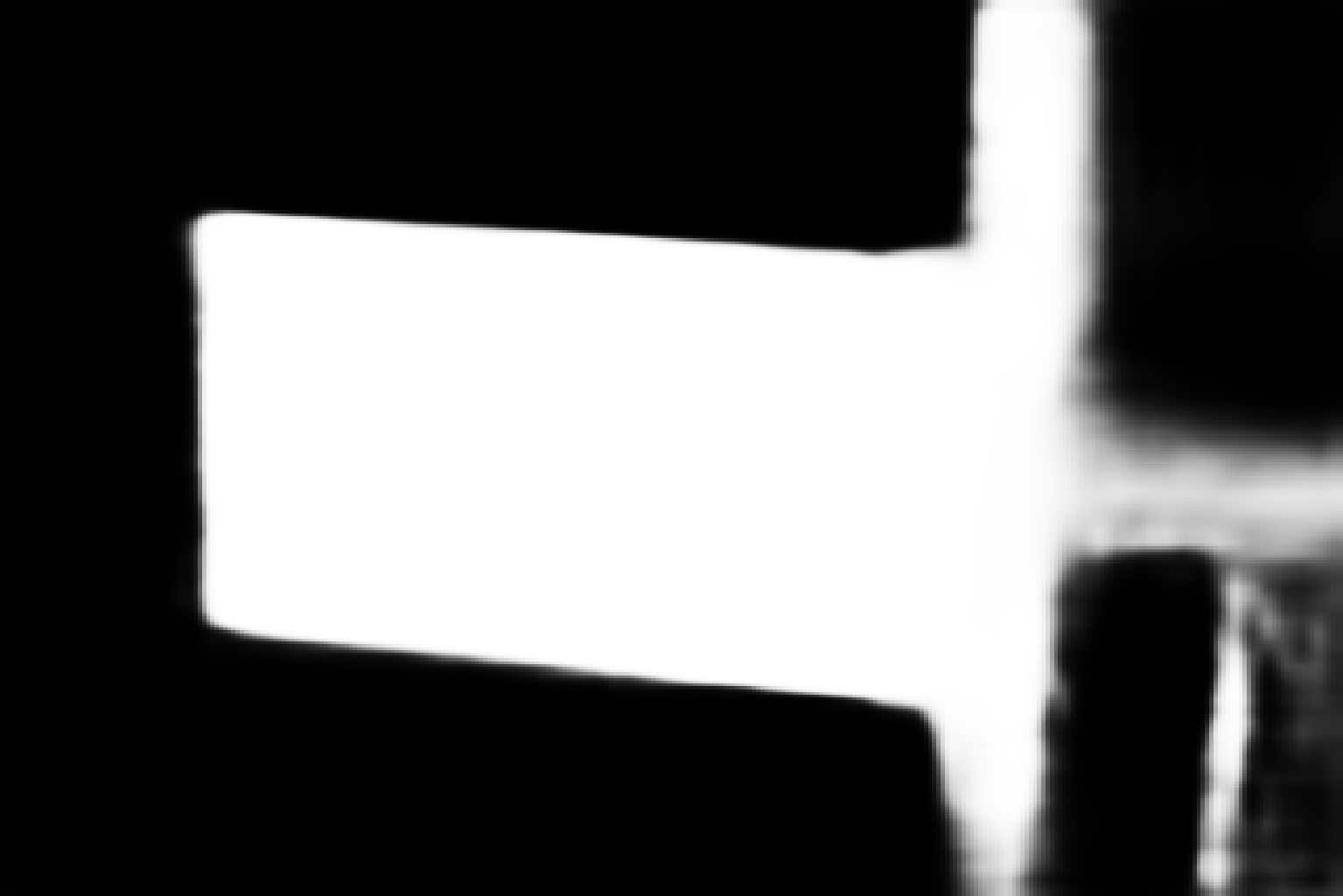}
	\end{subfigure}
    \begin{subfigure}{0.12\linewidth}
		\centering
		\includegraphics[width=\linewidth]{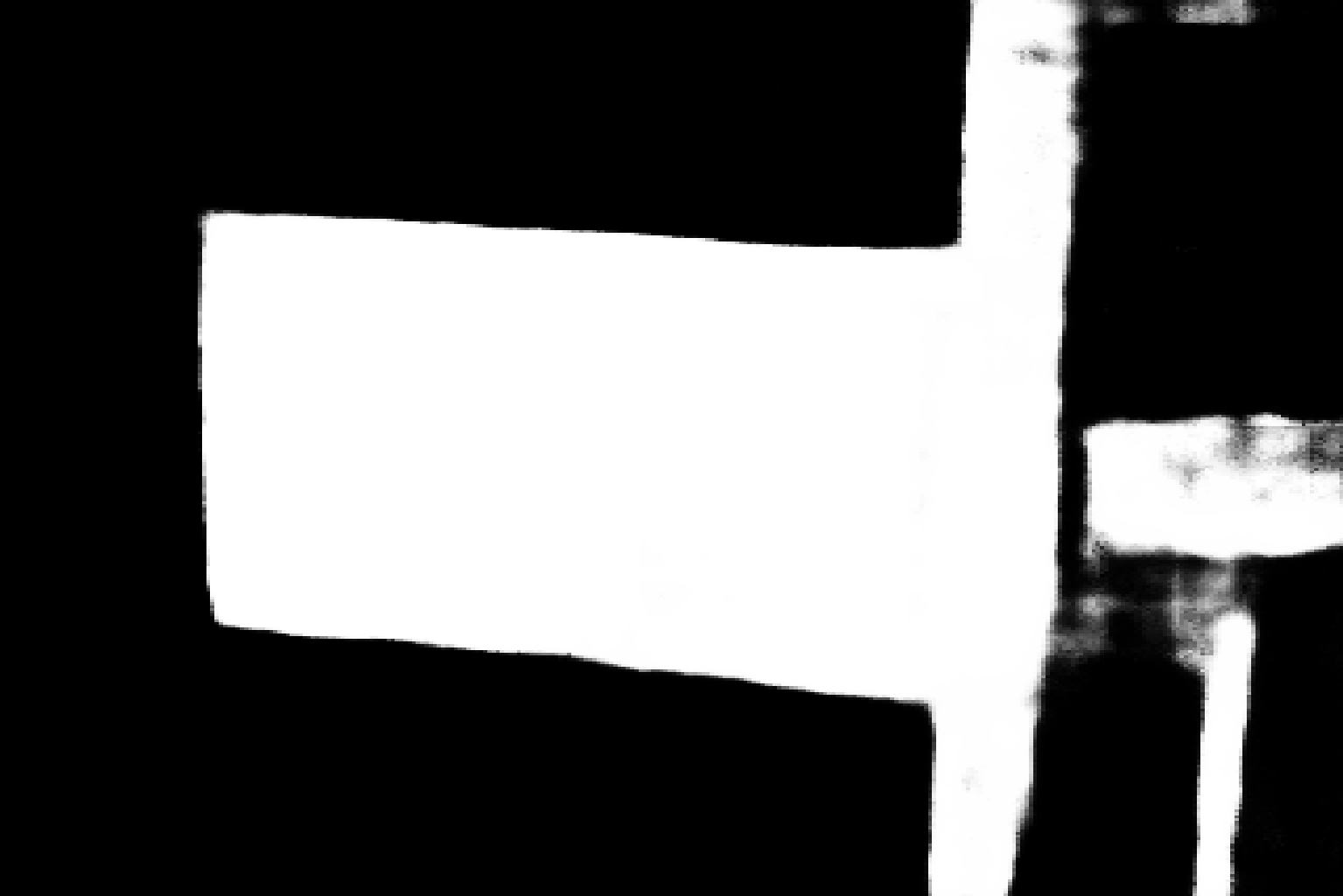}
	\end{subfigure}
    \begin{subfigure}{0.12\linewidth}
		\centering
		\includegraphics[width=\linewidth]{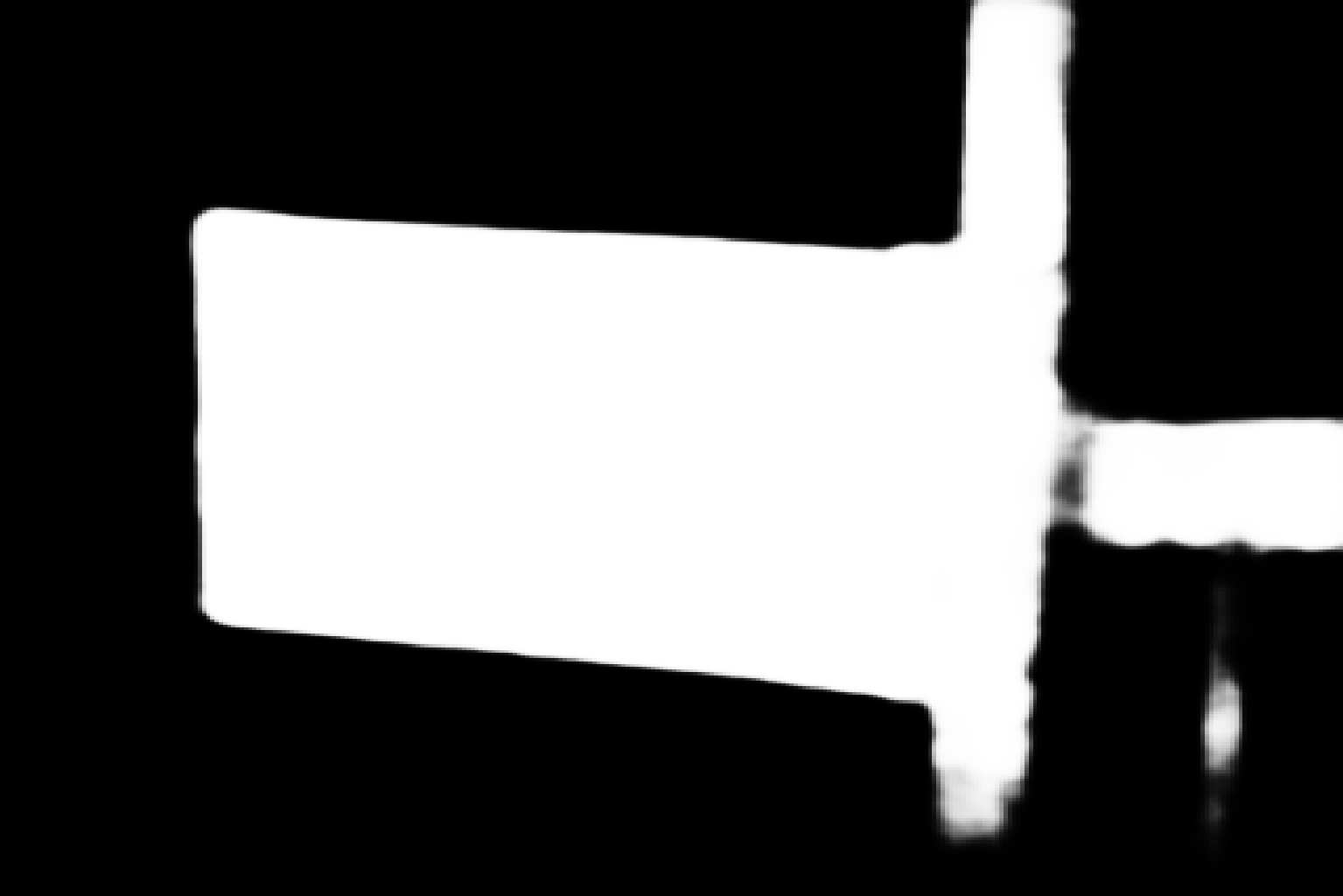}
	\end{subfigure}
    \begin{subfigure}{0.12\linewidth}
		\centering
		\includegraphics[width=\linewidth]{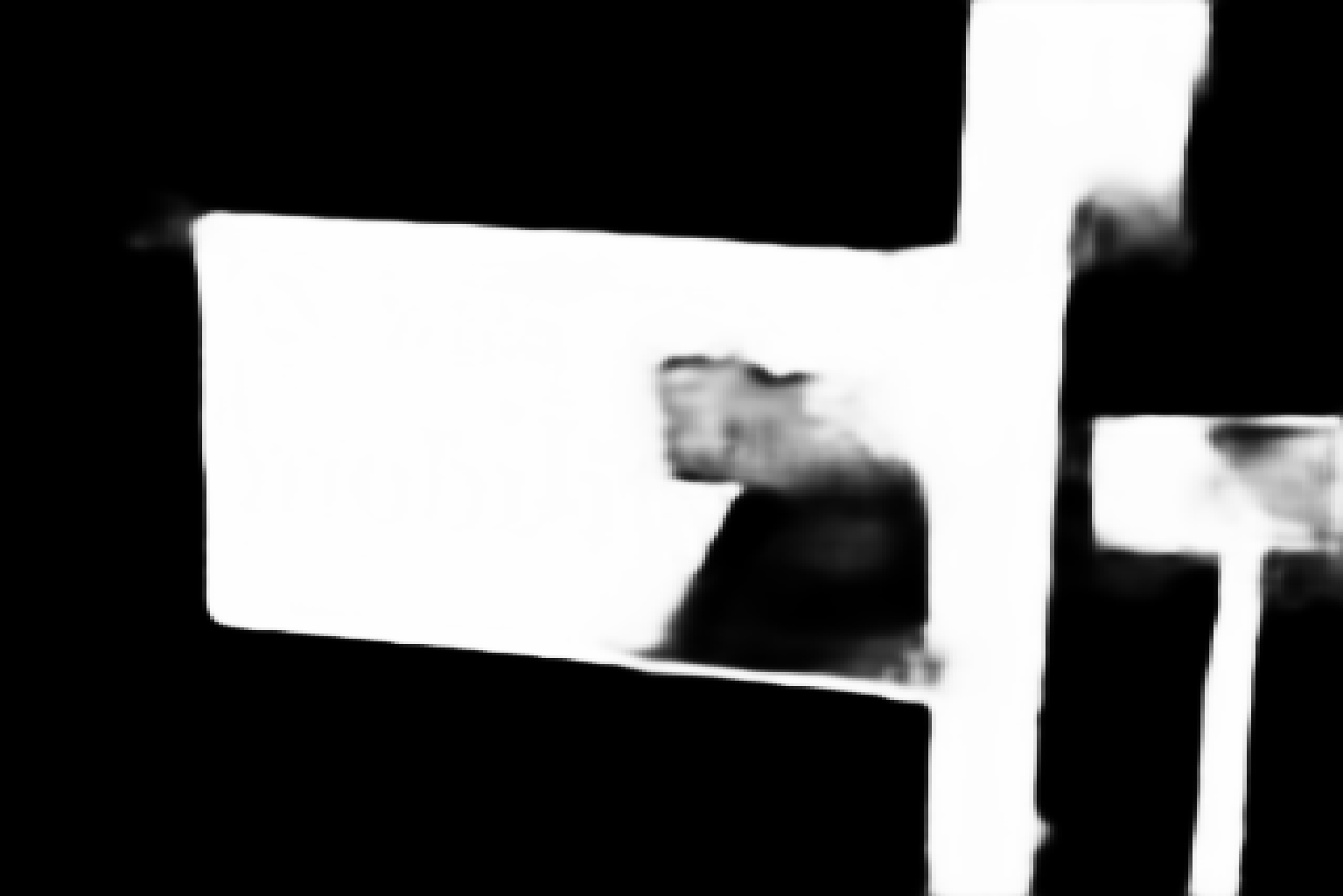}
	\end{subfigure}
    \begin{subfigure}{0.12\linewidth}
		\centering
		\includegraphics[width=\linewidth]{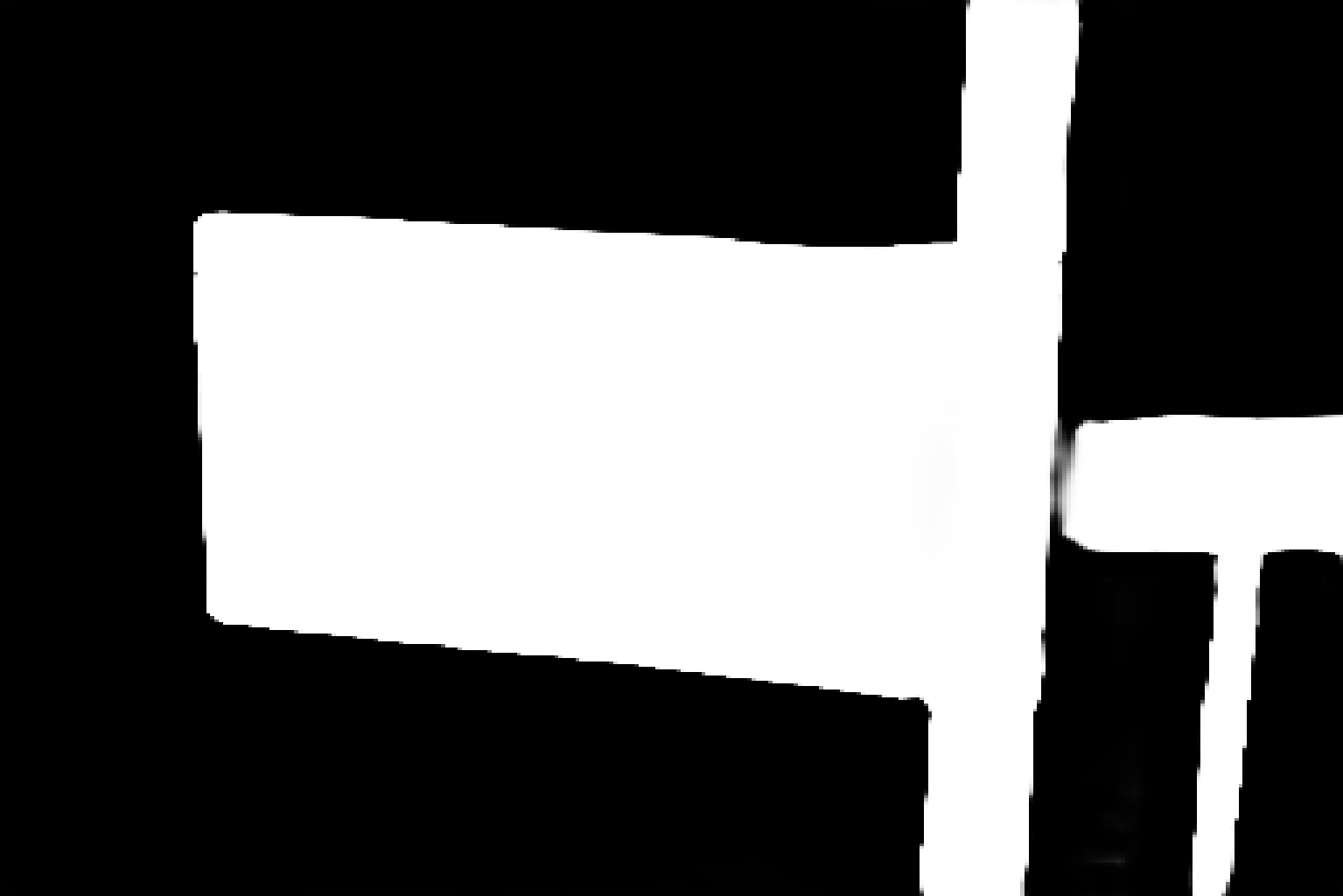}
	\end{subfigure}
    \begin{subfigure}{0.12\linewidth}
		\centering
		\includegraphics[width=\linewidth]{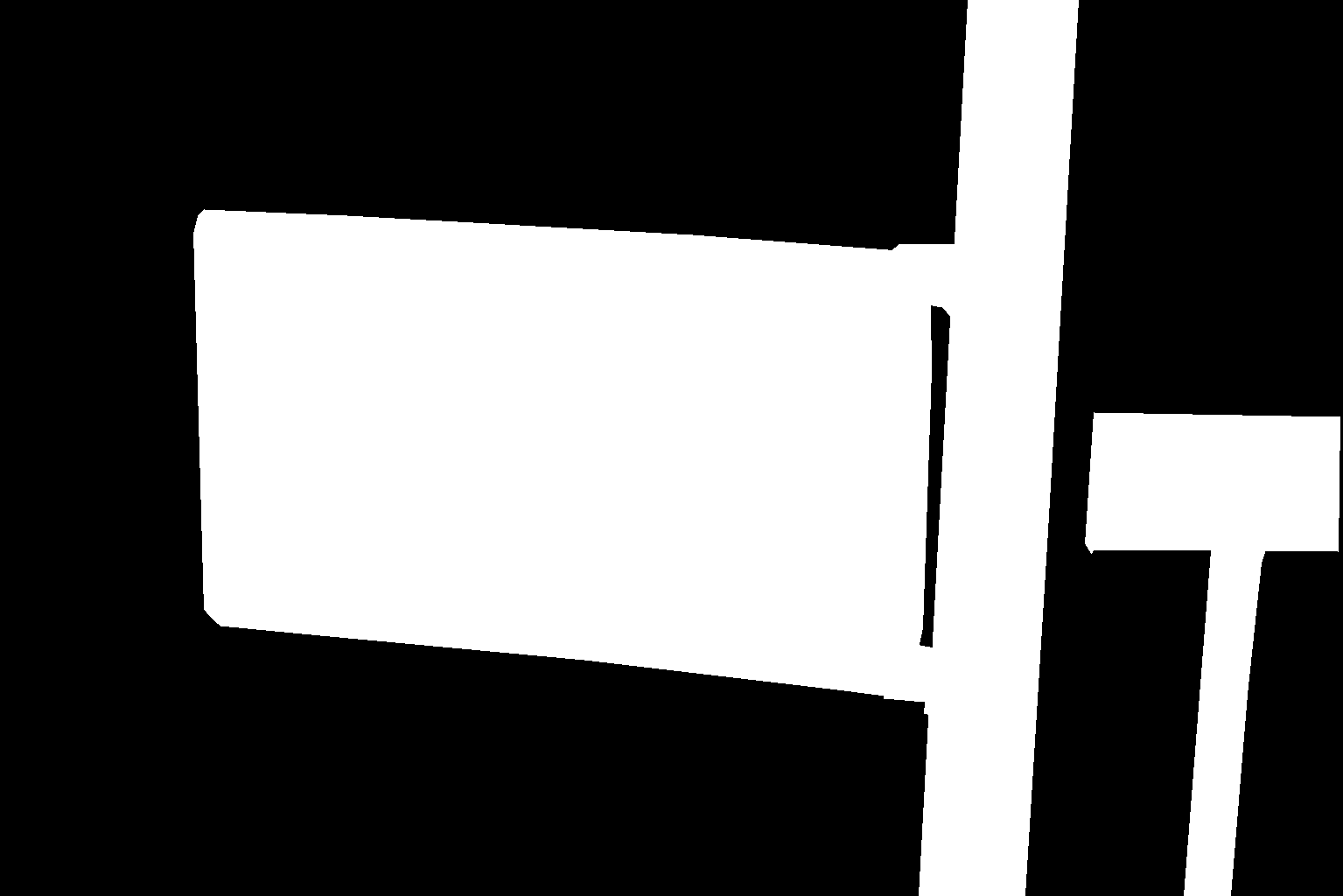}
	\end{subfigure}
 
    \begin{subfigure}{0.12\linewidth}
		\centering
		\includegraphics[width=\linewidth]{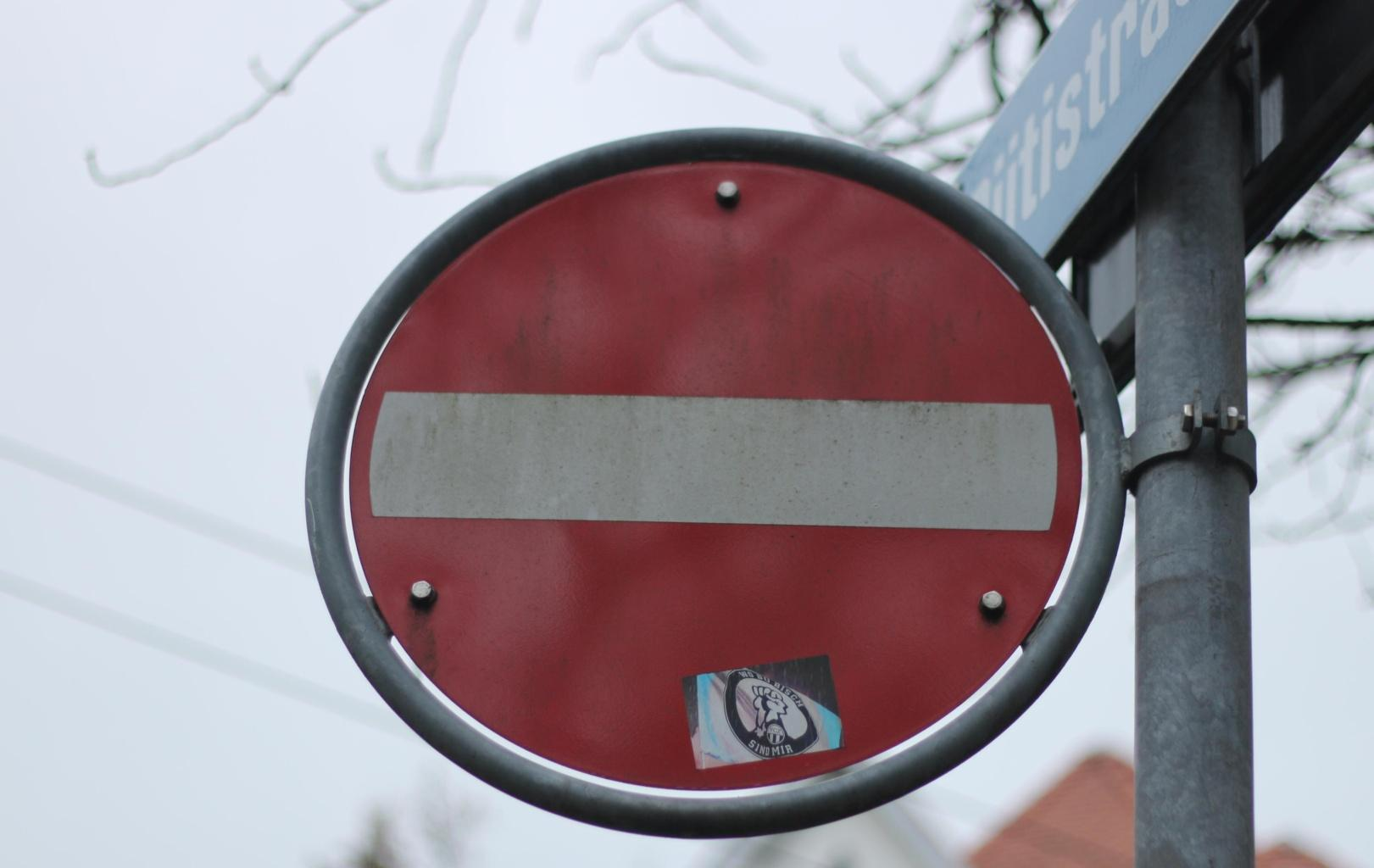}
	\end{subfigure}
    \begin{subfigure}{0.12\linewidth}
		\centering
		\includegraphics[width=\linewidth]{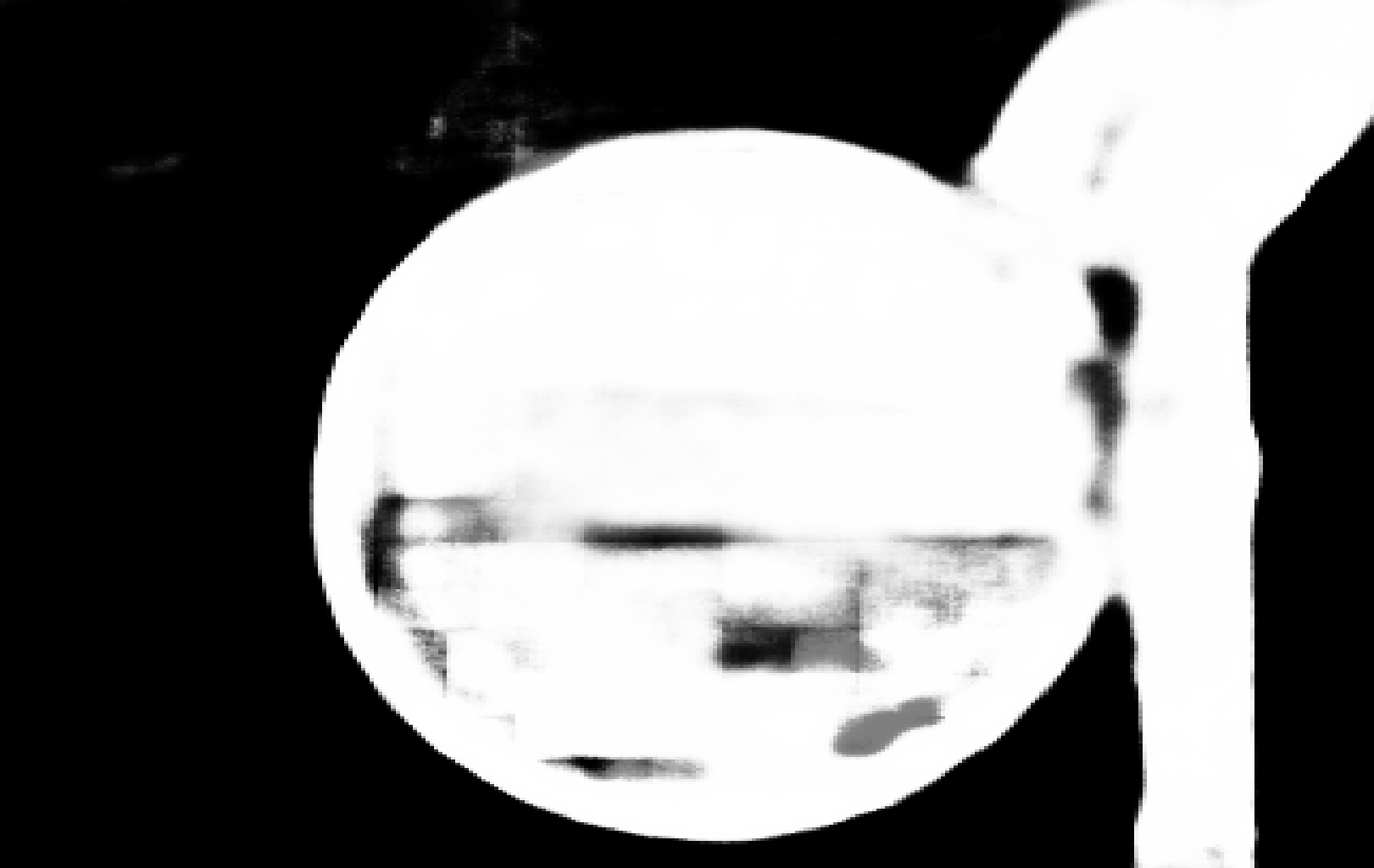}
	\end{subfigure}
    \begin{subfigure}{0.12\linewidth}
		\centering
		\includegraphics[width=\linewidth]{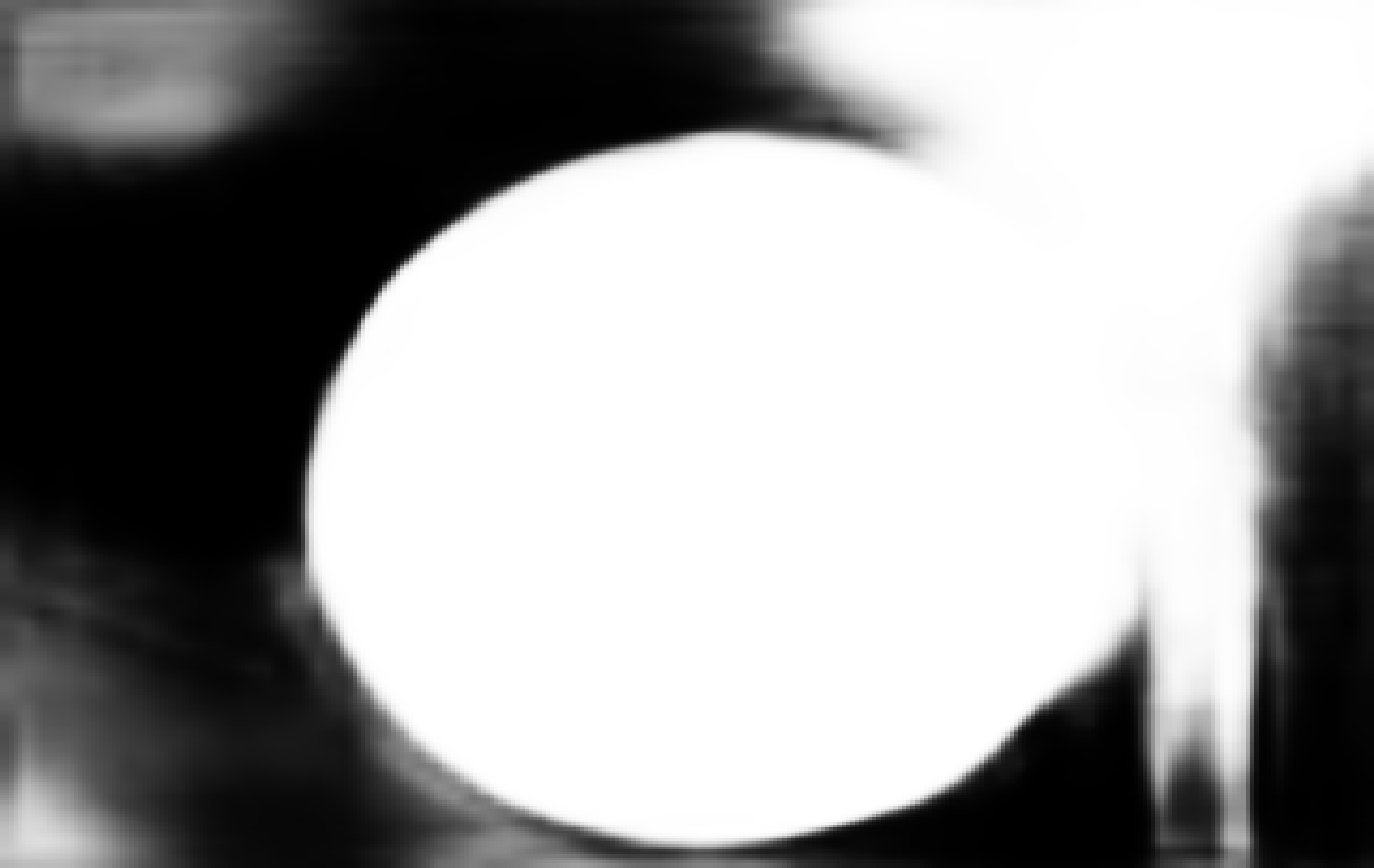}
	\end{subfigure}
    \begin{subfigure}{0.12\linewidth}
		\centering
		\includegraphics[width=\linewidth]{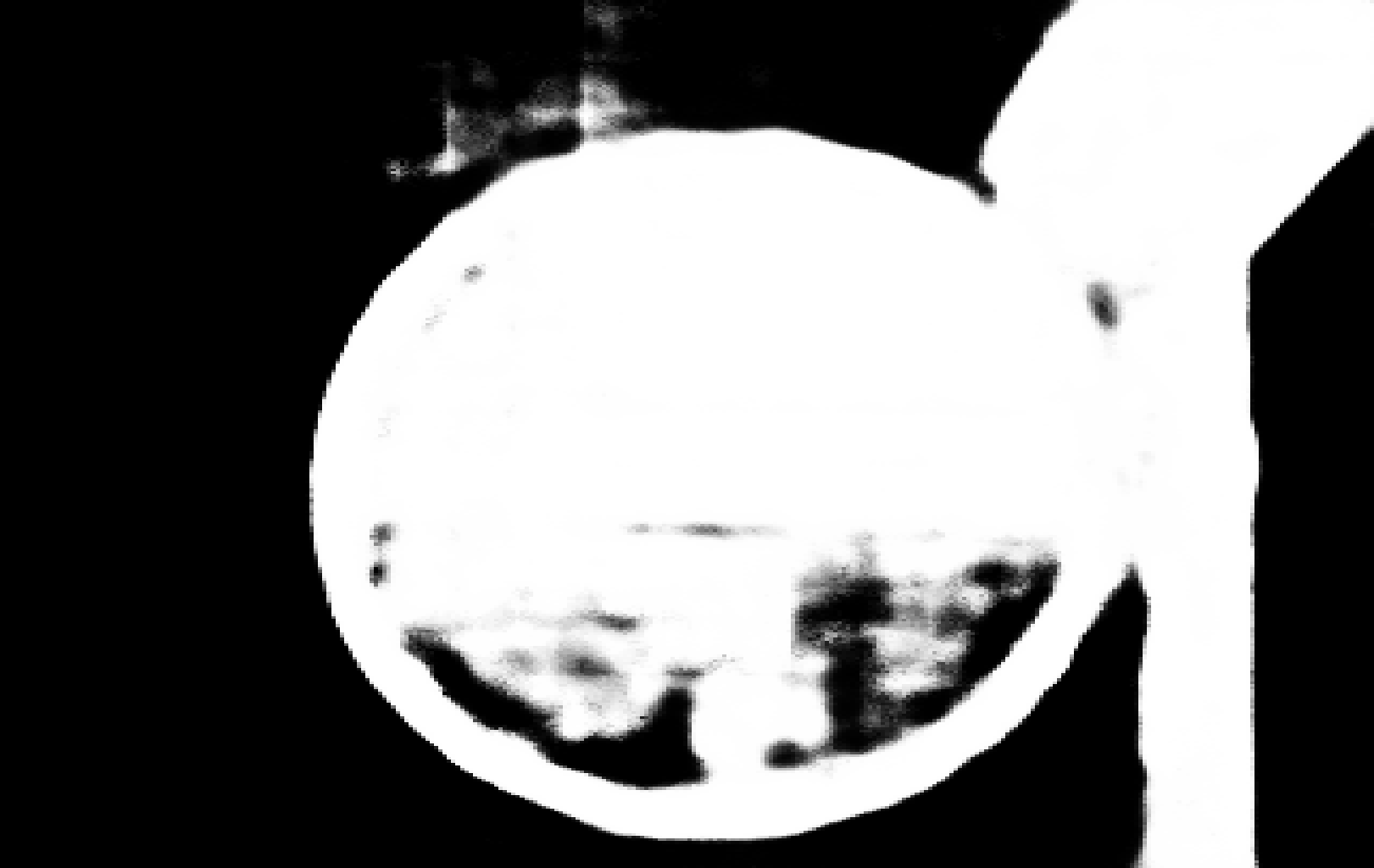}
	\end{subfigure}
    \begin{subfigure}{0.12\linewidth}
		\centering
		\includegraphics[width=\linewidth]{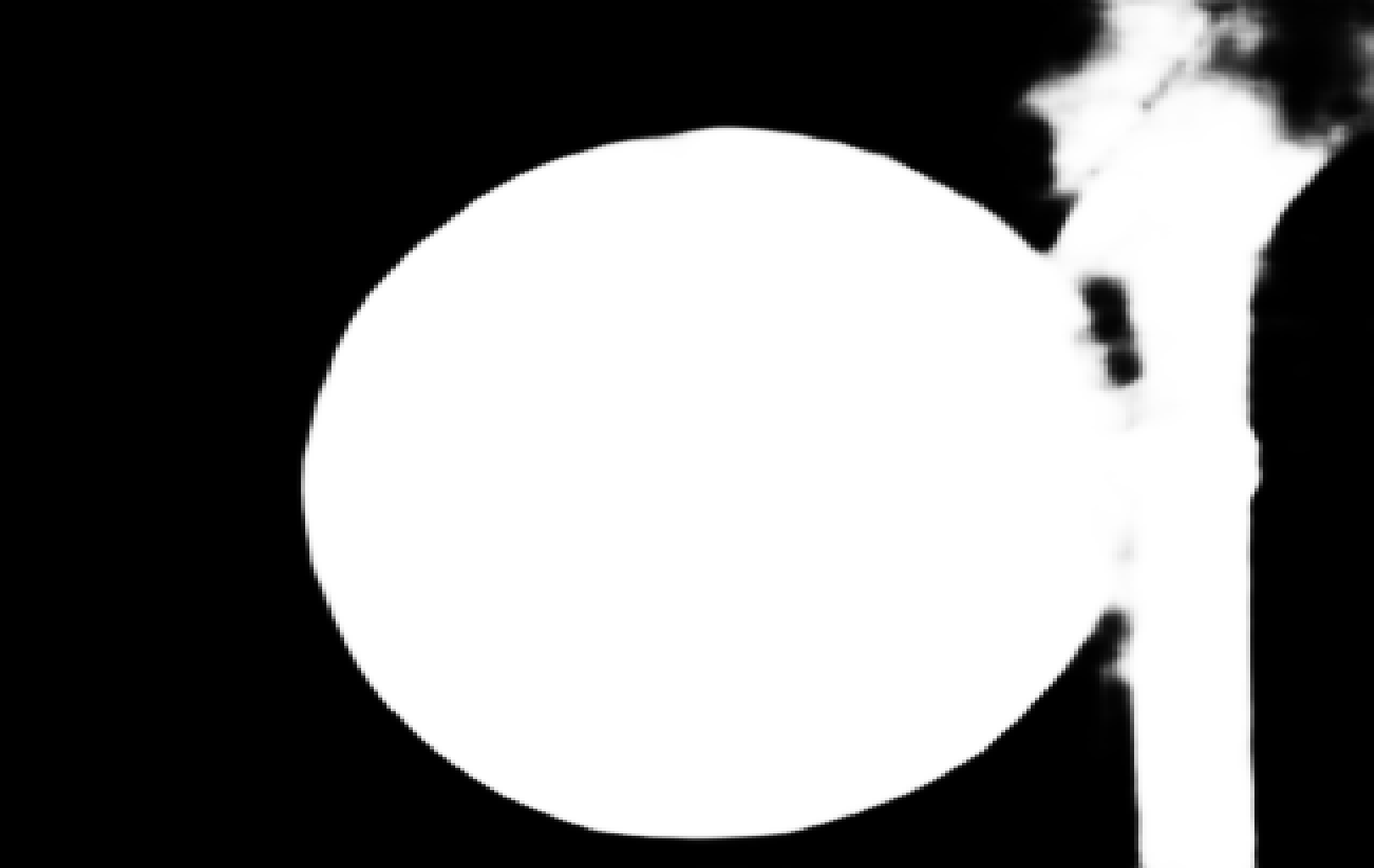}
	\end{subfigure}
    \begin{subfigure}{0.12\linewidth}
		\centering
		\includegraphics[width=\linewidth]{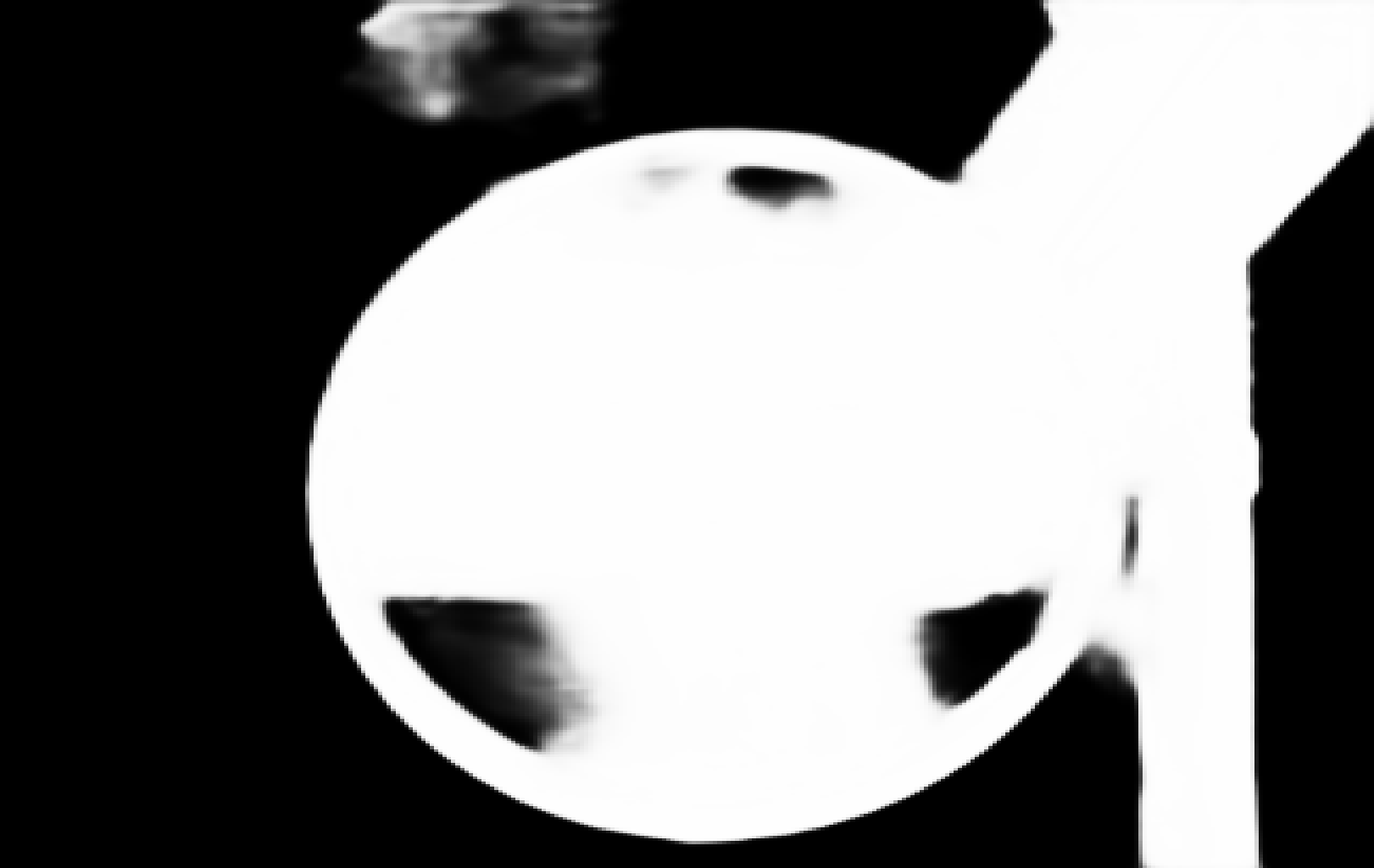}
	\end{subfigure}
    \begin{subfigure}{0.12\linewidth}
		\centering
		\includegraphics[width=\linewidth]{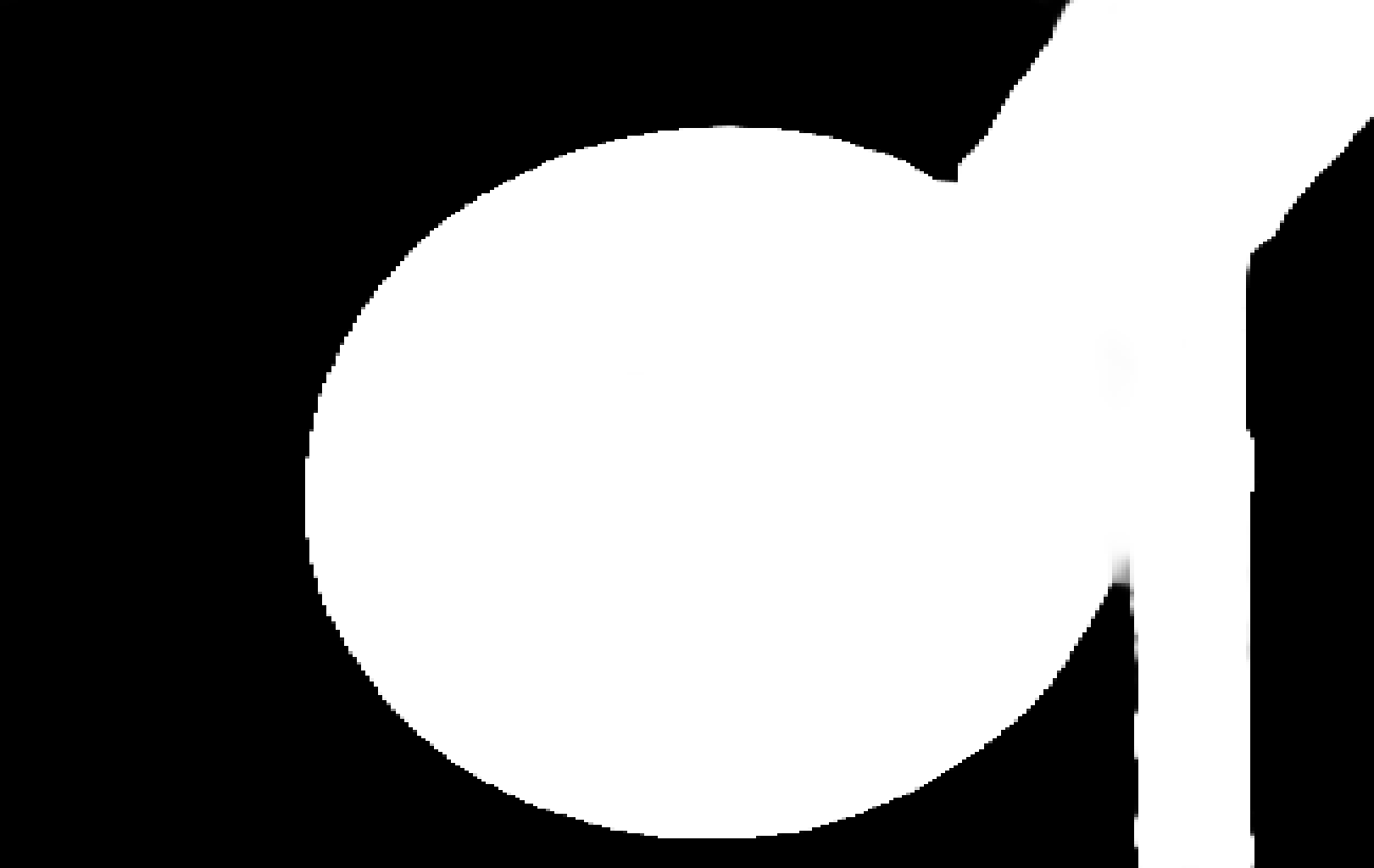}
	\end{subfigure}
    \begin{subfigure}{0.12\linewidth}
		\centering
		\includegraphics[width=\linewidth]{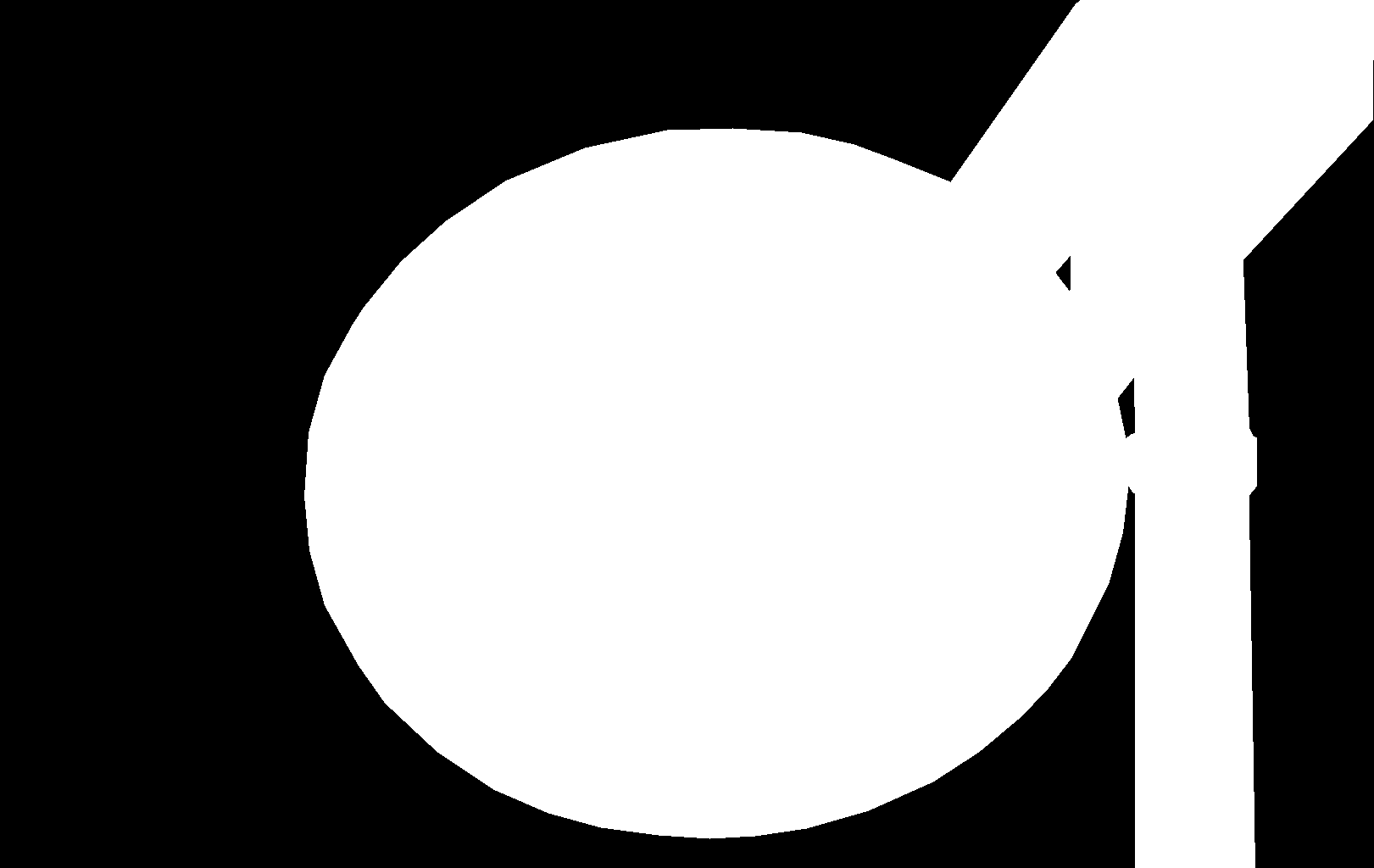}
	\end{subfigure}

    \begin{subfigure}{0.12\linewidth}
		\centering
		\includegraphics[width=\linewidth]{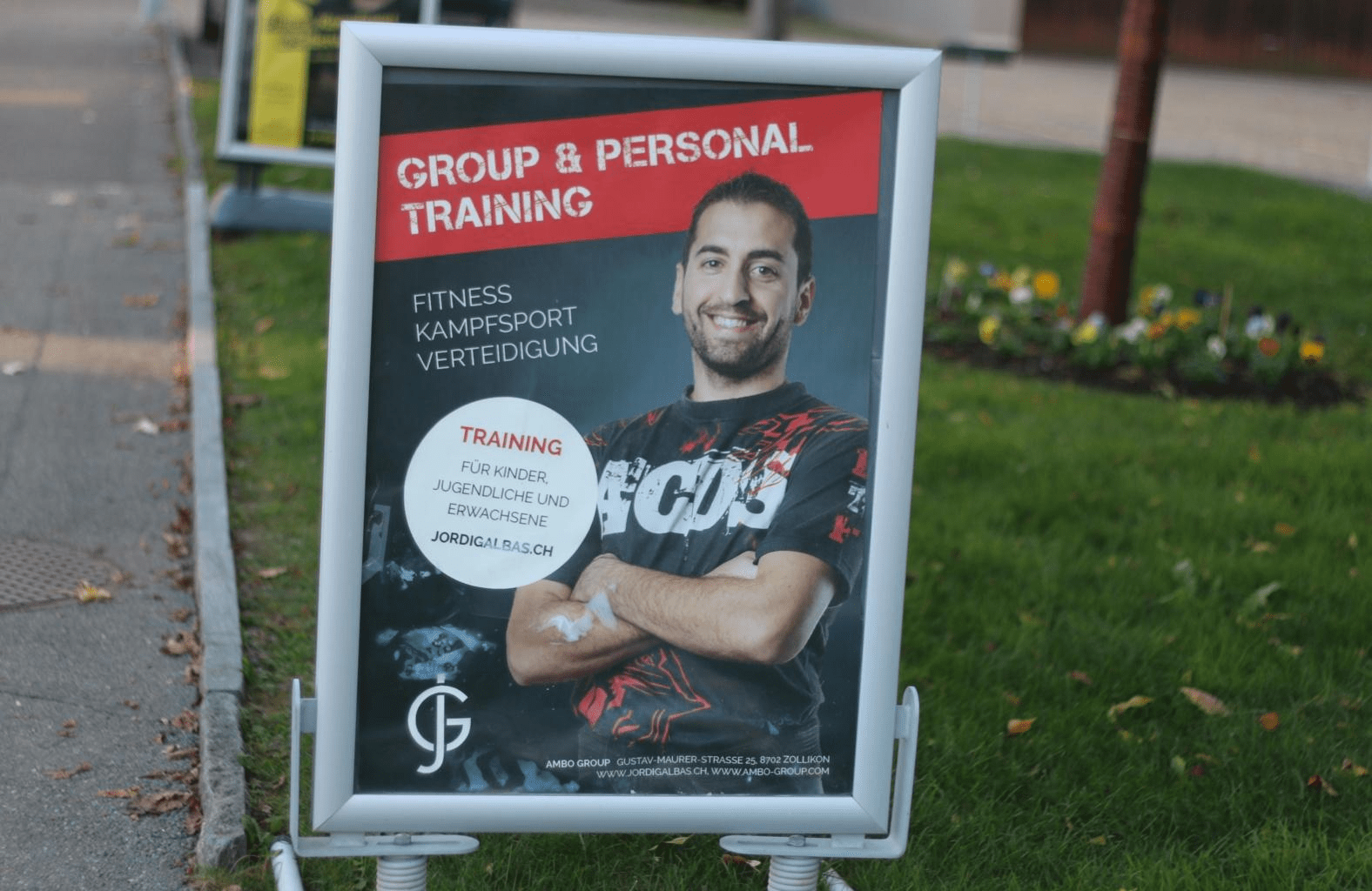}
	\end{subfigure}
    \begin{subfigure}{0.12\linewidth}
		\centering
		\includegraphics[width=\linewidth]{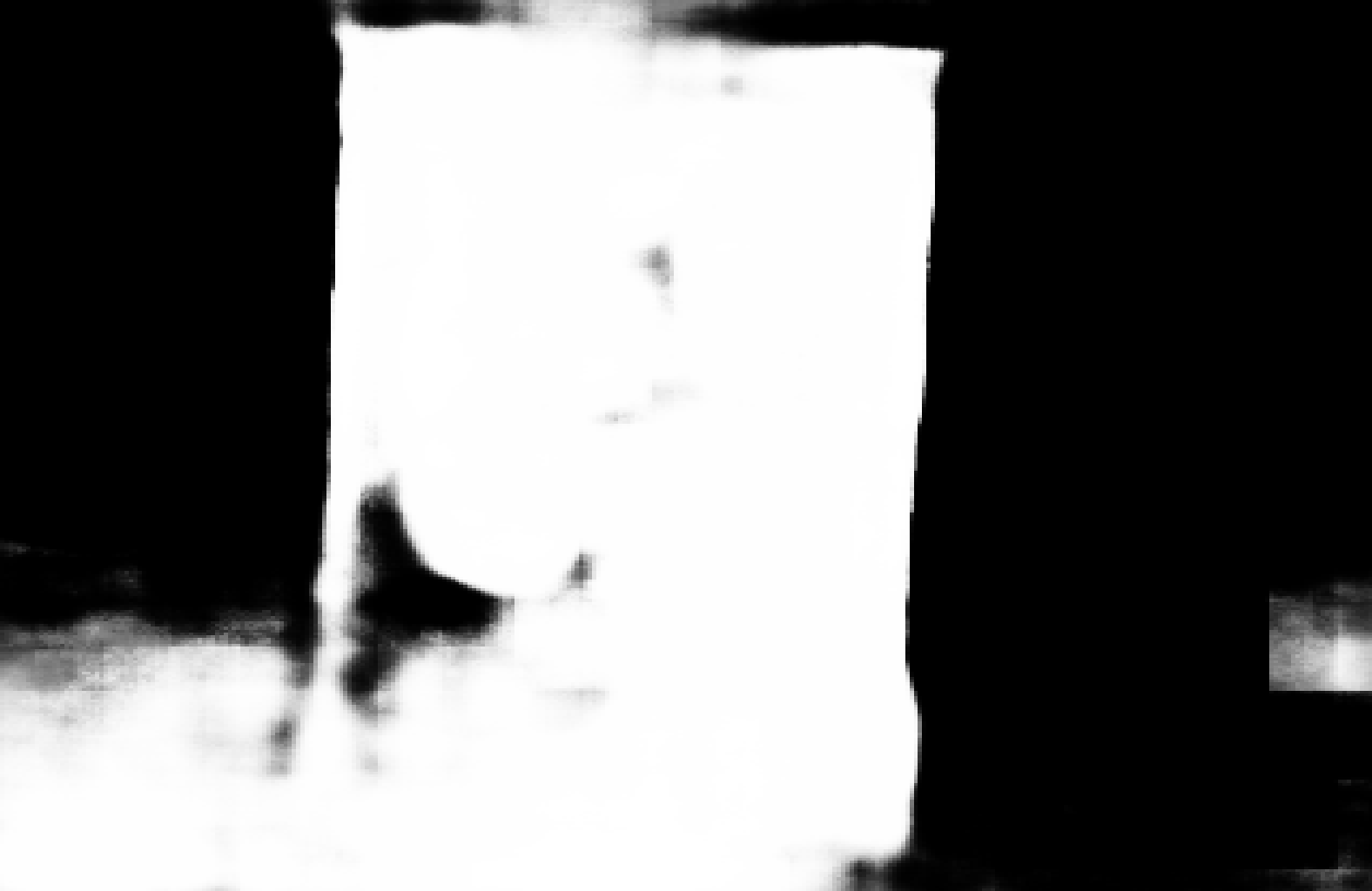}
	\end{subfigure}
    \begin{subfigure}{0.12\linewidth}
		\centering
		\includegraphics[width=\linewidth]{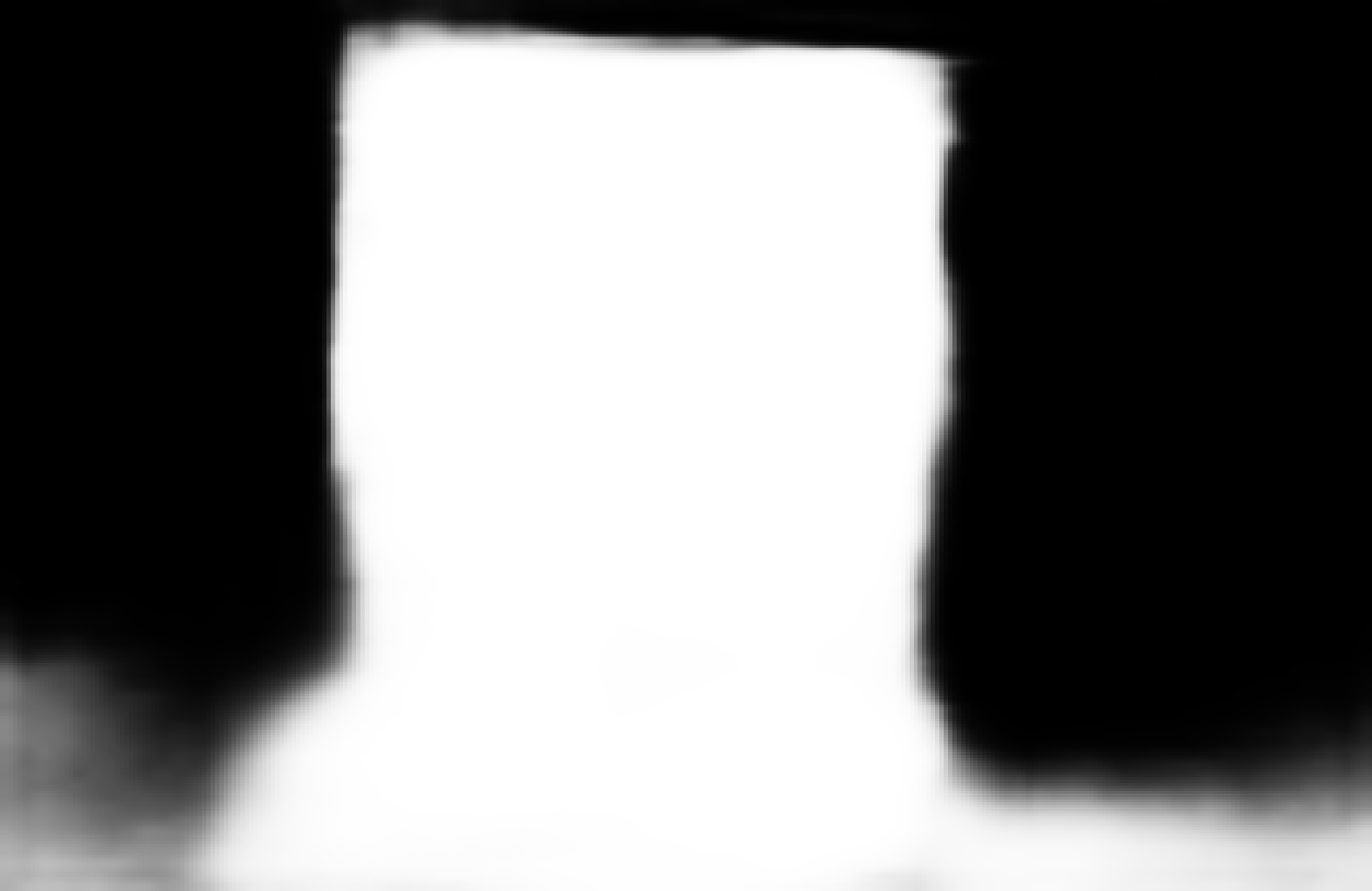}
	\end{subfigure}
    \begin{subfigure}{0.12\linewidth}
		\centering
		\includegraphics[width=\linewidth]{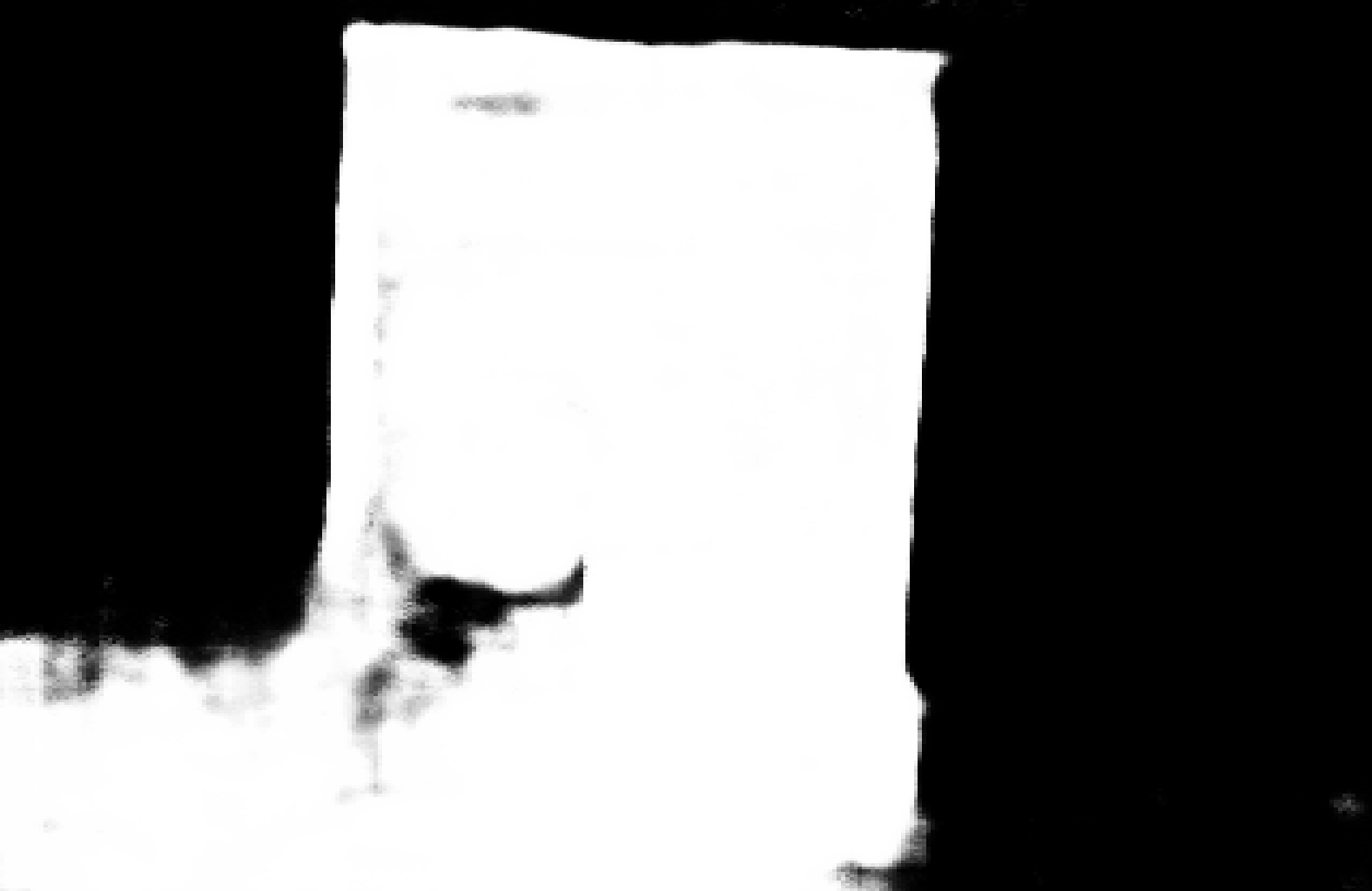}
	\end{subfigure}
    \begin{subfigure}{0.12\linewidth}
		\centering
		\includegraphics[width=\linewidth]{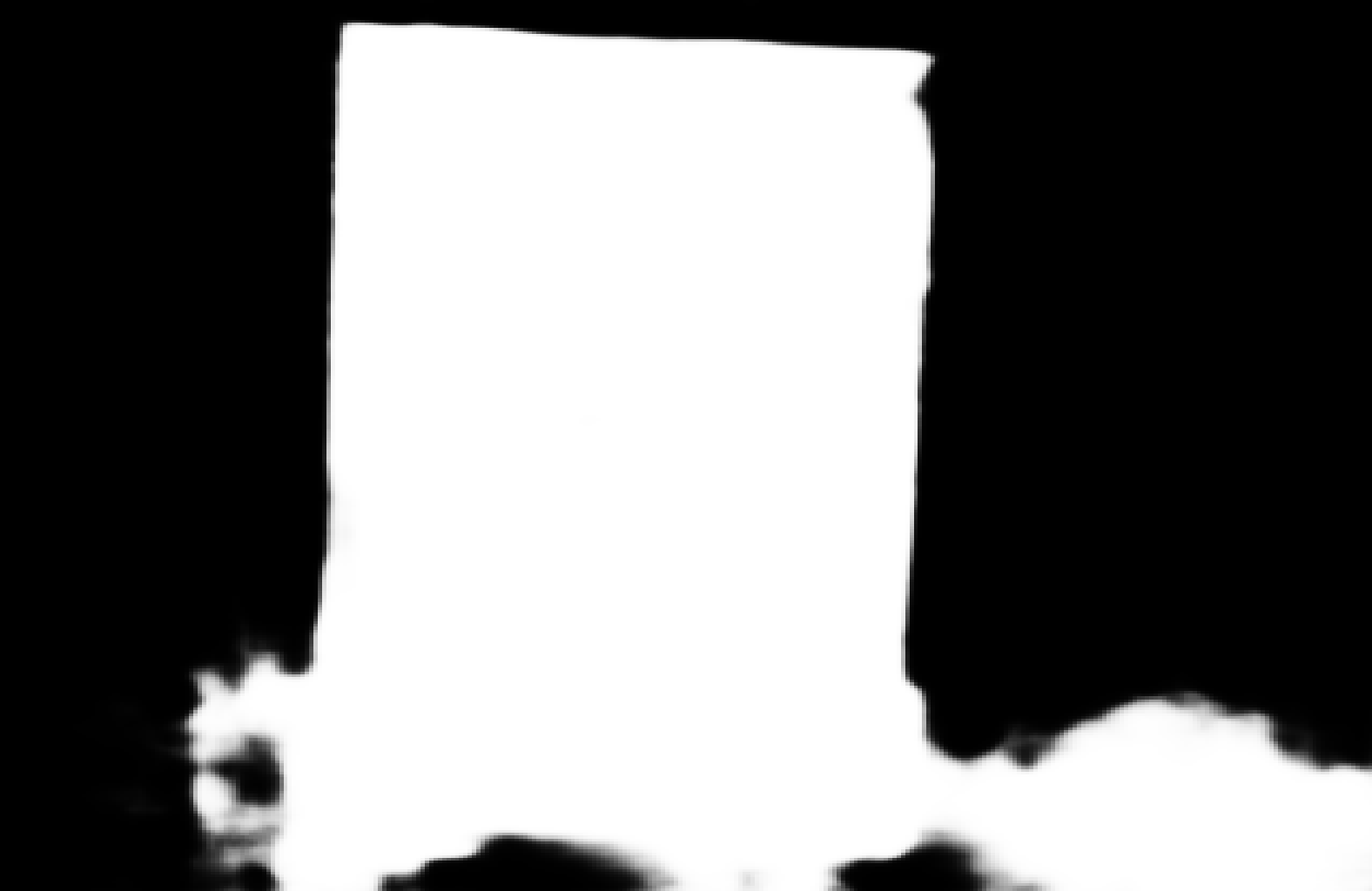}
	\end{subfigure}
    \begin{subfigure}{0.12\linewidth}
		\centering
		\includegraphics[width=\linewidth]{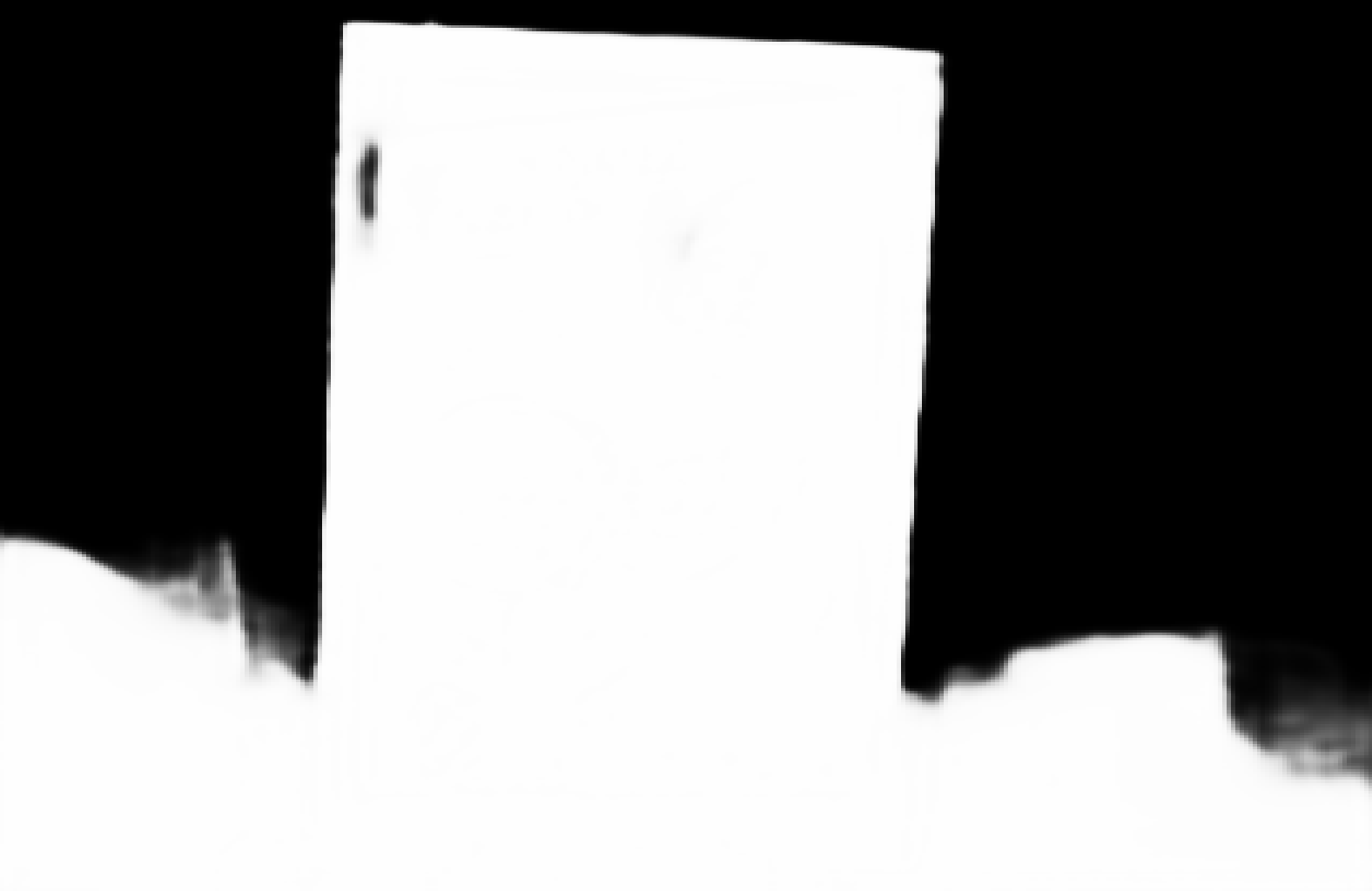}
	\end{subfigure}
    \begin{subfigure}{0.12\linewidth}
		\centering
		\includegraphics[width=\linewidth]{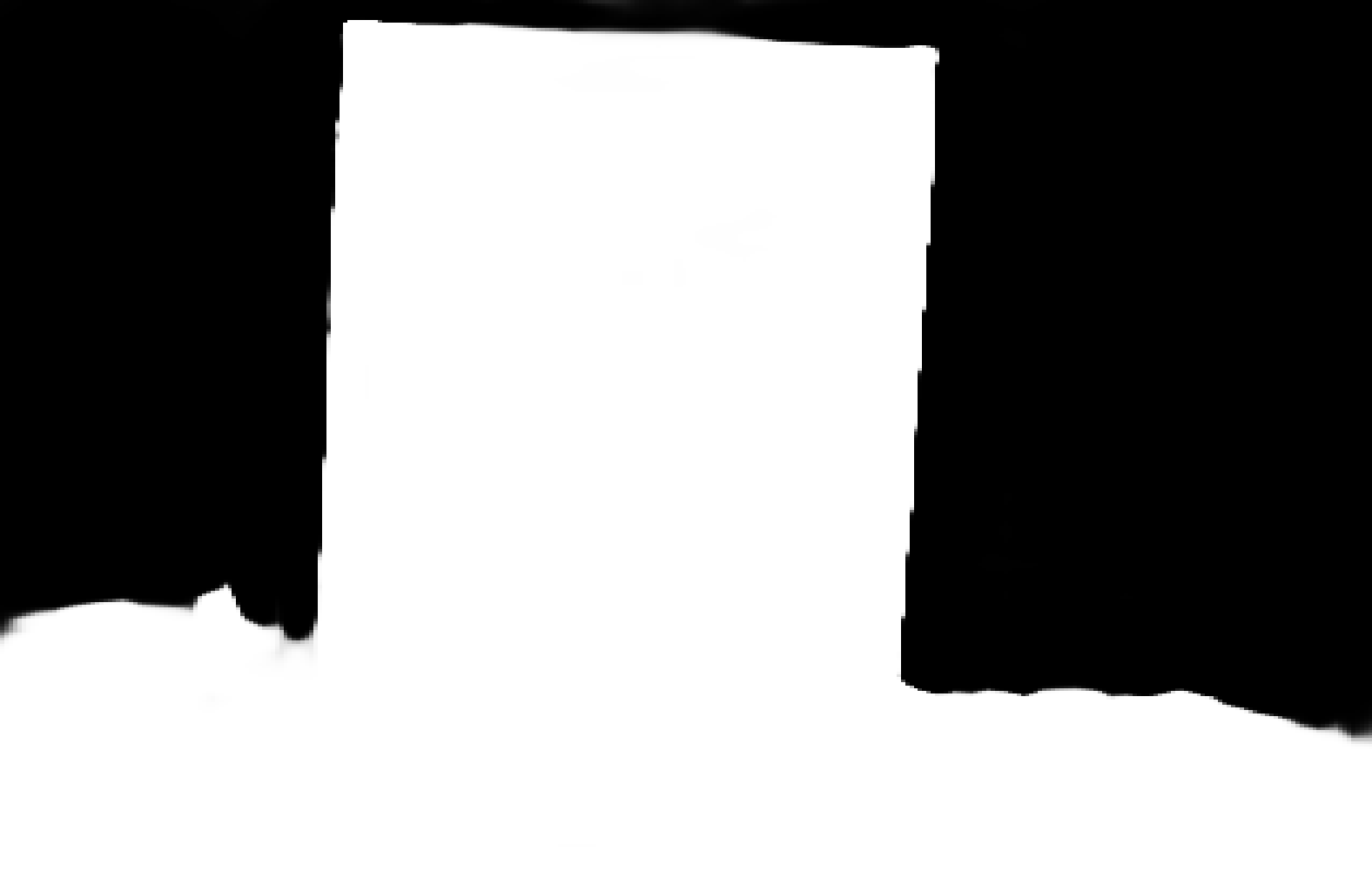}
	\end{subfigure}
    \begin{subfigure}{0.12\linewidth}
		\centering
		\includegraphics[width=\linewidth]{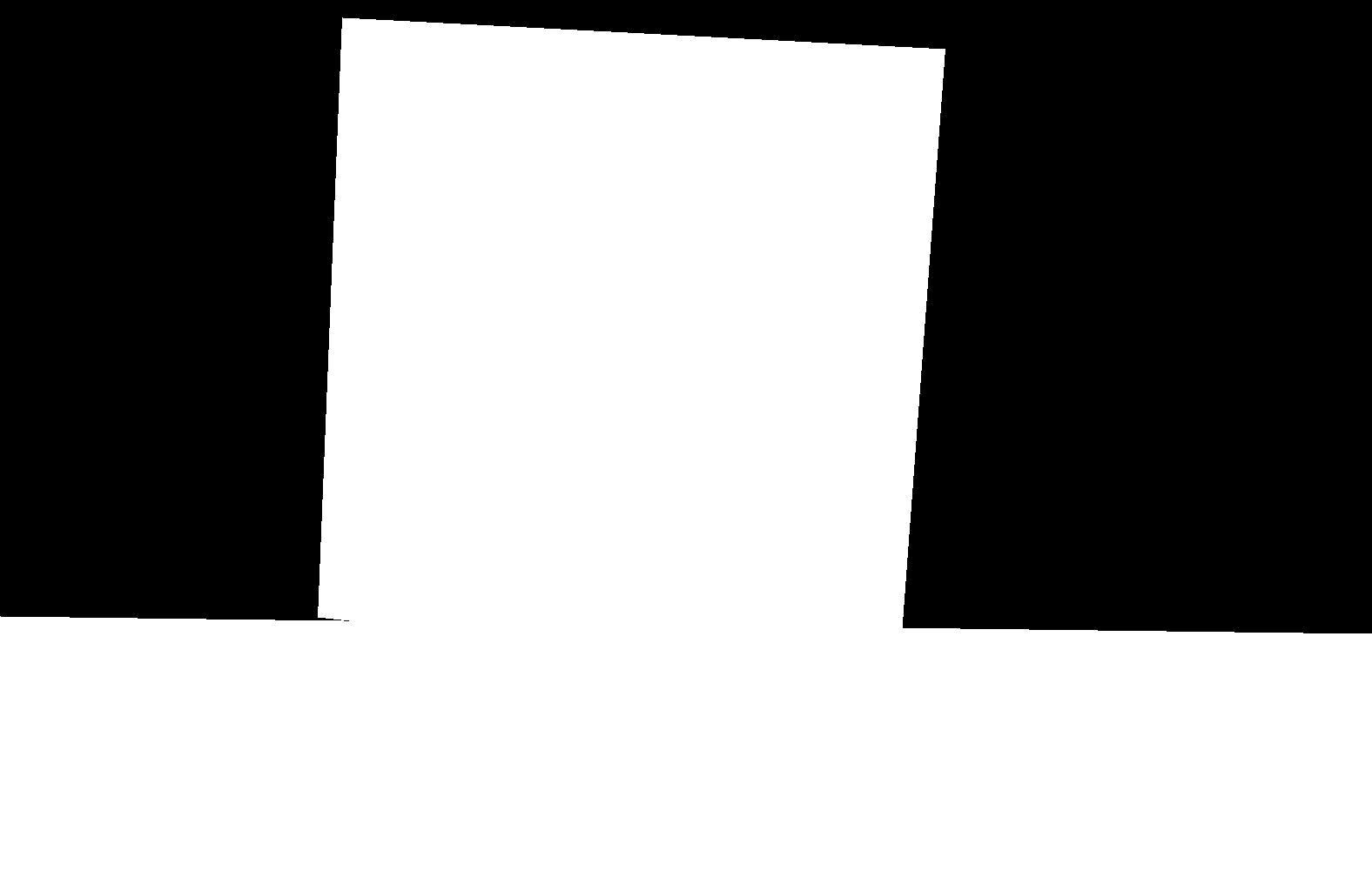}
	\end{subfigure}
 
    \begin{subfigure}{0.12\linewidth}
		\centering
		\includegraphics[width=\linewidth]{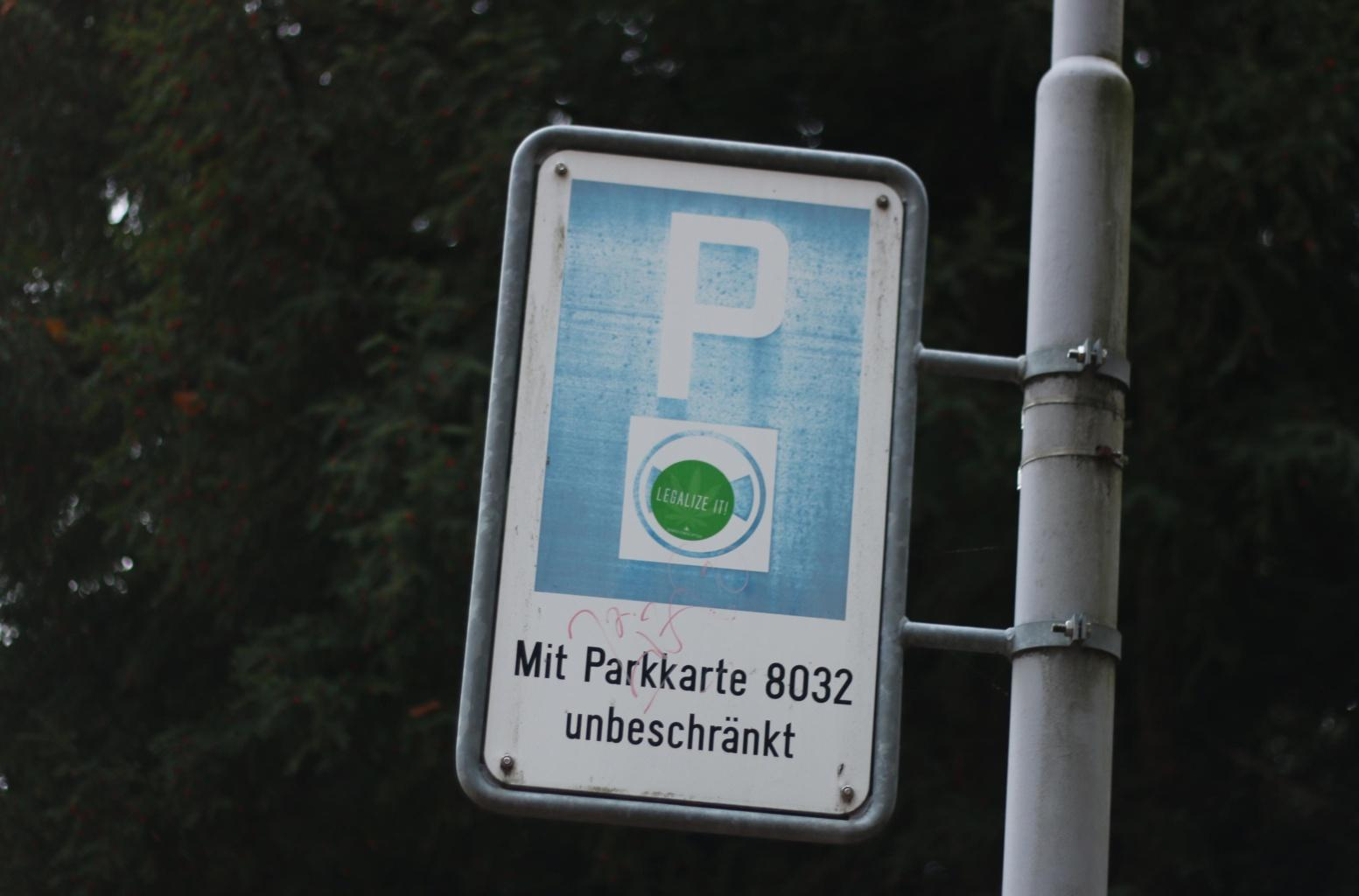}
		\caption{Images}
	\end{subfigure}
    \begin{subfigure}{0.12\linewidth}
		\centering
		\includegraphics[width=\linewidth]{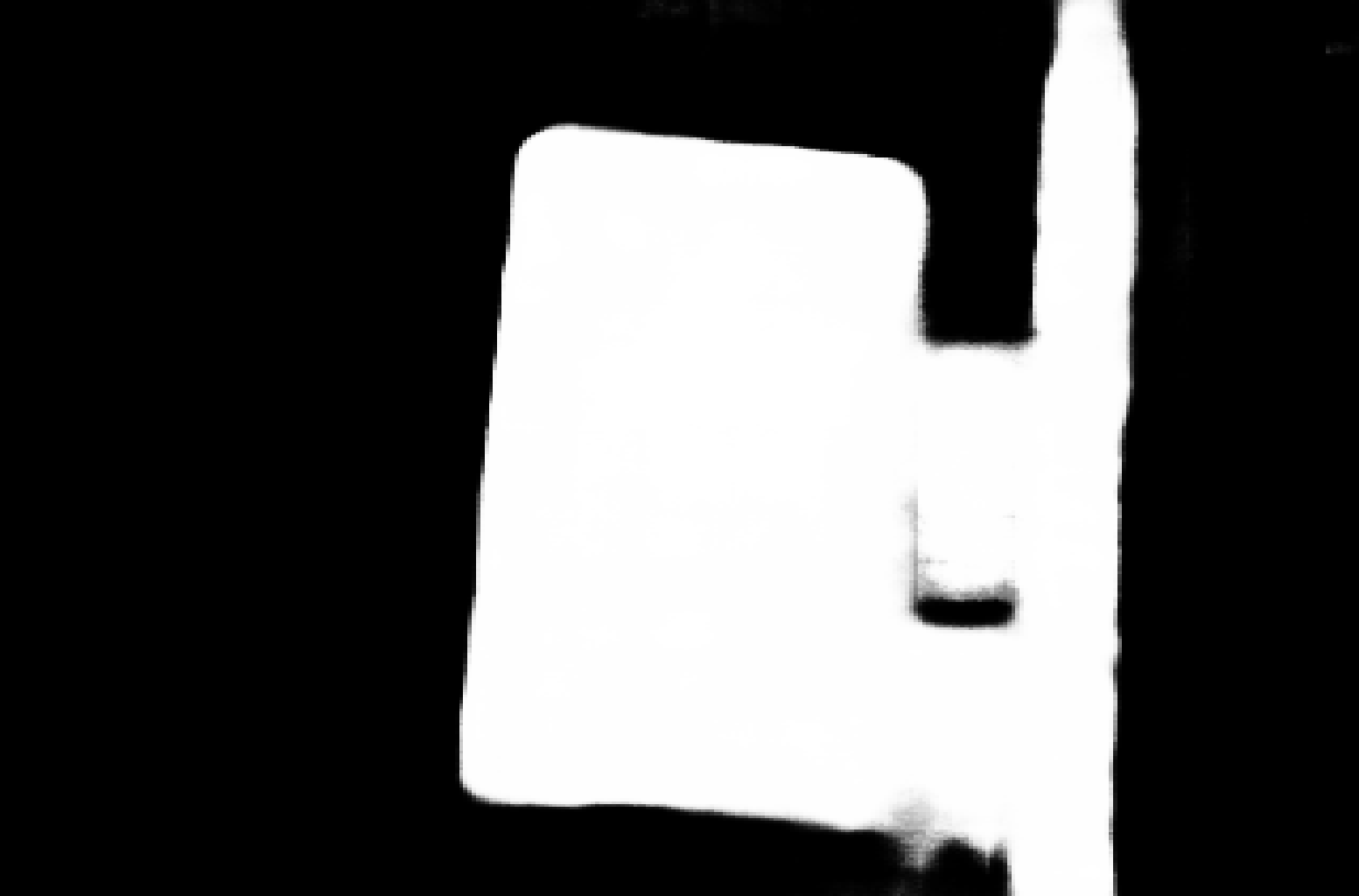}
		\caption{CENet \cite{zhao2019enhancing}}
	\end{subfigure}
    \begin{subfigure}{0.12\linewidth}
		\centering
		\includegraphics[width=\linewidth]{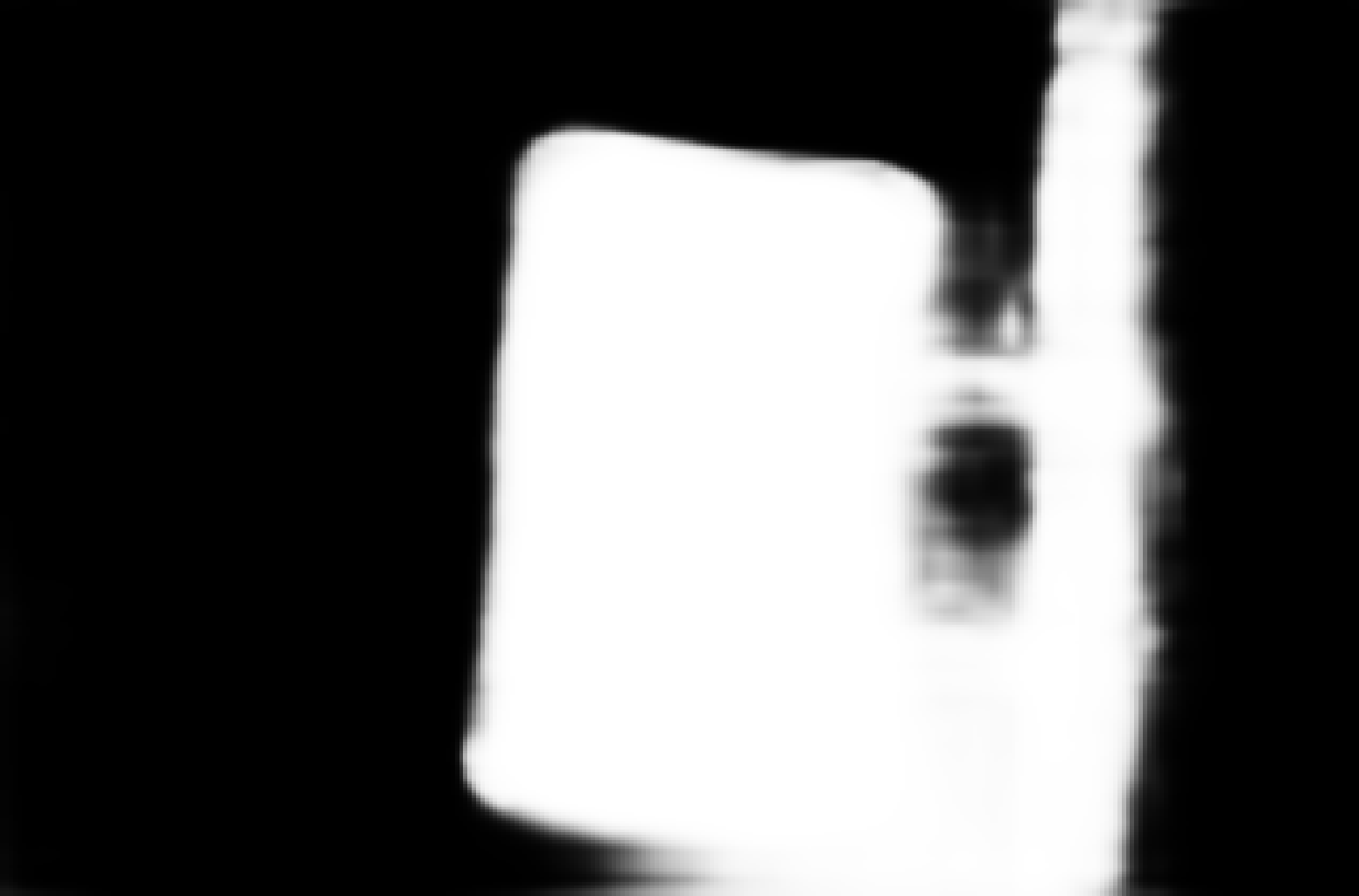}
		\caption{BR2Net \cite{tang2020br}}
	\end{subfigure}
    \begin{subfigure}{0.12\linewidth}
		\centering
		\includegraphics[width=\linewidth]{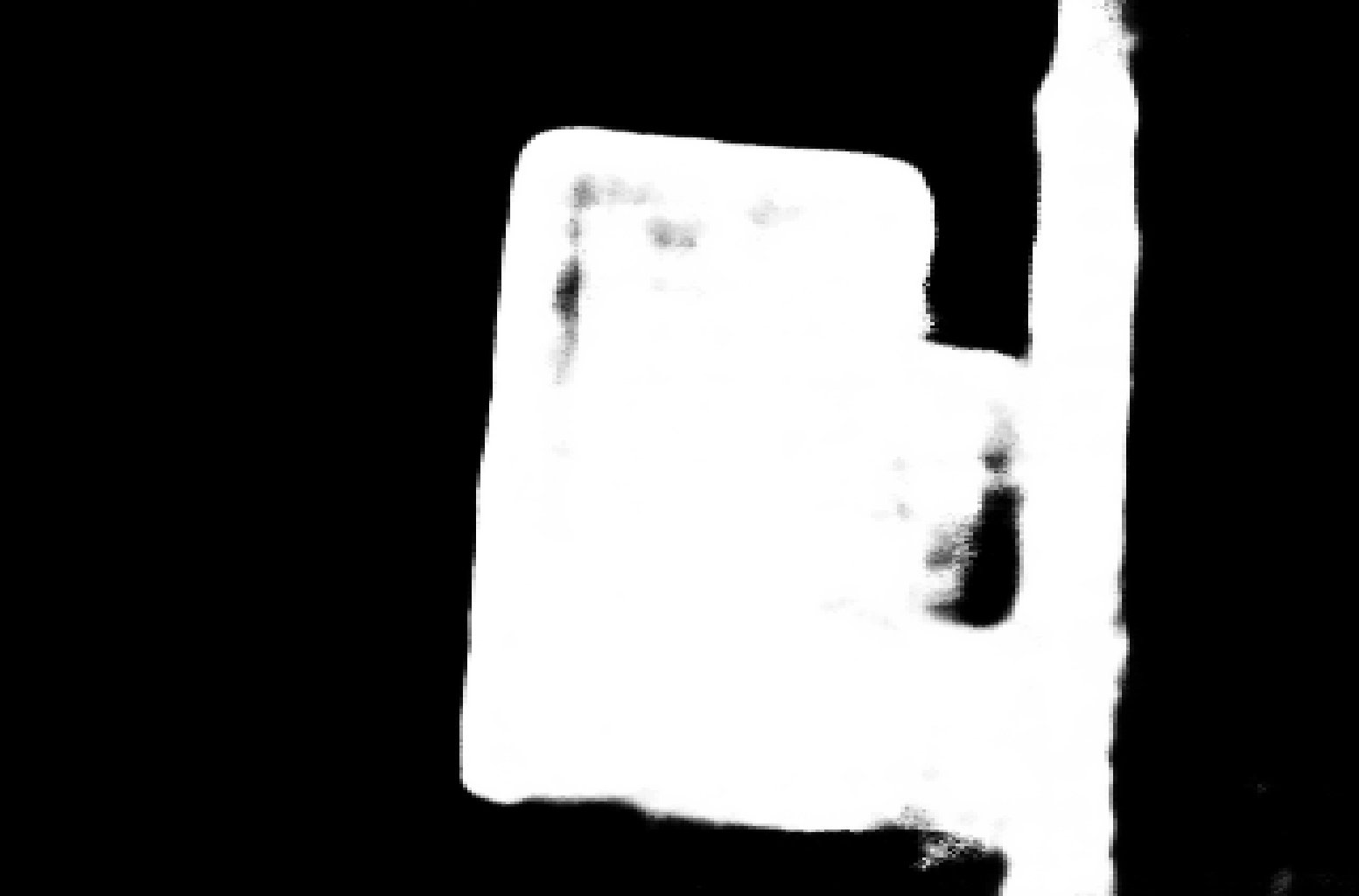}
		\caption{AENet \cite{zhao2021defocus}}
	\end{subfigure}
    \begin{subfigure}{0.12\linewidth}
		\centering
		\includegraphics[width=\linewidth]{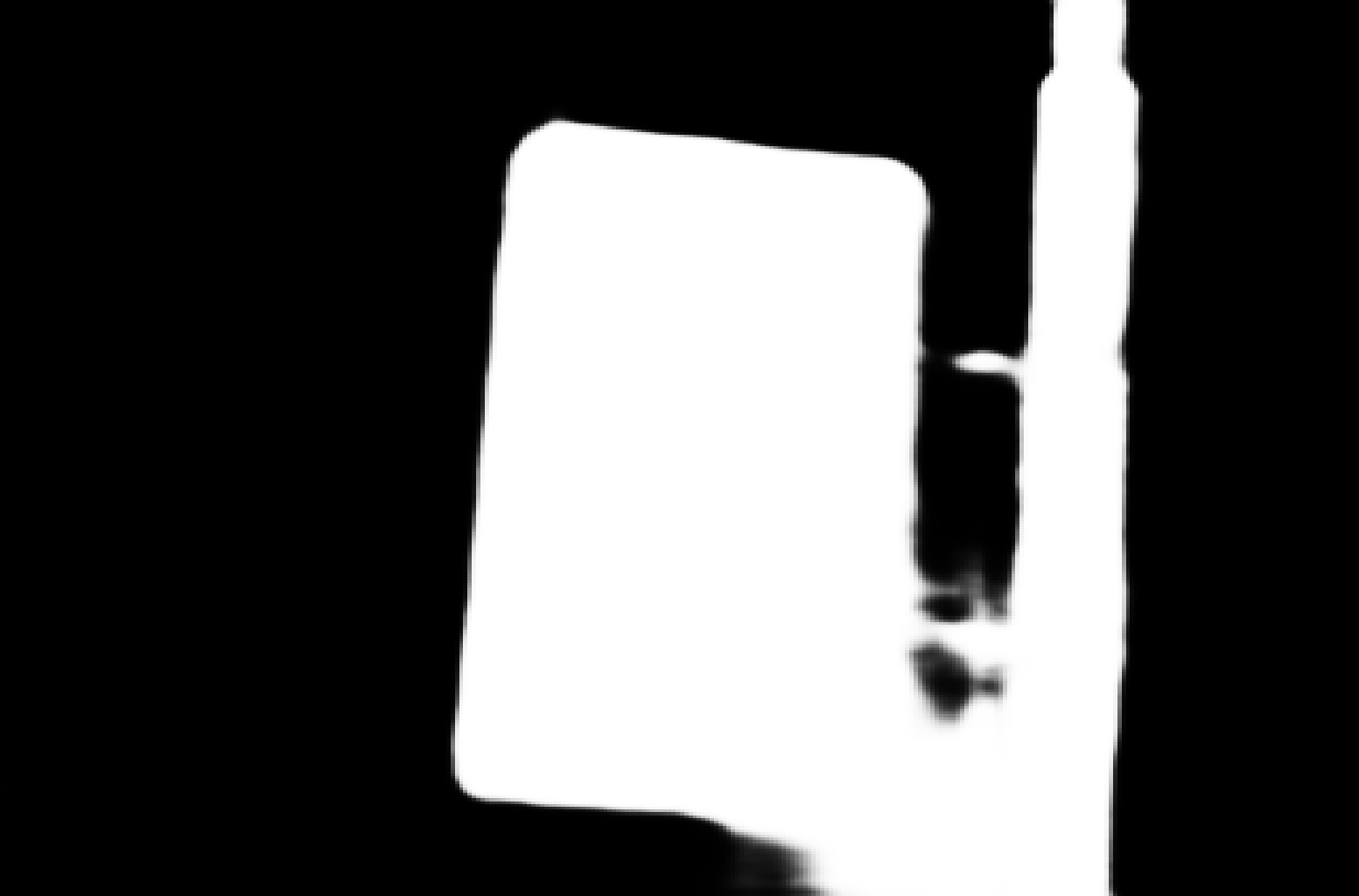}
		\caption{DD \cite{cun2020defocus}}
	\end{subfigure}
    \begin{subfigure}{0.12\linewidth}
		\centering
		\includegraphics[width=\linewidth]{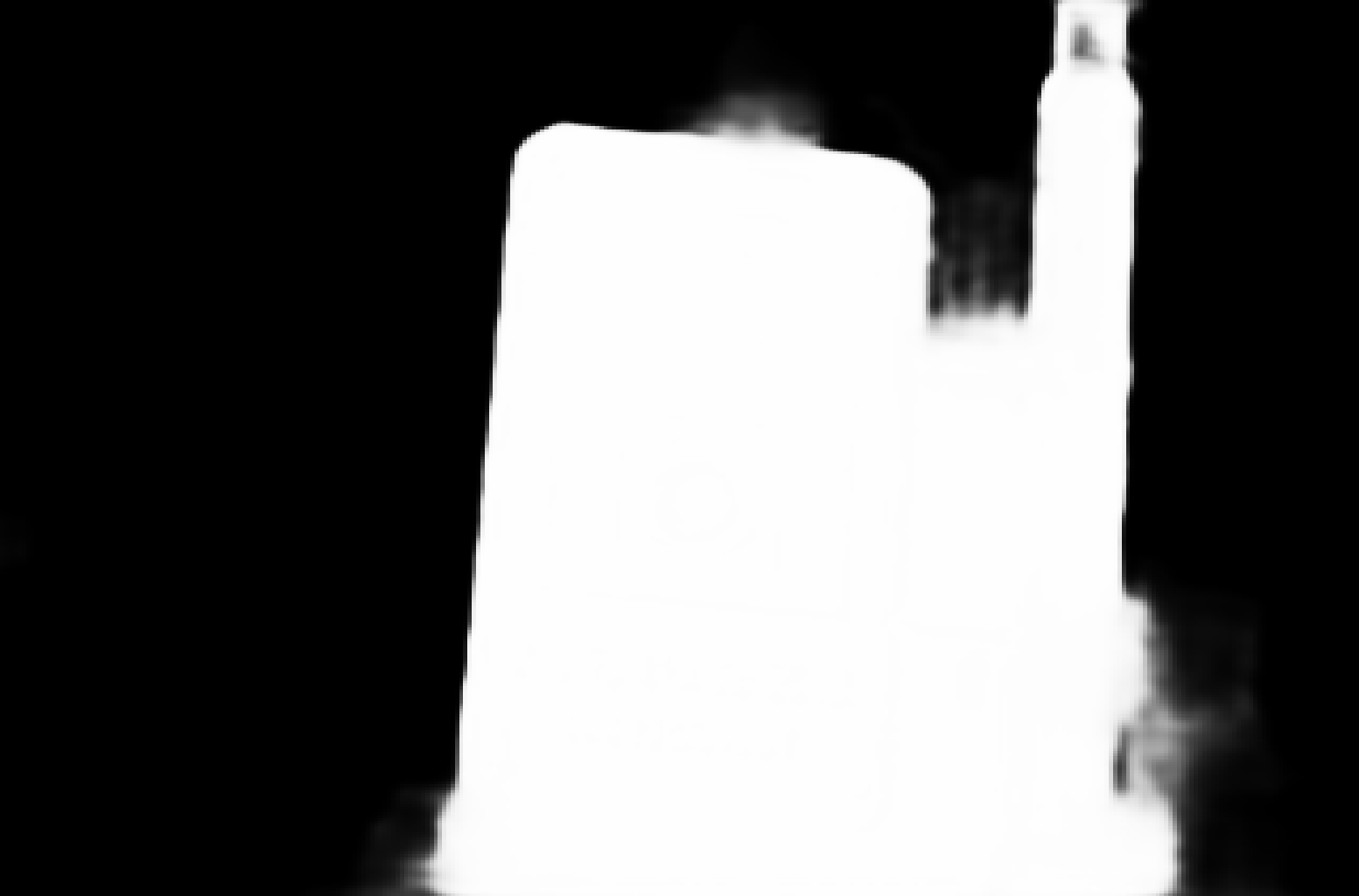}
		\caption{IS2CNet \cite{zhao2021image}}
	\end{subfigure}
    \begin{subfigure}{0.12\linewidth}
		\centering
		\includegraphics[width=\linewidth]{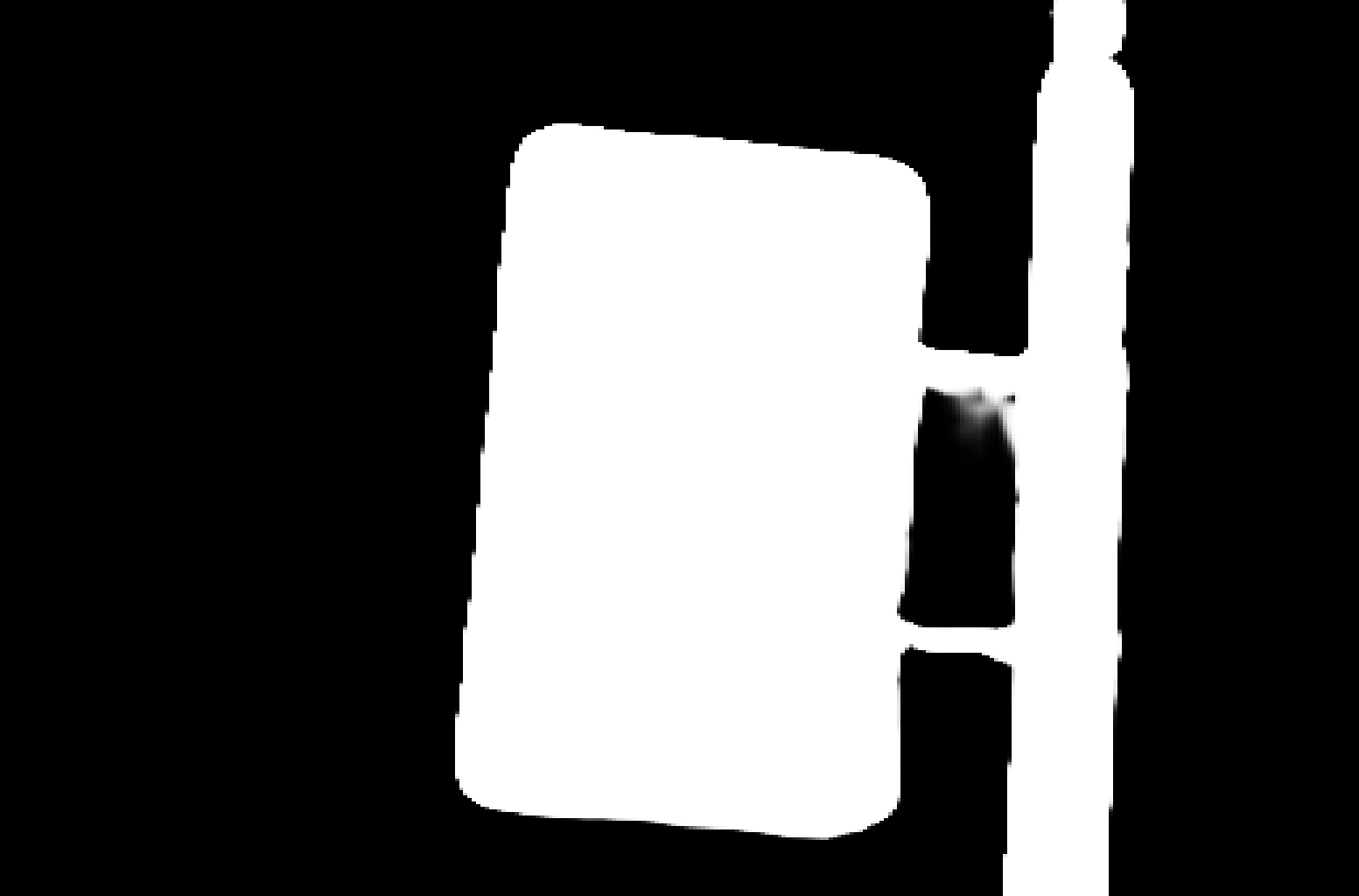}
		\caption{D-DFFNet}
	\end{subfigure}
    \begin{subfigure}{0.12\linewidth}
		\centering
		\includegraphics[width=\linewidth]{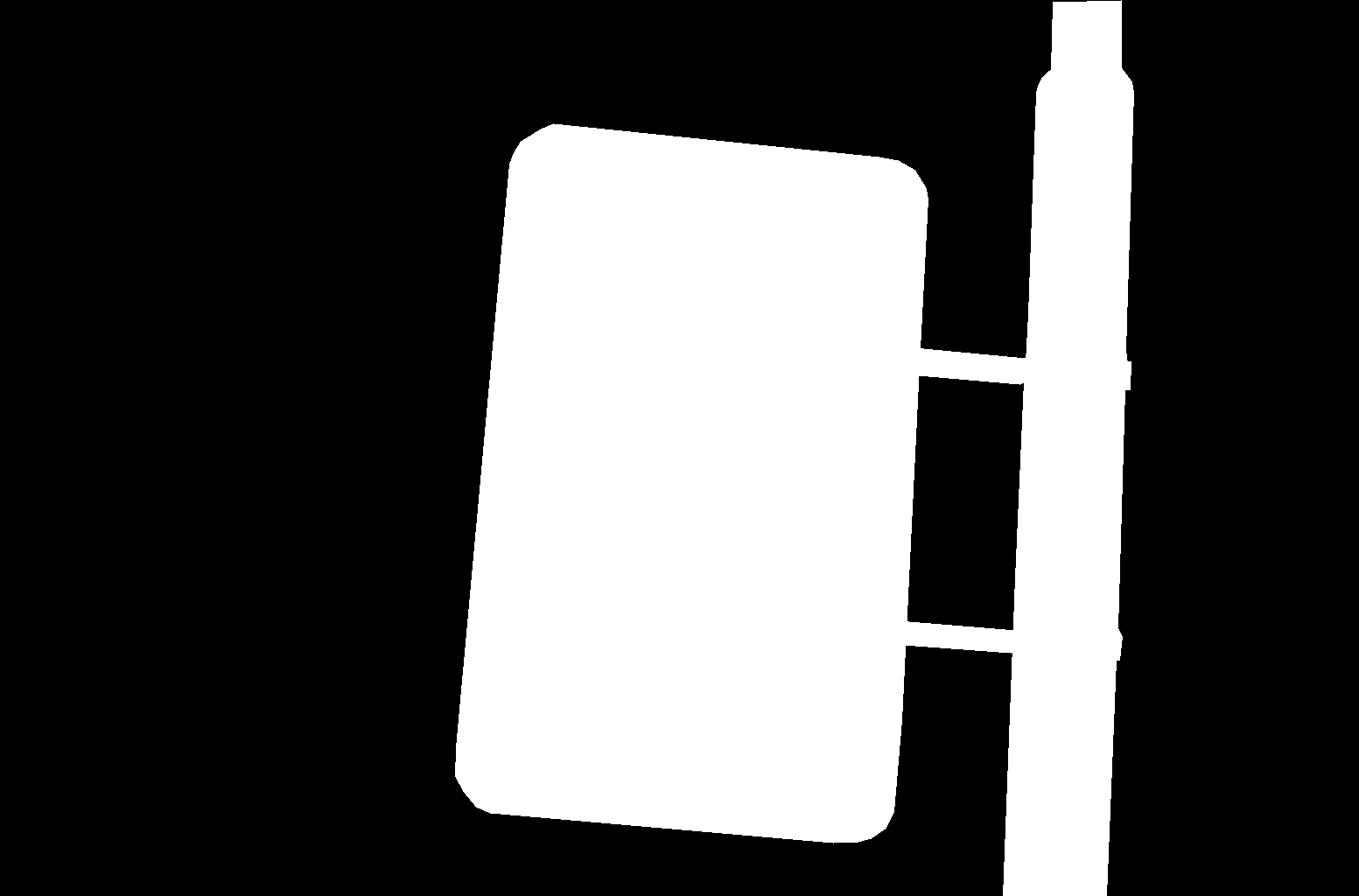}
		\caption{GTs}
	\end{subfigure}

\caption{Qualitative Comparison of Methods on EBD dataset.}
\label{compare on EBD}
\end{figure*}

\begin{figure*}[t]

	\begin{subfigure}{0.130\linewidth}
		\centering
		\includegraphics[width=\linewidth]{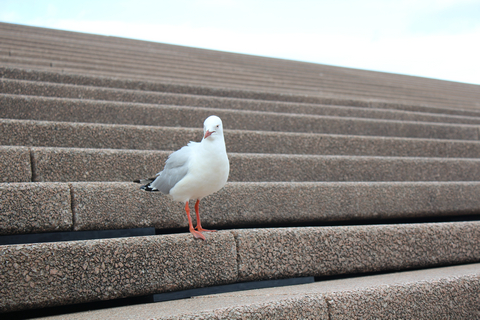}
	\end{subfigure}
	\centering
	\begin{subfigure}{0.130\linewidth}
		\centering
		\includegraphics[width=\linewidth]{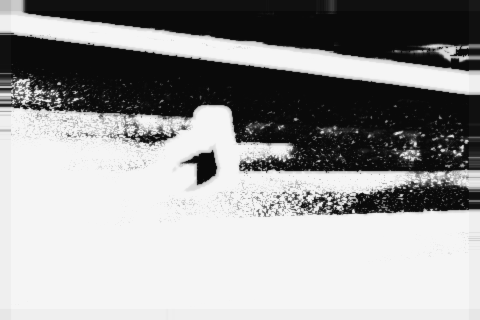}
	\end{subfigure}
	\centering
	\begin{subfigure}{0.130\linewidth}
		\centering
		\includegraphics[width=\linewidth]{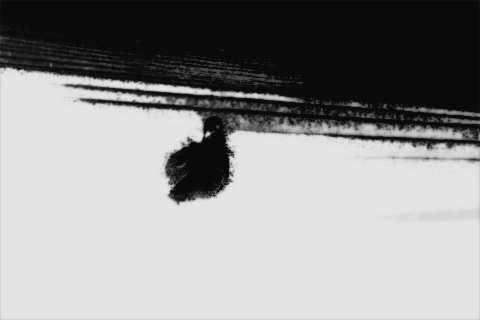}
	\end{subfigure}
	\begin{subfigure}{0.130\linewidth}
		\centering
		\includegraphics[width=\linewidth]{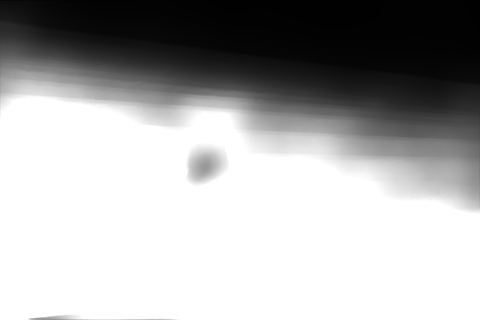}
	\end{subfigure}
	\begin{subfigure}{0.130\linewidth}
		\centering
		\includegraphics[width=\linewidth]{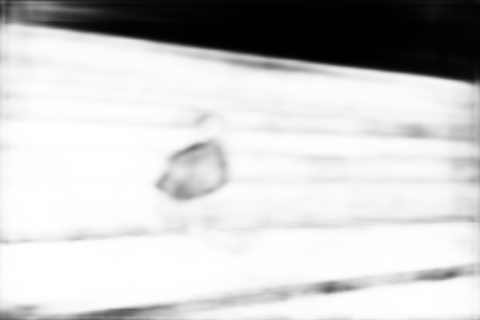}
	\end{subfigure}
	\begin{subfigure}{0.130\linewidth}
		\centering
		\includegraphics[width=\linewidth]{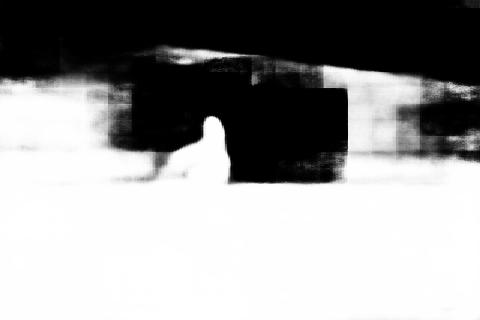}
	\end{subfigure}
	\begin{subfigure}{0.130\linewidth}
		\centering
		\includegraphics[width=\linewidth]{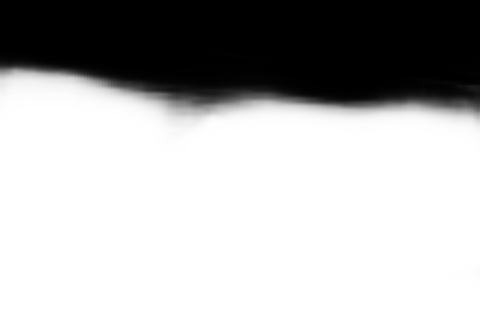}
	\end{subfigure}

	\begin{subfigure}{0.130\linewidth}
		\includegraphics[width=\linewidth]{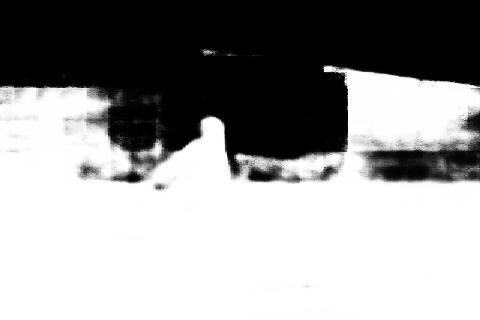}
	\end{subfigure}
	\begin{subfigure}{0.130\linewidth}
		\includegraphics[width=\linewidth]{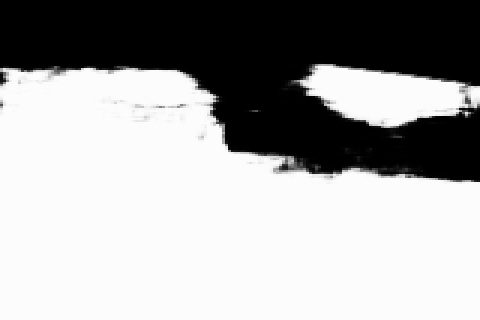}
	\end{subfigure}
	\begin{subfigure}{0.130\linewidth}
		\centering
		\includegraphics[width=\linewidth]{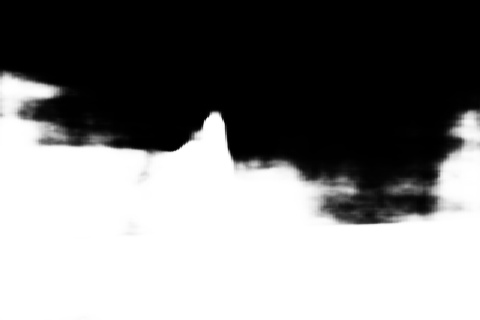}
	\end{subfigure}
	\begin{subfigure}{0.130\linewidth}
		\centering
		\includegraphics[width=\linewidth]{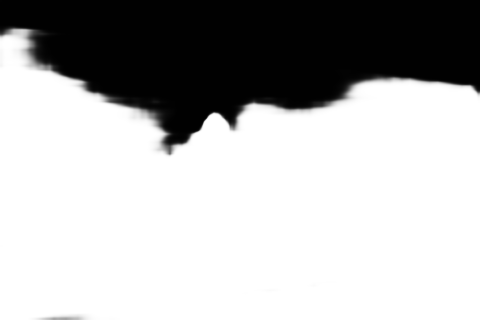}
	\end{subfigure}
	\begin{subfigure}{0.130\linewidth}
		\centering
		\includegraphics[width=\linewidth]{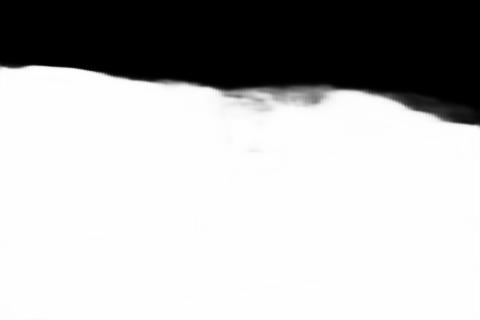}
	\end{subfigure}
	\begin{subfigure}{0.130\linewidth}
		\centering
		\includegraphics[width=\linewidth]{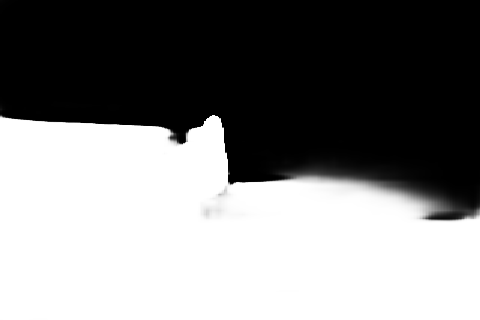}
	\end{subfigure}
	\begin{subfigure}{0.130\linewidth}
		\centering
		\includegraphics[width=\linewidth]{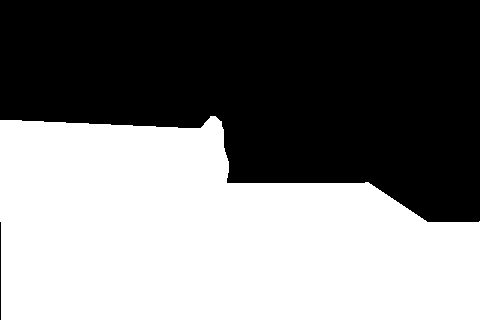}
	\end{subfigure}

	\begin{subfigure}{0.130\linewidth}
		\centering
		\includegraphics[width=\linewidth]{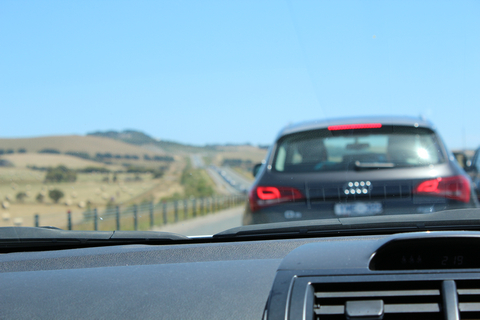}
	\end{subfigure}
	\centering
	\begin{subfigure}{0.130\linewidth}
		\centering
		\includegraphics[width=\linewidth]{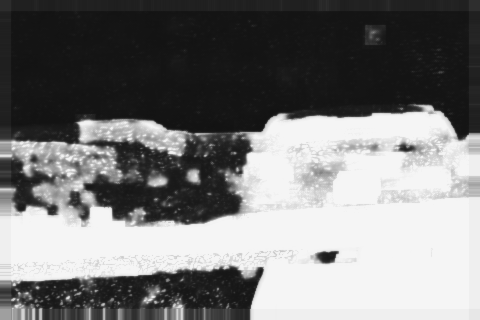}
	\end{subfigure}
	\centering
	\begin{subfigure}{0.130\linewidth}
		\centering
		\includegraphics[width=\linewidth]{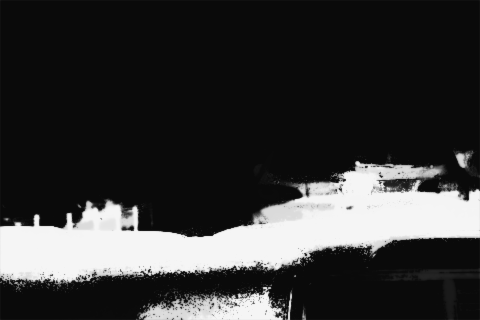}
	\end{subfigure}
	\begin{subfigure}{0.130\linewidth}
		\centering
		\includegraphics[width=\linewidth]{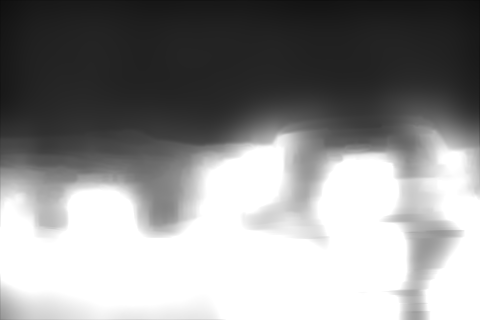}
	\end{subfigure}
	\begin{subfigure}{0.130\linewidth}
		\centering
		\includegraphics[width=\linewidth]{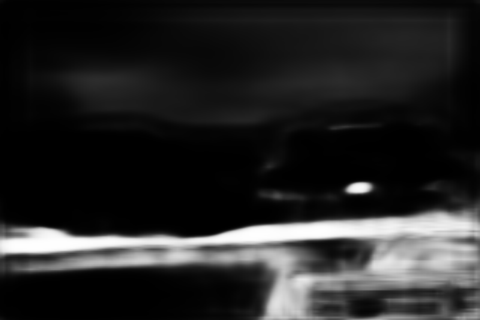}
	\end{subfigure}
	\begin{subfigure}{0.130\linewidth}
		\centering
		\includegraphics[width=\linewidth]{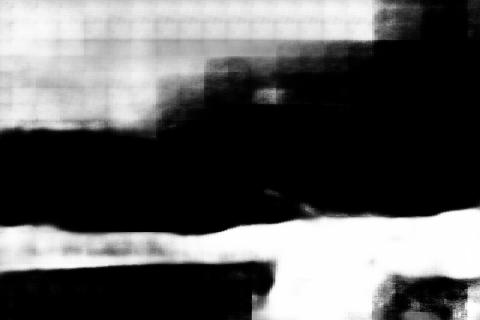}
	\end{subfigure}
	\begin{subfigure}{0.130\linewidth}
		\centering
		\includegraphics[width=\linewidth]{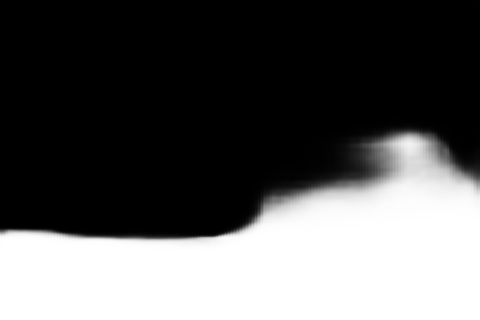}
	\end{subfigure}

	\begin{subfigure}{0.130\linewidth}
		\includegraphics[width=\linewidth]{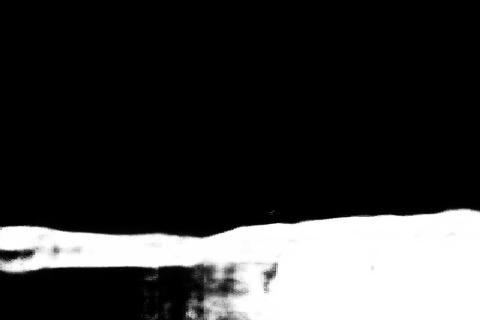}
	\end{subfigure}
	\begin{subfigure}{0.130\linewidth}
		\includegraphics[width=\linewidth]{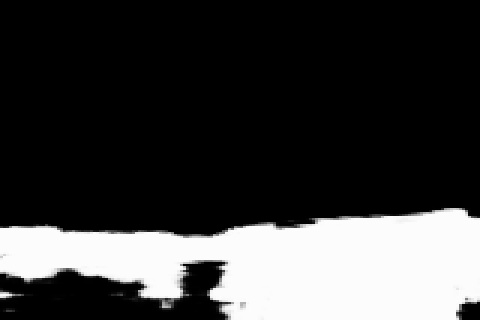}
	\end{subfigure}
	\begin{subfigure}{0.130\linewidth}
		\centering
		\includegraphics[width=\linewidth]{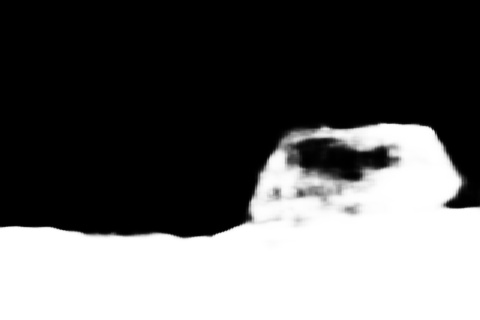}
	\end{subfigure}
	\begin{subfigure}{0.130\linewidth}
		\centering
		\includegraphics[width=\linewidth]{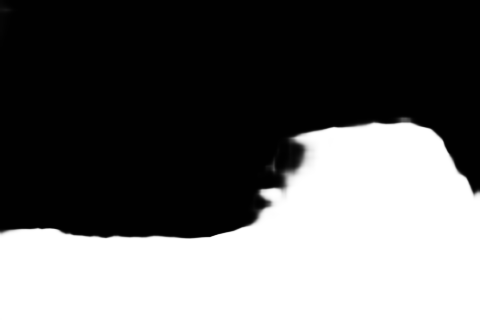}
	\end{subfigure}
	\begin{subfigure}{0.130\linewidth}
		\centering
		\includegraphics[width=\linewidth]{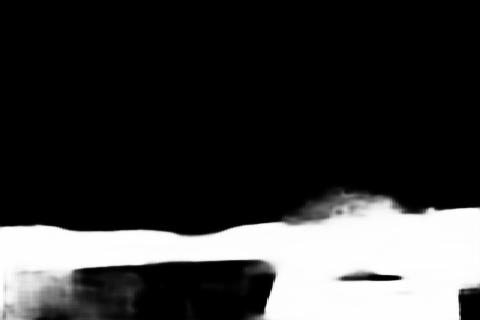}
	\end{subfigure}
	\begin{subfigure}{0.130\linewidth}
		\centering
		\includegraphics[width=\linewidth]{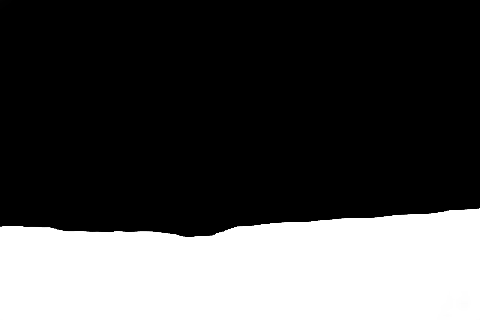}
	\end{subfigure}
	\begin{subfigure}{0.130\linewidth}
		\centering
		\includegraphics[width=\linewidth]{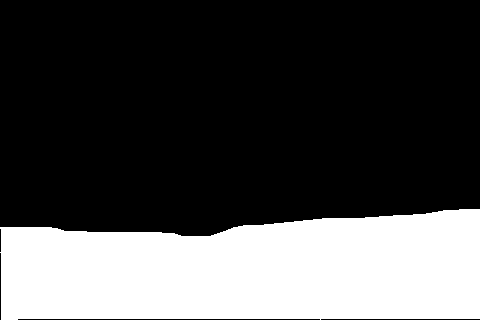}
	\end{subfigure}

	\begin{subfigure}{0.130\linewidth}
		\centering
		\includegraphics[width=\linewidth]{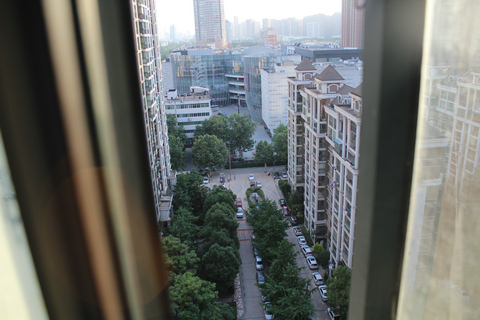}
	\end{subfigure}
	\centering
	\begin{subfigure}{0.130\linewidth}
		\centering
		\includegraphics[width=\linewidth]{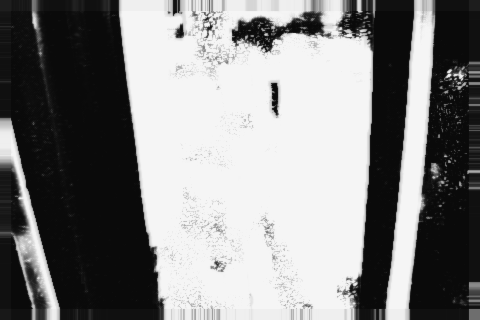}
	\end{subfigure}
	\centering
	\begin{subfigure}{0.130\linewidth}
		\centering
		\includegraphics[width=\linewidth]{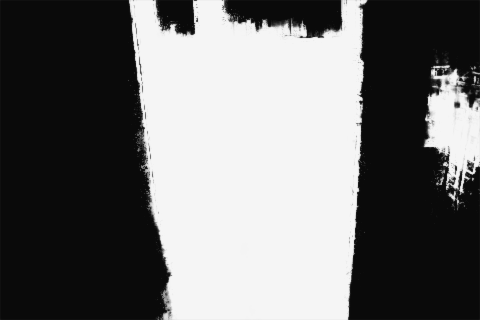}
	\end{subfigure}
	\begin{subfigure}{0.130\linewidth}
		\centering
		\includegraphics[width=\linewidth]{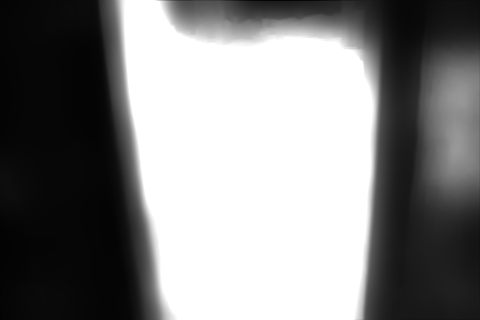}
	\end{subfigure}
	\begin{subfigure}{0.130\linewidth}
		\centering
		\includegraphics[width=\linewidth]{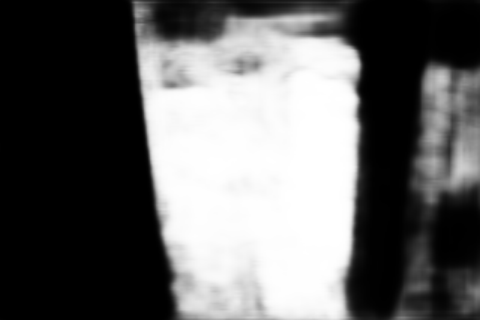}
	\end{subfigure}
	\begin{subfigure}{0.130\linewidth}
		\centering
		\includegraphics[width=\linewidth]{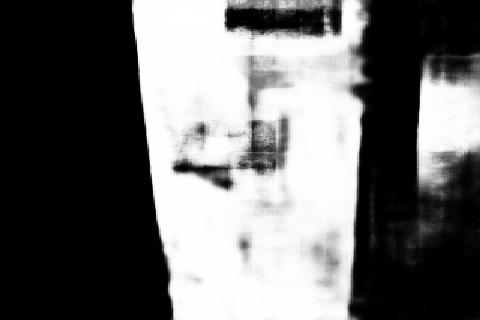}
	\end{subfigure}
	\begin{subfigure}{0.130\linewidth}
		\centering
		\includegraphics[width=\linewidth]{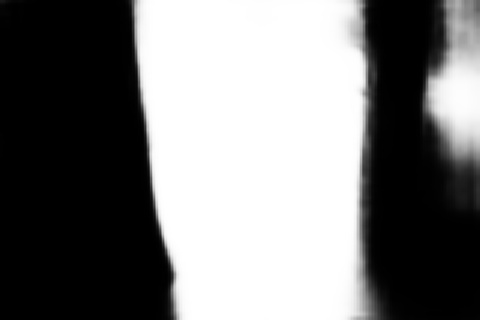}
	\end{subfigure}

	\begin{subfigure}{0.130\linewidth}
		\includegraphics[width=\linewidth]{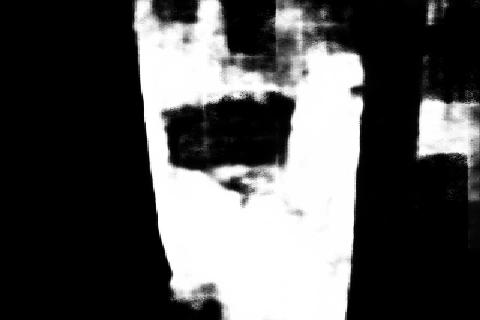}
	\end{subfigure}
	\begin{subfigure}{0.130\linewidth}
		\includegraphics[width=\linewidth]{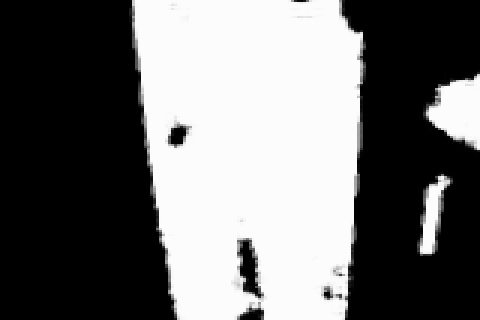}
	\end{subfigure}
	\begin{subfigure}{0.130\linewidth}
		\centering
		\includegraphics[width=\linewidth]{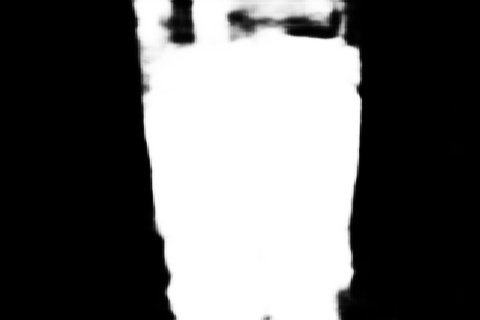}
	\end{subfigure}
	\begin{subfigure}{0.130\linewidth}
		\centering
		\includegraphics[width=\linewidth]{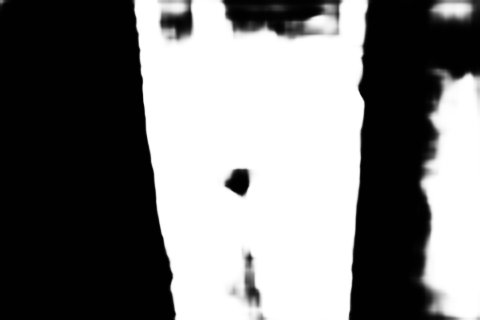}
	\end{subfigure}
	\begin{subfigure}{0.130\linewidth}
		\centering
		\includegraphics[width=\linewidth]{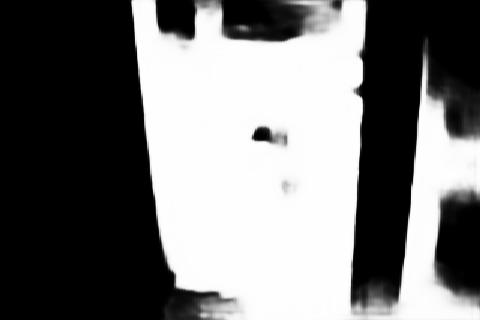}
	\end{subfigure}
	\begin{subfigure}{0.130\linewidth}
		\centering
		\includegraphics[width=\linewidth]{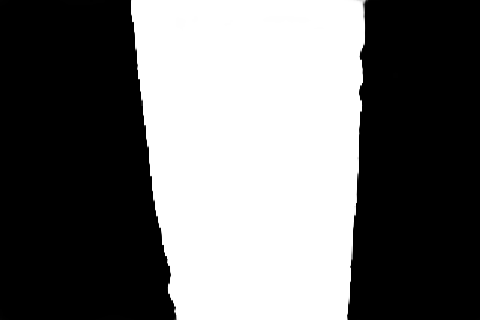}
	\end{subfigure}
	\begin{subfigure}{0.130\linewidth}
		\centering
		\includegraphics[width=\linewidth]{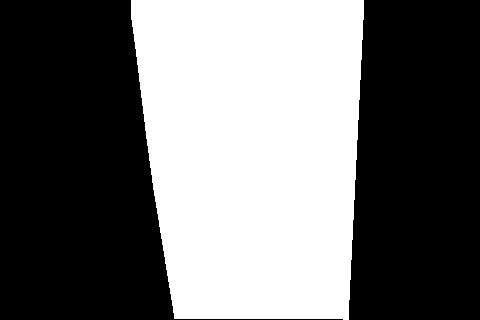}
	\end{subfigure}
	
	\begin{subfigure}{0.130\linewidth}
		\centering
		\includegraphics[width=\linewidth]{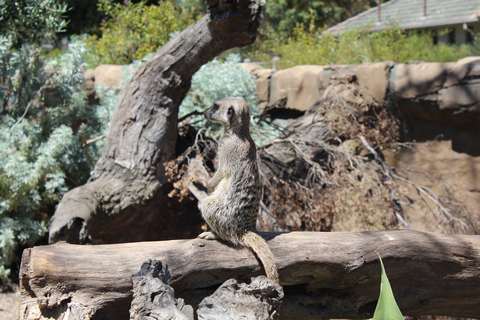}
	\end{subfigure}
	\centering
	\begin{subfigure}{0.130\linewidth}
		\centering
		\includegraphics[width=\linewidth]{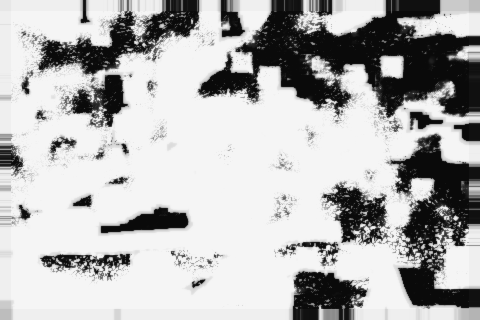}
	\end{subfigure}
	\centering
	\begin{subfigure}{0.130\linewidth}
		\centering
		\includegraphics[width=\linewidth]{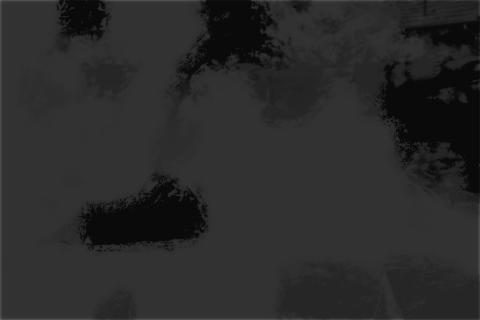}
	\end{subfigure}
	\begin{subfigure}{0.130\linewidth}
		\centering
		\includegraphics[width=\linewidth]{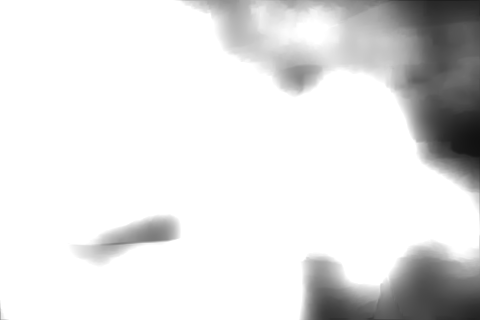}
	\end{subfigure}
	\begin{subfigure}{0.130\linewidth}
		\centering
		\includegraphics[width=\linewidth]{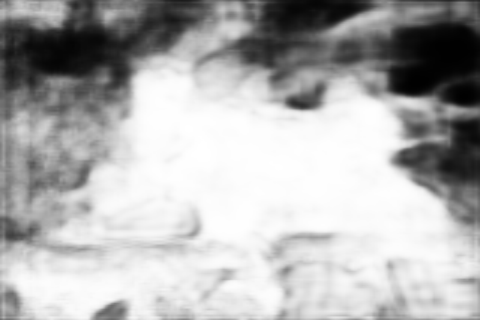}
	\end{subfigure}
	\begin{subfigure}{0.130\linewidth}
		\centering
		\includegraphics[width=\linewidth]{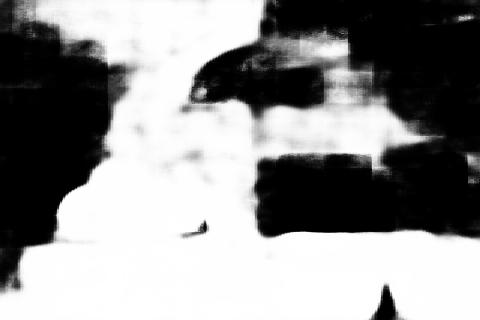}
	\end{subfigure}
	\begin{subfigure}{0.130\linewidth}
		\centering
		\includegraphics[width=\linewidth]{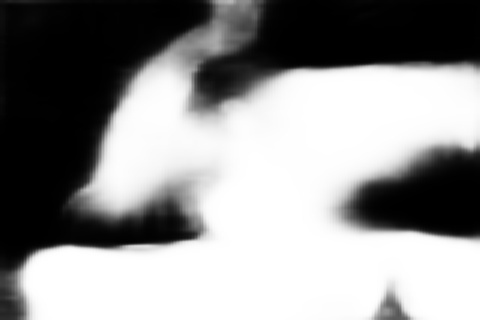}
	\end{subfigure}

	\begin{subfigure}{0.130\linewidth}
		\includegraphics[width=\linewidth]{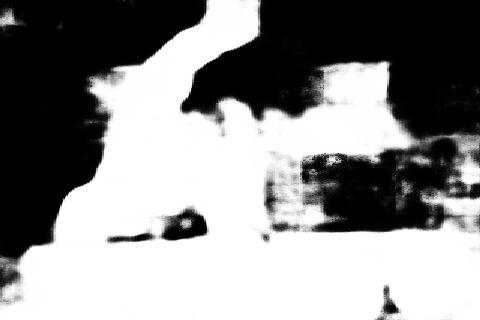}
	\end{subfigure}
	\begin{subfigure}{0.130\linewidth}
		\includegraphics[width=\linewidth]{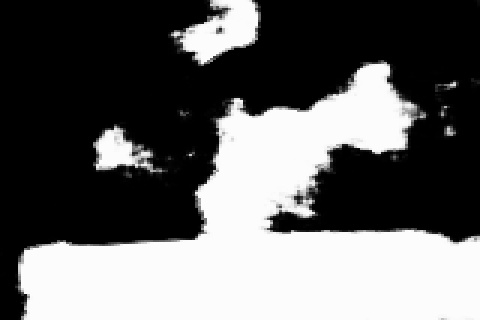}
	\end{subfigure}
	\begin{subfigure}{0.130\linewidth}
		\centering
		\includegraphics[width=\linewidth]{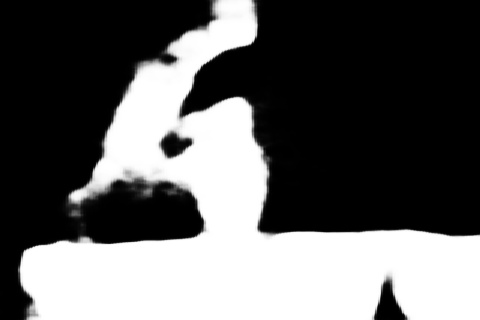}
	\end{subfigure}
	\begin{subfigure}{0.130\linewidth}
		\centering
		\includegraphics[width=\linewidth]{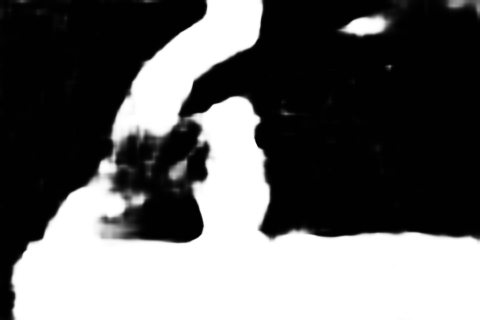}
	\end{subfigure}
	\begin{subfigure}{0.130\linewidth}
		\centering
		\includegraphics[width=\linewidth]{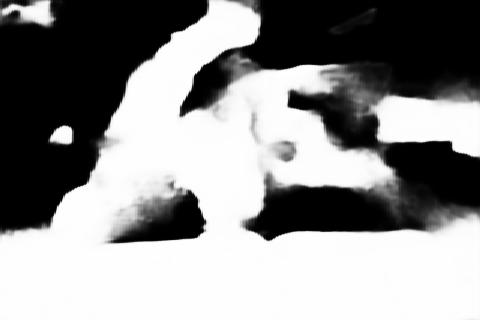}
	\end{subfigure}
	\begin{subfigure}{0.130\linewidth}
		\centering
		\includegraphics[width=\linewidth]{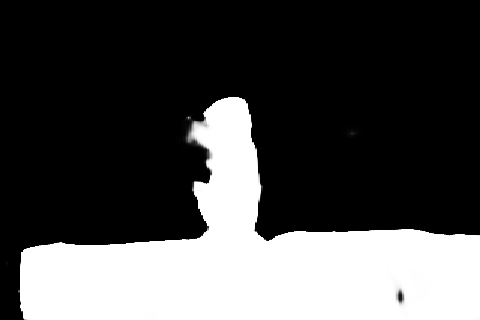}
	\end{subfigure}
	\begin{subfigure}{0.130\linewidth}
		\centering
		\includegraphics[width=\linewidth]{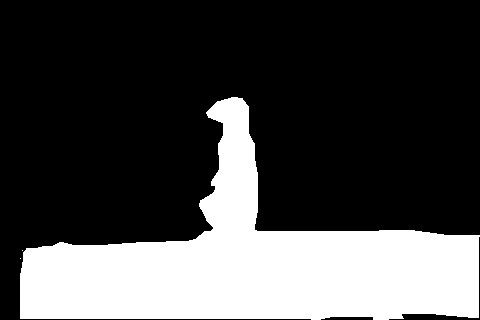}
	\end{subfigure}

	\begin{subfigure}{0.130\linewidth}
		\centering
		\includegraphics[width=\linewidth]{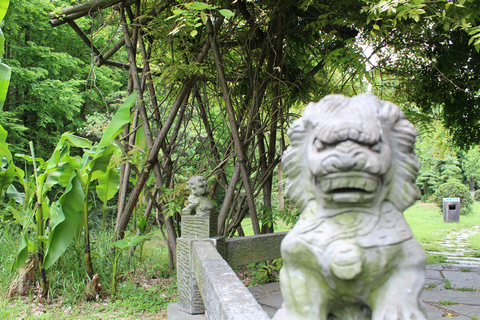}
	\end{subfigure}
	\begin{subfigure}{0.130\linewidth}
		\centering
		\includegraphics[width=\linewidth]{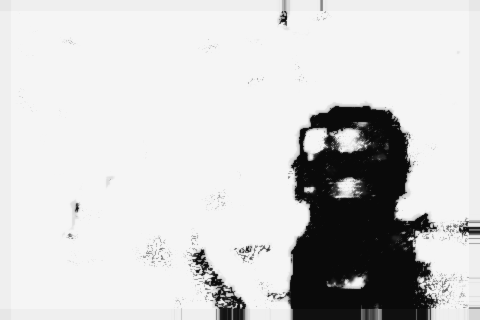}
	\end{subfigure}
	\begin{subfigure}{0.130\linewidth}
		\centering
		\includegraphics[width=\linewidth]{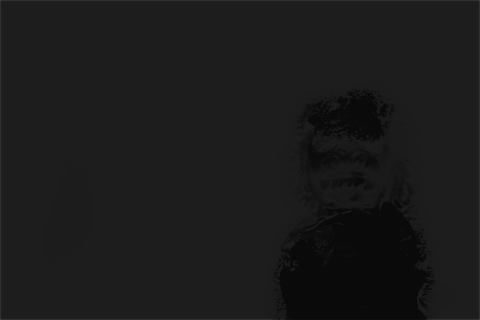}
	\end{subfigure}
	\begin{subfigure}{0.130\linewidth}
		\centering
		\includegraphics[width=\linewidth]{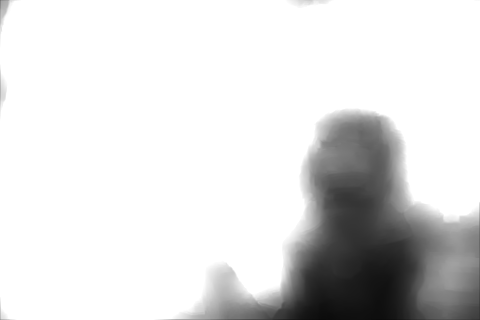}
	\end{subfigure}
	\begin{subfigure}{0.130\linewidth}
		\centering
		\includegraphics[width=\linewidth]{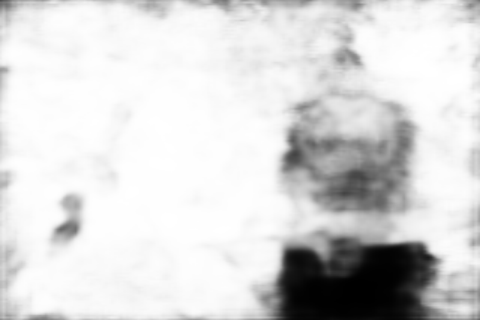}
	\end{subfigure}
	\begin{subfigure}{0.130\linewidth}
		\centering
		\includegraphics[width=\linewidth]{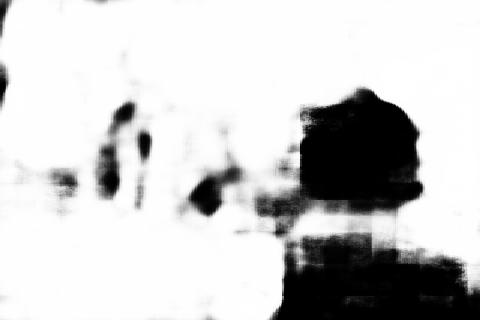}
	\end{subfigure}
	\begin{subfigure}{0.130\linewidth}
		\centering
		\includegraphics[width=\linewidth]{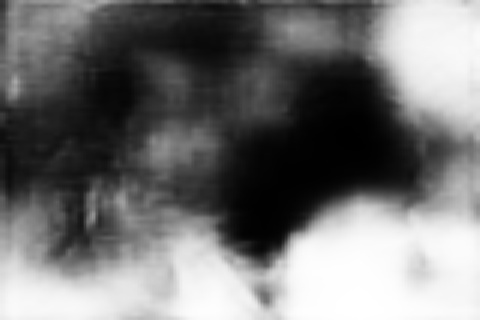}
	\end{subfigure}

	\begin{subfigure}{0.130\linewidth}
		\includegraphics[width=\linewidth]{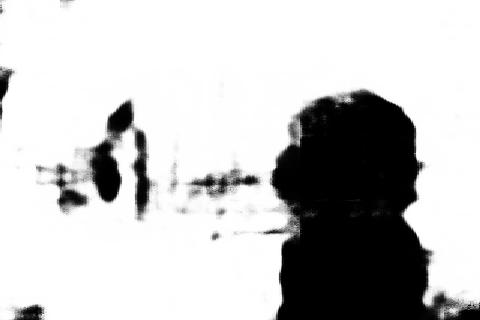}
	\end{subfigure}
	\begin{subfigure}{0.130\linewidth}
		\includegraphics[width=\linewidth]{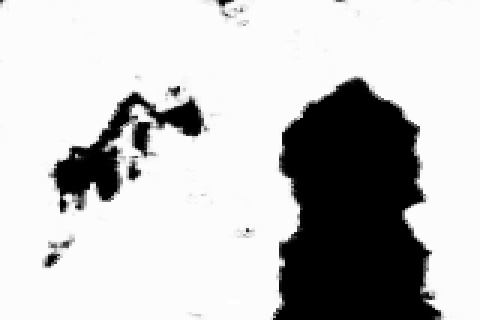}
	\end{subfigure}
	\begin{subfigure}{0.130\linewidth}
		\centering
		\includegraphics[width=\linewidth]{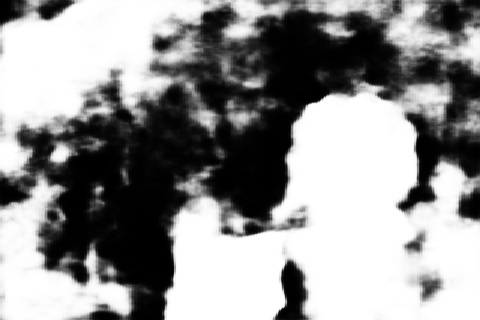}
	\end{subfigure}
	\begin{subfigure}{0.130\linewidth}
		\centering
		\includegraphics[width=\linewidth]{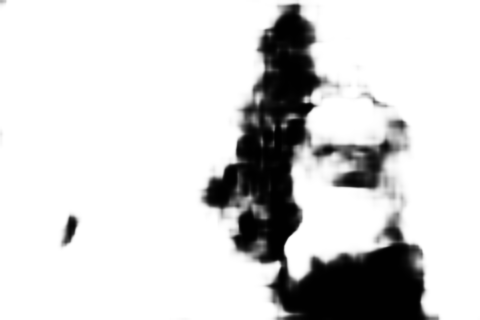}
	\end{subfigure}
	\begin{subfigure}{0.130\linewidth}
		\centering
		\includegraphics[width=\linewidth]{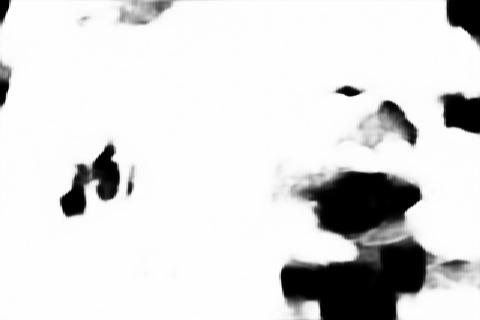}
	\end{subfigure}
	\begin{subfigure}{0.130\linewidth}
		\centering
		\includegraphics[width=\linewidth]{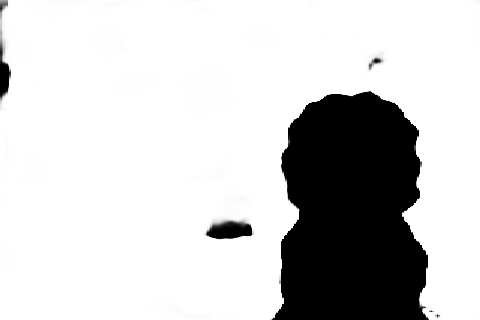}
	\end{subfigure}
	\begin{subfigure}{0.130\linewidth}
		\centering
		\includegraphics[width=\linewidth]{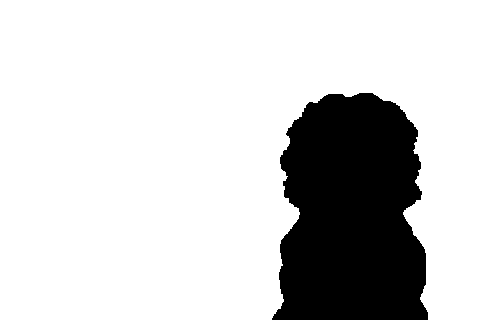}
	\end{subfigure}

	\begin{subfigure}{0.130\linewidth}
		\centering
		\includegraphics[width=\linewidth]{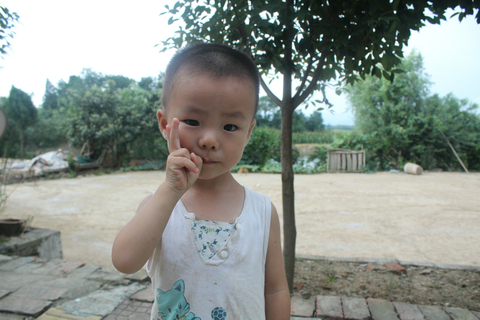}
		\caption{Images}
	\end{subfigure}
	\centering
	\begin{subfigure}{0.130\linewidth}
		\centering
		\includegraphics[width=\linewidth]{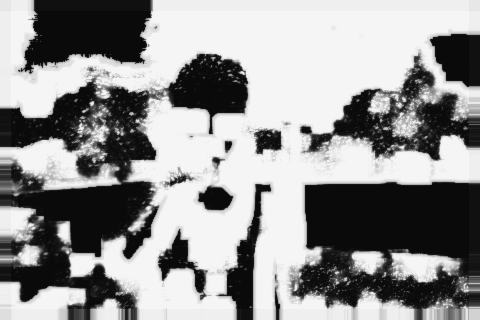}
		\caption{DBDF \cite{shi2014discriminative}}
	\end{subfigure}
	\centering
	\begin{subfigure}{0.130\linewidth}
		\centering
		\includegraphics[width=\linewidth]{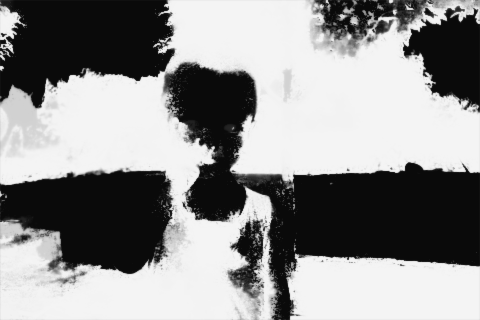}
		\caption{LBP \cite{yi2016lbp}}
	\end{subfigure}
	\begin{subfigure}{0.130\linewidth}
		\centering
		\includegraphics[width=\linewidth]{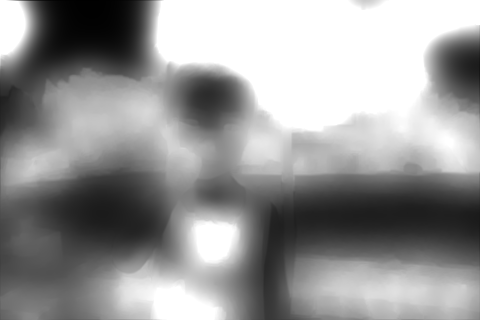}
		\caption{HiFST \cite{alireza2017spatially}}
	\end{subfigure}
	\begin{subfigure}{0.130\linewidth}
		\centering
		\includegraphics[width=\linewidth]{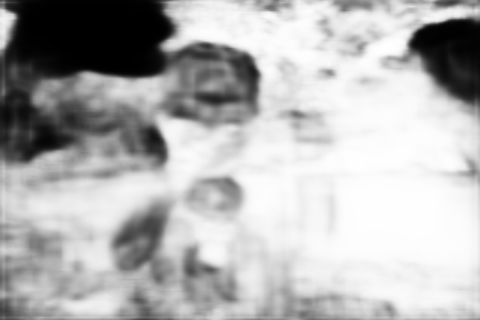}
		\caption{BTBNet \cite{zhao2018defocus}}
	\end{subfigure}
	\begin{subfigure}{0.130\linewidth}
		\centering
		\includegraphics[width=\linewidth]{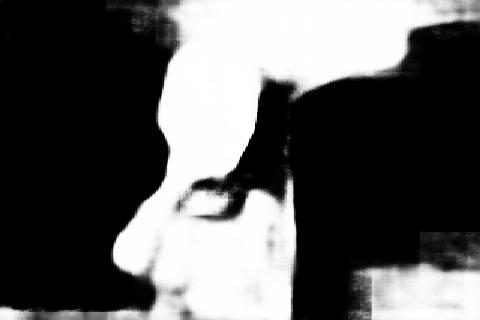}
		\caption{CENet \cite{zhao2019enhancing}}
	\end{subfigure}
	\begin{subfigure}{0.130\linewidth}
		\centering
		\includegraphics[width=\linewidth]{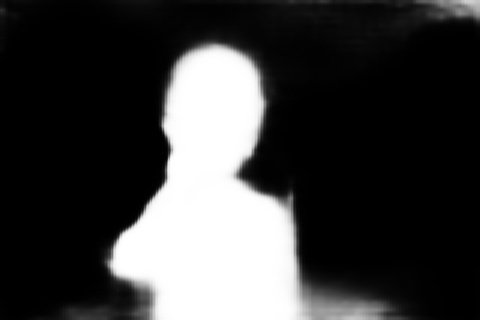}
		\caption{BR2Net \cite{tang2020br}}
	\end{subfigure}
    
	\begin{subfigure}{0.130\linewidth}
		\includegraphics[width=\linewidth]{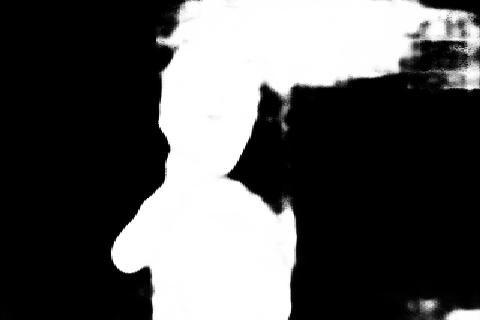}
		\caption{AENet \cite{zhao2021defocus}}
	\end{subfigure}
	\begin{subfigure}{0.130\linewidth}
		\includegraphics[width=\linewidth]{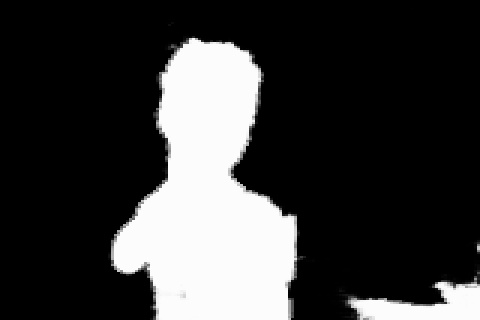}
		\caption{EFENet \cite{zhao2021defocus}}
	\end{subfigure}
	\begin{subfigure}{0.130\linewidth}
		\centering
		\includegraphics[width=\linewidth]{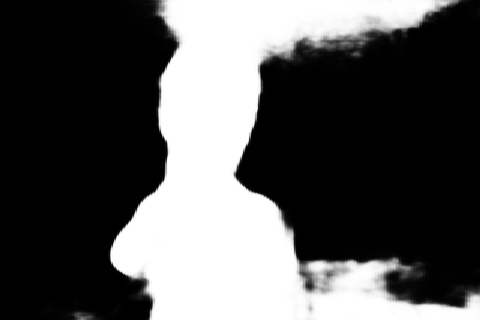}
		\caption{DD \cite{cun2020defocus}}
	\end{subfigure}
	\begin{subfigure}{0.130\linewidth}
		\centering
		\includegraphics[width=\linewidth]{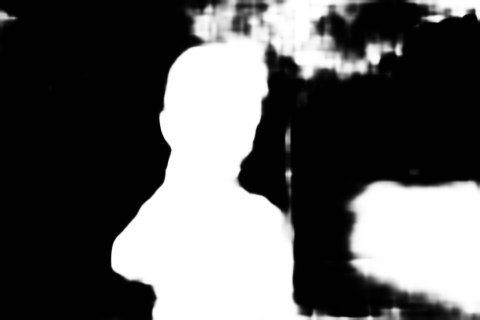}
		\caption{DefuNet \cite{tang2020defusionnet}}
	\end{subfigure}
	\begin{subfigure}{0.130\linewidth}
		\centering
		\includegraphics[width=\linewidth]{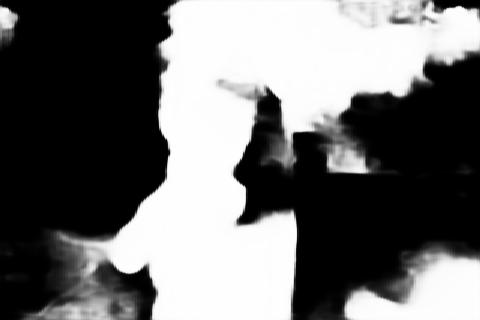}
		\caption{IS2CNet \cite{zhao2021image}}
	\end{subfigure}
	\begin{subfigure}{0.130\linewidth}
		\centering
		\includegraphics[width=\linewidth]{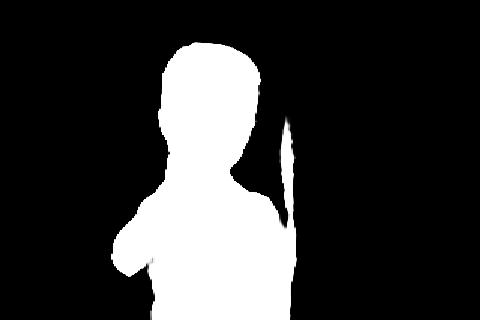}
		\caption{D-DFFNet}
	\end{subfigure}
	\begin{subfigure}{0.130\linewidth}
		\centering
		\includegraphics[width=\linewidth]{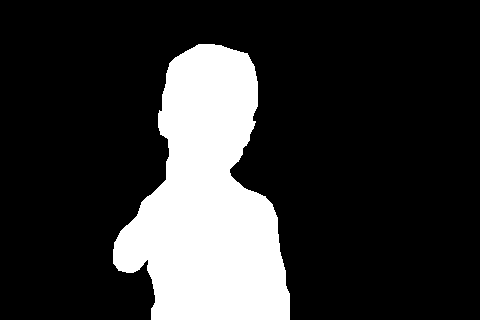}
		\caption{GTs}
	\end{subfigure}
	\hfill
	\caption{Qualitative Comparison of Methods on CTCUG dataset.}
	\label{compare on CTCUG}
\end{figure*}

\noindent
We present more visual comparisons of ours and other state-of-the-art DBD methods in Fig.~\ref{compare on EBD} and Fig.~\ref{compare on CTCUG}. Our results were generated by D-DFFNet, training on CUHK-TE-1.
In Fig.~\ref{compare on EBD}, the blue region in the first image, the table in the second image, the red region in the third and fourth images, and the dark background in the last image are all homogenous regions, our method with depth feature distillation shows better performance. In Fig.~\ref{compare on CTCUG}, the dark region in the second image and the ground of the last image are also homogeneous regions, and our method outperforms others in detecting those regions. 
The first image in Fig.~\ref{compare on CTCUG} presents a challenge for previous methods as the boundaries between the in-focus and out-of-focus regions are hard to locate. However, our method effectively combines DOF boundary information with depth feature distillation, allowing us to accurately locate the DOF regions with boundary details.
The third, fourth, and fifth images in Fig.~\ref{compare on CTCUG} have complex scenes, yet our method still shows superior prediction.

\section{Ablation Study}
\subsection{More Results of Depth Feature Distillation}
\begin{figure}[t]

	\begin{subfigure}{0.20\linewidth}
		\centering
		\includegraphics[width=0.9\linewidth]{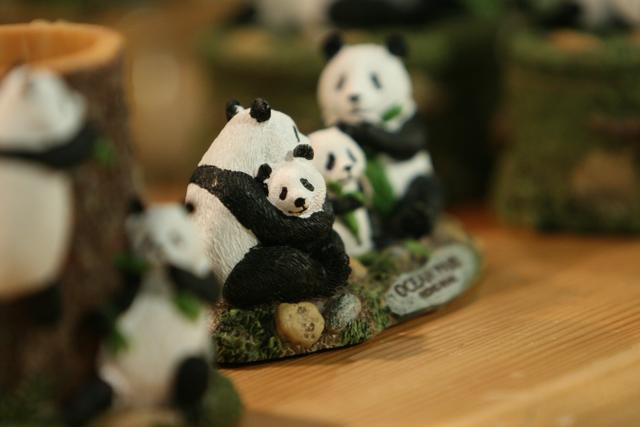}
	\end{subfigure}
	\centering
	\begin{subfigure}{0.20\linewidth}
		\centering
		\includegraphics[width=0.9\linewidth]{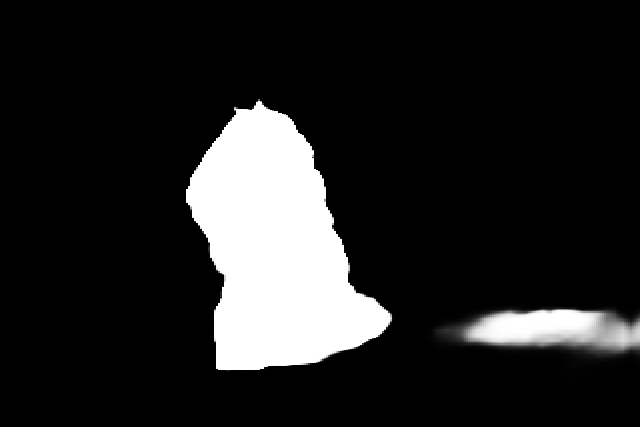}
	\end{subfigure}
	\begin{subfigure}{0.20\linewidth}
		\centering
		\includegraphics[width=0.9\linewidth]{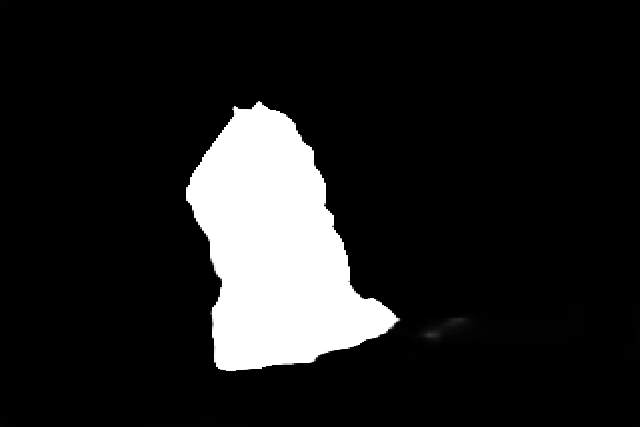}
	\end{subfigure}
	\begin{subfigure}{0.20\linewidth}
		\centering
		\includegraphics[width=0.9\linewidth]{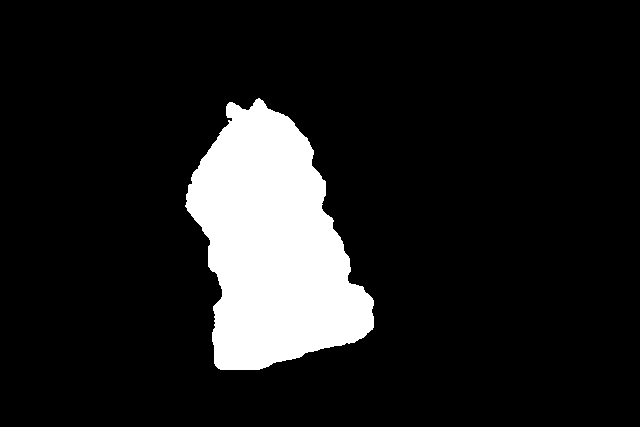}
	\end{subfigure}

	\begin{subfigure}{0.20\linewidth}
		\centering
		\includegraphics[width=0.9\linewidth]{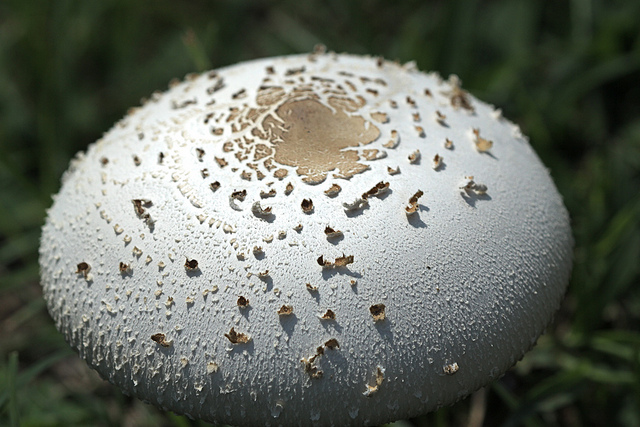}
	\end{subfigure}
	\centering
	\begin{subfigure}{0.20\linewidth}
		\centering
		\includegraphics[width=0.9\linewidth]{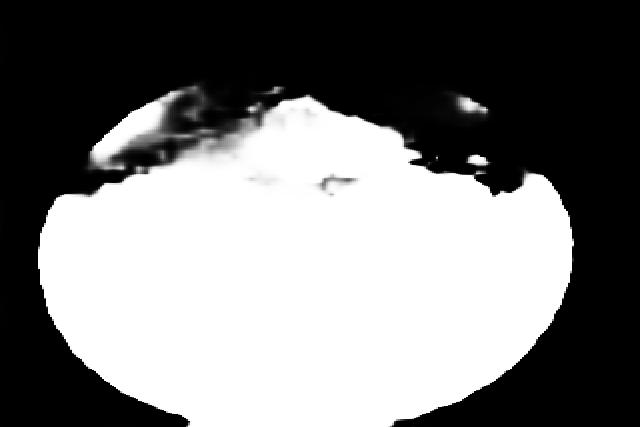}
	\end{subfigure}
	\begin{subfigure}{0.20\linewidth}
		\centering
		\includegraphics[width=0.9\linewidth]{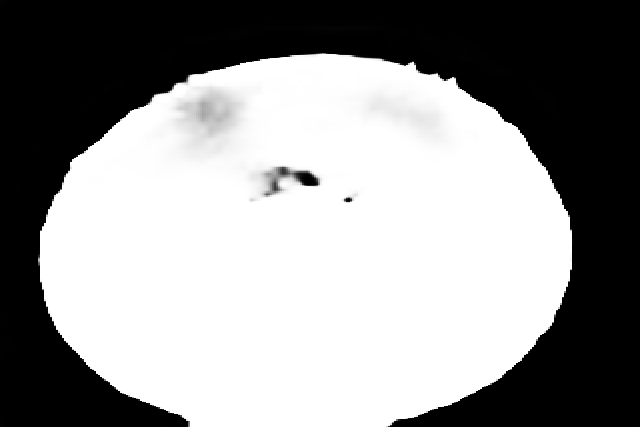}
	\end{subfigure}
	\begin{subfigure}{0.20\linewidth}
		\centering
		\includegraphics[width=0.9\linewidth]{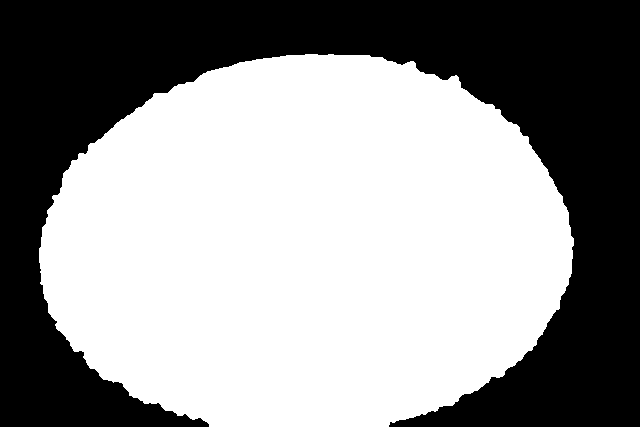}
	\end{subfigure}

	\begin{subfigure}{0.20\linewidth}
	\captionsetup{font={footnotesize}}
		\centering
		\includegraphics[width=0.9\linewidth]{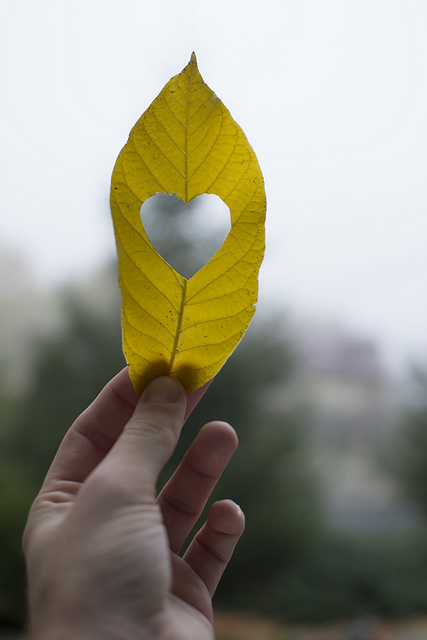}
		\caption{Images}
	\end{subfigure}
	\centering
	\begin{subfigure}{0.20\linewidth}
	\captionsetup{font={footnotesize}}
		\centering
		\includegraphics[width=0.9\linewidth]{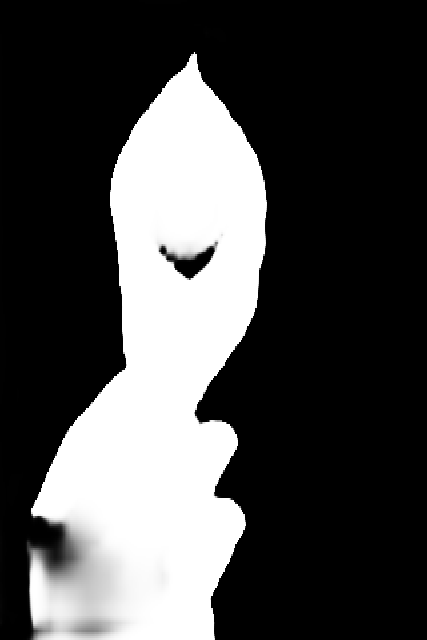}
		\caption{DFFNet}
	\end{subfigure}
	\begin{subfigure}{0.20\linewidth}
	\captionsetup{font={footnotesize}}
		\centering
		\includegraphics[width=0.9\linewidth]{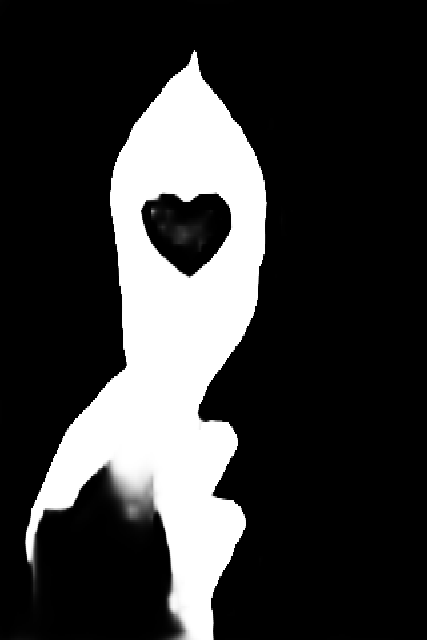}
		\caption{D-DFFNet}
	\end{subfigure}
	\begin{subfigure}{0.20\linewidth}
	\captionsetup{font={footnotesize}}
		\centering
		\includegraphics[width=0.9\linewidth]{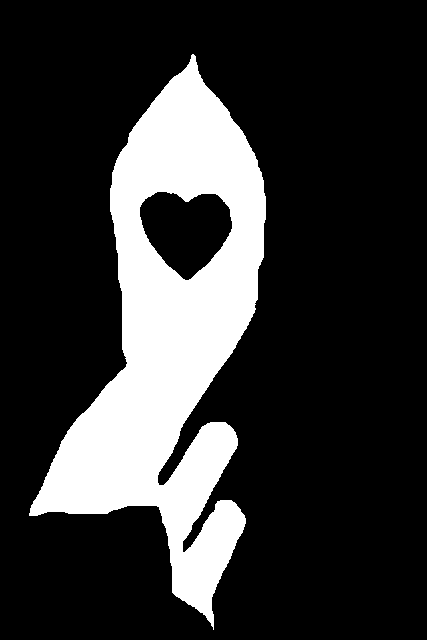}
		\caption{GTs}
	\end{subfigure}
\vspace{-0.5em}
\caption{Ablation study of depth feature distillation on CUHK-TE-1 dataset.}
\label{Ablation of depth feature distillation on CUHK}
\vspace{-1.5em}
\end{figure}

\begin{figure}[t]

	\begin{subfigure}{0.20\linewidth}
		\centering
		\includegraphics[width=0.9\linewidth]{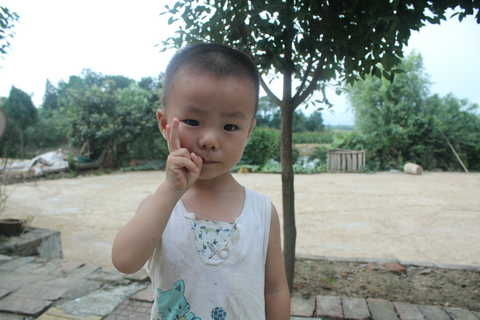}
	\end{subfigure}
	\centering
	\begin{subfigure}{0.20\linewidth}
		\centering
		\includegraphics[width=0.9\linewidth]{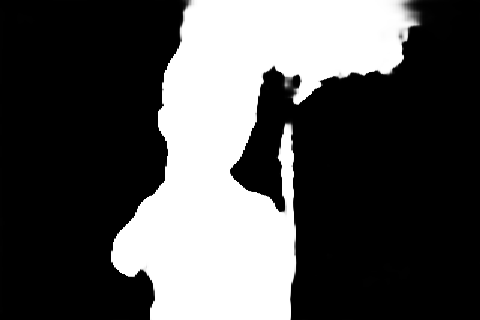}
	\end{subfigure}
	\begin{subfigure}{0.20\linewidth}
		\centering
		\includegraphics[width=0.9\linewidth]{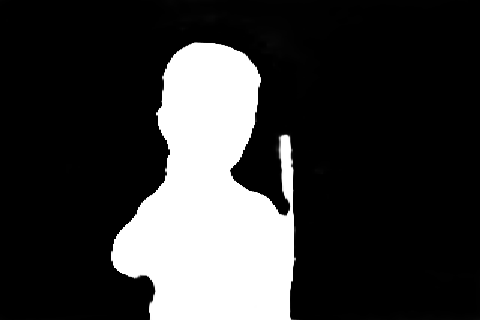}
	\end{subfigure}
	\begin{subfigure}{0.20\linewidth}
		\centering
		\includegraphics[width=0.9\linewidth]{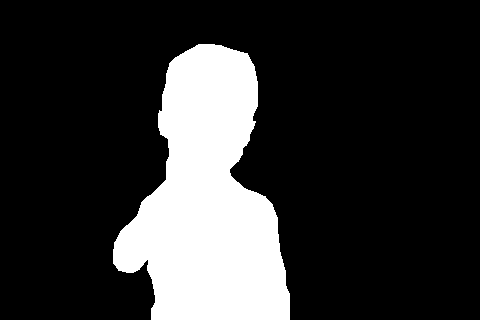}
	\end{subfigure}

	\begin{subfigure}{0.20\linewidth}
		\centering
		\includegraphics[width=0.9\linewidth]{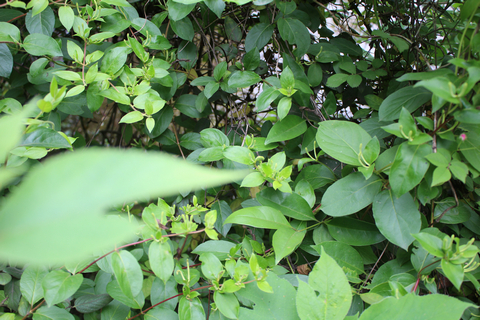}
	\end{subfigure}
	\centering
	\begin{subfigure}{0.20\linewidth}
		\centering
		\includegraphics[width=0.9\linewidth]{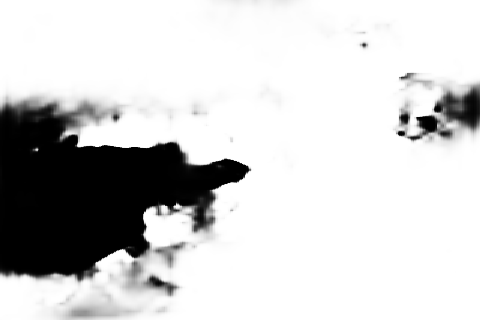}
	\end{subfigure}
	\begin{subfigure}{0.20\linewidth}
		\centering
		\includegraphics[width=0.9\linewidth]{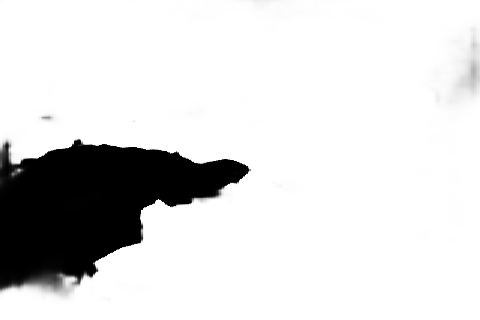}
	\end{subfigure}
	\begin{subfigure}{0.20\linewidth}
		\centering
		\includegraphics[width=0.9\linewidth]{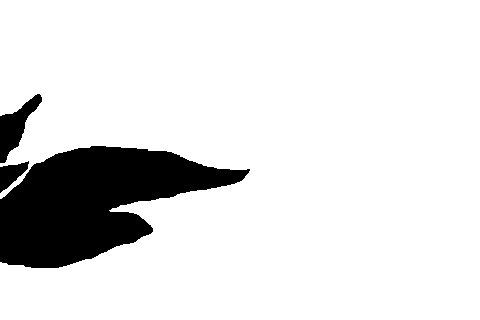}
	\end{subfigure}
	
	\begin{subfigure}{0.20\linewidth}
	\captionsetup{font={footnotesize}}
		\centering
		\includegraphics[width=0.9\linewidth]{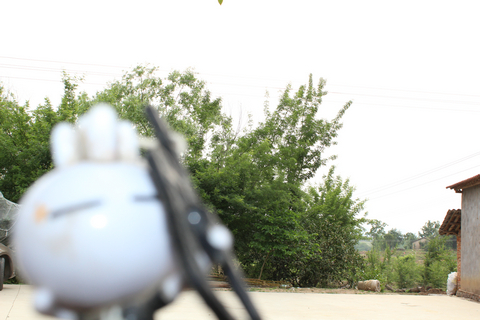}
		\caption{Images}
	\end{subfigure}
	\centering
	\begin{subfigure}{0.20\linewidth}
	\captionsetup{font={footnotesize}}
		\centering
		\includegraphics[width=0.9\linewidth]{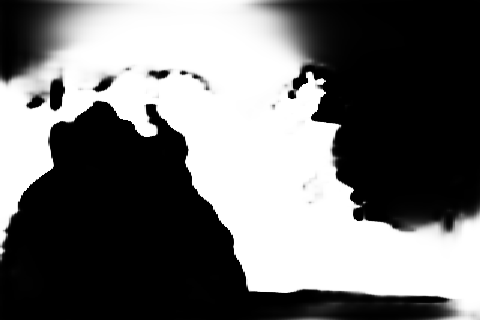}
		\caption{DFFNet}
	\end{subfigure}
	\begin{subfigure}{0.20\linewidth}
	\captionsetup{font={footnotesize}}
		\centering
		\includegraphics[width=0.9\linewidth]{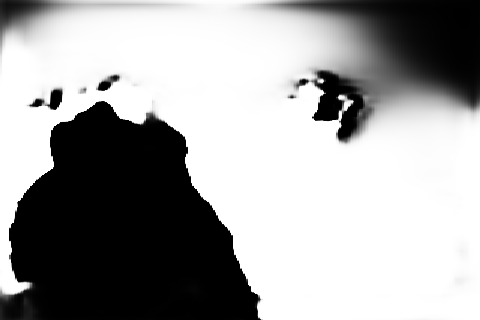}
		\caption{D-DFFNet}
	\end{subfigure}
	\begin{subfigure}{0.20\linewidth}
	\captionsetup{font={footnotesize}}
		\centering
		\includegraphics[width=0.9\linewidth]{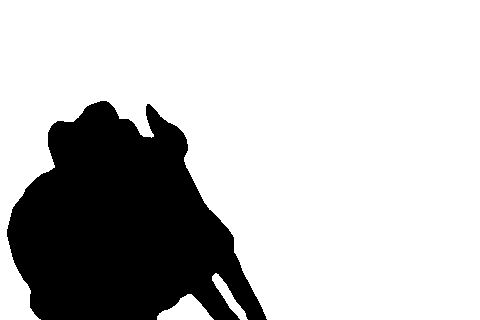}
		\caption{GTs}
	\end{subfigure}
\caption{Ablation study of depth feature distillation on CTCUG dataset.}
\label{Ablation of depth feature distillation on CTCUG}
\end{figure}

\noindent
We present more visual comparison to prove the effectiveness of depth feature distillation on three test datasets, in Fig.~\ref{Ablation of depth feature distillation}, Fig.~\ref{Ablation of depth feature distillation on CUHK}, and Fig.~\ref{Ablation of depth feature distillation on CTCUG}. Depth feature distillation can help separate defocus blur from homogeneous regions (petals of the first image and grass of the last image in Fig.~\ref{Ablation of depth feature distillation}, the sky of the last image in Fig.~\ref{Ablation of depth feature distillation on CTCUG}), and can help detect defocus blur on images that the foreground and the background have noticeable depth different (the second and the fourth images in Fig.~\ref{Ablation of depth feature distillation}, the last image in Fig.~\ref{Ablation of depth feature distillation on CUHK}).
\subsection{More Results of DOF-edge loss}
\begin{figure}[t]
	\begin{subfigure}{0.18\linewidth}
		\centering
		\includegraphics[width=0.90\linewidth]{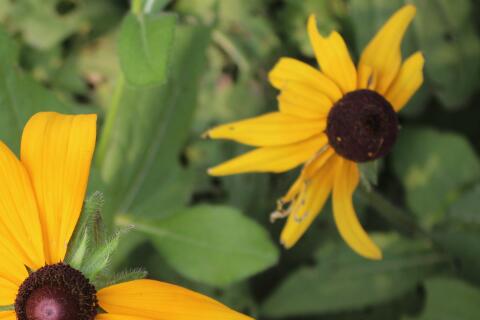}
	\end{subfigure}
	\begin{subfigure}{0.18\linewidth}
		\centering
		\includegraphics[width=0.90\linewidth]{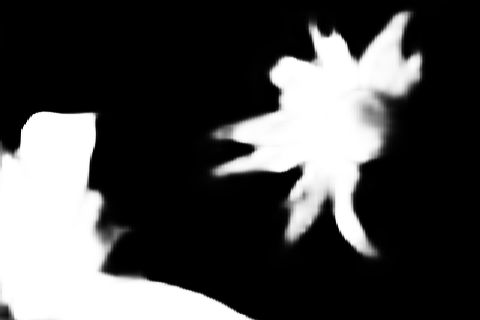}
	\end{subfigure}
	\centering
	\begin{subfigure}{0.18\linewidth}
		\centering
		\includegraphics[width=0.90\linewidth]{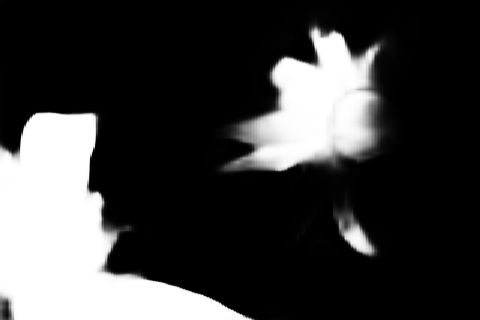}
	\end{subfigure}
	\begin{subfigure}{0.18\linewidth}
		\centering
		\includegraphics[width=0.90\linewidth]{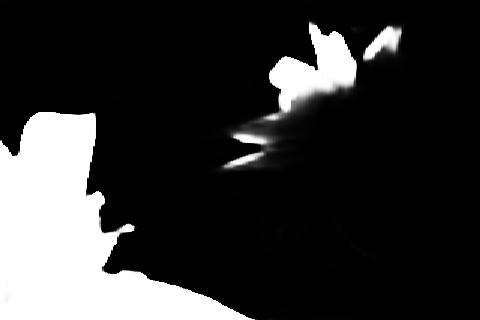}
	\end{subfigure}
	\begin{subfigure}{0.18\linewidth}
		\centering
		\includegraphics[width=0.90\linewidth]{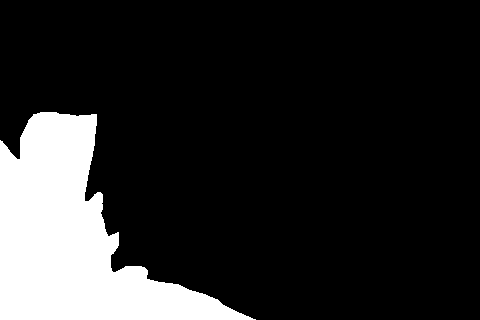}
	\end{subfigure}

	\begin{subfigure}{0.18\linewidth}
		\centering
		\includegraphics[width=0.90\linewidth]{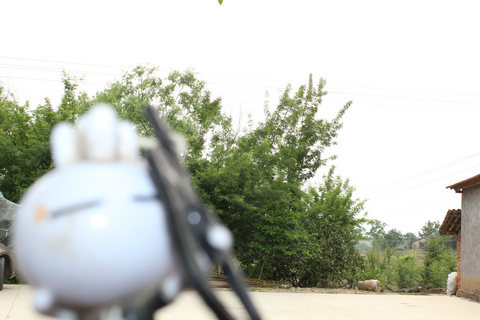}
	\end{subfigure}
	\begin{subfigure}{0.18\linewidth}
		\centering
		\includegraphics[width=0.90\linewidth]{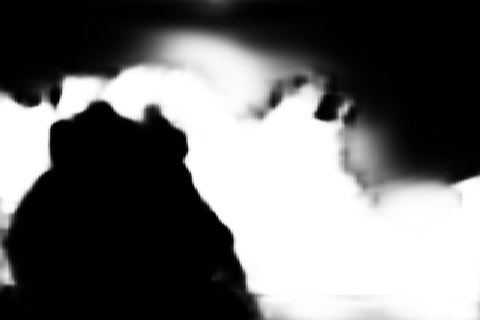}
	\end{subfigure}
	\centering
	\begin{subfigure}{0.18\linewidth}
		\centering
		\includegraphics[width=0.90\linewidth]{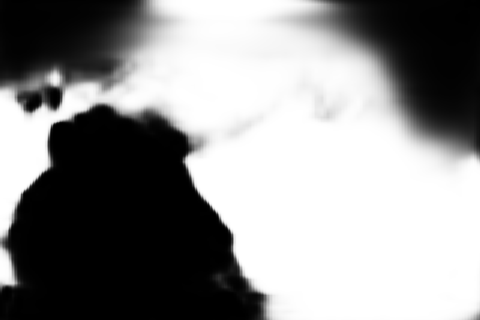}
	\end{subfigure}
	\begin{subfigure}{0.18\linewidth}
		\centering
		\includegraphics[width=0.90\linewidth]{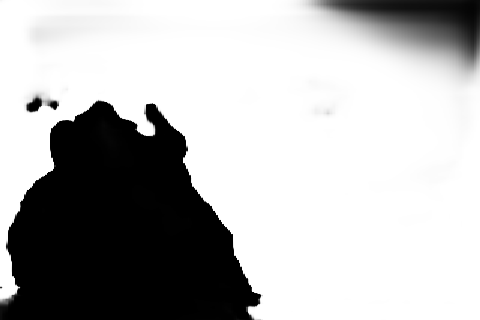}
	\end{subfigure}
	\begin{subfigure}{0.18\linewidth}
		\centering
		\includegraphics[width=0.90\linewidth]{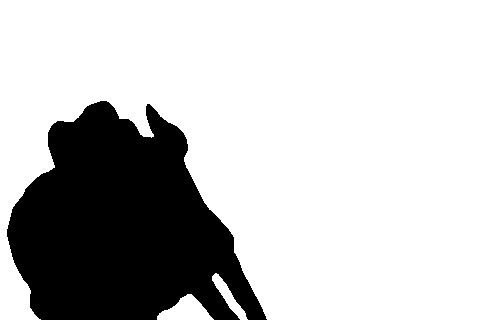}
	\end{subfigure}

	\begin{subfigure}{0.18\linewidth}
		\centering
		\includegraphics[width=0.90\linewidth]{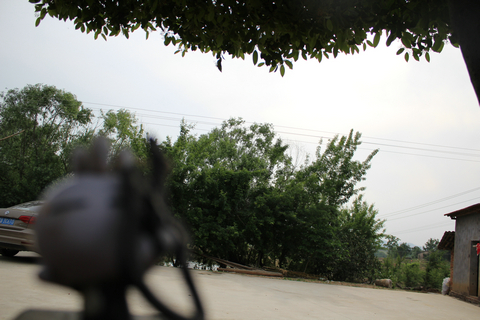}
	\end{subfigure}
	\begin{subfigure}{0.18\linewidth}
		\centering
		\includegraphics[width=0.90\linewidth]{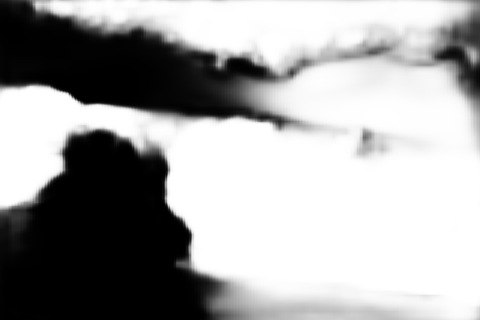}
	\end{subfigure}
	\centering
	\begin{subfigure}{0.18\linewidth}
		\centering
		\includegraphics[width=0.90\linewidth]{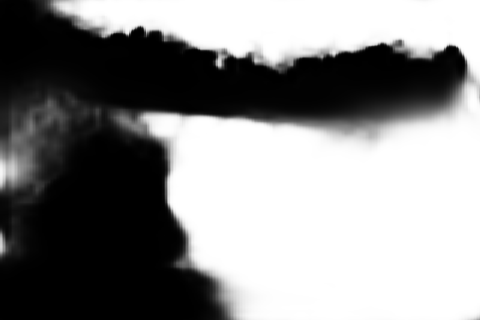}
	\end{subfigure}
	\begin{subfigure}{0.18\linewidth}
		\centering
		\includegraphics[width=0.90\linewidth]{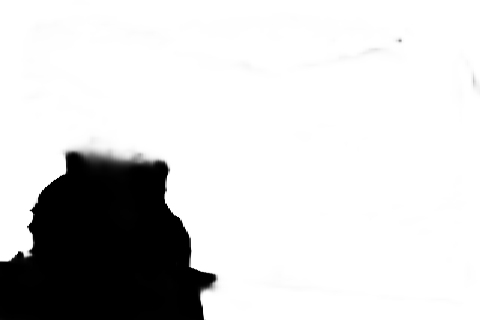}
	\end{subfigure}
	\begin{subfigure}{0.18\linewidth}
		\centering
		\includegraphics[width=0.90\linewidth]{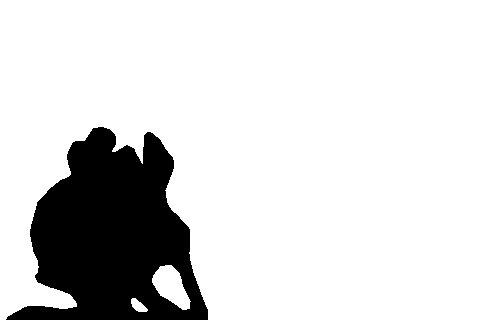}
	\end{subfigure}

	\begin{subfigure}{0.18\linewidth}
	\captionsetup{font={tiny}}
		\centering
		\includegraphics[width=0.90\linewidth]{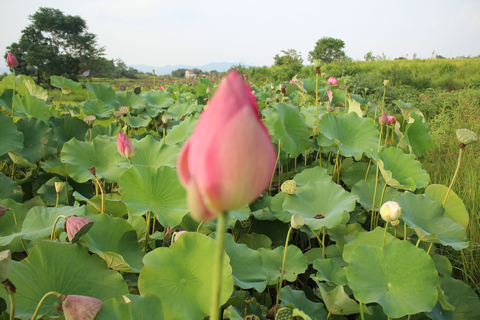}
		\caption{Images}
	\end{subfigure}
	\begin{subfigure}{0.18\linewidth}
	\captionsetup{font={tiny}}
		\centering
		\includegraphics[width=0.90\linewidth]{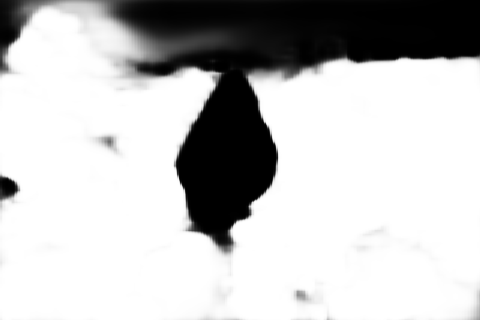}
		\caption{BCE}
	\end{subfigure}
	\centering
	\begin{subfigure}{0.18\linewidth}
	\captionsetup{font={tiny}}
		\centering
		\includegraphics[width=0.90\linewidth]{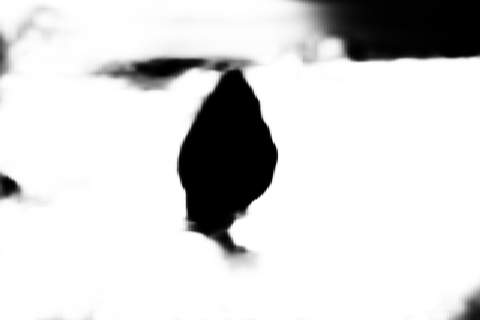}
		\caption{BCE-BCE}
	\end{subfigure}
	\begin{subfigure}{0.18\linewidth}
	\captionsetup{font={tiny}}
		\centering
		\includegraphics[width=0.90\linewidth]{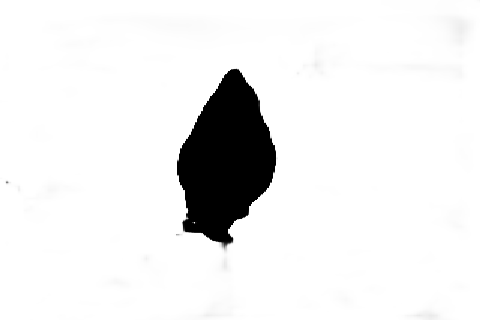}
		\caption{BCE-BCE\&EL}
	\end{subfigure}
	\begin{subfigure}{0.18\linewidth}
	\captionsetup{font={tiny}}
		\centering
		\includegraphics[width=0.90\linewidth]{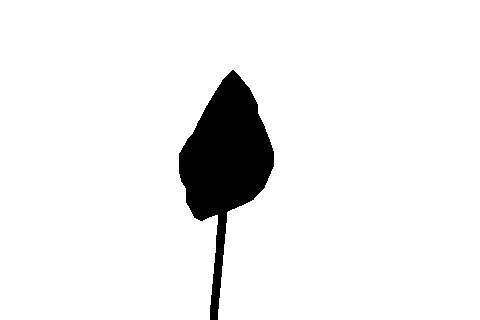}
		\caption{GTs}
	\end{subfigure}
\vspace{-0.5em}
\caption{Ablation study of DOF-edge loss. BCE represents training DFFNet using BCE loss. BCE-BCE represents training D-DFFNet using BCE loss in all two stages. BCE-BCE\&EL represents training D-DFFNet using BCE loss in stage 1, and using both BCE and DOF-edge loss in stage 2.}
\label{Ablation of edge loss}
\vspace{-1.5em}
\end{figure}

\noindent
In Fig.~\ref{Ablation of edge loss}, we present a visual comparison to prove the effectiveness of DOF-edge loss. Training D-DFFNet using BCE and DOF-edge loss in the second stage can help locate the DOF regions, and get better predictions than using only BCE loss. This demonstrates that DOF-edge loss can enhance the performance of depth feature representation. Moreover, predictions with BCE and DOF-edge loss have sharper boundaries than with only BCE loss. 
\subsection{Effectiveness of dense feature fusion module (DFFM).} 
\noindent
From Table~\ref{main ablation study}, the performance on DUT-TE and CTCUG datasets decreases dramatically when we remove DFFM from our DFFNet, which proves the effectiveness of DFFM.

\begin{figure}[t]
	\begin{subfigure}{0.20\linewidth}
		\centering
		\includegraphics[width=0.9\linewidth]{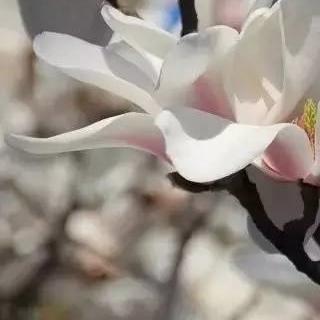}
	\end{subfigure}
	\centering
	\begin{subfigure}{0.20\linewidth}
		\centering
		\includegraphics[width=0.9\linewidth]{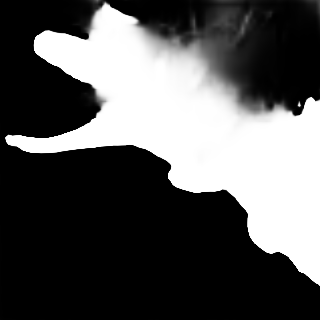}
	\end{subfigure}
	\begin{subfigure}{0.20\linewidth}
		\centering
		\includegraphics[width=0.9\linewidth]{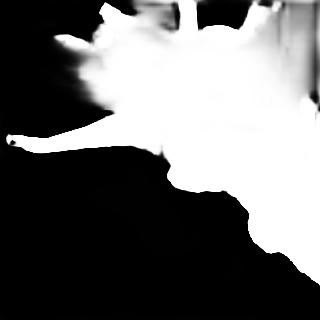}
	\end{subfigure}
	\begin{subfigure}{0.20\linewidth}
		\centering
		\includegraphics[width=0.9\linewidth]{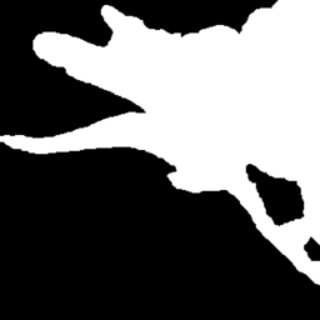}
	\end{subfigure}

	\begin{subfigure}{0.20\linewidth}
		\centering
		\includegraphics[width=0.9\linewidth]{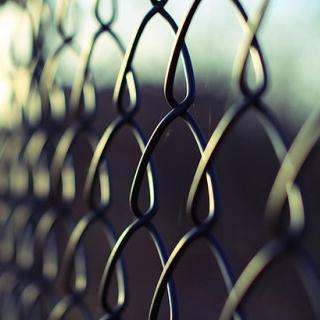}
	\end{subfigure}
	\centering
	\begin{subfigure}{0.20\linewidth}
		\centering
		\includegraphics[width=0.9\linewidth]{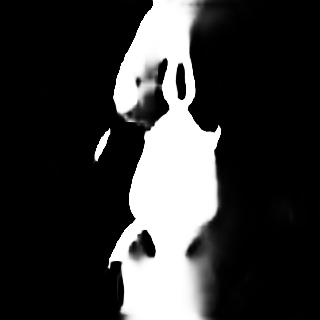}
	\end{subfigure}
	\begin{subfigure}{0.20\linewidth}
		\centering
		\includegraphics[width=0.9\linewidth]{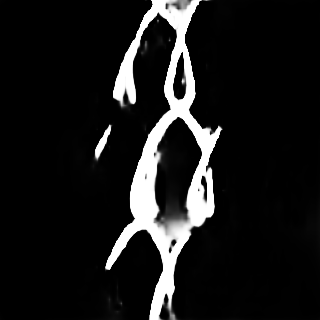}
	\end{subfigure}
	\begin{subfigure}{0.20\linewidth}
		\centering
		\includegraphics[width=0.9\linewidth]{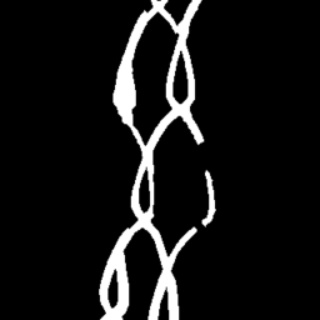}
	\end{subfigure}

	\begin{subfigure}{0.20\linewidth}
		\centering
		\includegraphics[width=0.9\linewidth]{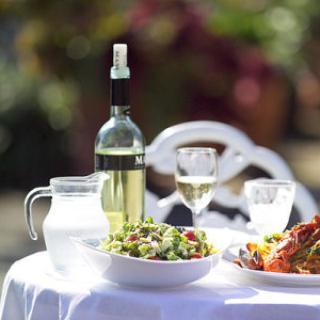}
	\end{subfigure}
	\centering
	\begin{subfigure}{0.20\linewidth}
		\centering
		\includegraphics[width=0.9\linewidth]{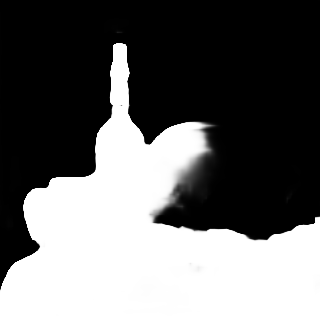}
	\end{subfigure}
	\begin{subfigure}{0.20\linewidth}
		\centering
		\includegraphics[width=0.9\linewidth]{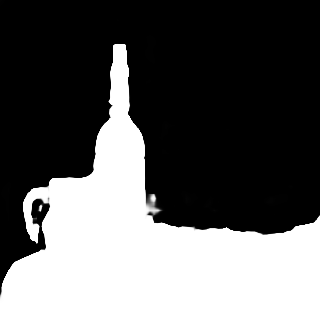}
	\end{subfigure}
	\begin{subfigure}{0.20\linewidth}
		\centering
		\includegraphics[width=0.9\linewidth]{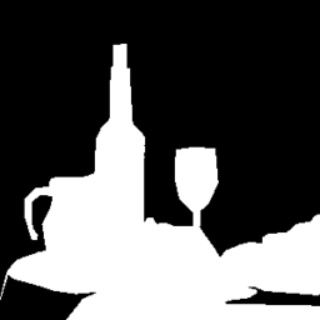}
	\end{subfigure}

	\begin{subfigure}{0.20\linewidth}
		\centering
		\includegraphics[width=0.9\linewidth]{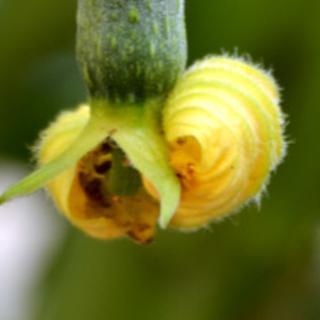}
	\end{subfigure}
	\centering
	\begin{subfigure}{0.20\linewidth}
		\centering
		\includegraphics[width=0.9\linewidth]{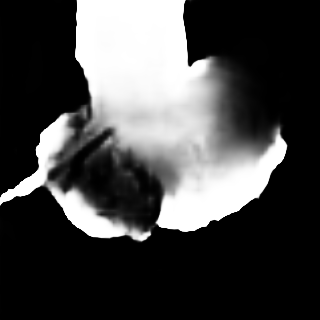}
	\end{subfigure}
	\begin{subfigure}{0.20\linewidth}
		\centering
		\includegraphics[width=0.9\linewidth]{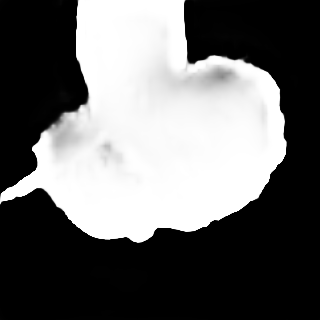}
	\end{subfigure}
	\begin{subfigure}{0.20\linewidth}
		\centering
		\includegraphics[width=0.9\linewidth]{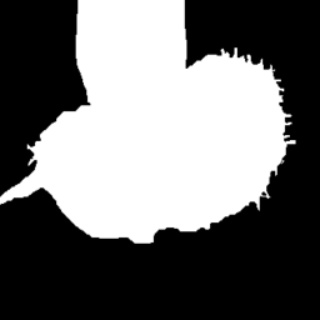}
	\end{subfigure}

	\begin{subfigure}{0.20\linewidth}
		\centering
		\includegraphics[width=0.9\linewidth]{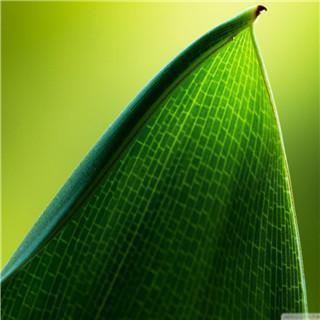}
	\end{subfigure}
	\centering
	\begin{subfigure}{0.20\linewidth}
		\centering
		\includegraphics[width=0.9\linewidth]{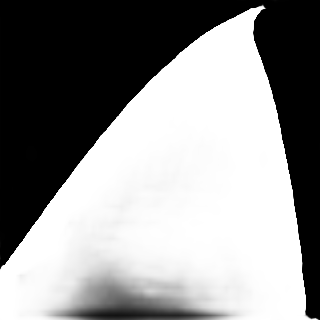}
	\end{subfigure}
	\begin{subfigure}{0.20\linewidth}
		\centering
		\includegraphics[width=0.9\linewidth]{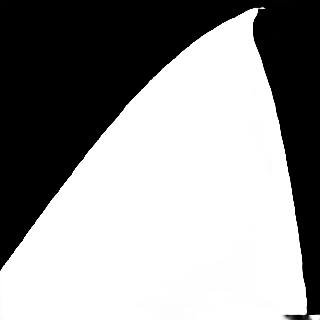}
	\end{subfigure}
	\begin{subfigure}{0.20\linewidth}
		\centering
		\includegraphics[width=0.9\linewidth]{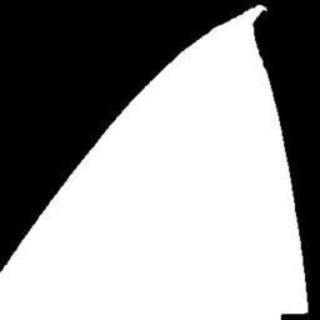}
	\end{subfigure}

	\begin{subfigure}{0.20\linewidth}
	\captionsetup{font={footnotesize}}
		\centering
		\includegraphics[width=0.9\linewidth]{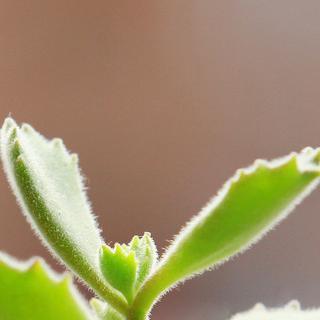}
		\caption{Images}
	\end{subfigure}
	\centering
	\begin{subfigure}{0.20\linewidth}
	\captionsetup{font={footnotesize}}
		\centering
		\includegraphics[width=0.9\linewidth]{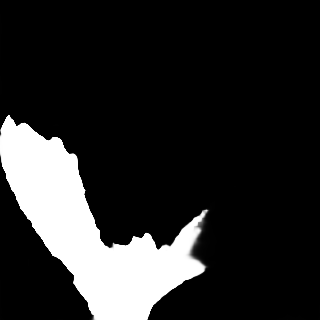}
		\caption{DFFNet}
	\end{subfigure}
	\begin{subfigure}{0.20\linewidth}
	\captionsetup{font={footnotesize}}
		\centering
		\includegraphics[width=0.9\linewidth]{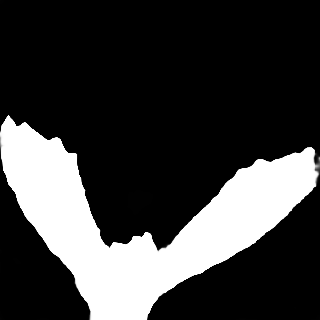}
		\caption{D-DFFNet}
	\end{subfigure}
	\begin{subfigure}{0.20\linewidth}
	\captionsetup{font={footnotesize}}
		\centering
		\includegraphics[width=0.9\linewidth]{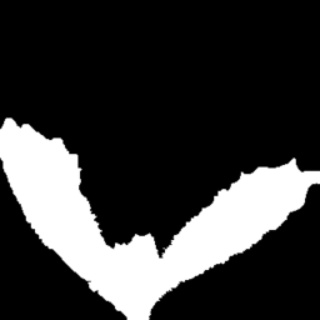}
		\caption{GTs}
	\end{subfigure}
\vspace{-0.5em}
\caption{Ablation study of depth feature distillation on DUT-TE dataset.}
\label{Ablation of depth feature distillation}
\vspace{-1.5em}
\end{figure}

\begin{figure}[t]
    \begin{subfigure}{0.20\linewidth}
        \centering
        \includegraphics[width=0.90\linewidth]{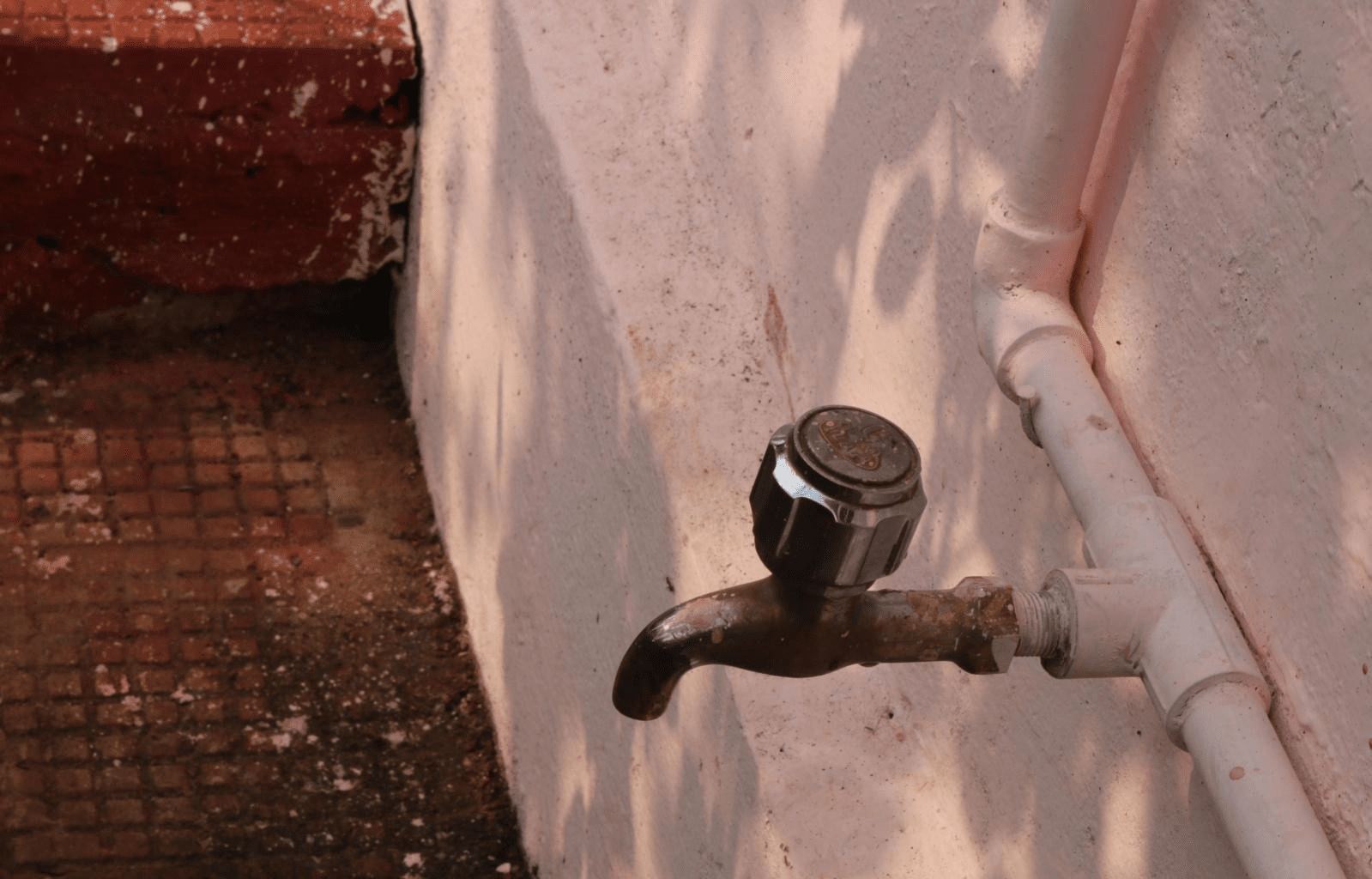}
    \end{subfigure}
    \centering
    \begin{subfigure}{0.20\linewidth}
        \centering
        \includegraphics[width=0.90\linewidth]{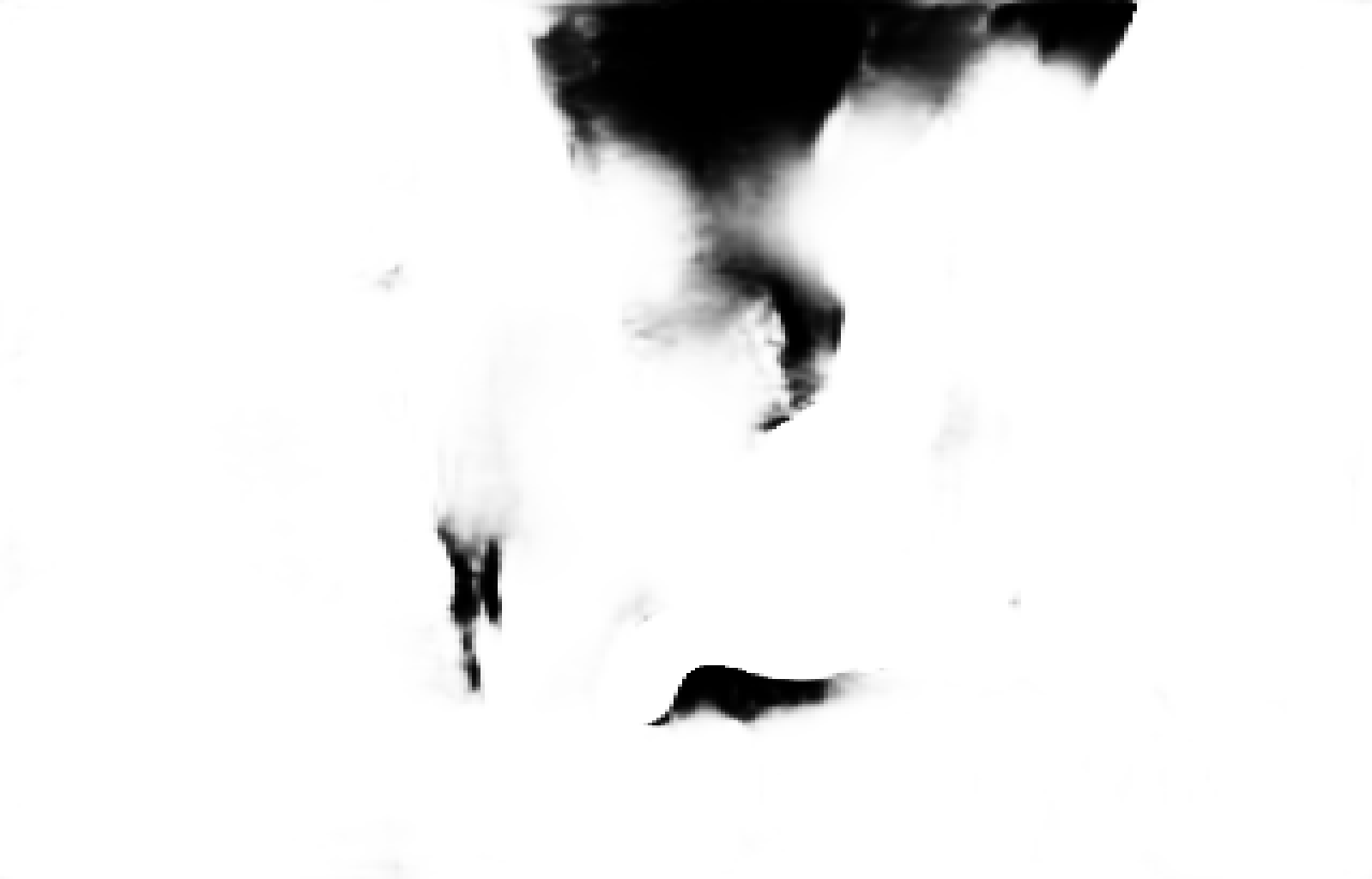}
    \end{subfigure}
    \begin{subfigure}{0.20\linewidth}
        \centering
        \includegraphics[width=0.90\linewidth]{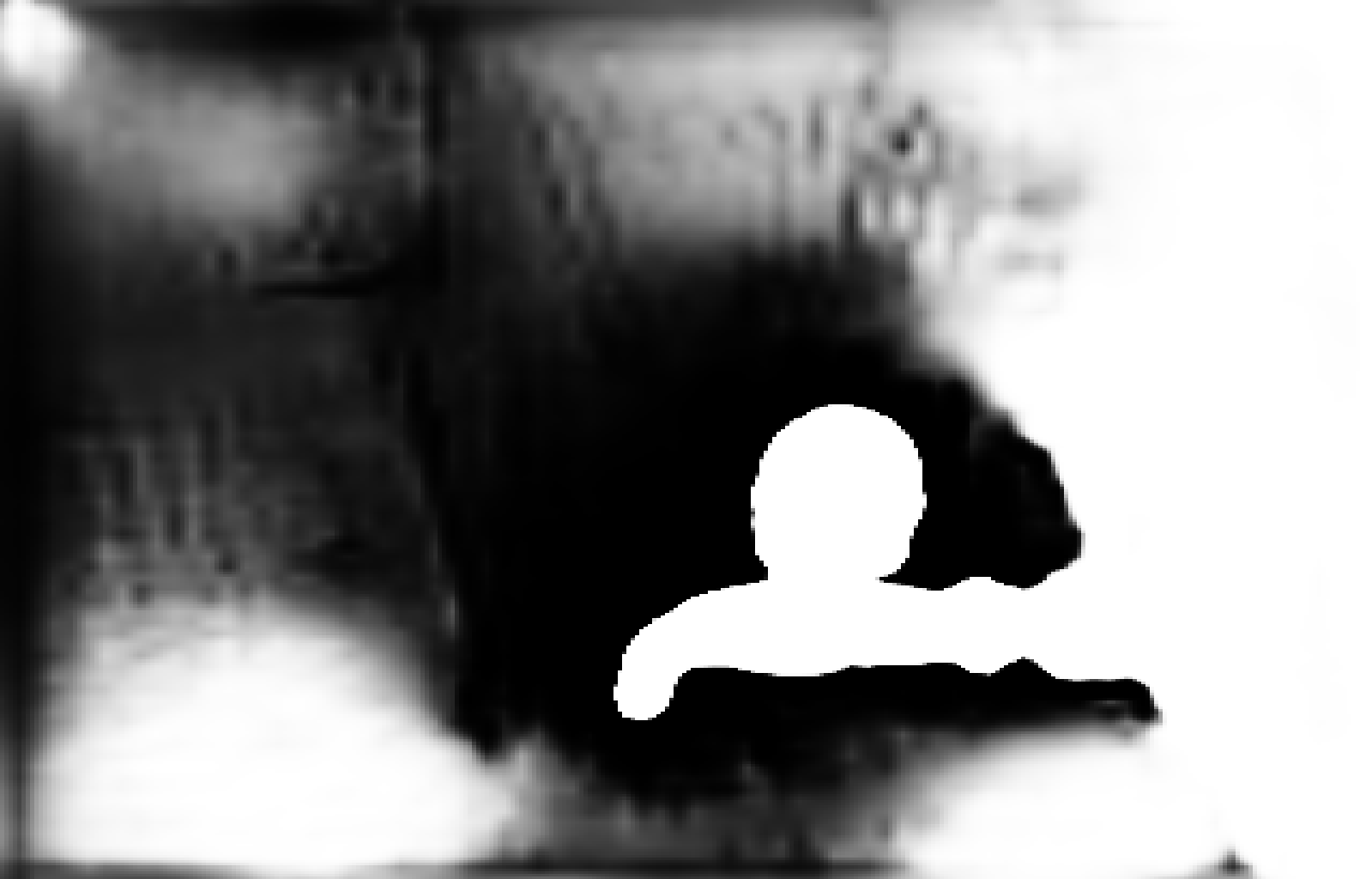}
    \end{subfigure}
    \begin{subfigure}{0.20\linewidth}
        \centering
        \includegraphics[width=0.90\linewidth]{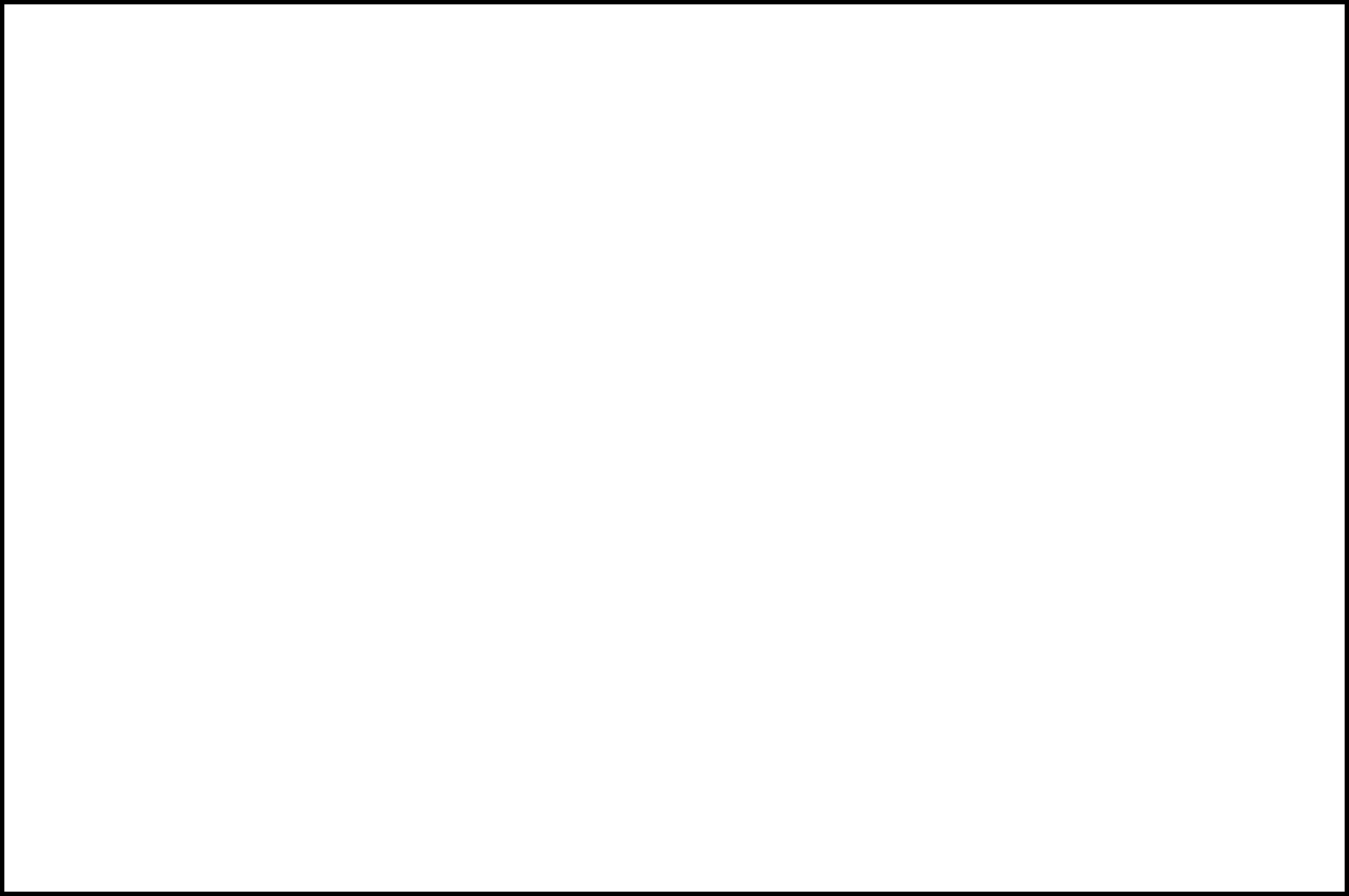}
    \end{subfigure}
 
    \begin{subfigure}{0.20\linewidth}
		\centering
		\includegraphics[width=0.90\linewidth]{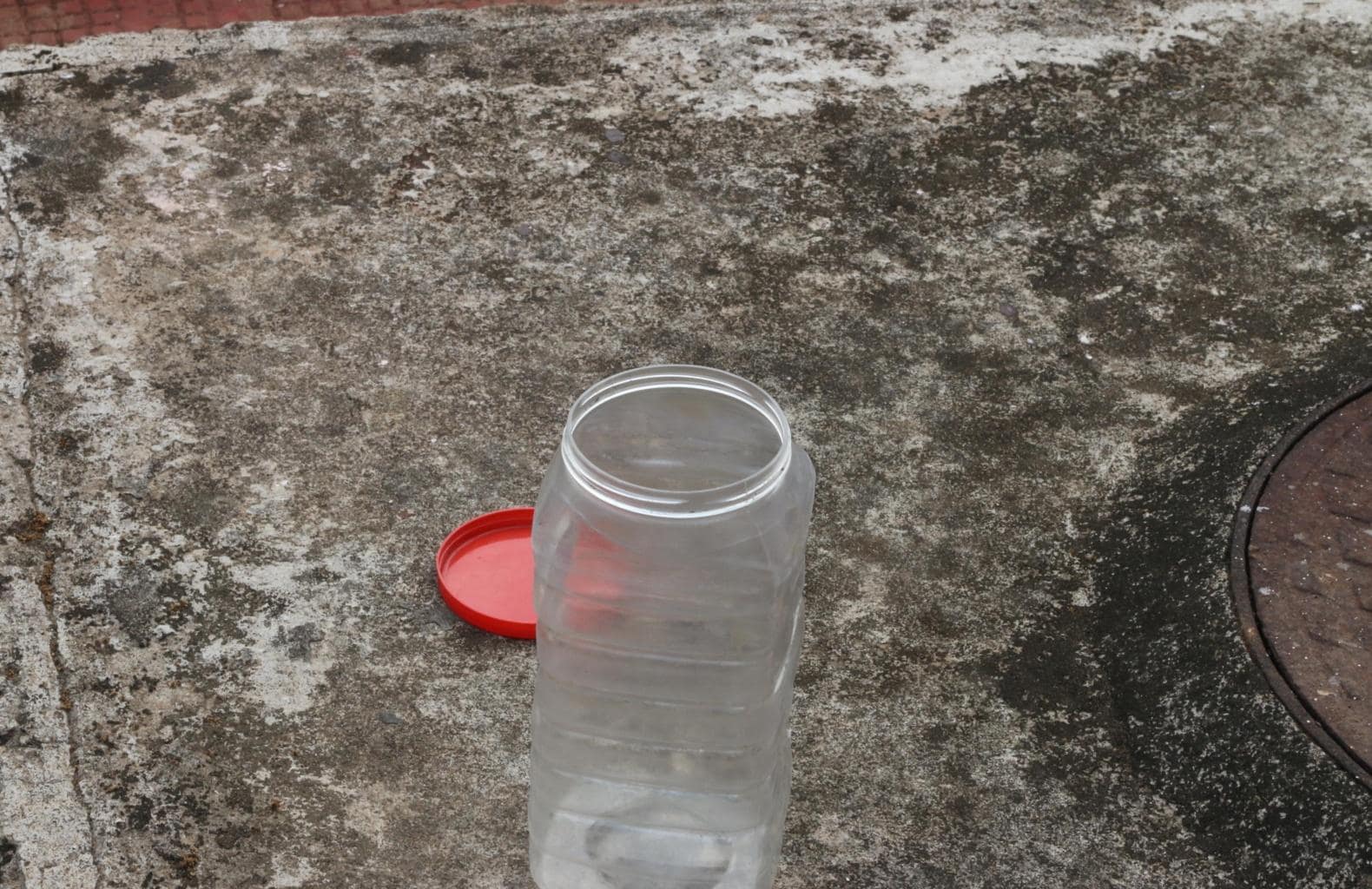}
	\end{subfigure}
	\centering
	\begin{subfigure}{0.20\linewidth}
		\centering
		\includegraphics[width=0.90\linewidth]{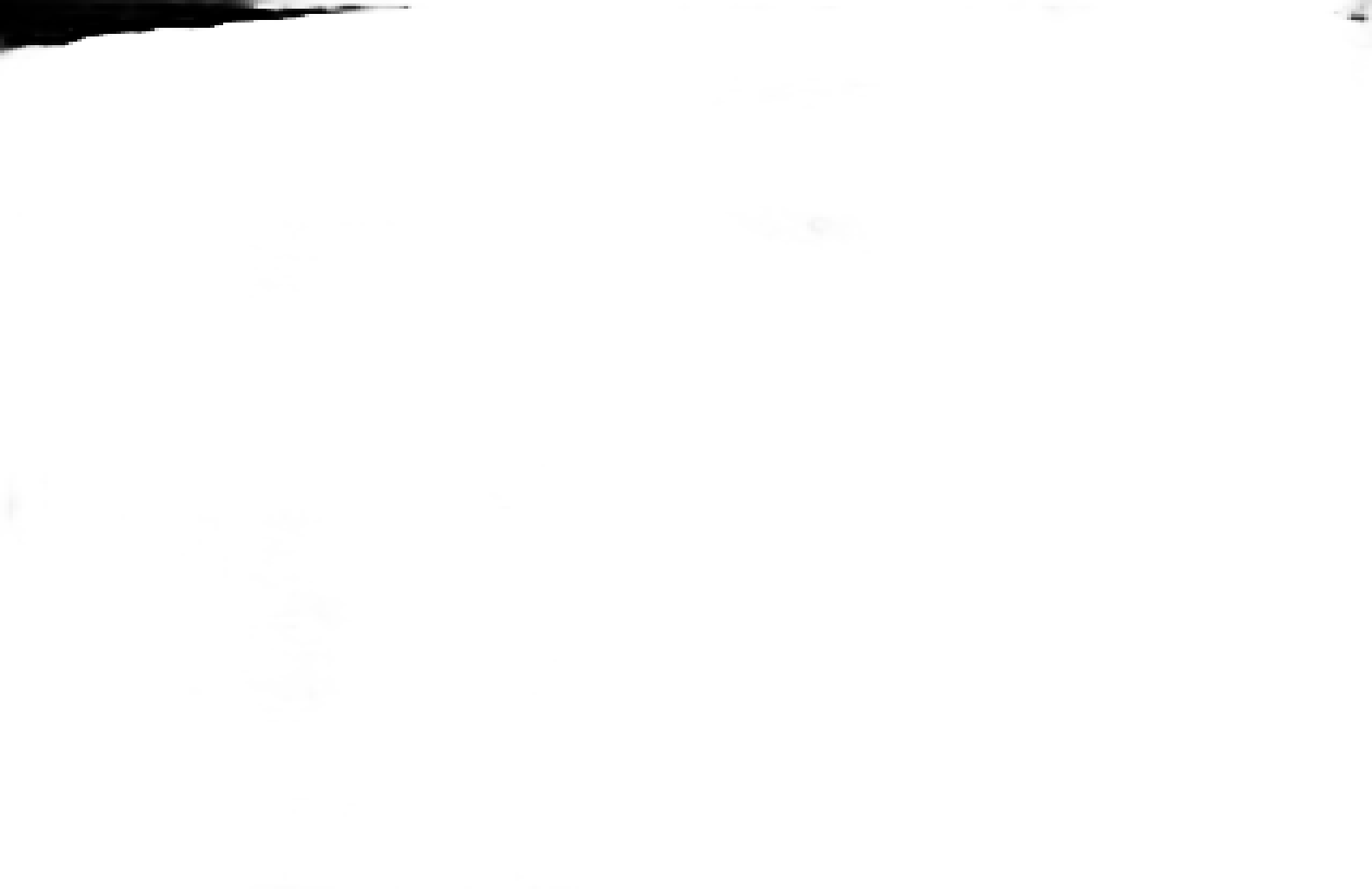}
	\end{subfigure}
	\begin{subfigure}{0.20\linewidth}
		\centering
		\includegraphics[width=0.90\linewidth]{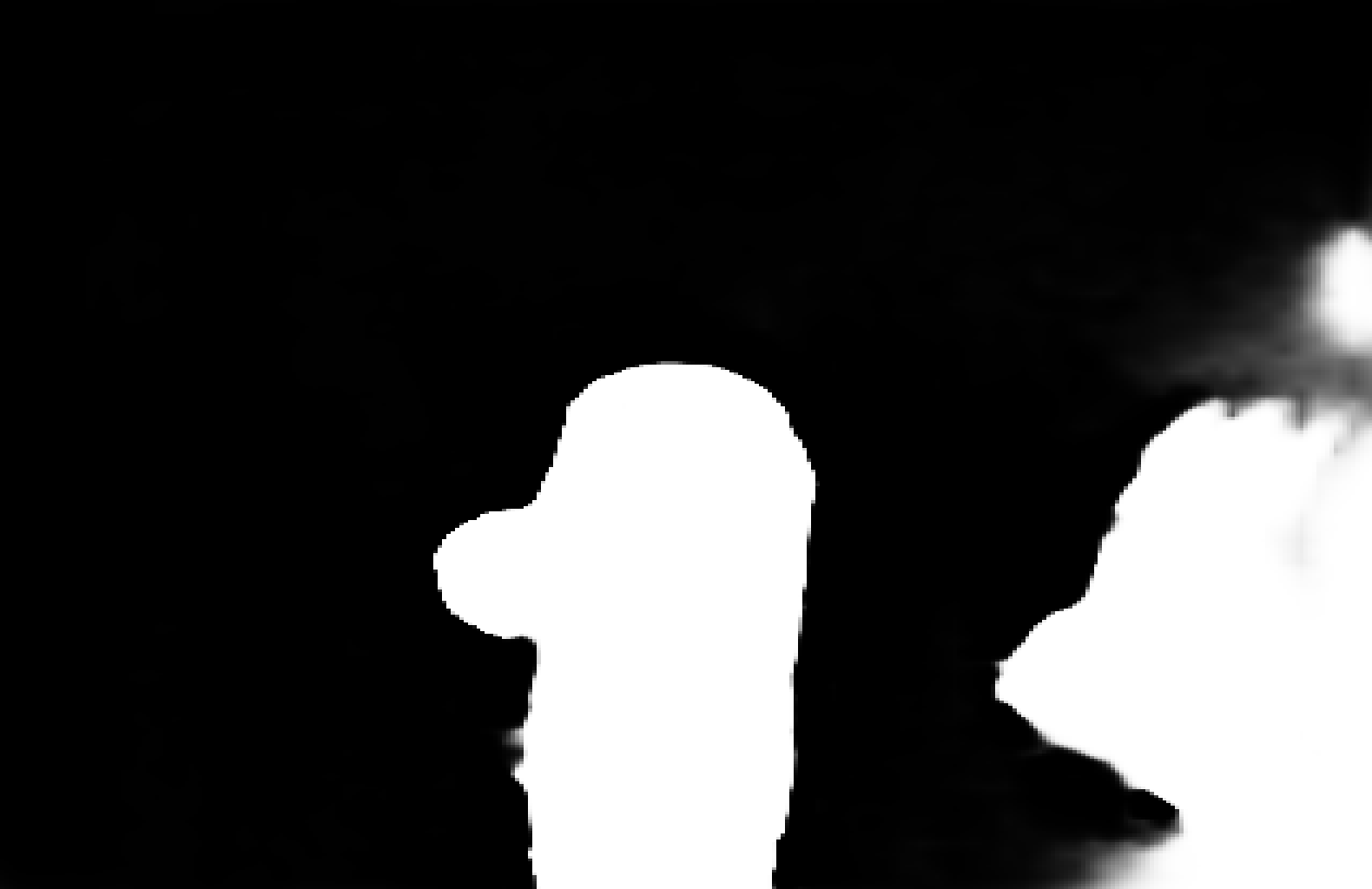}
	\end{subfigure}
	\begin{subfigure}{0.20\linewidth}
		\centering
		\includegraphics[width=0.90\linewidth]{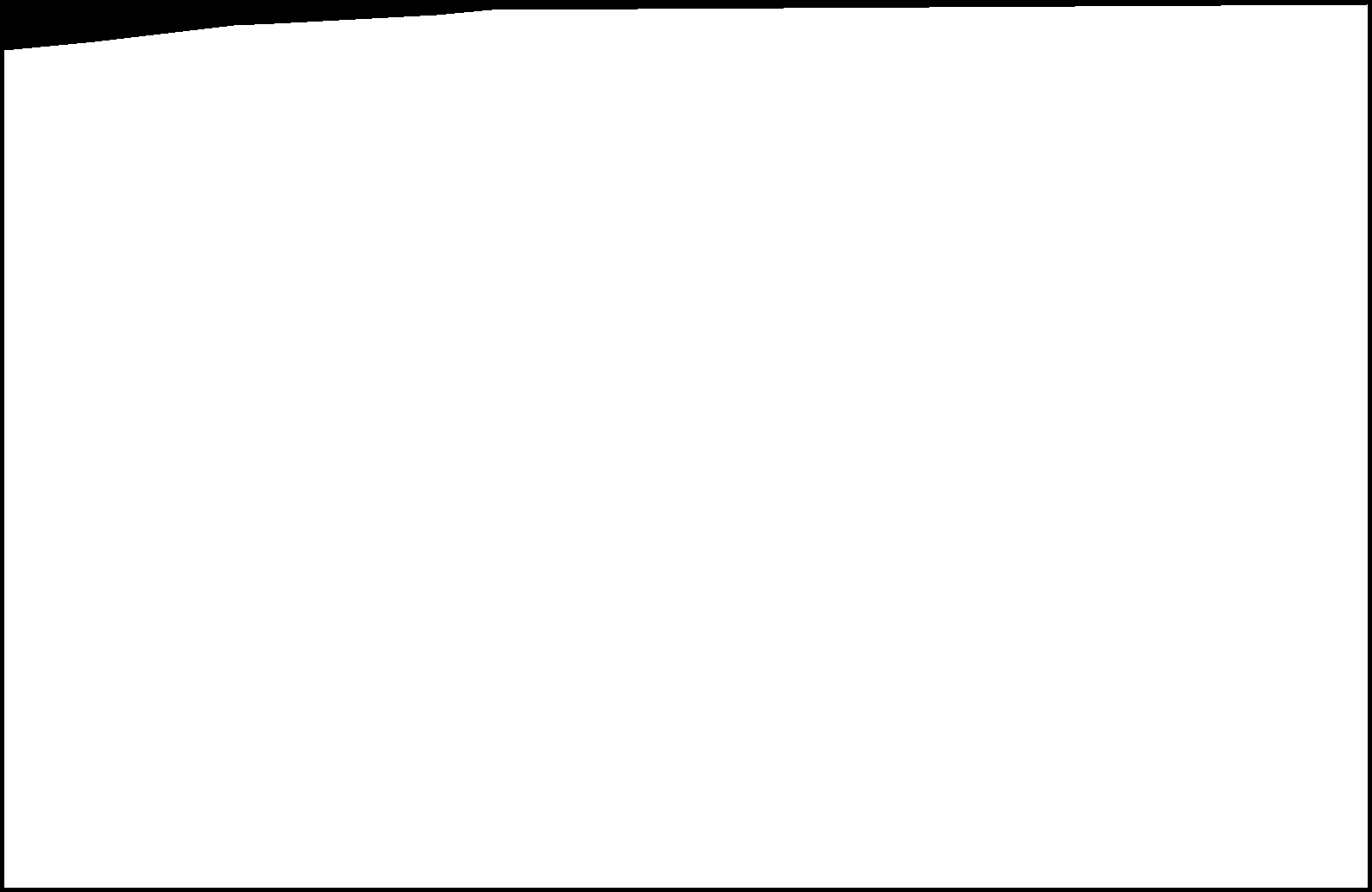}
	\end{subfigure}
 
	\begin{subfigure}{0.20\linewidth}
		\centering
		\includegraphics[width=0.90\linewidth]{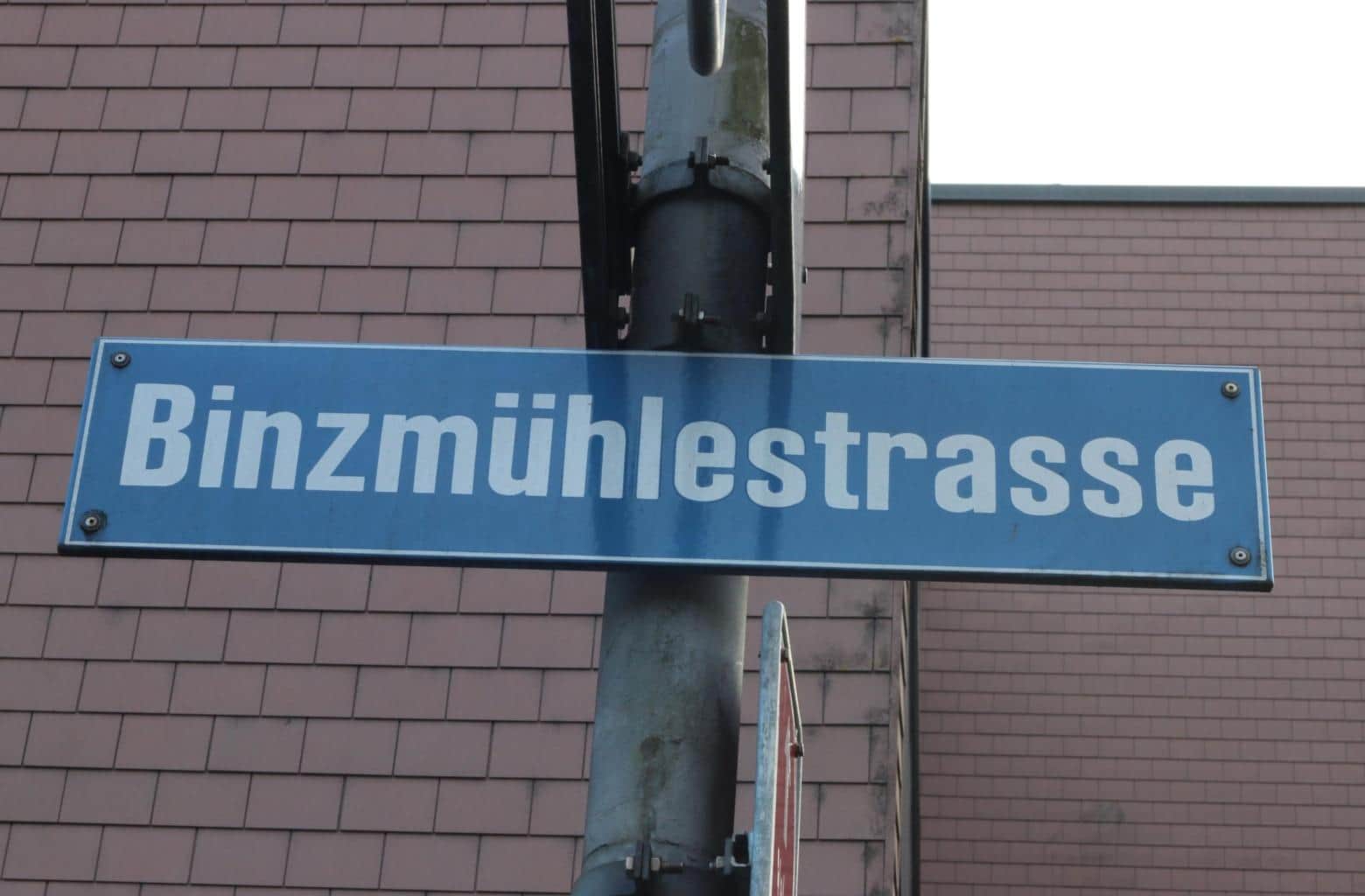}
	\end{subfigure}
	\centering
	\begin{subfigure}{0.20\linewidth}
		\centering
		\includegraphics[width=0.90\linewidth]{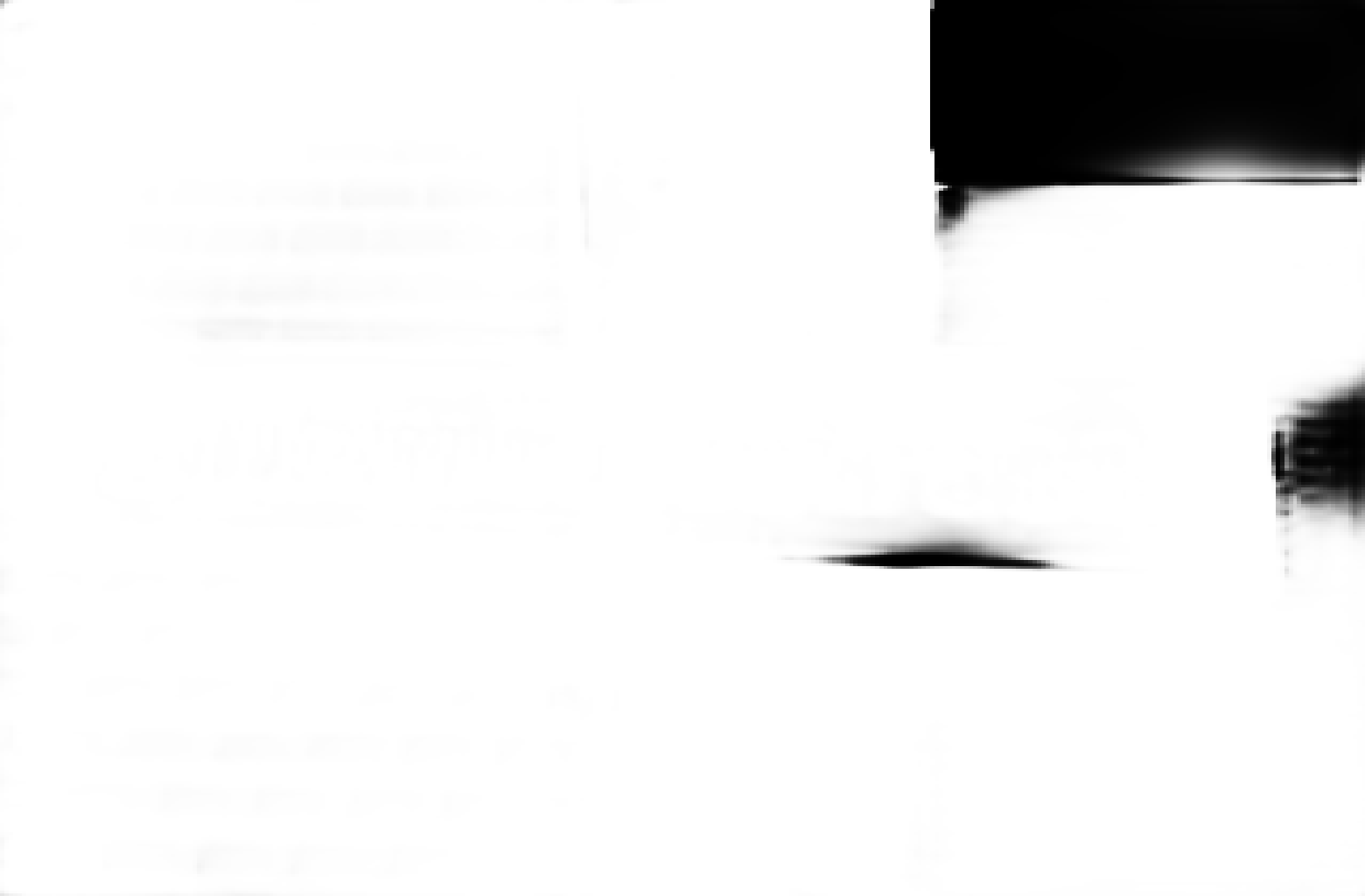}
	\end{subfigure}
	\begin{subfigure}{0.20\linewidth}
		\centering
		\includegraphics[width=0.90\linewidth]{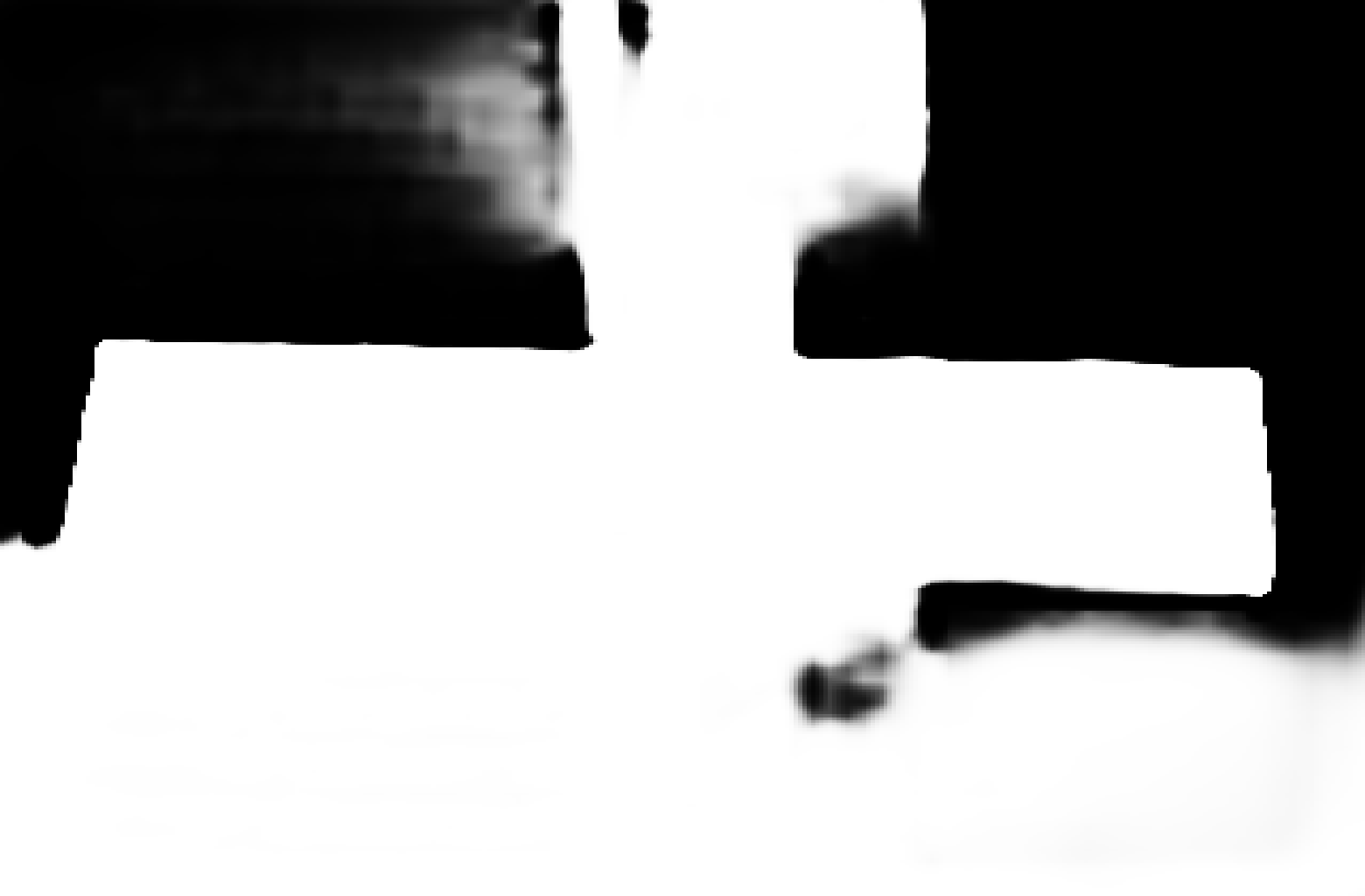}
	\end{subfigure}
	\begin{subfigure}{0.20\linewidth}
		\centering
		\includegraphics[width=0.90\linewidth]{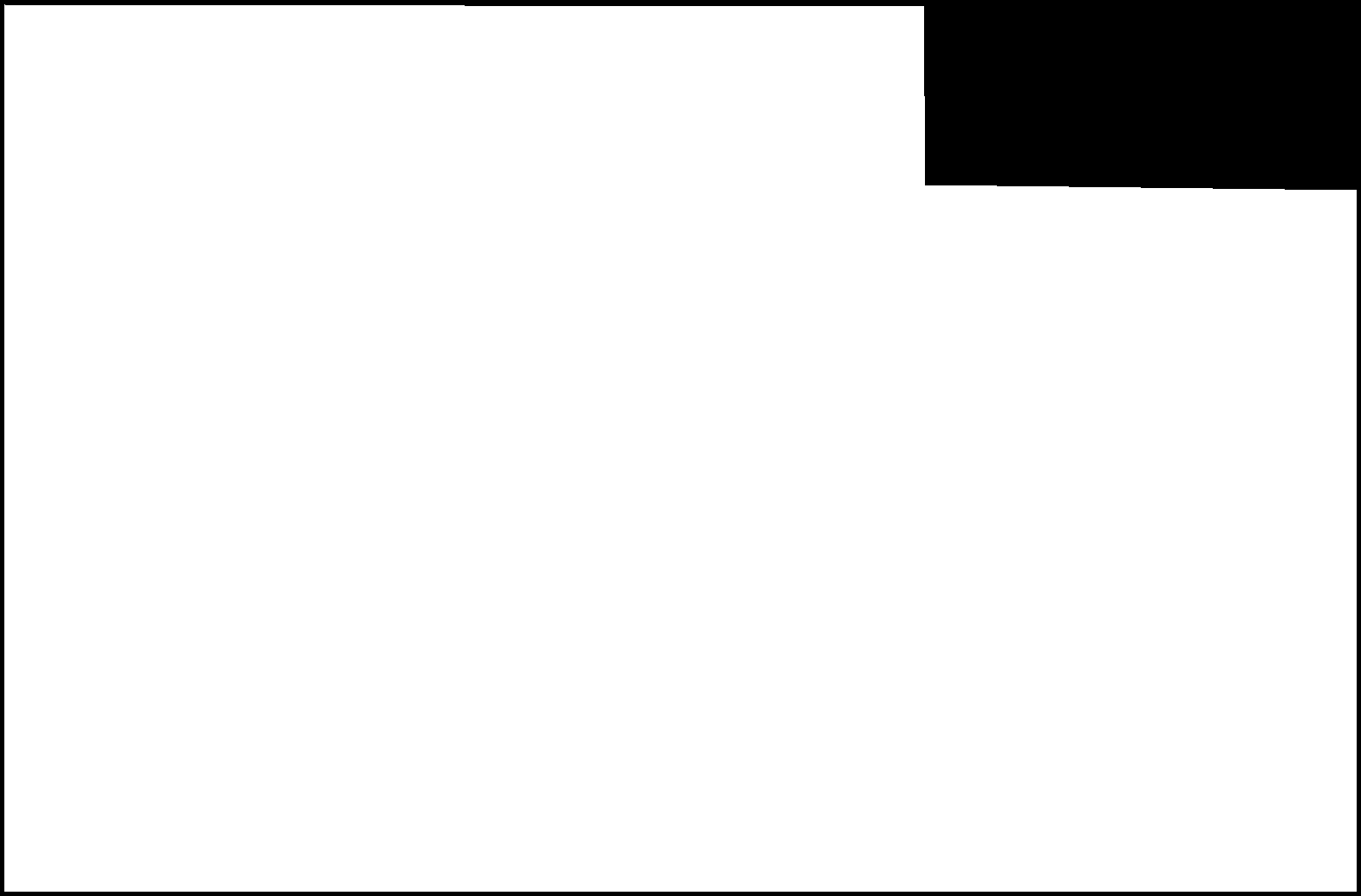}
	\end{subfigure}
 
	\begin{subfigure}{0.20\linewidth}
		\centering
		\includegraphics[width=0.90\linewidth]{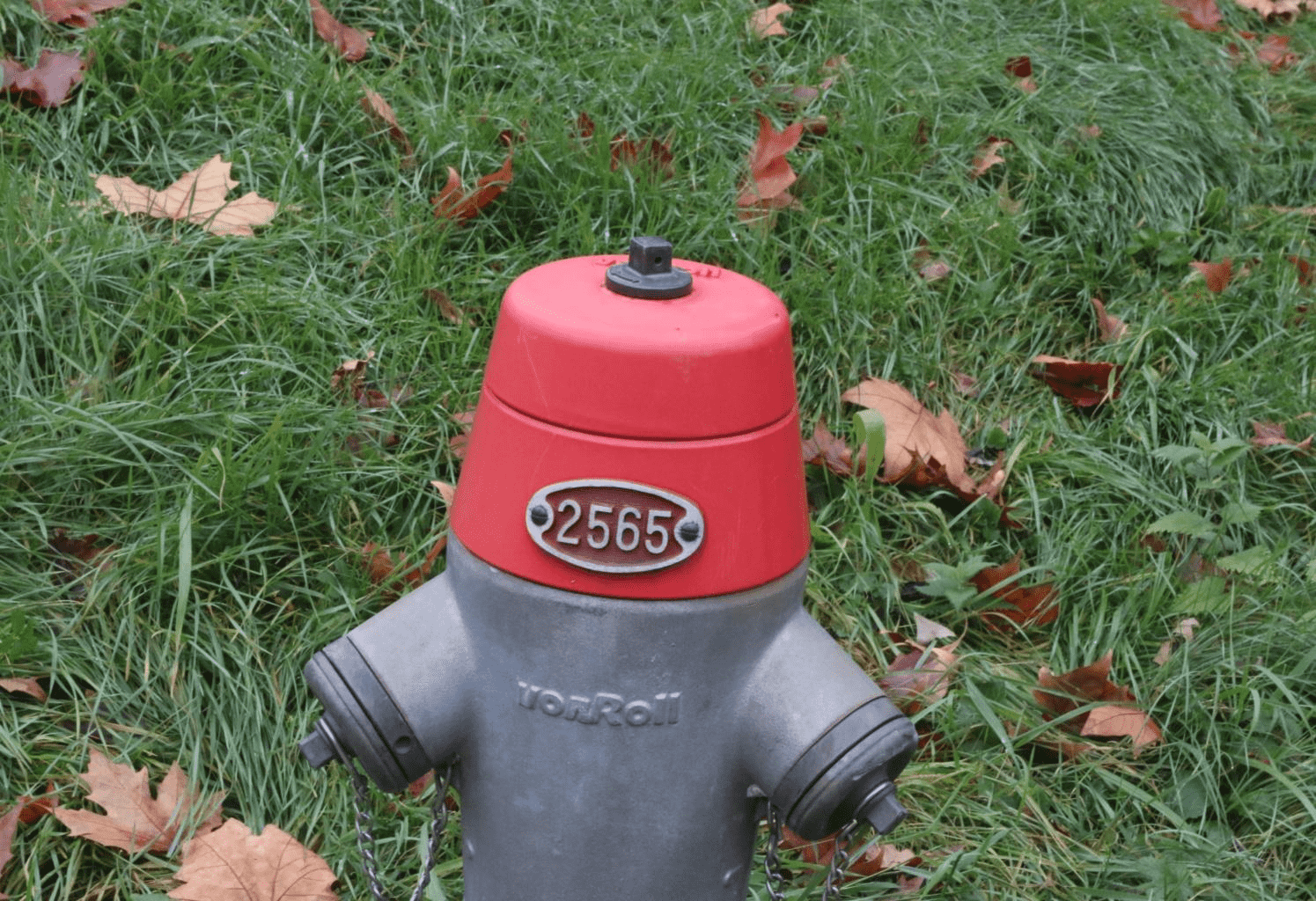}
	\end{subfigure}
	\centering
	\begin{subfigure}{0.20\linewidth}
		\centering
		\includegraphics[width=0.90\linewidth]{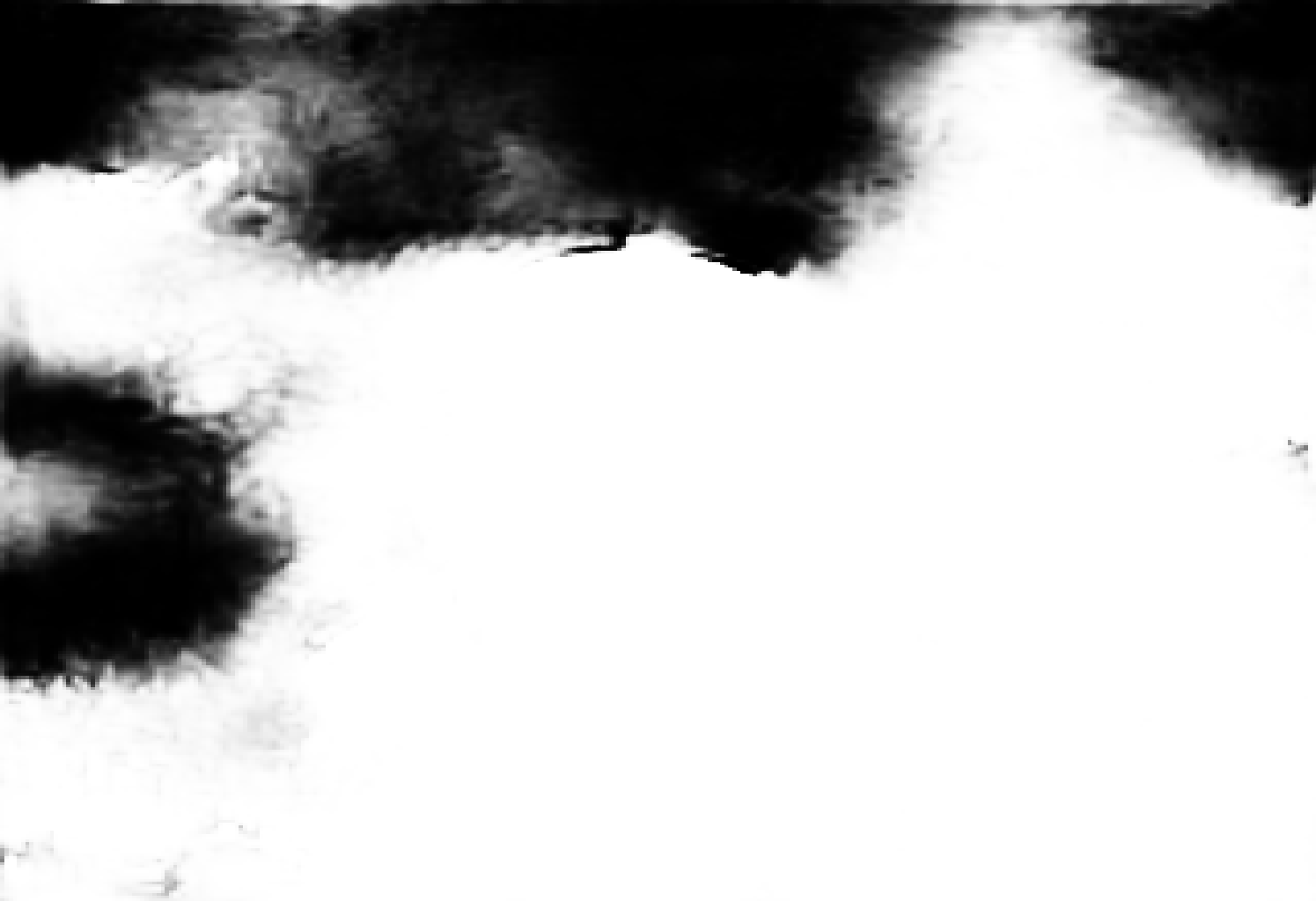}
	\end{subfigure}
	\begin{subfigure}{0.20\linewidth}
		\centering
		\includegraphics[width=0.90\linewidth]{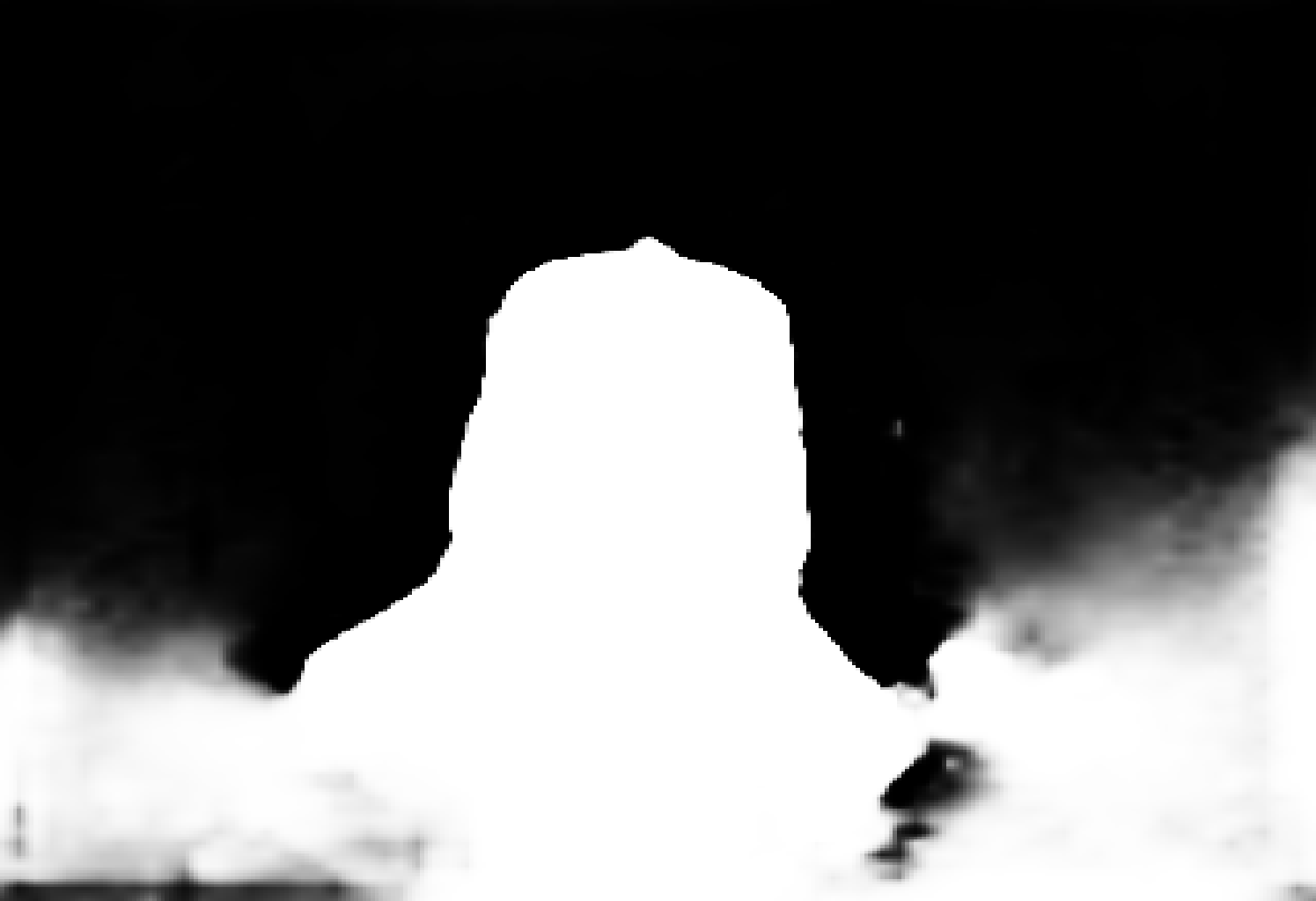}
	\end{subfigure}
	\begin{subfigure}{0.20\linewidth}
		\centering
		\includegraphics[width=0.90\linewidth]{Fig/vgg/white.png}
	\end{subfigure}
 
	\begin{subfigure}{0.20\linewidth}
		\centering
		\includegraphics[width=0.90\linewidth]{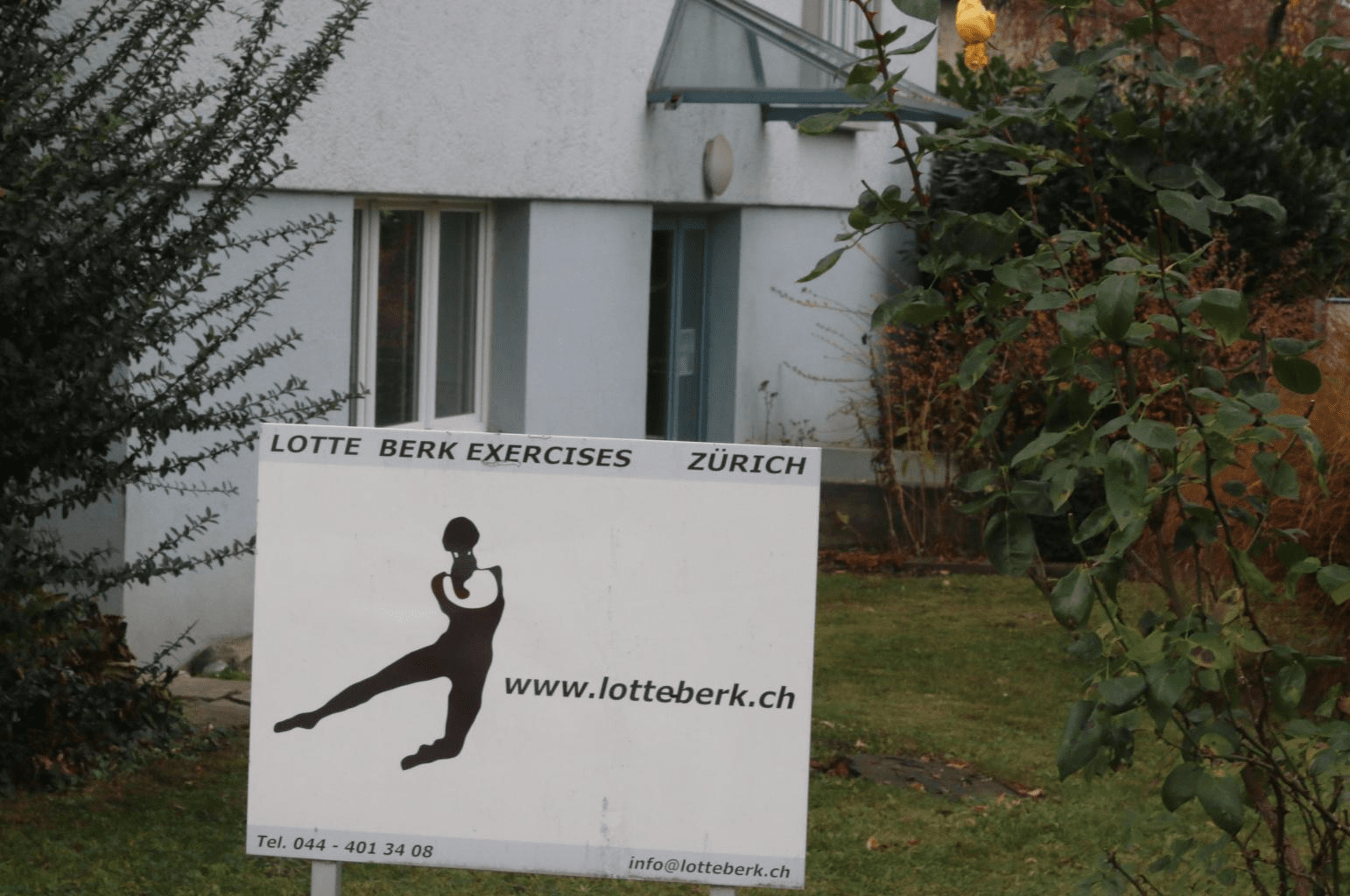}
	\end{subfigure}
	\centering
	\begin{subfigure}{0.20\linewidth}
		\centering
		\includegraphics[width=0.90\linewidth]{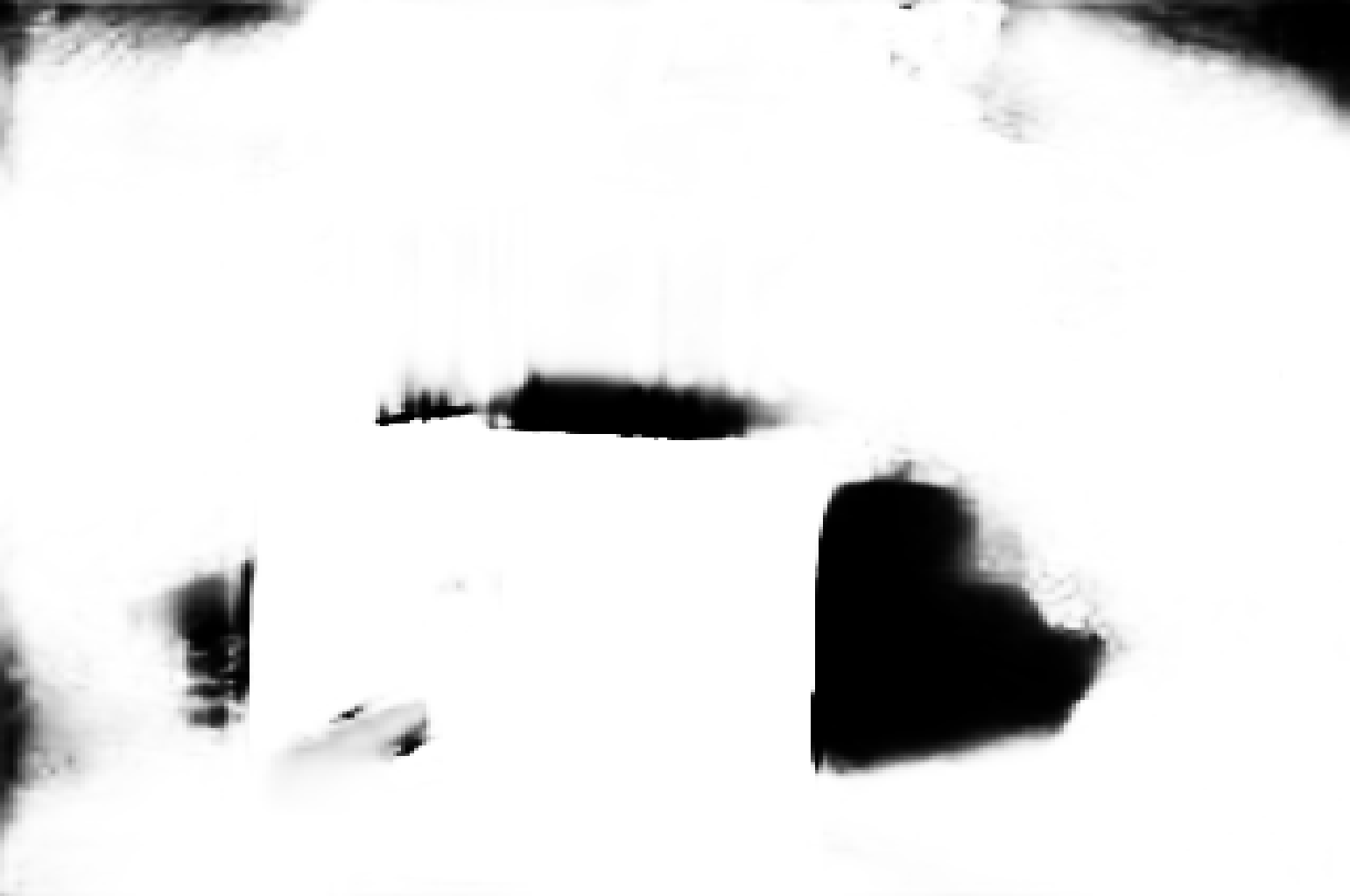}
	\end{subfigure}
	\begin{subfigure}{0.20\linewidth}
		\centering
		\includegraphics[width=0.90\linewidth]{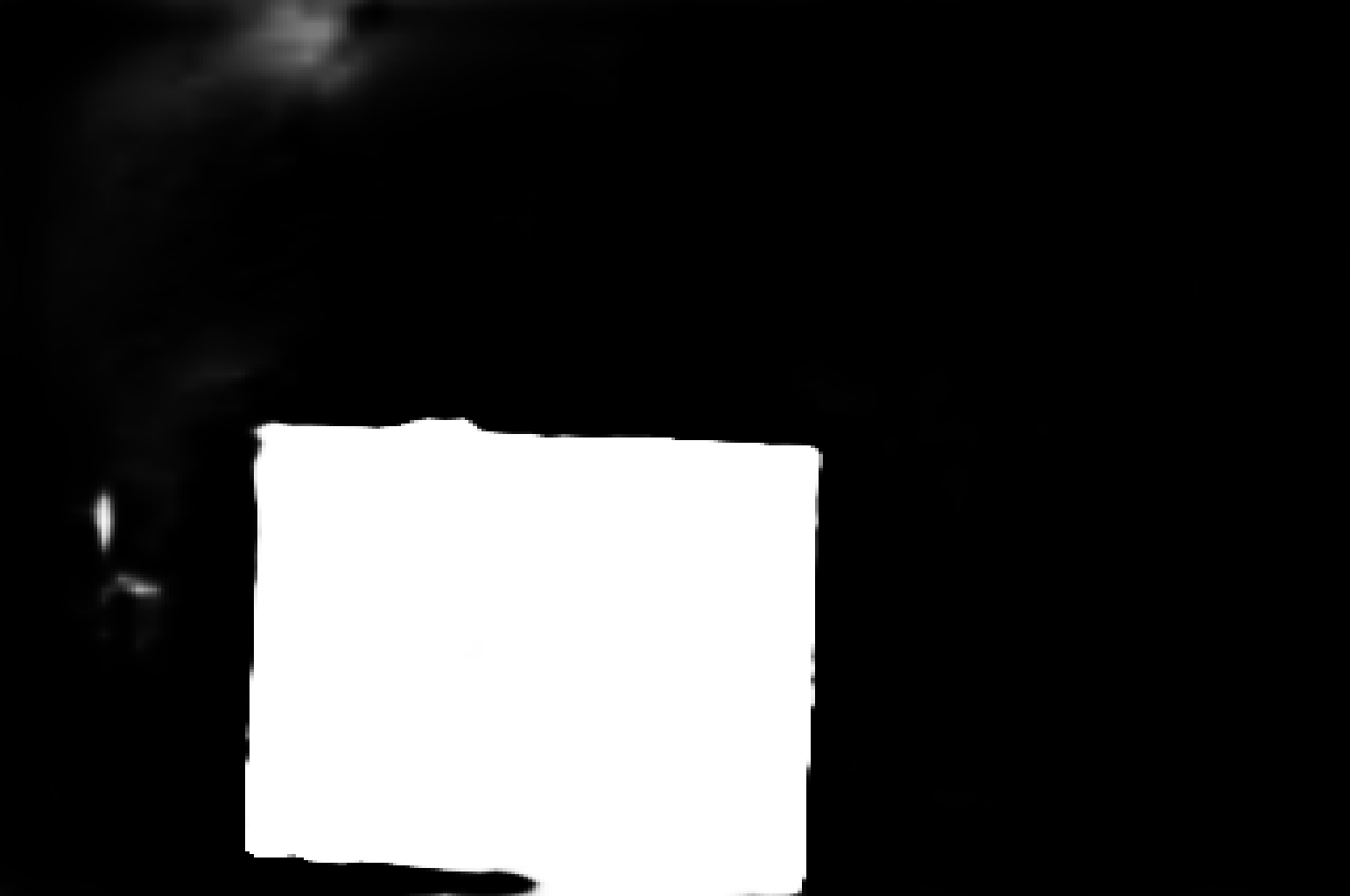}
	\end{subfigure}
	\begin{subfigure}{0.20\linewidth}
		\centering
		\includegraphics[width=0.90\linewidth]{Fig/vgg/white.png}
	\end{subfigure}

	\begin{subfigure}{0.20\linewidth}
		\centering
		\includegraphics[width=0.90\linewidth]{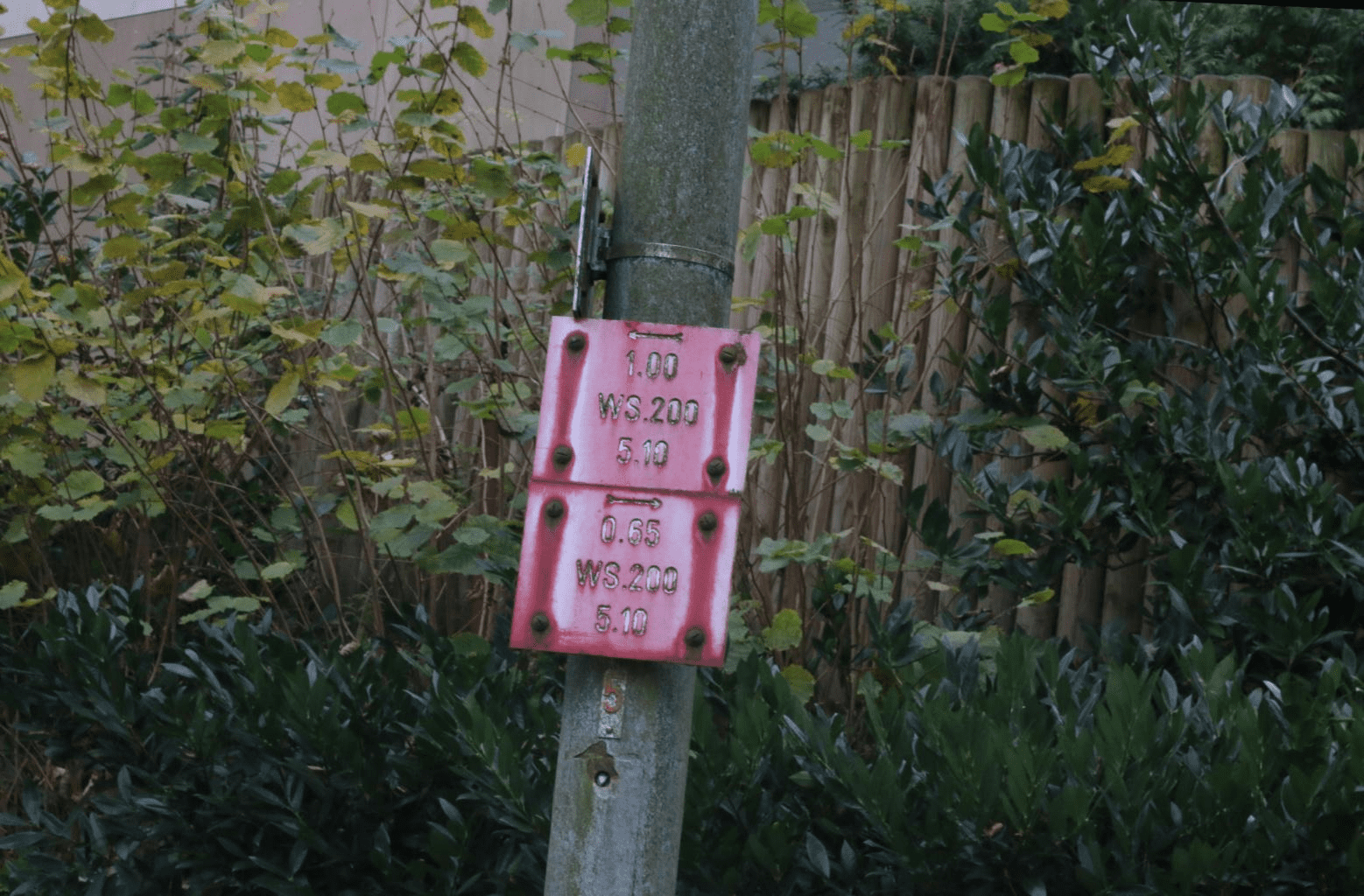}
	\end{subfigure}
	\centering
	\begin{subfigure}{0.20\linewidth}
		\centering
		\includegraphics[width=0.90\linewidth]{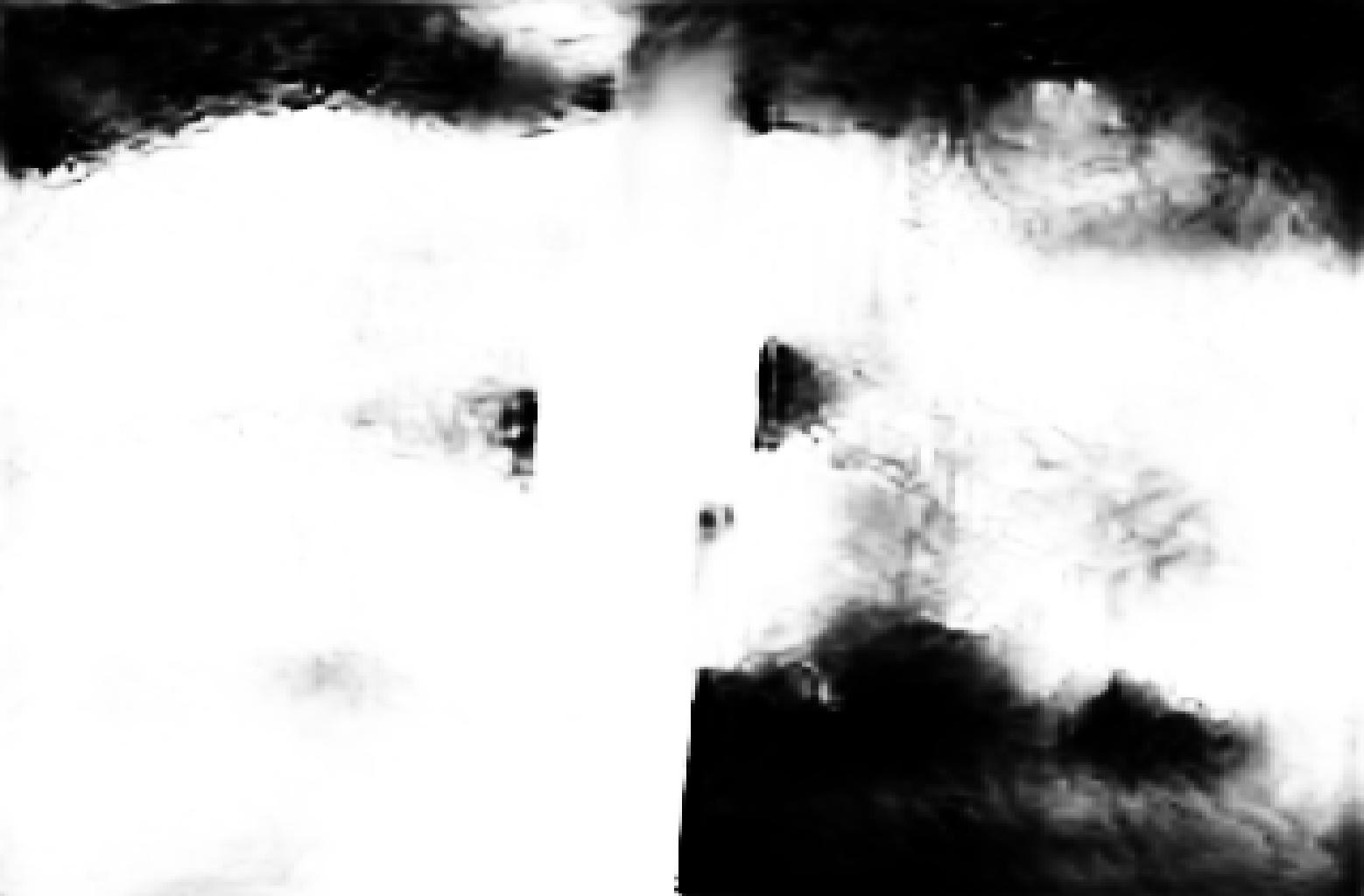}
	\end{subfigure}
	\begin{subfigure}{0.20\linewidth}
		\centering
		\includegraphics[width=0.90\linewidth]{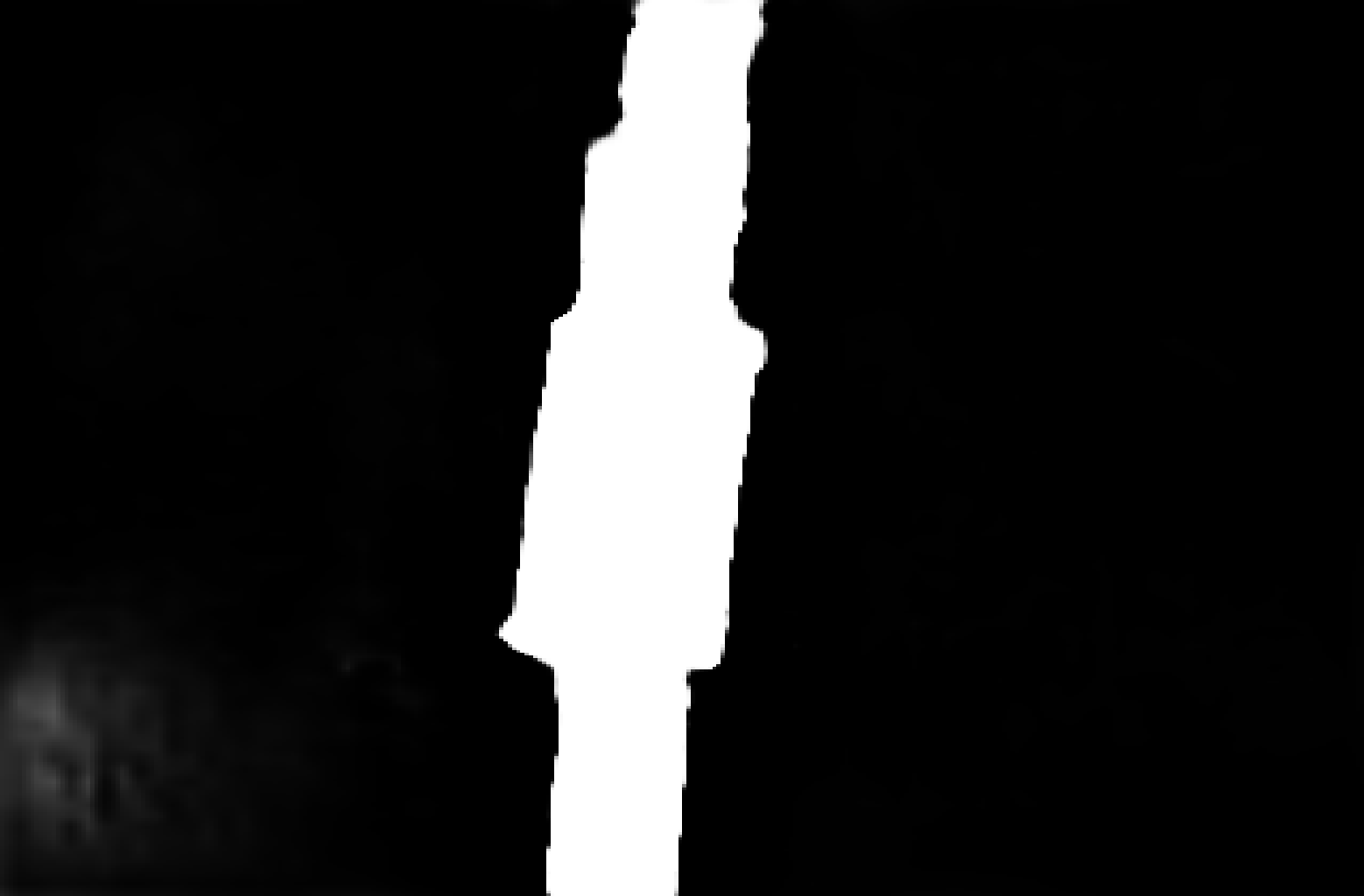}
	\end{subfigure}
	\begin{subfigure}{0.20\linewidth}
		\centering
		\includegraphics[width=0.90\linewidth]{Fig/vgg/white.png}
	\end{subfigure}

	\begin{subfigure}{0.20\linewidth}
		\centering
		\includegraphics[width=0.90\linewidth]{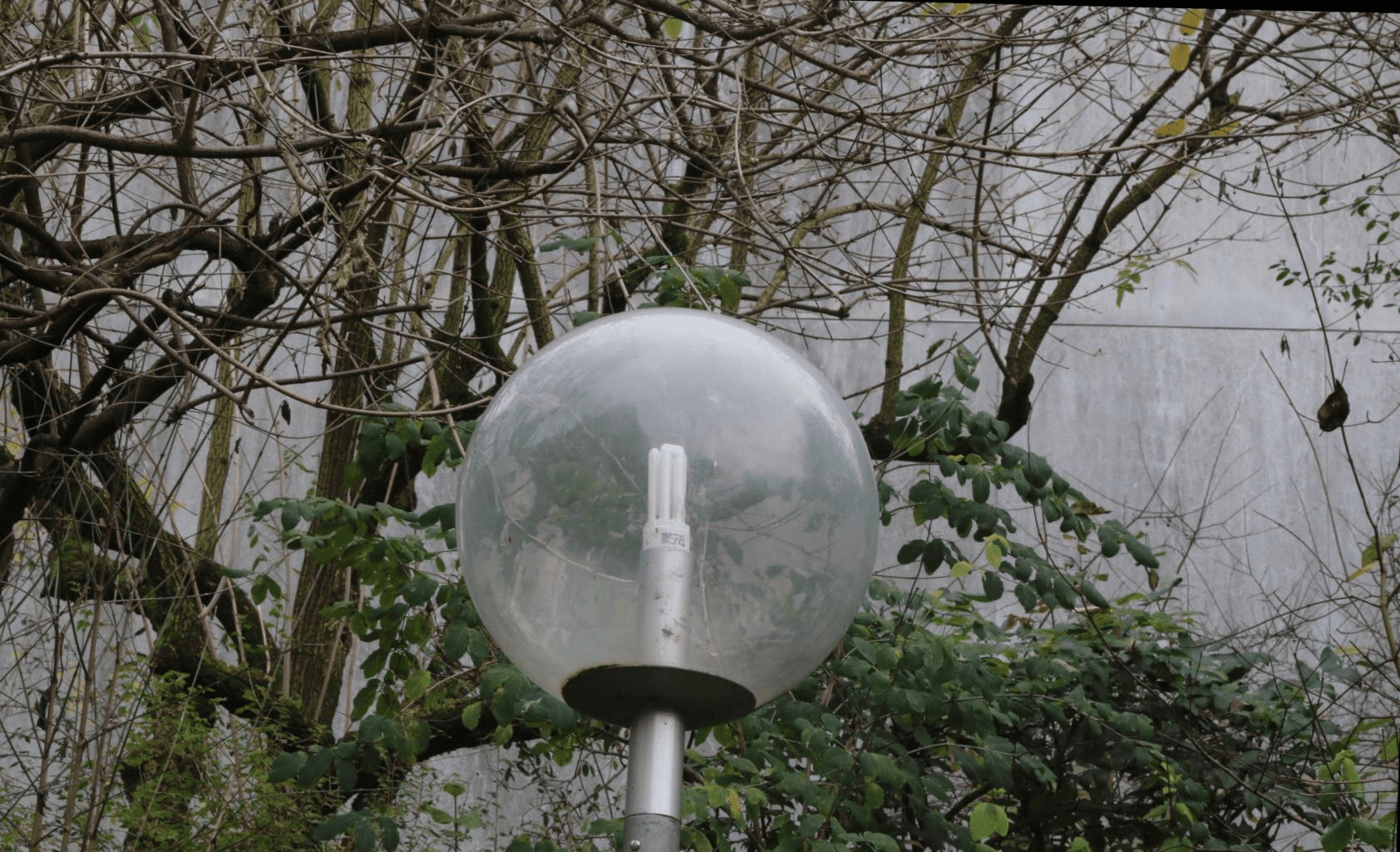}
	\end{subfigure}
	\centering
	\begin{subfigure}{0.20\linewidth}
		\centering
		\includegraphics[width=0.90\linewidth]{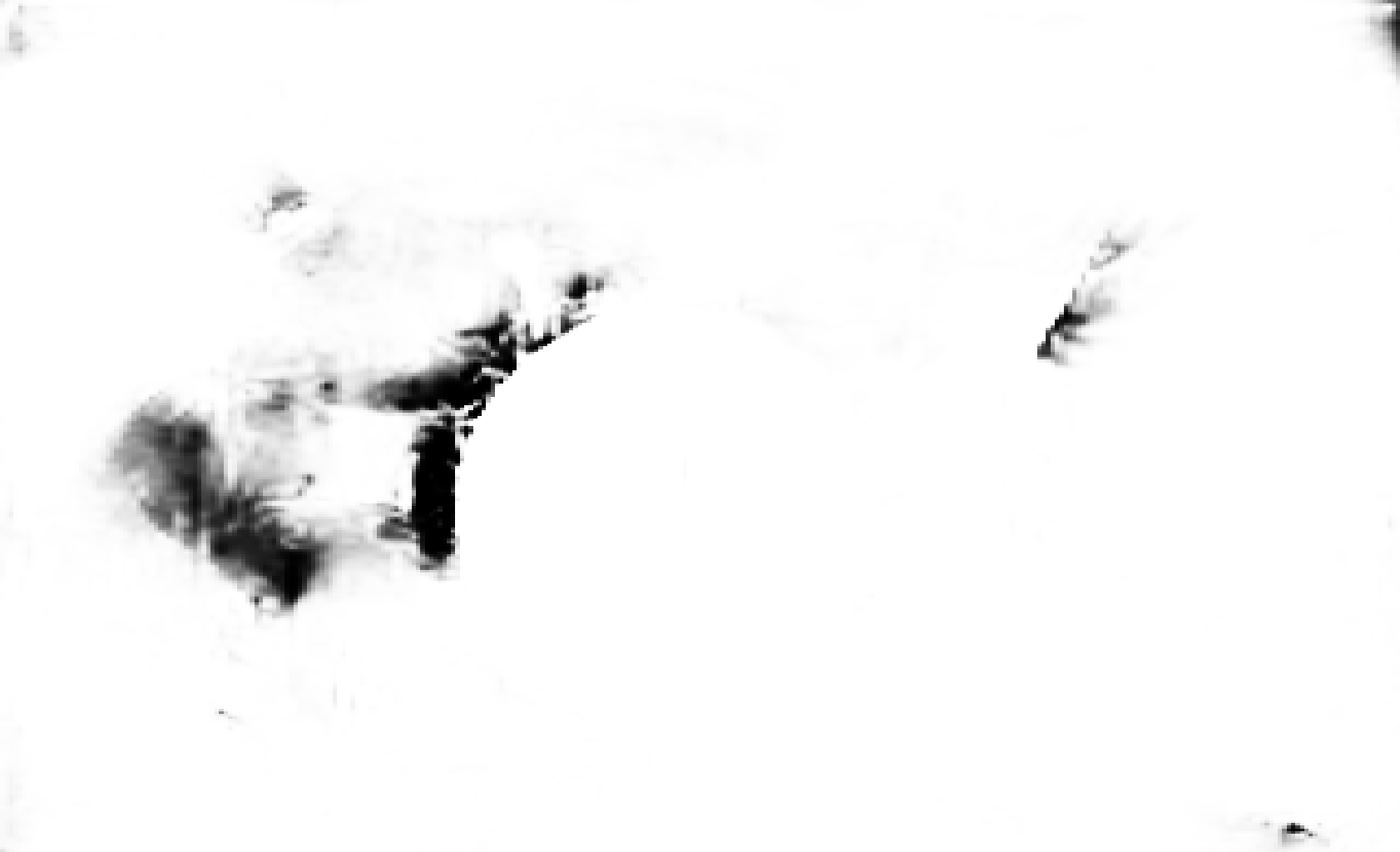}
	\end{subfigure}
	\begin{subfigure}{0.20\linewidth}
		\centering
		\includegraphics[width=0.90\linewidth]{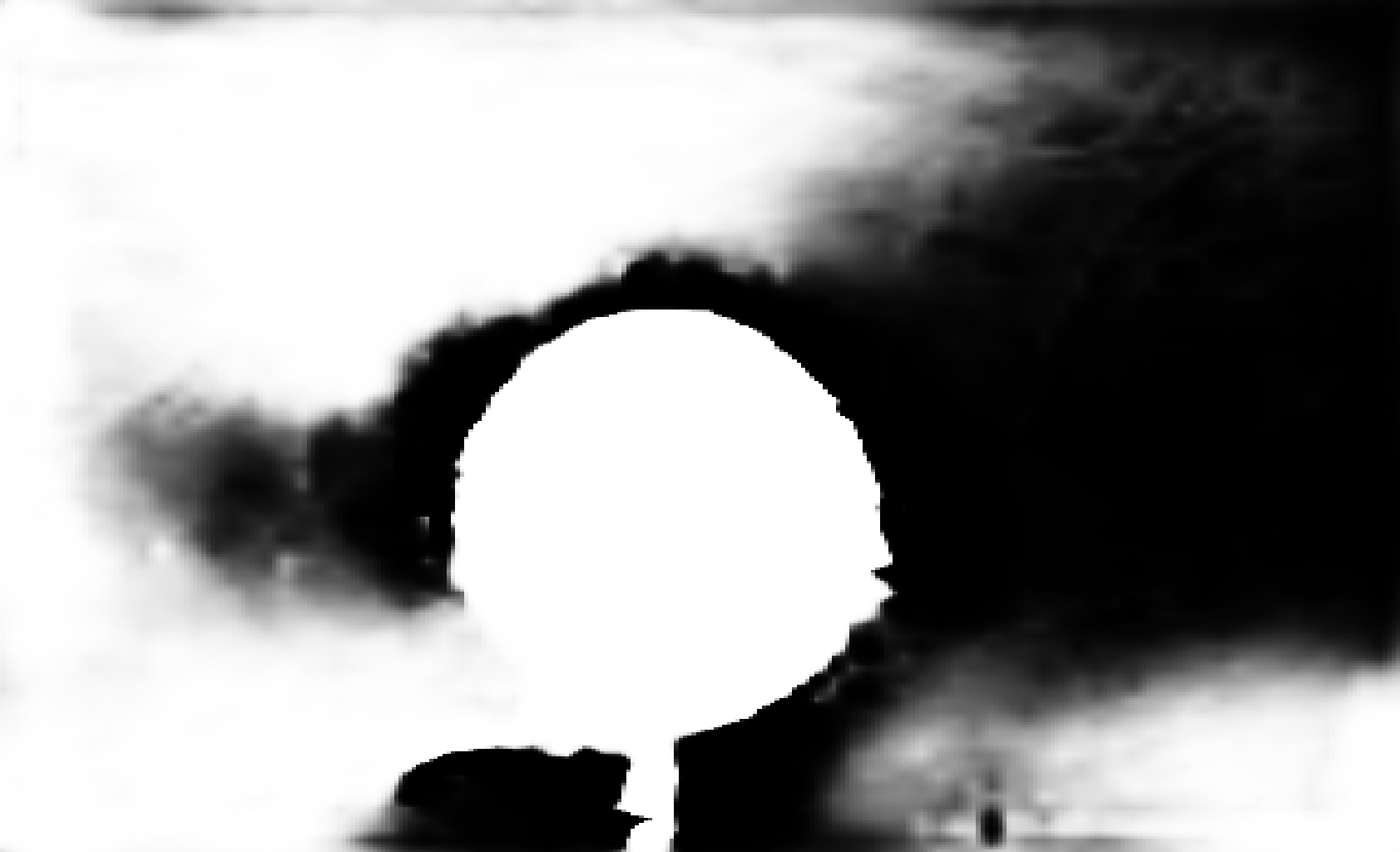}
	\end{subfigure}
	\begin{subfigure}{0.20\linewidth}
		\centering
		\includegraphics[width=0.90\linewidth]{Fig/vgg/white.png}
	\end{subfigure}

	\begin{subfigure}{0.20\linewidth}
	\captionsetup{font={scriptsize }}
		\centering
		\includegraphics[width=0.90\linewidth]{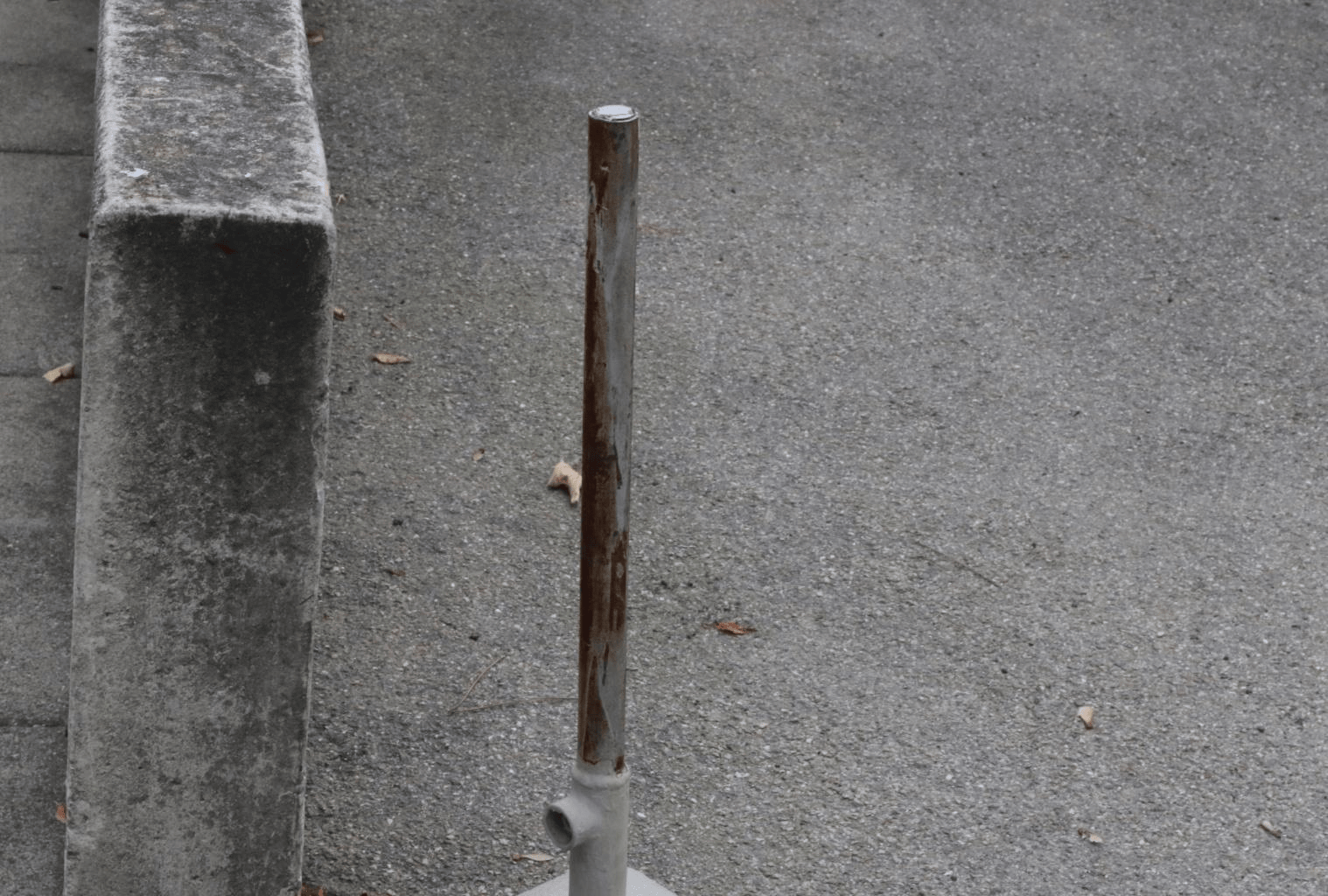}
		\caption{BCE}
	\end{subfigure}
	\centering
	\begin{subfigure}{0.20\linewidth}
	\captionsetup{font={scriptsize }}
		\centering
		\includegraphics[width=0.90\linewidth]{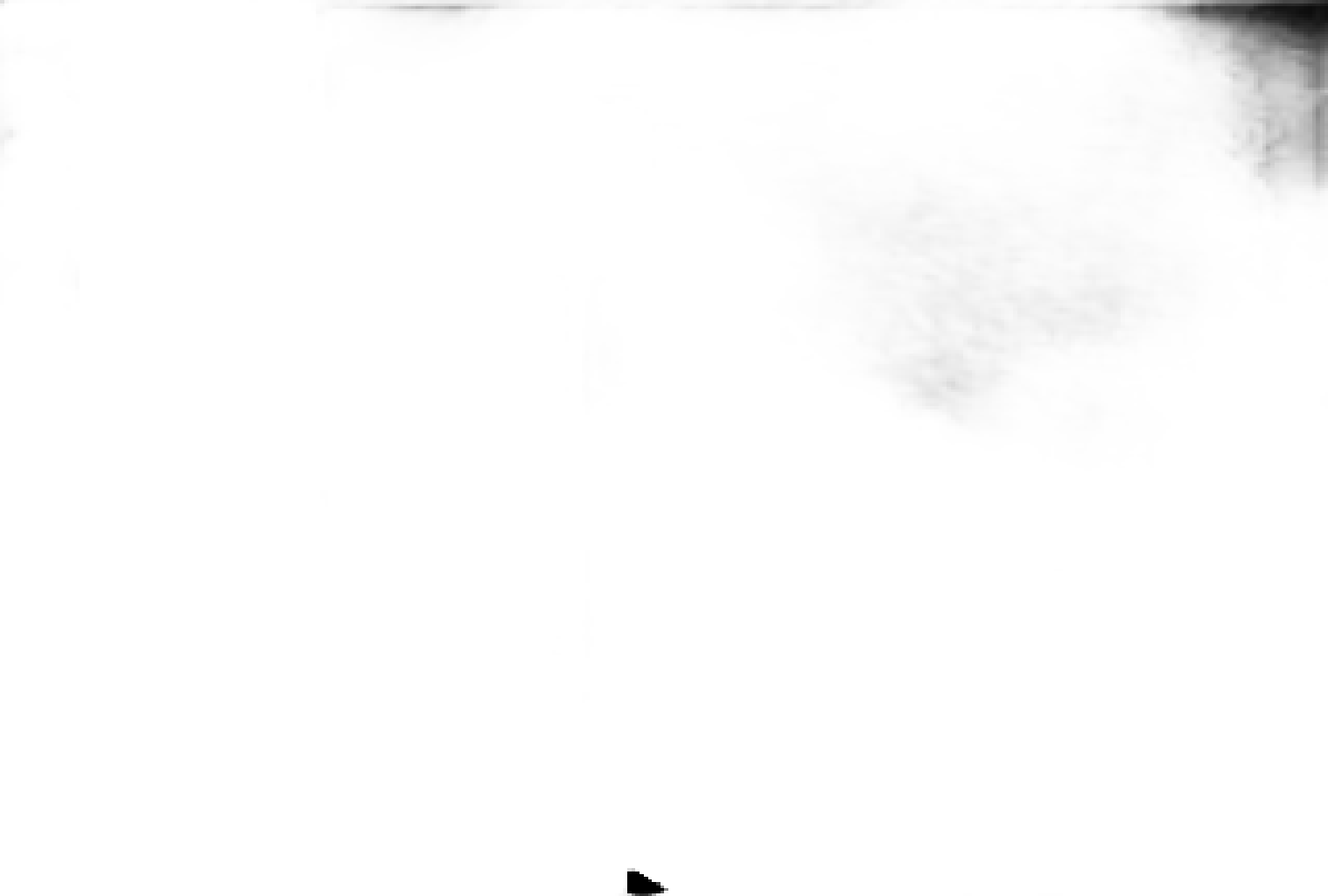}
		\caption{Vgg16}
	\end{subfigure}
	\begin{subfigure}{0.20\linewidth}
	\captionsetup{font={scriptsize }}
		\centering
		\includegraphics[width=0.90\linewidth]{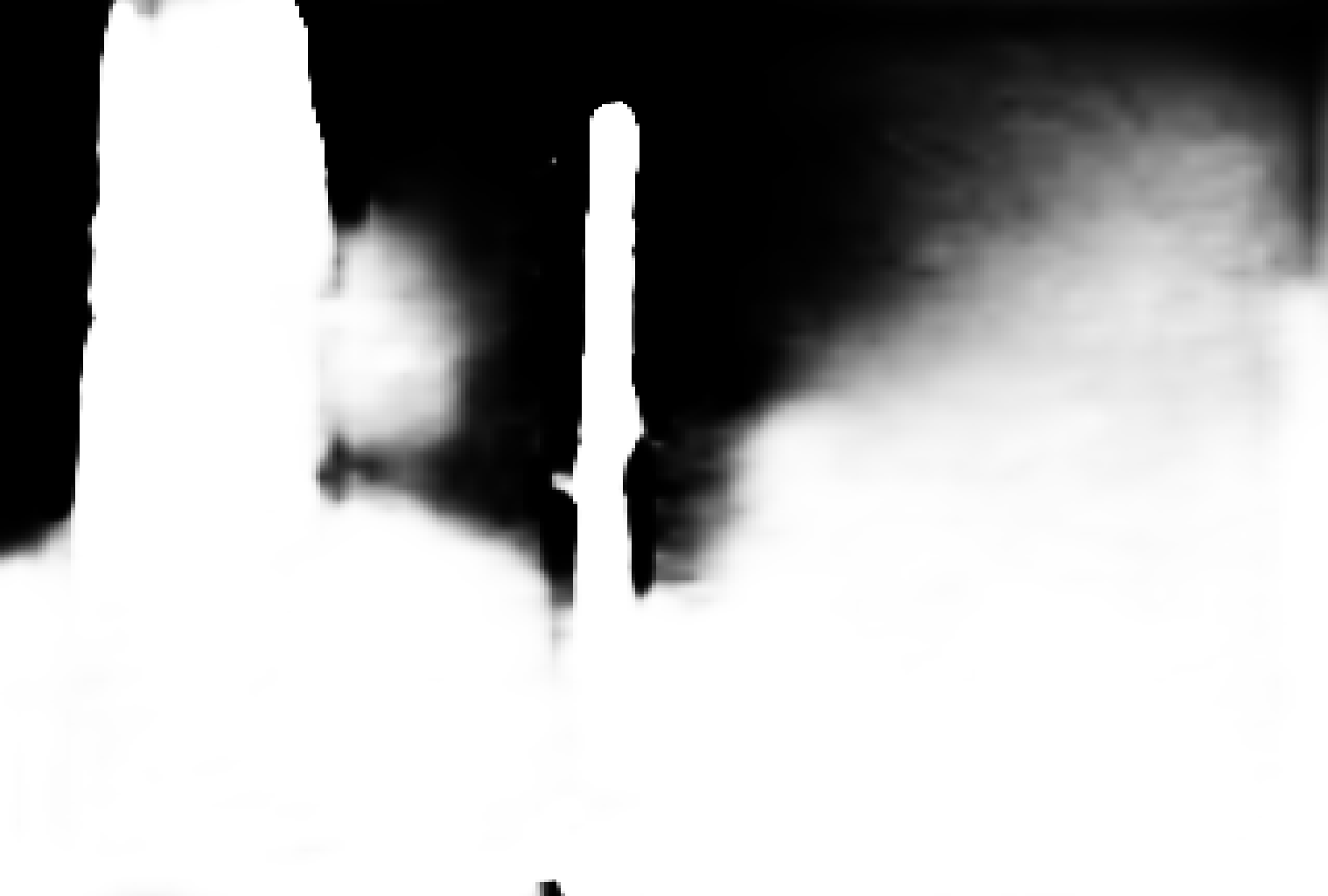}
		\caption{ResNest101}
	\end{subfigure}
	\begin{subfigure}{0.20\linewidth}
	\captionsetup{font={scriptsize }}
		\centering
		\includegraphics[width=0.90\linewidth]{Fig/vgg/white.png}
		\caption{GTs}
	\end{subfigure}
\caption{Comparison of D-DFFNet using ResNest101 and Vgg16 as backbone on wide-DOF images from EBD dataset.}
\label{future work}
\end{figure}

\end{appendices}

\end{document}